\documentclass{article}

    \PassOptionsToPackage{numbers, sort&compress}{natbib}

 \usepackage[eandd, final]{neurips_2026}

\usepackage[utf8]{inputenc} %
\usepackage[T1]{fontenc}    %
\usepackage{hyperref}       %
\usepackage{url}            %
\usepackage{booktabs}       %
\usepackage{amsfonts}       %
\usepackage{nicefrac}       %
\usepackage{microtype}      %
\usepackage{xcolor}         %
\usepackage{wrapfig}

\definecolor{linkblue}{rgb}{0.08, 0.30, 0.60}
\definecolor{citegreen}{rgb}{0.00, 0.40, 0.20}
\hypersetup{
    colorlinks=true,
    linkcolor=linkblue,
    citecolor=citegreen,
    urlcolor=linkblue,
    filecolor=linkblue,
}

\usepackage{amsmath}
\usepackage{amssymb}
\usepackage{mathtools}
\usepackage{amsthm}

\theoremstyle{plain}
\newtheorem{theorem}{Theorem}[section]

\newtheorem{lemma}[theorem]{Lemma}

\theoremstyle{definition}

\theoremstyle{remark}

\makeatletter
\newtheorem*{rep@theorem}{\rep@title}
\newcommand{\newreptheorem}[2]{%
	\newenvironment{rep#1}[1]{%
		\def\rep@title{{\normalfont \textbf{#2} \ref{##1}}}%
		\begin{rep@theorem}}%
		{\end{rep@theorem}}}
\makeatother
\newreptheorem{lemma}{Lemma}

\usepackage{enumitem}
\usepackage{dsfont}
\usepackage{color,soul}
\usepackage{subfig}
\usepackage{dsfont}

\usepackage[most]{tcolorbox}
\newtcolorbox{rqbox}[1][]{%
  enhanced, 
  breakable,
  colback=gray!10,
  colframe=black,
  boxrule=0.5pt,
  arc=2pt,
  auto outer arc,
  left=0pt,
  right=0pt,
  top=0pt,     %
  bottom=0pt,  %
  #1
}

\newcommand{\loss}{\mathtt{loss}}

\newcommand{\ct}{\mathtt{ctx}}
\newcommand{\recm}{\mathtt{rec}}
\newcommand{\cf}{\mathtt{cf}}
\newcommand{\lang}{L}

\makeatletter
\def\namedlabel#1#2{\begingroup
    #2%
    \def\@currentlabel{#2}%
    \phantomsection\label{#1}\endgroup
}
\makeatother

\title{Understanding the Interplay between Memorization and Learning in Large Language Models}

\author{%
  \textbf{Bishwamittra Ghosh\textsuperscript{1}, Soumi Das\textsuperscript{1}, Qinyuan Wu\textsuperscript{1}, Mohammad Aflah Khan\textsuperscript{1},} \\
  \textbf{Deepak Garg\textsuperscript{1}, Krishna P.~Gummadi\textsuperscript{1}, Evimaria Terzi\textsuperscript{2}} \\[0.5em]
  \normalfont\textsuperscript{1}Max Planck Institute for Software Systems, Germany \\
  \normalfont\textsuperscript{2}Boston University, USA
}

\begin{document}

\maketitle

\begin{abstract}
    We investigate foundational questions about the interplay between memorization and learning in large language models (LLMs). \textit{When an LLM generates a string, can we disentangle the roles of rote memorization and contextual learning?} Rote memorization results in an LLM regurgitating specific details of a training string, while contextual learning leads to the generation of new strings that follow generalizable patterns of the underlying language. 
    Related questions of interest include: \textit{can an LLM avoid memorization when optimally learning a language?}, and if not, \textit{can we design training schemes to reduce memorization and improve learning?}

    To address these questions, we propose \textit{contextual memorization}, a new measure that precisely disentangles memorization from learning. Using this measure, we establish that different training strings are memorized at different epochs in training, and that memorization of \textit{some} training strings is unavoidable when optimally learning a language. Finally, we propose memorization-aware training, a scheme that dynamically down-samples training strings based on their memorization, and in doing so, equalizes learning across strings, reduces memorization, and improves learning. Our conclusions are supported by extensive experiments on multiple LLMs using \textit{formal languages} as a controlled testbed, enabling precise language specification, exact string sampling, and elimination of data contamination; and further validated on natural languages.\footnote{Code: {\url{https://github.com/bishwamittra/formaLLM}}, and \href{https://huggingface.co/datasets/formallm/formal-grammar-llm-benchmark}{dataset}.}

\end{abstract}

\section{Introduction}
\label{sec:introduction}

{
\small
\begin{quote}
\centering
\emph{``Every teacher knows that there is a profound difference between a student learning a lesson by rote and learning it with understanding, or meaningfully.''\\}
-- Herbert Simon in The Sciences of the Artificial, 1996
\end{quote}
}

The unsupervised pre-training and fine-tuning of LLMs can result in both \textit{rote memorization} of specific details of individual training strings~\citep{bender2021dangers}, as well as \textit{contextual learning} of patterns of the language represented by the training strings~\citep{bubeck2023sparks}. 
Rote memorization can lead to regurgitation of training strings, while contextual learning can lead to generation of new strings not seen during training. 
So memorization is often considered the ugly cousin of contextual learning; an undesirable side effect of learning that should be avoided~\citep{hernandez2022scaling}. 
In this paper, we center our investigations on the \textit{interplay between memorization and learning}. 
Specifically, we ask (a) during training, can we determine the epoch (iteration) when a specific training string begins to be memorized by an LLM? (b) can an LLM avoid memorizing any and all training strings, when optimally learning a language?, and (c) if not, can we design training schemes for LLMs to reduce memorization and improve learning?

Our investigation begins by re-examining how researchers \emph{operationalize} memorization, i.e., whether and how they distinguish between cases where the generation of a string by an LLM is attributed to memorization versus contextual learning. Motivated by capturing privacy risks, the majority of measures estimate memorization by how well the LLM \textit{recollects} training strings~\citep{carlini2021extracting,tirumala2022memorization}. 
However, there can be cases when such recollection is not based on memorization. 
For example, consider asking an LLM to \texttt{count from $1$ to $1000$}.
Many LLMs today will likely recollect \texttt{$ 1, 2 ,\cdots, 1000 $} based on simple reasoning, i.e., contextual learning, and without needing to be trained on the exact input-output string.
To call such recollection a \textit{gray area} of memorization, as suggested in~\citet{NEURIPS2024_66453d57}, risks misclassifying contextual learning as memorization-based recollection.

\paragraph{The Case for Contextual Measures: Disentangling Memorization from Learning.}
How else could one quantify memorization?
Let's first conduct a thought experiment to illustrate a challenging desideratum for memorization measures.
Imagine an English speaker and a German speaker commit a paragraph in German to memory.
When recollecting the paragraph, do the two speakers rely on memorization to the same or different extents? 
Intuitively, the German speaker understands the syntax and semantics of the tokens in the paragraph, while the English speaker sees the paragraph as a sequence of alphabet tokens. 
Even before reading the paragraph, given some prefix, the former is more likely to predict the next token correctly than the latter. 
So it stands to reason that the extent of memorization involved in recollecting the paragraph is higher for the English speaker than the German speaker. 
A good memorization measure should therefore account for the LLM's ability to predict the next token based on context. We use ``context'' broadly, beyond the input prefix, to mean generalizable patterns of the language that enable generation of unseen strings. We believe the growing ability of LLMs to produce new content (e.g., solving open problems in mathematics) stems from contextual learning and not necessarily from memorization. %

\begin{wrapfigure}{r}{0.4\textwidth}
   \centering
   \includegraphics[scale=0.5]{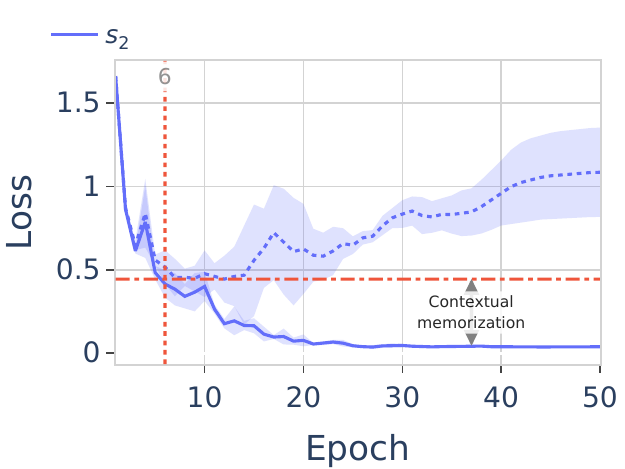}
   \caption{\textbf{Contextual Memorization.} The solid curve is the training loss and the dotted curve is the contextual loss of a string $ s $, when it is inside and outside the training dataset $ D \sim L $, respectively    (details in Section~\ref{sec:operationalization_of_measures}). Contextual memorization begins when training loss drops below the lowest contextual loss, i.e., the LLM generates $ s $ better than it would from learning the underlying language.}
   \label{fig:intro_contextual_memorization}
\end{wrapfigure}

Thus, we propose  \textit{contextual memorization}, a measure that can disentangle the effects of context-based recall from memorization-based recall. 
The key intuition, shown in  Figure~\ref{fig:intro_contextual_memorization}, is the following: for each string $s$ in the training dataset $D\sim L$ from a language $ L $, we first estimate the \textit{optimal} contextual recollection of $ s $, obtained by repeatedly training over a dataset $D'$ that excludes $s$ from $D$ and finding $ s $'s best recollection. This threshold represents the best extent of recollecting $ s $ by learning the language from all other strings except $s$.
We declare $s$ as being contextually memorized, if its recollection when included in training exceeds its optimal contextual recollection.
The motivation for contextual measures is thus rooted in concerns that memorization is an undesirable form of learning, i.e., it represents a type of \emph{local over-fitting} to string $s$ that harms \emph{local generalization}~\citep{van2021memorization}.

Contextual memorization differs from existing notions namely, \textit{recollection-based} and \textit{counterfactual} memorization both in their motivating perspectives and operationalization, as detailed in Section~\ref{sec:quantification_memorization}. They result in conflicts when determining the memorization of training strings under the same training dynamics. In addition, we show that estimates using recollection-based memorization by LLMs today paint a greatly exaggerated picture of training data being memorized. Many claimed memorization instances would be filtered out by contextual memorization (Appendix~\ref{sec:predictable_memorization}).

\paragraph{Learning--Memorization Trade-offs.} Given that memorization is a local phenomenon measured at the level of individual strings $s$ in the training dataset $D$ and learning is a global phenomenon measured over a test dataset also sampled from $L$, a natural question that arises is \emph{can we learn a language $L$ without memorizing any strings $s$ in $L$?} 
Based on extensive analysis using different memorization measures, we conclude in Section~\ref{sec:language_memorization} that \emph{learning without memorization is infeasible in LLMs}, a result in the generative setting, unlike the classification setting in~\citet{feldman2020does}. 
The key underlying intuition is the following: every string $s$ in $D$ has its own training epoch $e_s$, when it begins to be memorized and these vary significantly across different strings. 
The training epoch $e^{*}$ corresponding to globally optimal learning often occurs after some (and often many) strings have been memorized. 

\paragraph{Memorization-aware Training.} Since memorization and learning often occur simultaneously, a natural question is \textit{can we improve their trade-off, i.e., improve learning stability and uniformity while reducing memorization?} We propose memorization-aware training (MAT), where training strings are down-sampled once they are already learned and likely to be memorized. In experiments, MAT indeed reduces memorization, and improves learning. Intuitively, memorization is reduced because already learned training strings are temporarily excluded from training, preventing them from being memorized further. On the other hand, learning is improved because the LLM is trained more on \textit{harder} non-memorized strings than \textit{easier} memorized strings, achieving more stable and uniform learning across test strings, and often better learning on average (Section~\ref{sec:memorization_aware_training}).

\paragraph{Controlled Experiments.} We propose using formal languages as a testbed for studying memorization and learning in LLMs, where an LLM is trained on strings from a formal language and evaluated on test strings from the same language. Formal language~\citep{chomsky1956three} provides a precise specification of a language, where training and test strings are sampled from a known distribution. With access to the underlying grammar, we effectively control the entropy of the language, probing the nuances of different memorization measures. Also, their synthetic nature avoids data contamination, enabling a precise analysis of memorization and learning. For completeness, we validate our results on natural language datasets in Appendix~\ref{sec:app_experiments} and~\ref{sec:mat_appendix}, where contextual memorization is indeed relevant due to unequal contextual learning of different strings, and MAT is effective in improving learning.

\section{Motivation and Related Work}
\label{sec:related_work_main}
\textit{Measures of Memorization.} We classify memorization measures into two categories: those that distinguish context-based from memorization-based recall, and those that do not. Our proposed contextual memorization falls into the first category alongside \textit{counterfactual memorization}~\citep{feldman2020does,Zhang2021CounterfactualMI}, but differs from it in two important ways. (i) Contextual memorization targets learning risks, while counterfactual memorization targets privacy risks specifically for membership inference in the training data. (ii) Contextual memorization applies a stricter threshold than counterfactual memorization, as shown in Lemma~\ref{lm:contextual_vs_counterfactual_memorization} and Figure~\ref{fig:contextual_vs_counterfactual}.

The second category, \textit{recollection-based memorization}, entangles memorization-based recall with contextual learning. The motivation is also related to privacy: LLMs that easily recollect training data pose higher privacy risks, risks that recollection-based measures are designed to detect~\citep{carlini2021extracting,NEURIPS2024_66453d57}. 

\textit{Training Dynamics.} 
We study memorization and learning through the training dynamics of an LLM, identifying when individual strings begin to be memorized and whether learning becomes non-uniform and unstable across strings. In contrast, many prior studies~\citep{carlini2021extracting,carlini2022quantifying,NEURIPS2024_66453d57,Zhang2021CounterfactualMI} ask whether a pre-trained LLM has memorized some strings without access to training traces, overlooking how measures evolve during training and their interplay with learning.

\textit{Formal Language and LLMs.} Prior research considers formal languages to study LLMs' representational capacity~\citep{deletang2022neural,bhattamishra2020ability}, learning difficulty for specific transformer architectures~\citep{borenstein2024languages,hahn2020theoretical,cotterell2018all}, or comparing different learning modes in LLMs~\citep{ghosh2026fine-tuning}. In contrast, we use formal languages as a controlled testbed to study the interplay between memorization and learning, which is novel.

We discuss extended related work in Appendix~\ref{sec:related_work}.

\section{Research Questions \& Experimental Setup}
\label{sec:setup}

Memorization arises when an LLM repeatedly trains on a dataset: there is a training epoch when each string begins to be memorized, a local event, after which its memorization score increases until maximized. Learning, by contrast, is a global phenomenon measured by how well the LLM generates unseen test strings from the same language. We address three key research questions (RQs).

\paragraph{Formal Setup.}
An LLM $M$ is trained on $D_{\text{train}}$ and evaluated on $D_{\text{test}}$ over multiple epochs, where $D_{\text{train}}, D_{\text{test}} \sim L$ are random samples of a language $L$ and may contain duplicate strings.

\begin{rqbox}
    \begin{itemize}[leftmargin=*,noitemsep]
        \item \namedlabel{itm:rq_begin_of_mem}{\textbf{RQ 1}}\textbf{:} When does $ M $ begin to memorize a training string $ s \in D_{\text{train}} $? 
        
        \item \namedlabel{itm:rq_mem_learn}{\textbf{RQ 2}}\textbf{:} Does $ M $ avoid memorizing any string in $ D_{\text{train}} $ before achieving optimal learning on $ D_{\text{test}} $?

        \item \namedlabel{itm:rq_mem_mitigation}{\textbf{RQ 3}}\textbf{:} Does memorization-aware training lead to stable and uniform learning on $ D_{\text{test}} $?
    \end{itemize}

\end{rqbox}

\ref{itm:rq_begin_of_mem} concerns the determination of memorization and can be answered using different measures; \ref{itm:rq_mem_learn} and \ref{itm:rq_mem_mitigation} concern the interplay between memorization and learning.
We answer~\ref{itm:rq_begin_of_mem} in Section~\ref{sec:quantification_memorization}, \ref{itm:rq_mem_learn} in Section~\ref{sec:language_memorization}, and \ref{itm:rq_mem_mitigation} in Section~\ref{sec:memorization_aware_training}.

 \begin{wrapfigure}{r}{0.5\textwidth}
    \centering
    \scalebox{0.75}
    {
    \small
    \begin{minipage}{0.3\textwidth}
        \centering
        \begin{align*}
            & \textcolor{red}{S}\;\textcolor{black}{\rightarrow}\;\textcolor{red}{A19}\;\textcolor{blue}{[1]}\\
            & \textcolor{red}{A19}\;\textcolor{black}{\rightarrow}\;\textcolor{red}{A18}\;\textcolor{red}{A16}\;\textcolor{blue}{[0.95]}\\
            & \textcolor{red}{A19}\;\textcolor{black}{\rightarrow}\;\textcolor{red}{A16}\;\textcolor{red}{A18}\;\textcolor{red}{A17}\;\textcolor{blue}{[0.05]}\\
            & \textcolor{red}{A18}\;\textcolor{black}{\rightarrow}\;\textcolor{red}{A15}\;\textcolor{red}{A14}\;\textcolor{red}{A13}\;\textcolor{blue}{[0.95]}\\
            & \textcolor{red}{A18}\;\textcolor{black}{\rightarrow}\;\textcolor{red}{A14}\;\textcolor{red}{A15}\;\textcolor{red}{A13}\;\textcolor{blue}{[0.05]}\\
            & \textcolor{red}{A17}\;\textcolor{black}{\rightarrow}\;\textcolor{red}{A14}\;\textcolor{red}{A13}\;\textcolor{red}{A15}\;\textcolor{blue}{[0.95]}\\
            & \textcolor{red}{A17}\;\textcolor{black}{\rightarrow}\;\textcolor{red}{A13}\;\textcolor{red}{A14}\;\textcolor{red}{A15}\;\textcolor{blue}{[0.05]}\\
            & \textcolor{red}{A16}\;\textcolor{black}{\rightarrow}\;\textcolor{red}{A14}\;\textcolor{red}{A15}\;\textcolor{blue}{[0.95]}\\
            & \textcolor{red}{A16}\;\textcolor{black}{\rightarrow}\;\textcolor{red}{A15}\;\textcolor{red}{A14}\;\textcolor{blue}{[0.05]}\\
            & \textcolor{red}{A15}\;\textcolor{black}{\rightarrow}\;\textcolor{red}{A11}\;\textcolor{red}{A12}\;\textcolor{red}{A10}\;\textcolor{blue}{[0.95]}\\
            & \textcolor{red}{A15}\;\textcolor{black}{\rightarrow}\;\textcolor{red}{A12}\;\textcolor{red}{A10}\;\textcolor{red}{A11}\;\textcolor{blue}{[0.05]}\\
        \end{align*}
    \end{minipage}%
    \begin{minipage}{0.33\textwidth}
        \centering
        \begin{align*}
            & \textcolor{red}{A14}\;\textcolor{black}{\rightarrow}\;\textcolor{red}{A11}\;\textcolor{red}{A10}\;\textcolor{red}{A12}\;\textcolor{blue}{[0.95]}\\
            & \textcolor{red}{A14}\;\textcolor{black}{\rightarrow}\;\textcolor{red}{A10}\;\textcolor{red}{A11}\;\textcolor{red}{A12}\;\textcolor{blue}{[0.05]}\\
            & \textcolor{red}{A13}\;\textcolor{black}{\rightarrow}\;\textcolor{red}{A10}\;\textcolor{red}{A12}\;\textcolor{red}{A11}\;\textcolor{blue}{[0.95]}\\
            & \textcolor{red}{A13}\;\textcolor{black}{\rightarrow}\;\textcolor{red}{A12}\;\textcolor{red}{A11}\;\textcolor{red}{A10}\;\textcolor{blue}{[0.05]}\\
            & \textcolor{red}{A12}\;\textcolor{black}{\rightarrow}\;\textcolor{teal}{8}\;\textcolor{teal}{7}\;\textcolor{blue}{[0.95]}\\
            & \textcolor{red}{A12}\;\textcolor{black}{\rightarrow}\;\textcolor{teal}{9}\;\textcolor{teal}{8}\;\textcolor{teal}{7}\;\textcolor{blue}{[0.05]}\\
            & \textcolor{red}{A11}\;\textcolor{black}{\rightarrow}\;\textcolor{teal}{6}\;\textcolor{teal}{5}\;\textcolor{blue}{[0.95]}\\
            & \textcolor{red}{A11}\;\textcolor{black}{\rightarrow}\;\textcolor{teal}{6}\;\textcolor{teal}{4}\;\textcolor{teal}{5}\;\textcolor{blue}{[0.05]}\\
            & \textcolor{red}{A10}\;\textcolor{black}{\rightarrow}\;\textcolor{teal}{3}\;\textcolor{teal}{1}\;\textcolor{blue}{[0.95]}\\
            & \textcolor{red}{A10}\;\textcolor{black}{\rightarrow}\;\textcolor{teal}{1}\;\textcolor{teal}{2}\;\textcolor{teal}{3}\;\textcolor{blue}{[0.05]}\\
        \end{align*}
    \end{minipage}%
    }

    \includegraphics[trim={2.5cm 0 2.5cm 2.5cm},clip,scale=0.63]{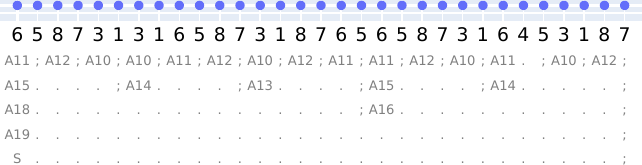}

    \caption{A probabilistic and hierarchical context-free grammar (top), and a sampled string generated by the grammar (bottom), adapted from~\citet{ghosh2026fine-tuning}. The grammar produces a skewed language, where some rules (patterns) are applied more frequently than others.}
    \label{fig:g1}
    
\end{wrapfigure}

\paragraph{Experimental Setup.} Knowing the language and its grammar rules can provide a vehicle for understanding whether
the generation of a string stems from learning the underlying language structure or from memorization.
 We consider \textit{probabilistic and hierarchical context-free languages} (PCFLs), which mimic the recursive structure of natural language~\citep{allen2023physics}. A language $L$ over alphabet $\mathbf{T}$ specifies a distribution $P_L: \mathbf{T}^\ast \rightarrow [0, 1]$ over all strings. A representative grammar and string are shown in Figure~\ref{fig:g1}.

Throughout, we use \textit{language entropy} as a key dimension, since adjusting entropy alters string frequencies, a factor central to studying many memorization measures~\citep{Zhang2021CounterfactualMI}. Formally, $H(L) = - \sum_{s \in \mathbf{T}^\ast} P_L(s)\log P_L(s)$~\citep{cover1991elements,carrasco1997accurate}.

We experiment with $18$ open-source LLMs from $6$ families: Mistral~\citep{jiang2023mistral7b}, Llama~\citep{dubey2024llama}, Qwen~\citep{yang2024qwen2}, Gemma~\citep{gemmateam2024gemmaopenmodelsbased}, Pythia~\citep{biderman2023pythia}, and Opt~\citep{zhang2022opt}, ranging from $0.5$B to $13$B parameters; all results are averaged over three runs. We experiment on $10$ languages $\{L_1, \dots, L_{10}\}$ of varying entropy, alphabet (e.g., numerical and Latin tokens), and size; formal definition of each language and training details are given in Appendix~\ref{sec:app_exp_setup}. Moreover, the main paper shows results on the Mistral-$ 7 $B model on a subset of formal languages; results on all models, remaining formal languages, and various natural languages are in the Appendix.

\section{On Operationalizing Memorization Notions}
\label{sec:quantification_memorization}

We propose operational measures for three notions of memorization, including a new notion of contextual memorization (Section~\ref{sec:definitions}), and show that applying them to the same training dynamics yields contradictory conclusions about when and in what order strings are memorized (Section~\ref{sec:operationalization_of_measures}). We start by dividing the research question~\ref{itm:rq_begin_of_mem} into the following sub-questions.

\begin{rqbox}
    \begin{itemize}[leftmargin=*,noitemsep]
        \item \namedlabel{itm:rq_begin_of_mem_1}{\textbf{RQ 1.1}}\textbf{:} At what epoch $ e_{s} $ does $ M $ begin to memorize $ s \in D_{\text{train}} $?
        \item \namedlabel{itm:rq_begin_of_mem_2}{\textbf{RQ 1.2}}\textbf{:} What is the score of memorization, $\mathtt{mem}(s, e) \in [0,1] $, of a string $ s $ at an epoch $ e \ge e_s $? Trivially, $\mathtt{mem}(s, e) = 0$ for any epoch $e < e_s$.
    \end{itemize}
\end{rqbox}

\subsection{Notions and Their Operational Measures}
\label{sec:definitions}

\paragraph{Contextual Memorization.} We propose contextual memorization to capture memorization as a learning risk. Language learning enables an LLM to generate strings \textit{beyond} its training data by acquiring the underlying language structure; by contrast, memorization is \textit{local over-fitting} to individual training strings~\citep{van2021memorization}. Memorization, in this sense, lets an LLM recollect a specific training string better than language learning alone could do. Contextual memorization therefore captures the gap between training recollection and the optimal recollection due to language learning.

Throughout, we use cross-entropy loss to measure recollection performance~\citep{mao2023cross}. Let $\loss(M_e(D), s)$ denote the loss of model $M$ recollecting string $s \in D$ at epoch $e$, where $D \equiv D_{\text{train}}$.

To compute \textit{optimal contextual loss}, we use counterfactual training on $D' = D \setminus \{s\}$~\citep{feldman2020does}. The optimal contextual loss is the lowest loss, $\loss^*(M(D'), s)$, that $s$ achieves without being in the training set. In a multi-epoch setup, we use the lowest loss across all epochs as a proxy, $\loss^*(M(D'), s) \approx \min_{e^*} \loss(M_{e^*}(D'), s) $. Since $s \notin D'$, its loss first decreases as the model learns the language, then rises as the model begins memorizing other strings in $D'$. Hence, contextual memorization begins at epoch $e = e^{\ct}_s$ when the training loss drops below the optimal contextual loss, $\loss(M_{e}(D), s) < \min_{e^*} \loss(M_{e^*}(D'), s)$. For $e \ge e^{\ct}_s$, the memorization score is

{
\begin{equation}
\mathtt{mem}^{\ct}(s, e, D) \triangleq \frac{\min_{e^*} \loss(M_{e^*}(D'), s) - \loss(M_{e}(D), s)}{\min_{e^*} \loss(M_{e^*}(D'), s)} \in [0,1]
\end{equation}
}

Since $\mathtt{mem}^{\ct}(s, e, D)$ depends on $D$, expected memorization averages over datasets sampled from $L$:
\begin{equation}
\mathtt{mem}^{\ct}(s, e) \triangleq \mathbb{E}_{D \sim L, D \owns s}[\mathtt{mem}^{\ct}(s, e, D)].
\label{eq:expected_mem_ctx}
\end{equation}

\paragraph{Efficient Computation of Contextual Memorization.}
Contextual memorization requires retraining on $D \setminus \{s\}$ per string, which is expensive. Can we avoid retraining and still compute optimal contextual loss of $s$? We empirically observe that same-frequency strings under a formal language have comparable losses on average, implying that they are similarly learnable (see frequency-grouped training loss in Figure~\ref{fig:mat_training_dynamics_base_train_loss} and test loss in Figure~\ref{fig:mat_training_dynamics_base_test_loss}). Under this observation, for a given training string $s \in D_{\text{train}}$, we use a test string $s' \in D_{\text{test}}$ from the same frequency group as a proxy for computing optimal contextual loss: since $s' \notin D_{\text{train}}$, its lowest loss approximates $\min_{e^*} \loss(M_{e^*}(D \setminus \{s\}), s)$. We validate this in Appendix~\ref{sec:validation_ctx}, finding low approximation error, and apply it in remaining sections.

\paragraph{Counterfactual Memorization.} A string $s$ is counterfactually memorized if its training loss deviates from its loss under counterfactual training on $D' = D \setminus \{s\}$. Unlike contextual memorization, the threshold is the epoch-wise $\loss(M_{e}(D'), s)$, not the minimum $\loss^*(M(D'), s)$. Memorization begins at epoch $e = e^{\cf}_s$ when $\loss(M_{e}(D), s) < \loss(M_{e}(D'), s)$, with score

{
\begin{equation*}
\mathtt{mem}^{\cf}(s, e, D) \triangleq \frac{\loss(M_{e}(D'), s) - \loss(M_{e}(D), s)}{\loss(M_{e}(D'), s)} \in [0,1].
\end{equation*}
}

Expected memorization is defined analogously to Eq.~\eqref{eq:expected_mem_ctx}. The following lemma states the precise relationship between contextual and counterfactual memorization.

\begin{lemma}
    \label{lm:contextual_vs_counterfactual_memorization}
    Contextual memorization is stricter than counterfactual memorization. The beginning epoch of contextual memorization never precedes the beginning epoch of counterfactual memorization, and contextual memorization score is a lower bound of counterfactual memorization score.
\end{lemma}

Intuitively, when the training loss of a string $ s $ deviates from the test loss of $ s $ in the counterfactual training, one can perform a membership inference attack, the indication that $ s $ is counterfactually memorized. However, \textit{test loss of $ s $ can still decrease beyond the deviating point due to learning}, as shown in Figure~\ref{fig:contextual_vs_counterfactual}. Put differently, contextual learning does not stop immediately when counterfactual memorization begins, resulting in a stricter threshold for contextual memorization. Contextual memorization therefore starts later, when the training loss of $s$ drops below the optimal contextual loss, the lowest test loss $s$ achieves under counterfactual training.

\paragraph{Recollection-based Memorization.} 
Let $\tau$ be a manually set loss threshold. Recollection memorization of $s$ begins at epoch $e = e^{\recm}_s$ when $\loss(M_e(D), s) < \tau$, with binary score $\mathtt{mem}^{\recm}(s, e, D) \triangleq \mathds{1}(\loss(M_e(D), s) < \tau)$.

Recollection-based memorization differs from both contextual and counterfactual memorization in its threshold: the former uses a fixed manual $\tau$, while the latter use string-learnability-adaptive thresholds derived from counterfactual training. Therefore, implementing recollection-based memorization is easier than the others. We present a detailed comparison in the Appendix~\ref{sec:measures_appendix}.

\subsection{Operationalizations Lead to Conflicting Outcomes}
\label{sec:operationalization_of_measures}

We demonstrate how different memorization measures yield conflicting outcomes under the same training dynamics (Table~\ref{tab:memorization_measure_characteristics}). To mimic the non-uniform distribution of natural languages, we use a low-entropy hierarchical formal language and examine three strings of decreasing frequency $\{s_0, s_1, s_2\}$, where $\mathtt{freq}(s_0) > \mathtt{freq}(s_1) > \mathtt{freq}(s_2)$. For each $s_i$, we train Mistral-$7$B on $D = D' \uplus \{\!\!\{ s_i^{(\mathtt{freq}(s_i))} \}\!\!\}$, where $D' \sim L$ excludes $s_i$; a separate model trained only on $D'$ is used for contextual and counterfactual memorization. Each experiment is repeated three times to assess robustness. We address \ref{itm:rq_begin_of_mem_1} below and defer \ref{itm:rq_begin_of_mem_2} to Appendix~\ref{sec:app_experiments}.

\begin{figure*}[t]
    \centering
    
    \subfloat[Recollection ($ 0.2 $)]{
        \includegraphics[scale=0.4]{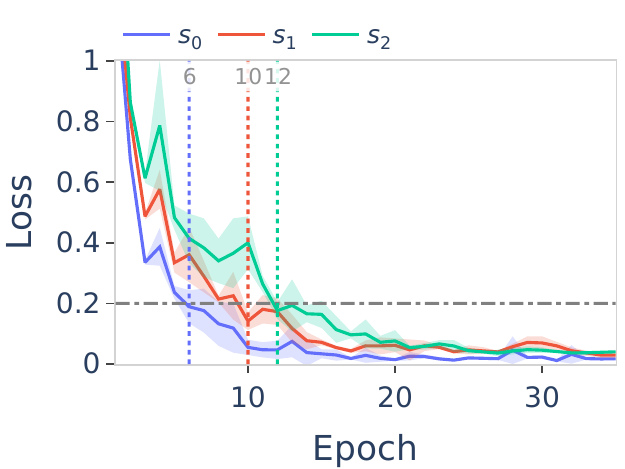}
        \label{fig:recollection_based_memorization}
    }\hfil
    \subfloat[Counterfactual]{
        \includegraphics[scale=0.4]{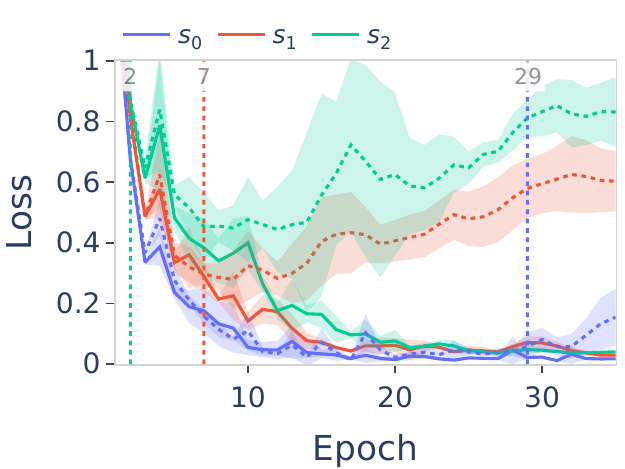}
        \label{fig:counterfactual_memorization}
    }\hfil
    \subfloat[Contextual]{
        \includegraphics[scale=0.4]{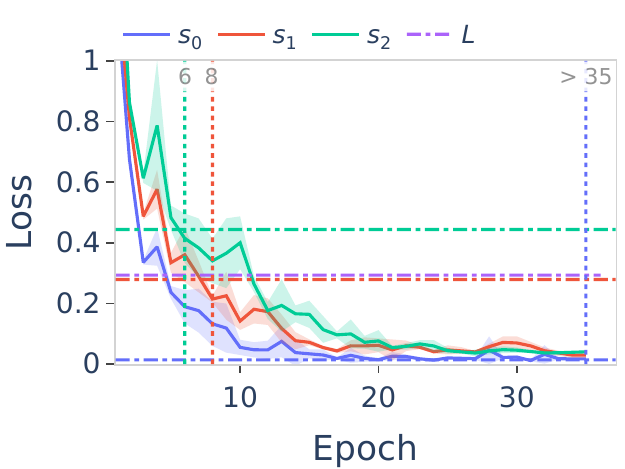}
        \label{fig:contextual_memorization}
    }

    \caption{\textbf{Disagreement Among Memorization Measures.} The beginning of memorization (vertical dotted lines) for strings $s_0, s_1, s_2$ of decreasing frequency. \textit{Recollection-based}: training loss (solid) drops below threshold $\tau = 0.2$ (Figure~\ref{fig:recollection_based_memorization}). \textit{Counterfactual}: training loss deviates from counterfactual test loss (dotted curves) (Figure~\ref{fig:counterfactual_memorization}). \textit{Contextual}: training loss drops below the string-specific optimal contextual loss (Figure~\ref{fig:contextual_memorization}). The optimal contextual loss of the mid-frequent string $s_1$ is usually close to the lowest average test loss over $L$.}
    \label{fig:begin_of_memorization}
\end{figure*}

\paragraph{Recollection-based measures are strongly correlated with the occurrence frequency of strings.}
In Figure~\ref{fig:recollection_based_memorization}, recollection-based memorization follows frequency order: $s_0$ is memorized first ($e^{\recm}_{s_0} = 6$), then $s_1$ ($e^{\recm}_{s_1} = 10$), then $s_2$ ($e^{\recm}_{s_2} = 12$). More frequent strings appear more often in training and thus reach the fixed threshold $\tau$ sooner. \textit{Therefore, the higher the frequency of a string, the earlier it is memorized by recollection,} consistent with prior studies~\citep{tirumala2022memorization,carlini2022quantifying}.

\paragraph{Contextual and counterfactual measures are uncorrelated, and at times, inversely correlated with the occurrence frequency of strings.} In Figures~\ref{fig:counterfactual_memorization} and~\ref{fig:contextual_memorization}, counterfactual and contextual memorization reverse the frequency order ($s_2 > s_1 > s_0$): more frequent strings have a lower optimal contextual loss and thus take more epochs to memorize. This is because a frequently generated string shares more of its production rules with other training strings than an infrequent one does (Figure~\ref{fig:g1}). So even when excluded from training, those shared rules are still learned from the remaining strings, lowering its optimal contextual loss. While the specific ordering of memorization depends on the language (with an exception observed in Figure~\ref{fig:begin_of_memorization_l_4}), the key point is that contextual and counterfactual memorization naturally yield per-string adaptive thresholds for memorization, unlike the fixed $\tau$ of recollection-based memorization. \textit{Therefore, different measures can disagree on when and in what order strings are memorized.}

\paragraph{Contextual memorization is stricter than counterfactual memorization, i.e., the training loss must drop below a lower threshold before contextual memorization begins.} In Figures~\ref{fig:counterfactual_memorization} and~\ref{fig:contextual_memorization}, counterfactual memorization begins no later than contextual memorization, and counterfactual memorization scores are higher than contextual memorization scores (Figure~\ref{fig:degree_of_memorization}). Both findings empirically support Lemma~\ref{lm:contextual_vs_counterfactual_memorization}. \textit{Therefore, counterfactual memorization usually precedes contextual memorization and produces higher memorization scores.}

These disagreements are not specific to formal languages. Across English, German, Chinese, Swahili, and Bengali languages, memorization measures differ on the beginning of memorization. Also different strings showcase different optimal contextual loss values, making the case for contextual memorization (Figures~\ref{fig:memorization_measures_nlp}--\ref{fig:nlp_bengali}).

\paragraph{Takeaway.} Contextual, counterfactual, and recollection-based memorization differ in motivation, data requirements, and outcomes. The advantage of contextual or counterfactual measures is that they use adaptive learnability-based thresholds. Recollection-based measures, on the other hand, use a fixed $\tau$, which is arbitrary, and conflates learning with memorization. Moreover, contextual memorization is preferred when the goal is to strictly separate memorization from contextual learning. Nevertheless, one consistent outcome across measures is that different strings begin to be memorized at different epochs, an observation motivating memorization-aware training (Section~\ref{sec:memorization_aware_training}).

\section{On  Learning and Memorization Trade-offs}
\label{sec:language_memorization}

Today, many perceive memorization as undesirable and antithetical to learning~\cite{hernandez2022scaling,zhang2021understanding,carlini2021extracting,carlini2022quantifying}. We revisit this assumption through~\ref{itm:rq_mem_learn}: \textit{Can an LLM avoid memorization when learning a language optimally?} To this end, we divide~\ref{itm:rq_mem_learn} into three sub-questions, define the memorization score of a dataset, and show the \textit{inherent conflict} among memorization measures in answering these questions.

\begin{rqbox}
    Let $ e^* $ be the epoch of optimal language learning, when the mean test loss on $ D_{\text{test}} $ is the lowest.

    \begin{itemize}[leftmargin=*,noitemsep]
      
    \item \namedlabel{itm:rq_mem_learn_1}{\textbf{RQ 2.1}}\textbf{:}  Can $ M $ avoid memorizing strings from $ D_{\text{train}} $ before reaching epoch $e^*$?

    \item \namedlabel{itm:rq_mem_learn_2}{\textbf{RQ 2.2}}\textbf{:} Are frequently repeated strings in $ D_{\text{train}} $ more likely to be memorized before epoch $e^*$?

    \item \namedlabel{itm:rq_mem_learn_3}{\textbf{RQ 2.3}}\textbf{:} Increasing training dataset size improves optimal learning, but do we risk memorizing more training strings due to more repeated strings? 

  \end{itemize}
\end{rqbox}

\paragraph{Memorization Score of a Dataset.} To answer these questions, we extend memorization from individual strings to a dataset. We define the \textit{fraction of memorized strings} $\mathtt{mem}_{\mathtt{frac}}(D, e) = \mathbb{E}_{s \sim D}[\mathds{1}(\mathtt{mem}(s, e) > 0)]$ and \textit{weighted memorization} $\mathtt{mem}_{\mathtt{weighted}}(D, e) = \mathbb{E}_{s \sim D}[\mathtt{mem}(s, e)]$, both in $[0,1]$. Note that $\mathtt{mem}_{\mathtt{frac}}$ is a special case of $\mathtt{mem}_{\mathtt{weighted}}$ when $\mathtt{mem}(s,e) \in \{0,1\}$.

\begin{wrapfigure}{r}{0.55\textwidth}
    \centering
    \captionsetup[subfigure]{justification=centering}
    \subfloat[$ L_1 $ (high entropy)]{
        \includegraphics[scale=0.35]{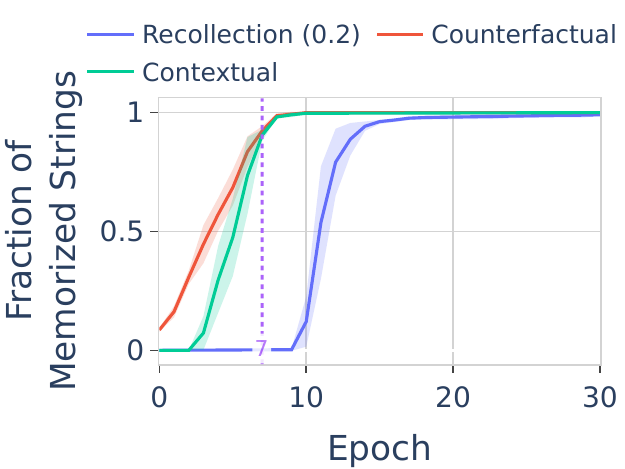}
        \label{fig:memorization_fraction_pcfg_cfg3b_eq_len_uniform_prob_mistral-7b_256_loss_memorization}
    }
    \subfloat[$ L_2 $ (low entropy)]{
        \includegraphics[scale=0.35]{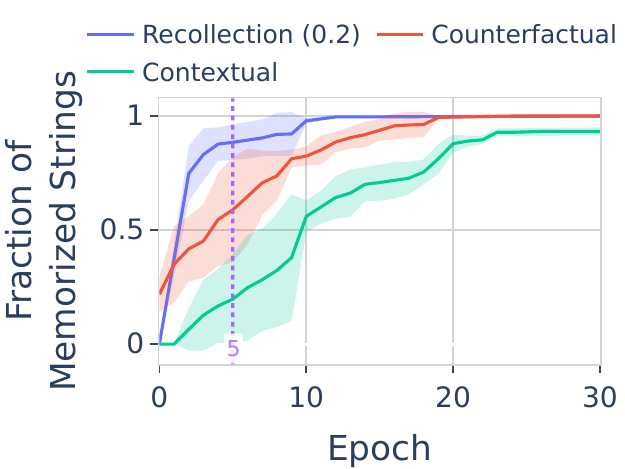}
        \label{fig:memorization_fraction_pcfg_cfg3b_eq_len_skewed_prob_mistral-7b_256_loss_memorization}
    }

    \subfloat[$ L_1 $ (high entropy)]{
        \includegraphics[scale=0.35]{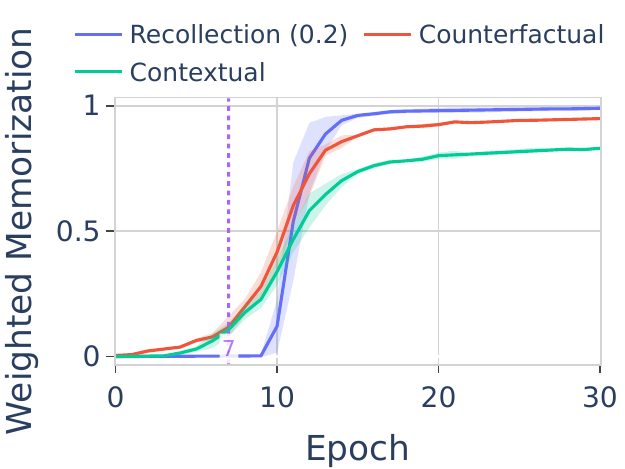}
        \label{fig:memorization_weighted_pcfg_cfg3b_eq_len_uniform_prob_mistral-7b_256_loss_memorization}
    }
    \subfloat[$ L_2 $ (low entropy)]{
        \includegraphics[scale=0.35]{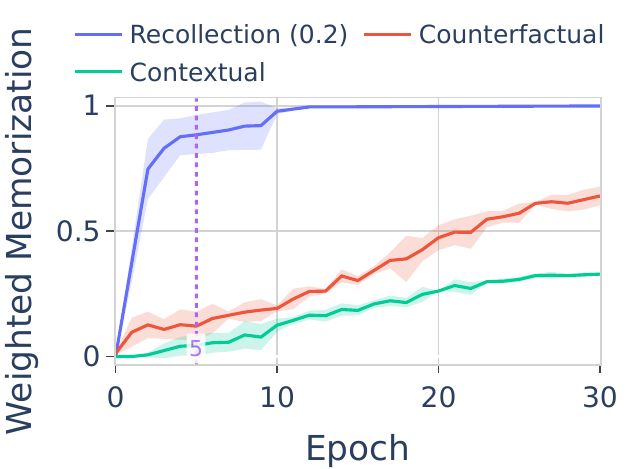}
        \label{fig:memorization_weighted_pcfg_cfg3b_eq_len_skewed_prob_mistral-7b_256_loss_memorization}
    }
    \caption{Memorization of training strings across languages of different entropy. The vertical dotted line marks $e^*$, the epoch of optimal learning (see Figure~\ref{fig:weighted_mem_vs_learning}). At $e^*$, memorization is nonzero.}
    \label{fig:language_memorization}

\end{wrapfigure}

\paragraph{Answer to~\protect\ref{itm:rq_mem_learn_1}: Memorization is unavoidable when optimally learning both high- and low-entropy languages.} 
In Figure~\ref{fig:language_memorization}, we compare memorization across languages of different entropy. The fraction of memorized strings increases monotonically in both languages, and weighted memorization confirms the same trend. At $e^*$, contextual and counterfactual measures show that nearly all strings are memorized in the high-entropy language and a sizable subset in the low-entropy language -- some memorization is unavoidable at optimal learning. Recollection-based measures, however, are inconsistent: $\tau = 0.2$ yields zero memorization in the high-entropy language but near-total memorization in the low-entropy language, as no fixed threshold generalizes across languages.

Across measures, a consistent pattern emerges. In the high-entropy language, all strings begin to be memorized within a narrow epoch band\footnote{Memorization band is a range of epochs $[e_{\text{min}}, e_{\text{max}}]$ containing the beginning of memorization of all training strings, $e_{\text{min}} \le e_s\le e_{\text{max}}, \forall s \sim L$.}. In contrast, in the low-entropy language, where strings occur with widely different frequencies, memorization spreads across a broad band. The optimal learning epoch $e^*$ typically falls within these \textit{memorization bands}, since good generalization requires fitting training strings to some degree, inevitably triggering memorization before $e^*$. \textit{Memorization is therefore unavoidable when optimally learning both high- and low-entropy languages.}

\begin{wrapfigure}{r}{0.55\textwidth}

    \subfloat[Recollection (0.2)]{
        \includegraphics[scale=0.35]{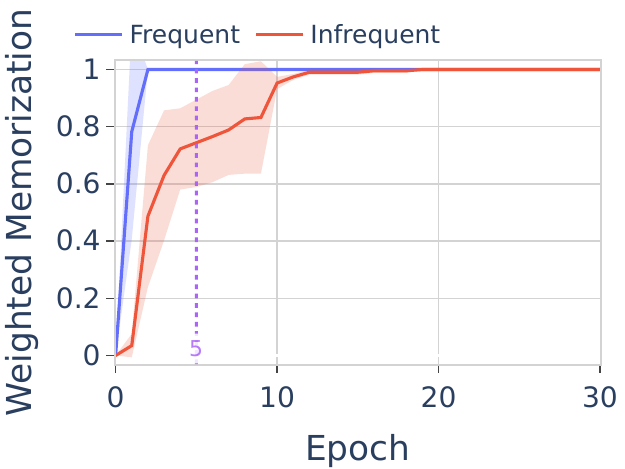}
    }
    \subfloat[Contextual]{
        \includegraphics[scale=0.35]{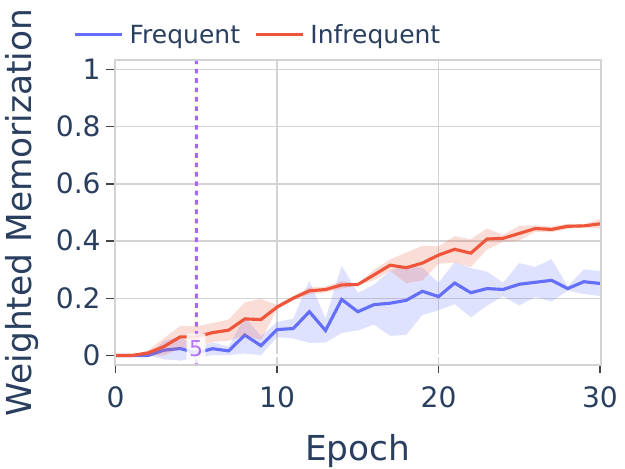}
        \label{fig:contextual_fraction_low_entropy_language}
    }
    \caption{Contradiction among measures on memorization of the top $10\%$ most frequent and bottom $10\%$ least frequent strings in low-entropy language $L_2$ (extended results in Figure~\ref{fig:low_entropy_language_contradiction_detailed_fraction}).}
    \label{fig:low_entropy_language_contradiction}

    \captionsetup[subfigure]{justification=centering}
    \subfloat[Language $ L_2 $]{
        \includegraphics[scale=0.35]{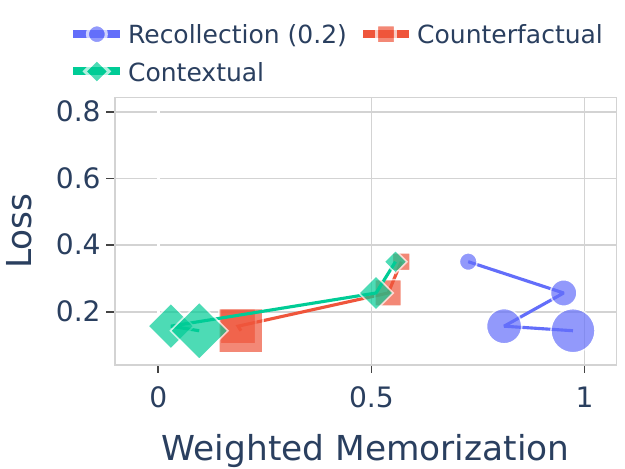}
    }
    \subfloat[Language $ L_4 $]{
        \includegraphics[scale=0.35]{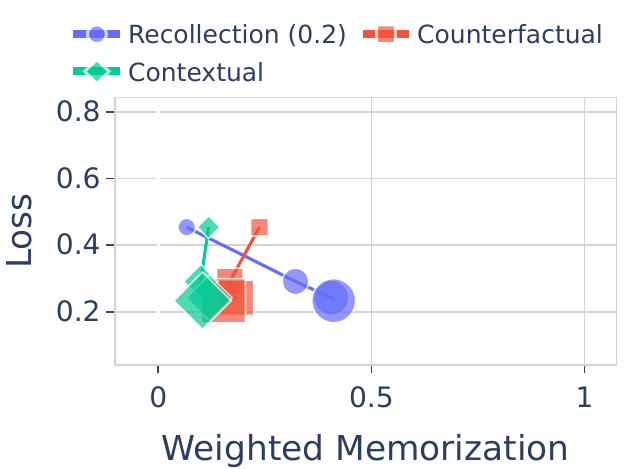}
    }
    \caption{Memorization vs.\ optimal learning (lowest test loss) as training dataset size varies in $ \{16, 64, 256, 1024\} $ (marker size $ \propto $ training size ). Improved learning lowers contextual and counterfactual memorization but raises recollection-based measures.}
    \label{fig:trade_offs_memorization_learning}
\end{wrapfigure}

\paragraph{Answer to~\protect\ref{itm:rq_mem_learn_2}: Frequently repeated strings are likely to be memorized more based on recollection, but both frequent and infrequent strings are almost equally susceptible to contextual and counterfactual memorization.} Since memorization occurs simultaneously with optimal learning, we focus on the low-entropy language and compare memorization of the top $10\%$ most frequent and bottom $10\%$ least frequent strings. Figure~\ref{fig:low_entropy_language_contradiction} reveals a contradiction: recollection-based measure identifies frequent strings as more susceptible to memorization, while contextual and counterfactual measures show nearly equal memorization across both groups. Figure~\ref{fig:contextual_fraction_low_entropy_language} even shows slightly lower contextual memorization for frequent strings, since their lower optimal contextual loss raises the bar for contextual memorization (Section~\ref{sec:quantification_memorization}). Hence, \textit{frequent and infrequent strings are almost equally susceptible to contextual and counterfactual memorization, while frequent strings appear more memorized under recollection-based measures}.

\paragraph{Answer to~\protect\ref{itm:rq_mem_learn_3}: Improved learning due to larger training datasets does not necessarily increase contextual and counterfactual memorization of repeated strings, but increases their recollection-based memorization.}
Larger training datasets improve learning (lower test loss), but does this increase memorization of repeated strings, especially in a low-entropy language where repetition grows with size? In Figure~\ref{fig:trade_offs_memorization_learning}, as training size increases, optimal learning improves while recollection-based memorization also increases -- more strings fall below the fixed threshold $\tau$ due to higher repetition. Contextual and counterfactual memorization, however, decrease: improved learning lowers the optimal contextual loss, raising the bar for contextual memorization.

As an \textit{analogy}, becoming fluent in a language by knowing more sentences increases our ability to produce new ones. Recollection-based measures flag higher fluency as higher memorization, a fallacy. Fluency, however, extends beyond known sentences, improving contextual recollection and thus lowering contextual memorization. In summary, \textit{improved learning can decrease contextual (and counterfactual) memorization, despite increasing recollection-based memorization.}

\paragraph{Takeaway.} The choice of memorization measure significantly affects conclusions about the memorization-learning trade-off. 
Current wisdom relies on recollection-based measures, whose findings depend on a manually chosen threshold $\tau$. 
Using contextual and counterfactual measures instead, we find that (a) some memorization is unavoidable at optimal learning, (b) string frequency has no strong correlation with its susceptibility to memorization, and (c) larger training datasets improve learning and decrease memorization, even when strings are repeated more. The next section dives deeper into the interplay between memorization and learning.

    {
    \color{red}

    }

\section{On Memorization-Aware Training: Towards Stable and Uniform Learning}
\label{sec:memorization_aware_training}

Prior sections show that during repeated training, the LLM memorizes different training strings at different epochs (Figure~\ref{fig:begin_of_memorization}). Consequently, test strings are learned non-uniformly; some are learned better than others at any given epoch. A training string often benefits test strings that are structurally close to it, while structurally distant test strings see little to no advantage (Appendix~\ref{sec:distance_dependent_influence}). Down-sampling such a training string, especially when the model has learned generalizable patterns from it and begins to memorize it, is therefore harmless to the learning of distant test strings. In the following, we discuss memorization-aware training (MAT), which is built on the interplay between memorization and learning.

\begin{figure}[t]
    \centering
    \subfloat[Train Loss]{
        \includegraphics[scale=0.35]{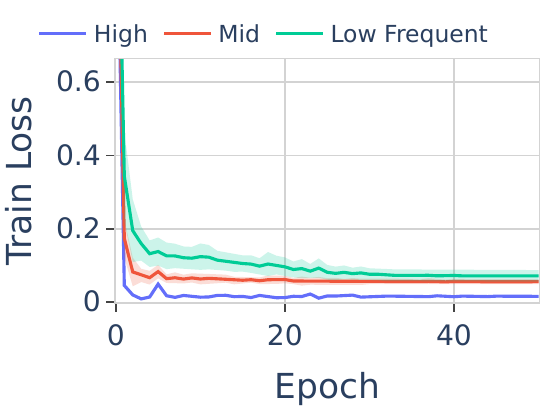}
        \label{fig:mat_training_dynamics_base_train_loss}
    }\hfil
    \subfloat[Test Loss]{
        \includegraphics[scale=0.35]{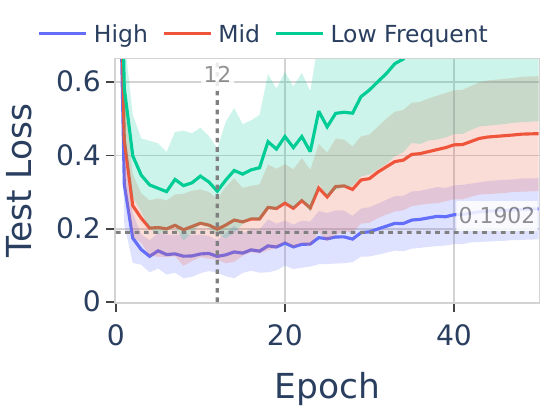}
        \label{fig:mat_training_dynamics_base_test_loss}
    }\hfil
    \subfloat[Contextual]{
        \includegraphics[scale=0.35]{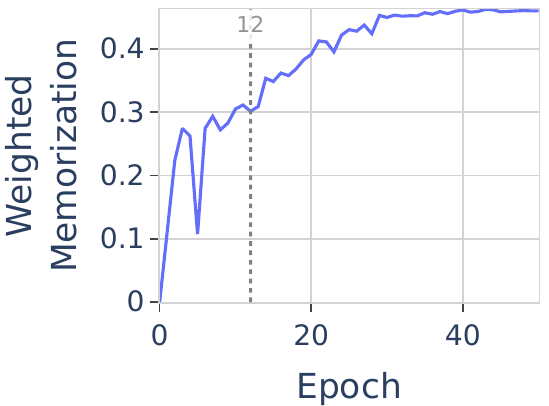}
        \label{fig:mat_training_dynamics_base_contextual_mem}
    }
    \caption{Base training dynamics on language $L_2$ using Mistral-7B, by frequency group. Optimal epoch and mean test loss are marked using vertical and horizontal lines, respectively (Figure~\ref{fig:mat_training_dynamics_base_test_loss}).}
    \label{fig:base_training_dynamics}

\end{figure}

\begin{figure}
    \subfloat[Train Strings Removed]{
        \includegraphics[scale=0.35]{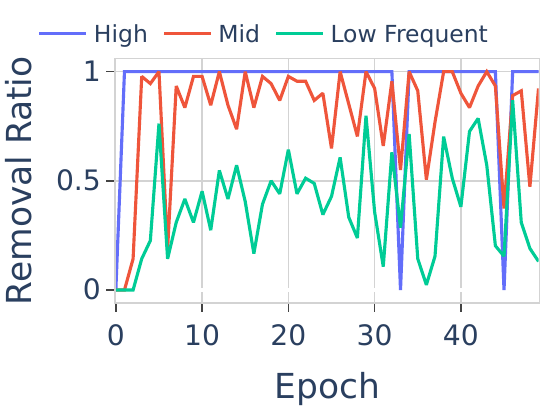}
        \label{fig:mat_training_dynamics_mat_strings_removed}
    }
    \subfloat[Train Loss]{
        \includegraphics[scale=0.35]{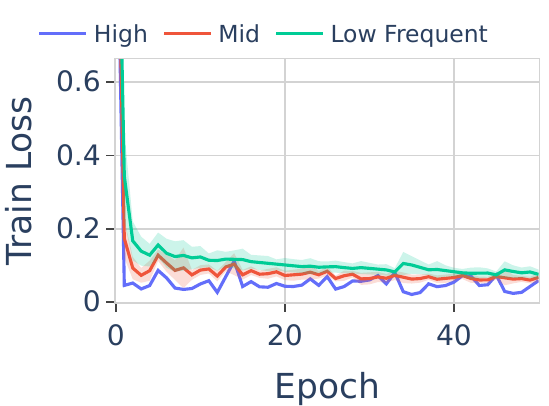}
        \label{fig:mat_training_dynamics_mat_train_loss}
    }
    \subfloat[Test Loss]{
        \includegraphics[scale=0.35]{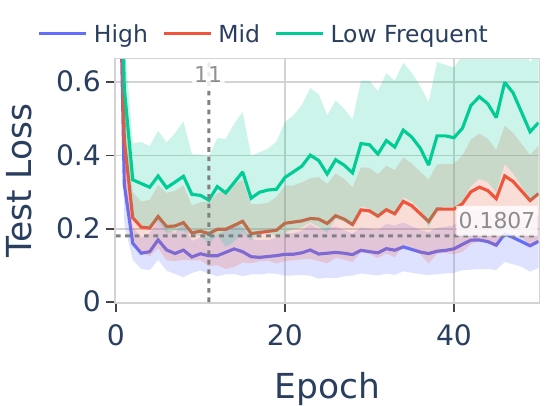}
        \label{fig:mat_training_dynamics_mat_test_loss}
    }
    \subfloat[Contextual]{
        \includegraphics[scale=0.35]{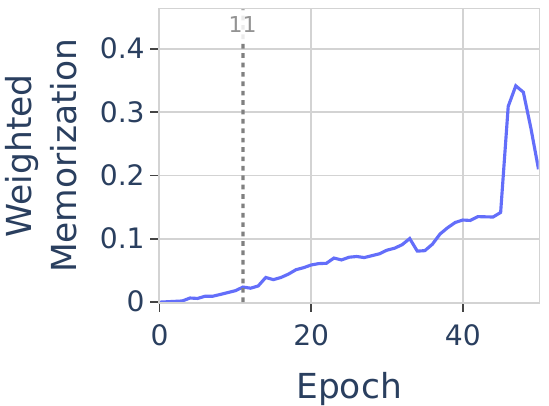}
        \label{fig:mat_training_dynamics_mat_contextual_mem}
    }
    \caption{Training dynamics under MAT in the same setting as in Figure~\ref{fig:base_training_dynamics}. We additionally show the fraction of training strings removed per frequency group in MAT (leftmost).}
    \label{fig:mat_training_dynamics}
\end{figure}

\paragraph{A Closer Look at Memorization-Learning Dynamics.} In the default ``base'' training, training strings remain fixed across epochs, and training loss decreases unevenly (Figure~\ref{fig:mat_training_dynamics_base_train_loss}), reflecting non-uniform memorization. On the learning side, the loss of an individual test string first decreases to a local minimum and increases thereafter, a learning instability, as shown in Figure~\ref{fig:intro_contextual_memorization}. As a result, test strings have varied losses at any given epoch, leading to non-uniform learning (Figure~\ref{fig:mat_training_dynamics_base_test_loss}). To this end, dynamically down-sampling memorized strings leads to more stable and uniform learning.

\paragraph{Down-sampling Heuristic for MAT.} At each epoch, we compute $x\%$, the fraction of test strings whose loss increases relative to the previous epoch. Ideally, we would identify and remove the training strings responsible for this degradation; however, tracing test loss increases to individual training strings is non-trivial. Instead, we down-sample the $x\%$ of training strings with the lowest loss; the LLM has already learned generalizable patterns from them, and they are likely memorized by recollection. If a down-sampled string is subsequently forgotten (its loss rises), it re-enters training at a future epoch. As an \textit{analogy}, memorizing easier chapters of a book does not help in learning harder ones. A better strategy is to skip \textit{already-learned} chapters and focus on harder ones, re-reading if forgotten, yielding more uniform learning, less memorization, and less reading time.

In Figure~\ref{fig:mat_training_dynamics_mat_strings_removed}, more frequent training strings are down-sampled first, followed by less-frequent ones. Consequently, training loss of higher-frequency strings fluctuates more (Figure~\ref{fig:mat_training_dynamics_mat_train_loss}). The result is learning stability: test loss of higher-frequency strings remains low even after the globally optimal learning epoch (Figure~\ref{fig:mat_training_dynamics_mat_test_loss}). The variance of test loss is also reduced, reflecting more uniform learning. As a byproduct, contextual memorization is reduced (Figure~\ref{fig:mat_training_dynamics_mat_contextual_mem}).

The computational overhead of MAT is low, requiring only a forward pass to compute string losses. We also study MAT*, a variant that removes training strings based on contextual loss rather than recollection loss, achieving learning stability across all frequency groups (Appendix~\ref{sec:mat_appendix}); however, the higher cost of computing contextual loss makes MAT* less practical.

\begin{figure}
    \centering
    \subfloat{\includegraphics[scale=0.4]{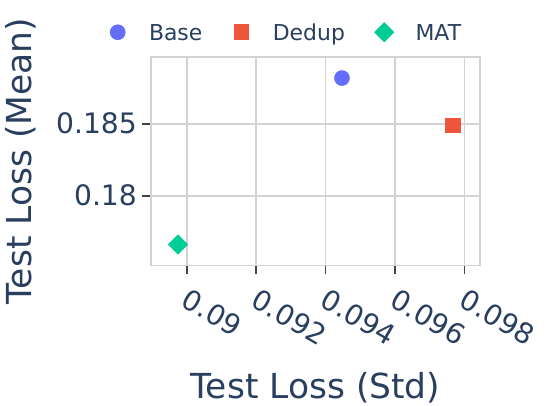}}\hfil
    \subfloat{\includegraphics[scale=0.4]{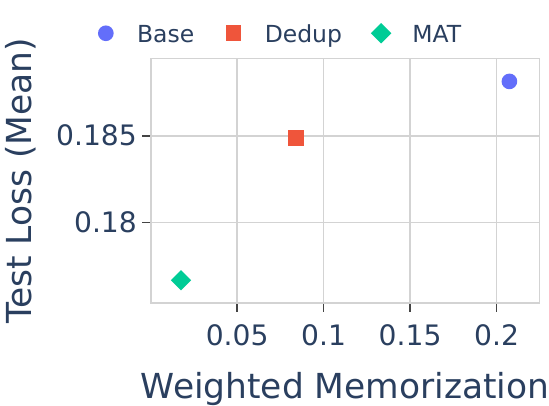}}\hfil
    \subfloat{\includegraphics[scale=0.4]{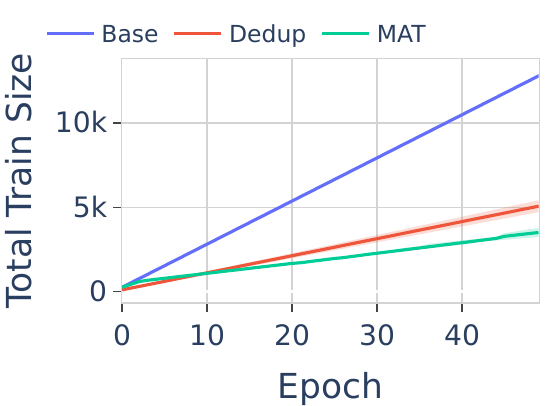}}

    \caption{MAT improves learning by reducing the variance of test loss (often with lower mean loss as well), while also reducing contextual memorization and training cost.}
    \label{fig:memorization_mitigation_fine-tuning}
\end{figure}

\paragraph{Answering~\protect\ref{itm:rq_mem_mitigation} affirmatively: MAT achieves stable and uniform learning.} In Figure~\ref{fig:memorization_mitigation_fine-tuning}, we compare different training schemes at the epoch of optimal learning. Across multiple languages, MAT consistently reduces the standard deviation of test loss, i.e., more uniform learning, and the mean test loss is lower or comparable to base training. Furthermore, at the optimal learning epoch, contextual memorization is lower in MAT compared to base training. Finally, the training cost, measured as the effective number of training strings, is significantly lower in MAT.

In contrast, existing deduplication schemes~\citep{kandpal2022deduplicating,lee2021deduplicating} do not perform as effectively. Deduplication
indeed makes string frequency uniform in the training data. Still, between two strings with the same frequency, one string can be more easily memorized than another, depending on the intrinsic complexity of the string. Deduplication cannot prevent this local over-fitting, causing learning instability and non-uniformity of test strings, as shown in Figure~\ref{fig:mat_training_dynamics_dedup_test_loss_appendix}. MAT, in contrast, down-samples dynamically when a string has already been learned, suggesting that dynamic down-sampling, not frequency balancing alone, drives the improvement. \textit{Thus, MAT improves learning stability and uniformity, reduces memorization, and lowers training cost.}

This finding holds beyond the base fine-tuning setup studied here. It is consistent across multiple formal languages, additional baselines such as prior data pruning~\citep{marion2023less}, a pre-training setup without string repetition, and natural language datasets. The memorization benefit also transfers when a held-out validation set replaces the test set for computing the down-sampling fraction, and using $ x\% $ as the default down-sampling rate is supported by a sensitivity analysis across more conservative and aggressive variants (Appendix~\ref{sec:mat_appendix}).

\section{Conclusions}

We propose contextual memorization in large language models, a learning-focused measure that precisely disentangles memorization from contextual learning, an ability to generate an unseen string by acquiring the underlying language structure. Memorization is a string-specific phenomenon, and different strings are memorized at different epochs in training. We demonstrate that learning a language optimally requires memorizing some training strings. Memorization-aware training, where likely memorized strings are down-sampled dynamically, improves learning-memorization tradeoffs by achieving stable and uniform learning while reducing memorization and training cost. We also demonstrate that formal languages offer a controllable testbed for analyzing the interplay between memorization and learning, with findings that extend to natural language.

\paragraph{Limitations.} Our analysis of the interplay between memorization and learning is organized around string frequency. However, frequency is not the only property that affects this interplay;
entropy, syntactic complexity, etc. may be others. A richer characterization of string difficulty could sharpen our analysis.
One of the drawbacks of contextual memorization is that it is computationally expensive and our approximations rely on the relationship between learnability and string frequency. Alternative approximation schemes would benefit this analysis. In our experiments,
 MAT implements a simple heuristic for down-sampling, which may be affected by the training data. Finally, our results on formal languages must be validated on a wider range of natural language settings. We elaborate on the limitations of our work in Appendix~\ref{sec:limitations}.

\clearpage

\bibliographystyle{abbrvnat}
\bibliography{main}

\newpage
\appendix
\clearpage
{

\section*{Table of Contents (Appendix)}
\label{sec:appendix_toc}

\makeatletter
\newcommand{\@plaintitleref}[5]{#3}
\newcommand{\plainnameref}[1]{%
  \@ifundefined{r@#1}{\textbf{??}}{%
    \expandafter\let\expandafter\plainnameref@tmp\csname r@#1\endcsname
    \expandafter\@plaintitleref\plainnameref@tmp
  }%
}
\makeatother

\begingroup
\renewcommand{\labelitemi}{}
\renewcommand{\labelitemii}{--}
\begin{itemize}[leftmargin=*, itemsep=0.3em, topsep=0.3em]
    \item \textbf{\S\ref{sec:limitations}~\plainnameref{sec:limitations}} (p.~\pageref{sec:limitations})
    \item \textbf{\S\ref{sec:broader_impact}~\plainnameref{sec:broader_impact}} (p.~\pageref{sec:broader_impact})
    \item \textbf{\S\ref{sec:related_work}~\plainnameref{sec:related_work}} (p.~\pageref{sec:related_work})
    \item \textbf{\S\ref{sec:measures_appendix}~\plainnameref{sec:measures_appendix}} (p.~\pageref{sec:measures_appendix})
    \begin{itemize}[leftmargin=1.5em, itemsep=0.1em, topsep=0.1em]
        \item \plainnameref{sec:operational_challenges} (p.~\pageref{sec:operational_challenges})
    \end{itemize}
    \item \textbf{\S\ref{sec:app_theory}~\plainnameref{sec:app_theory}} (p.~\pageref{sec:app_theory})
    \item \textbf{\S\ref{sec:app_exp_setup}~\plainnameref{sec:app_exp_setup}} (p.~\pageref{sec:app_exp_setup})
    \item \textbf{\S\ref{sec:validation_ctx}~\plainnameref{sec:validation_ctx}} (p.~\pageref{sec:validation_ctx})
    \item \textbf{\S\ref{sec:app_experiments}~\plainnameref{sec:app_experiments}} (p.~\pageref{sec:app_experiments})
    \begin{itemize}[leftmargin=1.5em, itemsep=0.1em, topsep=0.1em]
        \item \plainnameref{sec:nlp_results_appendix} (p.~\pageref{sec:nlp_results_appendix})
    \end{itemize}
    \item \textbf{\S\ref{sec:learning_memorization_tradeoffs_appendix}~\plainnameref{sec:learning_memorization_tradeoffs_appendix}} (p.~\pageref{sec:learning_memorization_tradeoffs_appendix})
    \item \textbf{\S\ref{sec:mat_appendix}~\plainnameref{sec:mat_appendix}} (p.~\pageref{sec:mat_appendix})
    \begin{itemize}[leftmargin=1.5em, itemsep=0.1em, topsep=0.1em]
        \item \S\ref{sec:mat_performance_finetuning}~\plainnameref{sec:mat_performance_finetuning} (p.~\pageref{sec:mat_performance_finetuning})
        \item \S\ref{sec:mat_threshold_sensitivity}~\plainnameref{sec:mat_threshold_sensitivity} (p.~\pageref{sec:mat_threshold_sensitivity})
        \item \S\ref{sec:mat_val_ablation}~\plainnameref{sec:mat_val_ablation} (p.~\pageref{sec:mat_val_ablation})
        \item \S\ref{sec:mat_vs_data_pruning}~\plainnameref{sec:mat_vs_data_pruning} (p.~\pageref{sec:mat_vs_data_pruning})
        \item \S\ref{sec:mat_pretraining}~\plainnameref{sec:mat_pretraining} (p.~\pageref{sec:mat_pretraining})
        \item \S\ref{sec:mat_nlp_appendix}~\plainnameref{sec:mat_nlp_appendix} (p.~\pageref{sec:mat_nlp_appendix})
        \item \S\ref{sec:distance_dependent_influence}~\plainnameref{sec:distance_dependent_influence} (p.~\pageref{sec:distance_dependent_influence})
    \end{itemize}
    \item \textbf{\S\ref{sec:predictable_memorization}~\plainnameref{sec:predictable_memorization}} (p.~\pageref{sec:predictable_memorization})
\end{itemize}
\endgroup

\clearpage
}

\section{Limitations}
\label{sec:limitations}

We study memorization in large language models from the learning perspective. The present manuscript has the following limitations that warrant future research.

\begin{itemize}[leftmargin=*]
    \item \textbf{Contextual memorization is computationally expensive.} Computing contextual memorization requires access to the training dataset and additional retraining per string, making it costly to apply at scale. Our heuristic approximation, mapping training strings to test strings of similar frequency, mitigates this cost but introduces an assumption: that string frequency is the primary determinant of learnability. This assumption may not hold when strings of the same frequency differ substantially in intrinsic difficulty, e.g., due to varying entropy or syntactic structure.

    \item \textbf{Memorization and learning analysis is organized around string frequency.} Throughout this work, string frequency serves as the main axis along which we study memorization and learning. However, frequency is not the only relevant property: strings with the same frequency may differ in entropy, syntactic complexity, or the amount of context required to generate them. A richer characterization of string difficulty, beyond frequency, could sharpen the memorization--learning trade-off analysis and lead to more targeted training interventions.

    \item \textbf{MAT is a heuristic that may be improved by training data influence.} The down-sampling criterion in MAT, removing the $x$\% of training strings with the lowest recollection loss, is a proxy for memorization, not a principled optimal strategy. A more principled alternative would use training data influence~\citep{koh2017understanding} to identify which training strings are responsible for test loss degradation, directly targeting the strings that hurt generalization. We leave this as a direction for future work. 

    \item \textbf{Results on formal languages must be validated on broader natural language settings.} Formal languages provide a controllable testbed where entropy, grammar, and string frequency can be precisely manipulated, avoiding data contamination. However, natural language is far more complex and heterogeneous. We include supplementary results on natural language datasets (Appendix~\ref{sec:app_experiments} and~\ref{sec:mat_appendix}), but a systematic study across diverse natural language corpora, model architectures, and training regimes is needed to confirm how well our findings generalize.

    \item \textbf{Multiple measures exist in the context of recollection-based memorization.} Many memorization measures related to recollection exist today, as detailed in Section~\ref{sec:related_work}. Fundamentally, these measures vary on the strictness of the threshold determining whether a string is memorized. In this paper, we operationalize recollection-based memorization using a fixed generative performance threshold via cross-entropy loss, capturing the essence of many such notions. A future study could consider more nuanced operationalizations, although we believe our key results would remain valid.
\end{itemize}

\section{Broader Impact}
\label{sec:broader_impact}
This work focuses on the conceptual foundations of memorization in large language models from the learning perspective, specifically on disentangling memorization from contextual learning. While our contributions are primarily theoretical and empirical on formal languages, the implications extend to societal, privacy, and security dimensions of generative AI. LLMs that memorize training data verbatim risk exposing sensitive or proprietary information present in their training corpora, raising concerns around data privacy and intellectual property. From a security perspective, memorization can be exploited through membership inference and training data extraction attacks. Our proposed measure of contextual memorization offers a more principled lens for identifying when a model has truly memorized a string versus learned the underlying language structure, a distinction that is directly relevant to auditing and mitigating these risks. Memorization-aware training further provides a practical mechanism to reduce unwanted memorization and improve the stability and uniformity of learning, with broader implications for any learning algorithm trained on sensitive data.

\clearpage

\section{Related Work Extended} 
\label{sec:related_work}

\begin{table}[t!]
    \caption{Characteristics of memorization measures.}
    \label{tab:memorization_measure_characteristics}

    \small
    
    \centering
        \begin{tabular}{llccc}
			\toprule
			\shortstack{Memorization\\Measure} & Motivation & \shortstack{Memorization\\Threshold} & \shortstack{Ease of\\Operationalization} & \shortstack{Strictness of \\Measure}\\
			\midrule
			Recollection & Disclosing private information & Manual & \textbf{Easy} & Variable \\
			Counterfactual & Differential privacy & \textbf{Adaptive} & Hard & Medium \\
			Contextual (ours) & Local over-fitting  & \textbf{Adaptive} & Hard & \textbf{High} \\
			\bottomrule
		\end{tabular}
    
\end{table}

\paragraph{Measures of Memorization.} Memorization of training data by LLMs has both desirable and undesirable implications. One example of desired memorization is factual knowledge~\citep{chang2024large,wu2025towards}. The research community however is focused more on the undesirable risks of memorization: privacy risk is a prominent one~\citep{carlini2021extracting,huang2022large,kim2023propile,jagielski2022measuring,bender2021dangers,henderson2023foundation,mueller2024llms,freeman2024exploring}. Most measures of memorization, therefore, associate memorization with recollection, i.e., \textit{is it easier for the LLM to generate sensitive private information encountered in the training data?} We broadly call them recollection-based memorization~\citep{NEURIPS2024_66453d57,biderman2024emergent}. Recollection-based memorization is often captured as perfect memorization~\citep{kandpal2022deduplicating}, verbatim or exact memorization~\citep{carlini2021extracting,carlini2019secret,tirumala2022memorization,mireshghallah2022empirical}, approximate memorization~\citep{ippolito2022preventing,peng2023near,duan2024uncovering}, entity memorization~\citep{zhou2024quantifying}, etc. For an extended taxonomy of memorization measures, we refer to recent papers~\citep{xiong2025landscape,satvaty2024undesirable}.
For example,~\citet{tirumala2022memorization} considered per-token training accuracy as the proxy of memorization: given a training string as a prompt, an LLM memorizes it if it recollects the next token in the string correctly.~\citet{carlini2022quantifying} proposed a relatively stringent measure by imposing an exact recollection of the next $ 50 $ tokens.
The choice of recollection threshold determines which strings are declared memorized, with consequences on the interpretation of memorization, as we demonstrate in Sections~\ref{sec:quantification_memorization} and~\ref{sec:language_memorization}.

In a related direction of capturing privacy risks of memorization,~\citet{Zhang2021CounterfactualMI} defined counterfactual memorization as the change in a model's generative performance when a string is included versus excluded from training~\citep{pappu2024measuring,feldman2020neural}. This approach specifically targets rare strings, which are less recollected, but can cause larger performance shifts in the counterfactual case. While many measures are proposed to capture the privacy risks of memorization, an unexplored direction is to capture memorization as a learning risk, which we address in this study.

From the learning perspective, the critical idea is that any string, including those not in the training set, can be recollected to some extent based on their context~\citep{haviv2022understanding,wang2024generalization,fu2024think,chen2025memorize,speicherrethinking,dong2024generalization,mccoy2023much}. Therefore, the memorization measure should differentiate between how much recollection is due to \textit{rote memorization} versus how much is due to \textit{contextual learning}, which is the ability to generate the same string by understanding the underlying language. Our proposed \textit{contextual memorization} precisely disentangles rote memorization from contextual learning, where the threshold for memorization is the optimal contextual recollection, the best recollection achievable by learning the language without training on that string.

\paragraph{Memorization and Learning.}Several prior works have investigated the interplay between memorization and learning~\citep{van2021memorization}. Notably, the seminal work of~\citet{feldman2020does}, in the classification setting, shows that memorization of classification labels, and in some cases interpolation, are necessary for achieving nearly optimal generalization. Here, we focus on generative models, such as LLMs, and empirically extend their analysis to the generative setting. Particularly, we show that some strings are inevitably memorized before the LLM reaches optimal learning, regardless of the measure of memorization applied (Section~\ref{sec:language_memorization}). Concurrent with our work,~\citet{liu2025language} find that verbatim generation is observed in LLMs even when the string is absent from the training data, and call for a critical rethinking of what defines \textit{membership} in the training data. Contextual memorization explains their finding: a non-member string can be generated via contextual understanding of the language, confirming that recollection is caused by both rote memorization (i.e., membership) and contextual learning (i.e., non-membership), and providing a precise way to separate them (Section~\ref{sec:quantification_memorization}).

\paragraph{Memorization and Learning Dynamics.} We study memorization of individual training strings by analyzing their learning dynamics. Related to ours,~\citet{biderman2024emergent} consider the problem of \textit{forecasting} whether a large model memorizes a specific string based on its memorization by a smaller and \textit{different} model. They apply a variant of recollection-based memorization for detection, but do not find any conclusive signal for forecasting; they even question the appropriateness of recollection-based memorization, similar to us. Our paper however differs with~\citet{biderman2024emergent} in two distinct ways: (i) We focus on a single model and its memorization of training strings, where we disentangle rote memorization-based recollection from contextual learning. (ii) We revisit different memorization measures, and demonstrate the limitations of recollection-based memorization compared to contextual/counterfactual memorization. Recently,~\citet{duanun2025covering} study how memorization changes throughout training and focus on strings which occurred only once throughout training. They find that the model memorizes rarely occurring strings instead of forgetting them, i.e., the model can still recall the string long after seeing it. Their measure of memorization is based on Levenshtein distance, which is also a variant of recollection-based memorization. Our paper thus questions their findings, where a rarely occurring string can still be recalled, not necessarily based on memorization but by contextual learning of the language from strings seen later in the training dynamics. Also related,~\citet{chang2024large} study learning dynamics but define learning as the step-wise acquisition and retention of injected factual knowledge, evaluated via probes, a fact-specific, probe-level notion. Our notion is distributional: the ability to generate unseen strings drawn from the same underlying language, placing learning and memorization on the same footing.

\paragraph{Memorization-aware Training.} Up-sampling or down-sampling training data dynamically is an effective training strategy to achieve specific learning goals~\cite{olmo2025olmo,Chrysakis2020OnlineCLA,Kim2022AnEEA,Imani2023HyperparameterOAA}. Related to our setup is training data deduplication methods proposed to reduce recollection-based memorization~\cite{kandpal2022deduplicating,lee2021deduplicating}. Deduplication down-samples frequent strings in a skewed language to make all strings equally occurring, yet cannot guarantee that all strings are equally memorized. Hence, deduplication cannot change the band of memorization where some strings are still memorized earlier than others, and the beginning of memorization has a spread. Due to non-uniform memorization in deduplication, the variance of test loss across test strings remains high. As a result, deduplication is not effective in improving learning or reducing memorization compared to our approach MAT. Also related is~\citet{marion2023less}, whose best-performing heuristic is perplexity-based down-sampling during training, similar to MAT; however, their approach relies on an auxiliary reference model rather than the training model itself, and their heuristics are not guided by learning dynamics. In contrast, MAT and MAT* down-sample training strings \textit{as needed during training}, informed by whether test string learning degrades, resulting in lower variance and higher stability of test loss (Section~\ref{sec:memorization_aware_training}). Moreover, both MAT and MAT* is related to curriculum and self-paced learning~\citep{bengio2009curriculum,kumar2010self}, which also select training examples based on difficulty; however, those works target faster convergence rather than uniform learning, and do not use memorization as the selection signal.

\section{Memorization Measures Extended}
\label{sec:measures_appendix}
\begin{figure}[h]
    \centering
    \subfloat{\includegraphics[scale=0.4]{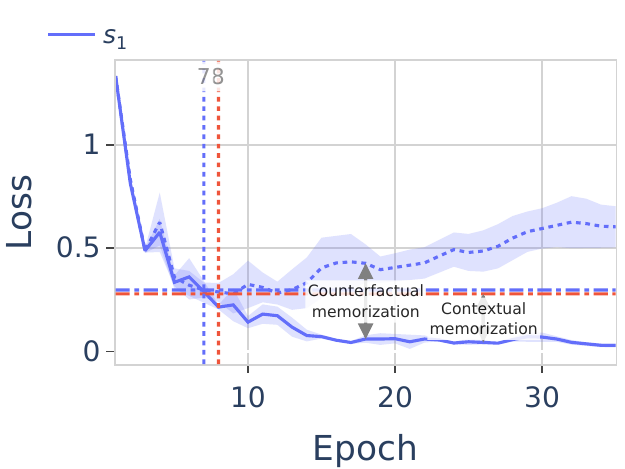}}
    \subfloat{\includegraphics[scale=0.4]{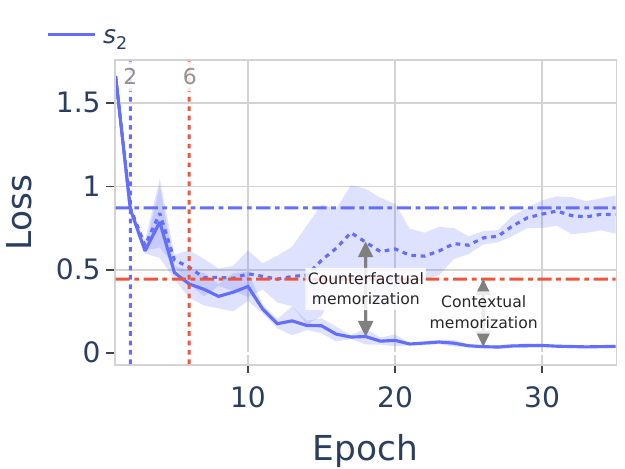}}

    \subfloat{\includegraphics[scale=0.4]{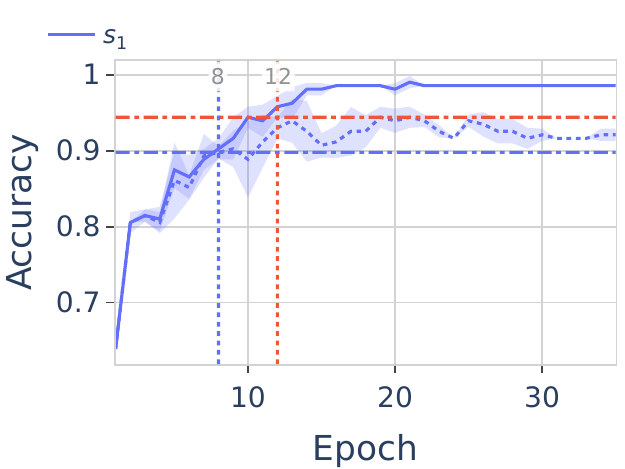}}
    \subfloat{\includegraphics[scale=0.4]{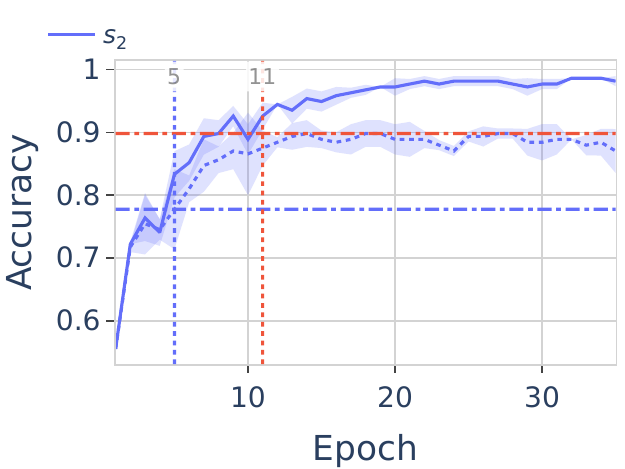}}

    \caption{Contextual memorization is stricter than counterfactual memorization. \textbf{Top row}: the contextual threshold (\textcolor{red}{red} dash-dot line) lies below the counterfactual threshold (\textcolor{blue}{blue} dash-dot line), since contextual learning keeps improving the recollection of $s$ even after the epoch at which counterfactual memorization already permits a membership-inference attack. Contextual memorization of $s$ therefore begins at the same epoch or later, with its score lower-bounded by the counterfactual score (length of the horizontal \textcolor{darkgray}{gray} arrow). \textbf{Bottom row}: the same relationship holds when measured by accuracy instead of loss; since higher accuracy indicates better recollection, the contextual threshold (\textcolor{red}{red}) now lies above the counterfactual threshold (\textcolor{blue}{blue}), and contextual memorization again begins at the same epoch or later.}
    \label{fig:contextual_vs_counterfactual}
\end{figure}

\paragraph{Contextual vs.\ Counterfactual Memorization.} Contextual memorization imposes the strictest threshold $ \min_{e^*} \loss(M_{e^*}(D'), s) $ before classifying a string as memorized, so that a string is only flagged as memorized when its recollection loss strictly falls below the optimal contextual loss. The counterfactual notion, in contrast, seeks the earliest evidence of memorization, using $  \loss(M_{e}(D'), s) $ as the threshold, to flag the earliest potential privacy risks. Considering their different motivations, we refrain from arguing which measure is better. Instead, we formalize their relationship in Lemma~\ref{lm:contextual_vs_counterfactual_memorization}.

\paragraph{Contextual vs. Recollection-based Memorization} Recollection-based memorization requires a fixed memorization threshold $ \tau $ to decide memorization. Contextual memorization, in contrast, derives the threshold adaptively based on how well a string $ s $ would be recollected when learning the underlying language $L$, represented by remaining data $D \setminus \{s\}$, avoiding any over or under estimation of memorization. Common phrases (e.g., `Good morning') are flagged as memorized under recollection-based measures due to high recollection. Yet they are unlikely contextually memorized -- the preceding context (e.g., time of day) enables recollection through language learning alone.

\subsection*{Operational Challenges}
\label{sec:operational_challenges}

\paragraph{Information Requirement Challenges.}  Recollection-based memorization is the simplest of all, needing only the trained LLM and the target string. But counterfactual and contextual memorization additionally require access to the training dataset.

\paragraph{Computational Challenges.} Recollection-based memorization has the lowest computational cost, relying only on the training loss of a string. But, counterfactual and contextual memorization require retraining the LLM separately without the target string, making them computationally expensive and less practical.

\paragraph{Implementation of Counterfactual Memorization.} In our experimental setup, the language $ L $ is a distribution from which $D$ is sampled, resulting in a precise estimation of counterfactual memorization. In contrast, \citet{Zhang2021CounterfactualMI} lacked access to $ L $ and relied on subset sampling, where $D \subset \mathcal{D}$ is drawn from a larger dataset $\mathcal{D}$. Unlike our setup, they did not define memorization per epoch.

Furthermore, counterfactual memorization can be efficiently approximated using the proposed technique for contextual memorization in Section~\ref{sec:definitions}. The only change is that counterfactual memorization considers the test loss of similarly frequent strings as the memorization threshold, while contextual memorization computes the lowest test loss as the memorization threshold.

\section{Theoretical Analysis of Memorization Measures}
\label{sec:app_theory}

\begin{replemma}{lm:contextual_vs_counterfactual_memorization}
    \it
    Contextual memorization is stricter than counterfactual memorization. The beginning epoch of contextual memorization never precedes the beginning epoch of counterfactual memorization, and contextual memorization score is a lower bound of counterfactual memorization score.
\end{replemma}

\begin{proof}
    We prove by considering cross-entropy loss as the proxy of recollection performance. Let $ \loss(M_{e}(D), s) $ be a recollection loss of the string $s$ at epoch $e$ as the LLM $M$ is repeatedly trained on the dataset $D$. In the counterfactual training when the string $ s $ is excluded from the training dataset $D'=D \setminus \{s\}$, let $ \loss(M_{e}(D'), s) $ be the counterfactual (or contextual) loss of the string $s$ at epoch $e$, and let $ \min_{e^*} \loss(M_{e^*}(D'), s) $ be the lowest contextual loss of the string $s$ in all epochs. Here, $ \loss(M_{e}(D'), s) $ and $ \min_{e^*} \loss(M_{e^*}(D'), s) $ are the threshold of counterfactual memorization and contextual memorization, respectively, and 
    \[
        \min_{e^*} \loss(M_{e^*}(D'), s) \le \loss(M_{e}(D'), s).
    \]

    We prove by contradiction. Let the beginning epoch of counterfactual and contextual memorization be $ e^{\cf}_s $ and $ e^{\ct}_s$, respectively. Suppose contextual memorization begins earlier than counterfactual memorization, i.e., $ e^{\ct}_s < e^{\cf}_s $. Then at epoch $ e^{\ct}_s $:

    \begin{align*}
        & \loss(M_{e^{\ct}_s}(D), s) < \min_{e^*} \loss(M_{e^*}(D'), s) \le \loss(M_{e^{\ct}_s}(D'), s) \\
        \Rightarrow\; & \loss(M_{e^{\ct}_s}(D), s) <  \loss(M_{e^{\ct}_s}(D'), s)
    \end{align*}

    But $ \loss(M_{e^{\ct}_s}(D), s) <  \loss(M_{e^{\ct}_s}(D'), s) $ implies that counterfactual memorization has begun at epoch $ e^{\ct}_s < e^{\cf}_s $, which is a contradiction. Therefore, contextual memorization cannot begin earlier than counterfactual memorization.

    On the other hand, contextual memorization can begin at the same epoch or later than counterfactual memorization, since for an
    epoch $e \ge e^{\cf}_s$,
    \begin{align*}
        \underbrace{\loss(M_{e}(D), s) \ge \min_{e^*} \loss(M_{e^*}(D'), s)}_{\text{contextual memorization not yet begun}} \text{ and }\underbrace{\loss(M_{e}(D), s) <
        \loss(M_{e}(D'), s)}_{\text{counterfactual memorization is active}}
    \end{align*}

    i.e., when the training loss diverges from the epoch-specific counterfactual loss, but remains above the lowest contextual loss.

    Furthermore, the counterfactual memorization score is no less than the contextual memorization score, since at any epoch $e \ge \max(e^{\cf}_s,e^{\ct}_s)$, i.e., after memorization begins for both, $\min_{e^*} \loss(M_{e^*}(D'), s) \le \loss(M_e(D'), s) $.
    \begin{align*}
        \underbrace{\frac{\min_{e^*} \loss(M_{e^*}(D'), s) - \loss(M_e(D), s)}{\min_{e^*} \loss(M_{e^*}(D'), s)}}_{\text{contextual memorization score}} \le \underbrace{\frac{\loss(M_e(D'), s) - \loss(M_e(D), s)}{\loss(M_e(D'), s)}}_{\text{counterfactual memorization score}}
    \end{align*}

    Therefore, counterfactual memorization overestimates the memorization score relative to contextual memorization, while reporting memorization at an earlier epoch.
    
\end{proof}

\section{Experimental Setup}
\label{sec:app_exp_setup}

Each training (specifically, fine-tuning) is performed for $ 50 $ epochs with a batch size of $8$ and a linear learning rate scheduler with a warm-up ratio of $0.05$. We fix the learning rate for Qwen, Gemma, and Llama-$ 3 $ families as $5\times 10^{-5}$, Mistral, Opt, and Llama-$ 2 $ families as $5\times 10^{-6}$, and Pythia family as $10^{-5}$. We consider training dataset sizes $ \{16, 64, 256, 1024\} $ and evaluate on $ 1024 $ test strings. In each training, we find the epoch of best learning according to the lowest cross-entropy loss on the test strings and report respective weighted memorization by different measures. All experiments are conducted in compute clusters with Python as the programming language (version $ 3.10 $), where we use $ 8 $x Nvidia H$ 100 $ $ 94 $GB NVL GPUs  and $ 2 $x AMD EPYC $ 9554 $ CPU @ $ 3.1 $ GHz, $ 2 $x$ 64 $ cores, and $ 24 $x $ 96 $GB RAM.

Below, we provide details of the formal languages used in our experiments, along with their formal definitions. Intuitively, we carefully design languages to show the robustness of our results across changing the entropy and token types of the language.

\paragraph{Formal Languages and Grammars.}

\begin{figure*}
\begin{minipage}{0.5\textwidth}
\centering

\begin{align*}
	& \textcolor{red}{S}\;\textcolor{black}{\rightarrow}\;\textcolor{red}{A16}\;\textcolor{blue}{[1]}\\
	& \textcolor{red}{A16}\;\textcolor{black}{\rightarrow}\;\textcolor{red}{A15}\;\textcolor{red}{A14}\;\textcolor{red}{A13}\;\textcolor{blue}{[0.50]}\\
	& \textcolor{red}{A16}\;\textcolor{black}{\rightarrow}\;\textcolor{red}{A13}\;\textcolor{red}{A15}\;\textcolor{red}{A14}\;\textcolor{blue}{[0.50]}\\
	& \textcolor{red}{A13}\;\textcolor{black}{\rightarrow}\;\textcolor{red}{A11}\;\textcolor{red}{A12}\;\textcolor{blue}{[0.50]}\\
	& \textcolor{red}{A13}\;\textcolor{black}{\rightarrow}\;\textcolor{red}{A12}\;\textcolor{red}{A11}\;\textcolor{blue}{[0.50]}\\
	& \textcolor{red}{A14}\;\textcolor{black}{\rightarrow}\;\textcolor{red}{A11}\;\textcolor{red}{A10}\;\textcolor{red}{A12}\;\textcolor{blue}{[0.50]}\\
	& \textcolor{red}{A14}\;\textcolor{black}{\rightarrow}\;\textcolor{red}{A10}\;\textcolor{red}{A11}\;\textcolor{red}{A12}\;\textcolor{blue}{[0.50]}\\
	& \textcolor{red}{A15}\;\textcolor{black}{\rightarrow}\;\textcolor{red}{A12}\;\textcolor{red}{A11}\;\textcolor{red}{A10}\;\textcolor{blue}{[0.50]}\\
	& \textcolor{red}{A15}\;\textcolor{black}{\rightarrow}\;\textcolor{red}{A11}\;\textcolor{red}{A12}\;\textcolor{red}{A10}\;\textcolor{blue}{[0.50]}\\
	& \textcolor{red}{A10}\;\textcolor{black}{\rightarrow}\;\textcolor{red}{A7}\;\textcolor{red}{A9}\;\textcolor{red}{A8}\;\textcolor{blue}{[0.50]}\\
	& \textcolor{red}{A10}\;\textcolor{black}{\rightarrow}\;\textcolor{red}{A9}\;\textcolor{red}{A8}\;\textcolor{red}{A7}\;\textcolor{blue}{[0.50]}\\
	& \textcolor{red}{A11}\;\textcolor{black}{\rightarrow}\;\textcolor{red}{A8}\;\textcolor{red}{A7}\;\textcolor{red}{A9}\;\textcolor{blue}{[0.50]}\\
	& \textcolor{red}{A11}\;\textcolor{black}{\rightarrow}\;\textcolor{red}{A7}\;\textcolor{red}{A8}\;\textcolor{red}{A9}\;\textcolor{blue}{[0.50]}\\
	& \textcolor{red}{A12}\;\textcolor{black}{\rightarrow}\;\textcolor{red}{A8}\;\textcolor{red}{A9}\;\textcolor{red}{A7}\;\textcolor{blue}{[0.50]}\\
	& \textcolor{red}{A12}\;\textcolor{black}{\rightarrow}\;\textcolor{red}{A9}\;\textcolor{red}{A7}\;\textcolor{red}{A8}\;\textcolor{blue}{[0.50]}\\
	& \textcolor{red}{A7}\;\textcolor{black}{\rightarrow}\;\textcolor{teal}{3}\;\textcolor{teal}{1}\;\textcolor{teal}{2}\;\textcolor{blue}{[0.50]}\\
	& \textcolor{red}{A7}\;\textcolor{black}{\rightarrow}\;\textcolor{teal}{1}\;\textcolor{teal}{2}\;\textcolor{teal}{3}\;\textcolor{blue}{[0.50]}\\
	& \textcolor{red}{A8}\;\textcolor{black}{\rightarrow}\;\textcolor{teal}{6}\;\textcolor{teal}{5}\;\textcolor{teal}{4}\;\textcolor{blue}{[0.50]}\\
	& \textcolor{red}{A8}\;\textcolor{black}{\rightarrow}\;\textcolor{teal}{6}\;\textcolor{teal}{4}\;\textcolor{teal}{5}\;\textcolor{blue}{[0.50]}\\
	& \textcolor{red}{A9}\;\textcolor{black}{\rightarrow}\;\textcolor{teal}{9}\;\textcolor{teal}{8}\;\textcolor{teal}{7}\;\textcolor{blue}{[0.50]}\\
	& \textcolor{red}{A9}\;\textcolor{black}{\rightarrow}\;\textcolor{teal}{8}\;\textcolor{teal}{7}\;\textcolor{teal}{9}\;\textcolor{blue}{[0.50]}\\
\end{align*}

\end{minipage}%
\begin{minipage}{0.5\textwidth}
\centering

\begin{align*}
	& \textcolor{red}{S}\;\textcolor{black}{\rightarrow}\;\textcolor{red}{A16}\;\textcolor{blue}{[1]}\\
	& \textcolor{red}{A16}\;\textcolor{black}{\rightarrow}\;\textcolor{red}{A15}\;\textcolor{red}{A14}\;\textcolor{red}{A13}\;\textcolor{blue}{[0.95]}\\
	& \textcolor{red}{A16}\;\textcolor{black}{\rightarrow}\;\textcolor{red}{A13}\;\textcolor{red}{A15}\;\textcolor{red}{A14}\;\textcolor{blue}{[0.05]}\\
	& \textcolor{red}{A13}\;\textcolor{black}{\rightarrow}\;\textcolor{red}{A11}\;\textcolor{red}{A12}\;\textcolor{blue}{[0.95]}\\
	& \textcolor{red}{A13}\;\textcolor{black}{\rightarrow}\;\textcolor{red}{A12}\;\textcolor{red}{A11}\;\textcolor{blue}{[0.05]}\\
	& \textcolor{red}{A14}\;\textcolor{black}{\rightarrow}\;\textcolor{red}{A11}\;\textcolor{red}{A10}\;\textcolor{red}{A12}\;\textcolor{blue}{[0.95]}\\
	& \textcolor{red}{A14}\;\textcolor{black}{\rightarrow}\;\textcolor{red}{A10}\;\textcolor{red}{A11}\;\textcolor{red}{A12}\;\textcolor{blue}{[0.05]}\\
	& \textcolor{red}{A15}\;\textcolor{black}{\rightarrow}\;\textcolor{red}{A12}\;\textcolor{red}{A11}\;\textcolor{red}{A10}\;\textcolor{blue}{[0.95]}\\
	& \textcolor{red}{A15}\;\textcolor{black}{\rightarrow}\;\textcolor{red}{A11}\;\textcolor{red}{A12}\;\textcolor{red}{A10}\;\textcolor{blue}{[0.05]}\\
	& \textcolor{red}{A10}\;\textcolor{black}{\rightarrow}\;\textcolor{red}{A7}\;\textcolor{red}{A9}\;\textcolor{red}{A8}\;\textcolor{blue}{[0.95]}\\
	& \textcolor{red}{A10}\;\textcolor{black}{\rightarrow}\;\textcolor{red}{A9}\;\textcolor{red}{A8}\;\textcolor{red}{A7}\;\textcolor{blue}{[0.05]}\\
	& \textcolor{red}{A11}\;\textcolor{black}{\rightarrow}\;\textcolor{red}{A8}\;\textcolor{red}{A7}\;\textcolor{red}{A9}\;\textcolor{blue}{[0.95]}\\
	& \textcolor{red}{A11}\;\textcolor{black}{\rightarrow}\;\textcolor{red}{A7}\;\textcolor{red}{A8}\;\textcolor{red}{A9}\;\textcolor{blue}{[0.05]}\\
	& \textcolor{red}{A12}\;\textcolor{black}{\rightarrow}\;\textcolor{red}{A8}\;\textcolor{red}{A9}\;\textcolor{red}{A7}\;\textcolor{blue}{[0.95]}\\
	& \textcolor{red}{A12}\;\textcolor{black}{\rightarrow}\;\textcolor{red}{A9}\;\textcolor{red}{A7}\;\textcolor{red}{A8}\;\textcolor{blue}{[0.05]}\\
	& \textcolor{red}{A7}\;\textcolor{black}{\rightarrow}\;\textcolor{teal}{3}\;\textcolor{teal}{1}\;\textcolor{teal}{2}\;\textcolor{blue}{[0.95]}\\
	& \textcolor{red}{A7}\;\textcolor{black}{\rightarrow}\;\textcolor{teal}{1}\;\textcolor{teal}{2}\;\textcolor{teal}{3}\;\textcolor{blue}{[0.05]}\\
	& \textcolor{red}{A8}\;\textcolor{black}{\rightarrow}\;\textcolor{teal}{6}\;\textcolor{teal}{5}\;\textcolor{teal}{4}\;\textcolor{blue}{[0.95]}\\
	& \textcolor{red}{A8}\;\textcolor{black}{\rightarrow}\;\textcolor{teal}{6}\;\textcolor{teal}{4}\;\textcolor{teal}{5}\;\textcolor{blue}{[0.05]}\\
	& \textcolor{red}{A9}\;\textcolor{black}{\rightarrow}\;\textcolor{teal}{9}\;\textcolor{teal}{8}\;\textcolor{teal}{7}\;\textcolor{blue}{[0.95]}\\
	& \textcolor{red}{A9}\;\textcolor{black}{\rightarrow}\;\textcolor{teal}{8}\;\textcolor{teal}{7}\;\textcolor{teal}{9}\;\textcolor{blue}{[0.05]}\\
\end{align*}

\end{minipage}%

\caption{Production rules of ${G}_1$ (left) and ${G}_2$ (right). Compared to $ G_1 $, the grammar $ G_2 $ generates more skewed distribution (or lower entropy) strings, since one out of two production rules for each non-terminal is selected with higher probability.}

\label{fig:grammar_g1_g2}
\end{figure*}

\begin{figure*}
\begin{minipage}{0.5\textwidth}
\centering

\begin{align*}
	& \textcolor{red}{S}\;\textcolor{black}{\rightarrow}\;\textcolor{red}{S5}\;\textcolor{blue}{[1]}\\
	& \textcolor{red}{S5}\;\textcolor{black}{\rightarrow}\;\textcolor{red}{B4}\;\textcolor{red}{C1_1}\;\textcolor{red}{E4}\;\textcolor{red}{T1_1}\;\textcolor{blue}{[0.25]}\\
	& \textcolor{red}{S5}\;\textcolor{black}{\rightarrow}\;\textcolor{red}{B4}\;\textcolor{red}{C1_2}\;\textcolor{red}{E4}\;\textcolor{red}{T1_2}\;\textcolor{blue}{[0.25]}\\
	& \textcolor{red}{S5}\;\textcolor{black}{\rightarrow}\;\textcolor{red}{B4}\;\textcolor{red}{C1_3}\;\textcolor{red}{E4}\;\textcolor{red}{T1_3}\;\textcolor{blue}{[0.25]}\\
	& \textcolor{red}{S5}\;\textcolor{black}{\rightarrow}\;\textcolor{red}{B4}\;\textcolor{red}{C1_4}\;\textcolor{red}{E4}\;\textcolor{red}{T1_4}\;\textcolor{blue}{[0.25]}\\
	& \textcolor{red}{B4}\;\textcolor{black}{\rightarrow}\;\textcolor{red}{B3}\;\textcolor{blue}{[0.3333]}\\
	& \textcolor{red}{B4}\;\textcolor{black}{\rightarrow}\;\textcolor{red}{B3}\;\textcolor{red}{B3}\;\textcolor{red}{B3}\;\textcolor{blue}{[0.3333]}\\
	& \textcolor{red}{B4}\;\textcolor{black}{\rightarrow}\;\textcolor{red}{B3}\;\textcolor{red}{B3}\;\textcolor{blue}{[0.3333]}\\
	& \textcolor{red}{B3}\;\textcolor{black}{\rightarrow}\;\textcolor{red}{B2}\;\textcolor{blue}{[0.3333]}\\
	& \textcolor{red}{B3}\;\textcolor{black}{\rightarrow}\;\textcolor{red}{B2}\;\textcolor{blue}{[0.3333]}\\
	& \textcolor{red}{B3}\;\textcolor{black}{\rightarrow}\;\textcolor{red}{B2}\;\textcolor{red}{B2}\;\textcolor{blue}{[0.3333]}\\
	& \textcolor{red}{B2}\;\textcolor{black}{\rightarrow}\;\textcolor{red}{B1}\;\textcolor{blue}{[0.3333]}\\
	& \textcolor{red}{B2}\;\textcolor{black}{\rightarrow}\;\textcolor{red}{B1}\;\textcolor{blue}{[0.3333]}\\
	& \textcolor{red}{B2}\;\textcolor{black}{\rightarrow}\;\textcolor{red}{B1}\;\textcolor{red}{B1}\;\textcolor{red}{B1}\;\textcolor{blue}{[0.3333]}\\
	& \textcolor{red}{B1}\;\textcolor{black}{\rightarrow}\;\textcolor{teal}{2}\;\textcolor{teal}{9}\;\textcolor{teal}{3}\;\textcolor{blue}{[0.3333]}\\
	& \textcolor{red}{B1}\;\textcolor{black}{\rightarrow}\;\textcolor{teal}{9}\;\textcolor{teal}{6}\;\textcolor{teal}{1}\;\textcolor{blue}{[0.3333]}\\
	& \textcolor{red}{B1}\;\textcolor{black}{\rightarrow}\;\textcolor{teal}{1}\;\textcolor{teal}{8}\;\textcolor{teal}{6}\;\textcolor{blue}{[0.3333]}\\
	& \textcolor{red}{E4}\;\textcolor{black}{\rightarrow}\;\textcolor{red}{E3}\;\textcolor{blue}{[0.3333]}\\
	& \textcolor{red}{E4}\;\textcolor{black}{\rightarrow}\;\textcolor{red}{E3}\;\textcolor{red}{E3}\;\textcolor{blue}{[0.3333]}\\
	& \textcolor{red}{E4}\;\textcolor{black}{\rightarrow}\;\textcolor{red}{E3}\;\textcolor{red}{E3}\;\textcolor{red}{E3}\;\textcolor{blue}{[0.3333]}\\
	& \textcolor{red}{E3}\;\textcolor{black}{\rightarrow}\;\textcolor{red}{E2}\;\textcolor{blue}{[0.3333]}\\
	& \textcolor{red}{E3}\;\textcolor{black}{\rightarrow}\;\textcolor{red}{E2}\;\textcolor{red}{E2}\;\textcolor{blue}{[0.3333]}\\
	& \textcolor{red}{E3}\;\textcolor{black}{\rightarrow}\;\textcolor{red}{E2}\;\textcolor{blue}{[0.3333]}\\
	& \textcolor{red}{E2}\;\textcolor{black}{\rightarrow}\;\textcolor{red}{E1}\;\textcolor{red}{E1}\;\textcolor{blue}{[0.3333]}\\
	& \textcolor{red}{E2}\;\textcolor{black}{\rightarrow}\;\textcolor{red}{E1}\;\textcolor{blue}{[0.3333]}\\
	& \textcolor{red}{E2}\;\textcolor{black}{\rightarrow}\;\textcolor{red}{E1}\;\textcolor{red}{E1}\;\textcolor{red}{E1}\;\textcolor{blue}{[0.3333]}\\
	& \textcolor{red}{E1}\;\textcolor{black}{\rightarrow}\;\textcolor{teal}{5}\;\textcolor{teal}{6}\;\textcolor{teal}{5}\;\textcolor{teal}{9}\;\textcolor{blue}{[0.3333]}\\
	& \textcolor{red}{E1}\;\textcolor{black}{\rightarrow}\;\textcolor{teal}{1}\;\textcolor{teal}{8}\;\textcolor{teal}{6}\;\textcolor{teal}{6}\;\textcolor{blue}{[0.3333]}\\
	& \textcolor{red}{E1}\;\textcolor{black}{\rightarrow}\;\textcolor{teal}{1}\;\textcolor{teal}{5}\;\textcolor{teal}{1}\;\textcolor{teal}{5}\;\textcolor{blue}{[0.3333]}\\
	& \textcolor{red}{T1_1}\;\textcolor{black}{\rightarrow}\;\textcolor{teal}{1}\;\textcolor{blue}{[1]}\\
	& \textcolor{red}{T1_2}\;\textcolor{black}{\rightarrow}\;\textcolor{teal}{2}\;\textcolor{blue}{[1]}\\
	& \textcolor{red}{T1_3}\;\textcolor{black}{\rightarrow}\;\textcolor{teal}{3}\;\textcolor{blue}{[1]}\\
	& \textcolor{red}{T1_4}\;\textcolor{black}{\rightarrow}\;\textcolor{teal}{4}\;\textcolor{blue}{[1]}\\
	& \textcolor{red}{C1_1}\;\textcolor{black}{\rightarrow}\;\textcolor{teal}{5}\;\textcolor{blue}{[1]}\\
	& \textcolor{red}{C1_2}\;\textcolor{black}{\rightarrow}\;\textcolor{teal}{6}\;\textcolor{blue}{[1]}\\
	& \textcolor{red}{C1_3}\;\textcolor{black}{\rightarrow}\;\textcolor{teal}{7}\;\textcolor{blue}{[1]}\\
	& \textcolor{red}{C1_4}\;\textcolor{black}{\rightarrow}\;\textcolor{teal}{8}\;\textcolor{blue}{[1]}\\
	& \textcolor{red}{C1_5}\;\textcolor{black}{\rightarrow}\;\textcolor{teal}{9}\;\textcolor{blue}{[1]}\\
\end{align*}

\end{minipage}%
\begin{minipage}{0.5\textwidth}
\centering

\begin{align*}
	& \textcolor{red}{S}\;\textcolor{black}{\rightarrow}\;\textcolor{red}{S5}\;\textcolor{blue}{[1]}\\
	& \textcolor{red}{S5}\;\textcolor{black}{\rightarrow}\;\textcolor{red}{B4}\;\textcolor{red}{C1_1}\;\textcolor{red}{E4}\;\textcolor{red}{T1_1}\;\textcolor{blue}{[0.25]}\\
	& \textcolor{red}{S5}\;\textcolor{black}{\rightarrow}\;\textcolor{red}{B4}\;\textcolor{red}{C1_2}\;\textcolor{red}{E4}\;\textcolor{red}{T1_2}\;\textcolor{blue}{[0.25]}\\
	& \textcolor{red}{S5}\;\textcolor{black}{\rightarrow}\;\textcolor{red}{B4}\;\textcolor{red}{C1_3}\;\textcolor{red}{E4}\;\textcolor{red}{T1_3}\;\textcolor{blue}{[0.25]}\\
	& \textcolor{red}{S5}\;\textcolor{black}{\rightarrow}\;\textcolor{red}{B4}\;\textcolor{red}{C1_4}\;\textcolor{red}{E4}\;\textcolor{red}{T1_4}\;\textcolor{blue}{[0.25]}\\
	& \textcolor{red}{B4}\;\textcolor{black}{\rightarrow}\;\textcolor{red}{B3}\;\textcolor{blue}{[0.3333]}\\
	& \textcolor{red}{B4}\;\textcolor{black}{\rightarrow}\;\textcolor{red}{B3}\;\textcolor{red}{B3}\;\textcolor{red}{B3}\;\textcolor{blue}{[0.3333]}\\
	& \textcolor{red}{B4}\;\textcolor{black}{\rightarrow}\;\textcolor{red}{B3}\;\textcolor{red}{B3}\;\textcolor{blue}{[0.3333]}\\
	& \textcolor{red}{B3}\;\textcolor{black}{\rightarrow}\;\textcolor{red}{B2}\;\textcolor{blue}{[0.3333]}\\
	& \textcolor{red}{B3}\;\textcolor{black}{\rightarrow}\;\textcolor{red}{B2}\;\textcolor{blue}{[0.3333]}\\
	& \textcolor{red}{B3}\;\textcolor{black}{\rightarrow}\;\textcolor{red}{B2}\;\textcolor{red}{B2}\;\textcolor{blue}{[0.3333]}\\
	& \textcolor{red}{B2}\;\textcolor{black}{\rightarrow}\;\textcolor{red}{B1}\;\textcolor{blue}{[0.3333]}\\
	& \textcolor{red}{B2}\;\textcolor{black}{\rightarrow}\;\textcolor{red}{B1}\;\textcolor{blue}{[0.3333]}\\
	& \textcolor{red}{B2}\;\textcolor{black}{\rightarrow}\;\textcolor{red}{B1}\;\textcolor{red}{B1}\;\textcolor{red}{B1}\;\textcolor{blue}{[0.3333]}\\
	& \textcolor{red}{B1}\;\textcolor{black}{\rightarrow}\;\textcolor{teal}{2}\;\textcolor{teal}{9}\;\textcolor{teal}{3}\;\textcolor{blue}{[0.95]}\\
	& \textcolor{red}{B1}\;\textcolor{black}{\rightarrow}\;\textcolor{teal}{9}\;\textcolor{teal}{6}\;\textcolor{teal}{1}\;\textcolor{blue}{[0.025]}\\
	& \textcolor{red}{B1}\;\textcolor{black}{\rightarrow}\;\textcolor{teal}{1}\;\textcolor{teal}{8}\;\textcolor{teal}{6}\;\textcolor{blue}{[0.025]}\\
	& \textcolor{red}{E4}\;\textcolor{black}{\rightarrow}\;\textcolor{red}{E3}\;\textcolor{blue}{[0.3333]}\\
	& \textcolor{red}{E4}\;\textcolor{black}{\rightarrow}\;\textcolor{red}{E3}\;\textcolor{red}{E3}\;\textcolor{blue}{[0.3333]}\\
	& \textcolor{red}{E4}\;\textcolor{black}{\rightarrow}\;\textcolor{red}{E3}\;\textcolor{red}{E3}\;\textcolor{red}{E3}\;\textcolor{blue}{[0.3333]}\\
	& \textcolor{red}{E3}\;\textcolor{black}{\rightarrow}\;\textcolor{red}{E2}\;\textcolor{blue}{[0.3333]}\\
	& \textcolor{red}{E3}\;\textcolor{black}{\rightarrow}\;\textcolor{red}{E2}\;\textcolor{red}{E2}\;\textcolor{blue}{[0.3333]}\\
	& \textcolor{red}{E3}\;\textcolor{black}{\rightarrow}\;\textcolor{red}{E2}\;\textcolor{blue}{[0.3333]}\\
	& \textcolor{red}{E2}\;\textcolor{black}{\rightarrow}\;\textcolor{red}{E1}\;\textcolor{red}{E1}\;\textcolor{blue}{[0.3333]}\\
	& \textcolor{red}{E2}\;\textcolor{black}{\rightarrow}\;\textcolor{red}{E1}\;\textcolor{blue}{[0.3333]}\\
	& \textcolor{red}{E2}\;\textcolor{black}{\rightarrow}\;\textcolor{red}{E1}\;\textcolor{red}{E1}\;\textcolor{red}{E1}\;\textcolor{blue}{[0.3333]}\\
	& \textcolor{red}{E1}\;\textcolor{black}{\rightarrow}\;\textcolor{teal}{5}\;\textcolor{teal}{6}\;\textcolor{teal}{5}\;\textcolor{teal}{9}\;\textcolor{blue}{[0.95]}\\
	& \textcolor{red}{E1}\;\textcolor{black}{\rightarrow}\;\textcolor{teal}{1}\;\textcolor{teal}{8}\;\textcolor{teal}{6}\;\textcolor{teal}{6}\;\textcolor{blue}{[0.025]}\\
	& \textcolor{red}{E1}\;\textcolor{black}{\rightarrow}\;\textcolor{teal}{1}\;\textcolor{teal}{5}\;\textcolor{teal}{1}\;\textcolor{teal}{5}\;\textcolor{blue}{[0.025]}\\
	& \textcolor{red}{T1_1}\;\textcolor{black}{\rightarrow}\;\textcolor{teal}{1}\;\textcolor{blue}{[1]}\\
	& \textcolor{red}{T1_2}\;\textcolor{black}{\rightarrow}\;\textcolor{teal}{2}\;\textcolor{blue}{[1]}\\
	& \textcolor{red}{T1_3}\;\textcolor{black}{\rightarrow}\;\textcolor{teal}{3}\;\textcolor{blue}{[1]}\\
	& \textcolor{red}{T1_4}\;\textcolor{black}{\rightarrow}\;\textcolor{teal}{4}\;\textcolor{blue}{[1]}\\
	& \textcolor{red}{C1_1}\;\textcolor{black}{\rightarrow}\;\textcolor{teal}{5}\;\textcolor{blue}{[1]}\\
	& \textcolor{red}{C1_2}\;\textcolor{black}{\rightarrow}\;\textcolor{teal}{6}\;\textcolor{blue}{[1]}\\
	& \textcolor{red}{C1_3}\;\textcolor{black}{\rightarrow}\;\textcolor{teal}{7}\;\textcolor{blue}{[1]}\\
	& \textcolor{red}{C1_4}\;\textcolor{black}{\rightarrow}\;\textcolor{teal}{8}\;\textcolor{blue}{[1]}\\
	& \textcolor{red}{C1_5}\;\textcolor{black}{\rightarrow}\;\textcolor{teal}{9}\;\textcolor{blue}{[1]}\\
\end{align*}

\end{minipage}%

\caption{Production rules of ${G}_3$ (left) and ${G}_4$ (right). Compared to $ G_3 $, the grammar $ G_4 $ generates more skewed distribution (or lower entropy) of strings, since one out of three production rules of non-terminal $ \textcolor{red}{B1} $ and $ \textcolor{red}{E1} $ is selected with higher probability.}

\label{fig:grammar_g3_g4}
\end{figure*}

\begin{figure*}
\begin{minipage}{0.5\textwidth}
\centering
\begin{align*}
	& \textcolor{red}{S}\;\textcolor{black}{\rightarrow}\;\textcolor{red}{A16}\;\textcolor{blue}{[1]}\\
	& \textcolor{red}{A16}\;\textcolor{black}{\rightarrow}\;\textcolor{red}{A15}\;\textcolor{red}{A13}\;\textcolor{blue}{[0.50]}\\
	& \textcolor{red}{A16}\;\textcolor{black}{\rightarrow}\;\textcolor{red}{A13}\;\textcolor{red}{A15}\;\textcolor{red}{A14}\;\textcolor{blue}{[0.50]}\\
	& \textcolor{red}{A13}\;\textcolor{black}{\rightarrow}\;\textcolor{red}{A11}\;\textcolor{red}{A12}\;\textcolor{blue}{[0.50]}\\
	& \textcolor{red}{A13}\;\textcolor{black}{\rightarrow}\;\textcolor{red}{A12}\;\textcolor{red}{A11}\;\textcolor{blue}{[0.50]}\\
	& \textcolor{red}{A14}\;\textcolor{black}{\rightarrow}\;\textcolor{red}{A11}\;\textcolor{red}{A10}\;\textcolor{red}{A12}\;\textcolor{blue}{[0.50]}\\
	& \textcolor{red}{A14}\;\textcolor{black}{\rightarrow}\;\textcolor{red}{A10}\;\textcolor{red}{A11}\;\textcolor{red}{A12}\;\textcolor{blue}{[0.50]}\\
	& \textcolor{red}{A15}\;\textcolor{black}{\rightarrow}\;\textcolor{red}{A12}\;\textcolor{red}{A11}\;\textcolor{red}{A10}\;\textcolor{blue}{[0.50]}\\
	& \textcolor{red}{A15}\;\textcolor{black}{\rightarrow}\;\textcolor{red}{A11}\;\textcolor{red}{A12}\;\textcolor{red}{A10}\;\textcolor{blue}{[0.50]}\\
	& \textcolor{red}{A10}\;\textcolor{black}{\rightarrow}\;\textcolor{red}{A7}\;\textcolor{red}{A9}\;\textcolor{red}{A8}\;\textcolor{blue}{[0.50]}\\
	& \textcolor{red}{A10}\;\textcolor{black}{\rightarrow}\;\textcolor{red}{A9}\;\textcolor{red}{A8}\;\textcolor{red}{A7}\;\textcolor{blue}{[0.50]}\\
	& \textcolor{red}{A11}\;\textcolor{black}{\rightarrow}\;\textcolor{red}{A8}\;\textcolor{red}{A7}\;\textcolor{red}{A9}\;\textcolor{blue}{[0.50]}\\
	& \textcolor{red}{A11}\;\textcolor{black}{\rightarrow}\;\textcolor{red}{A7}\;\textcolor{red}{A8}\;\textcolor{red}{A9}\;\textcolor{blue}{[0.50]}\\
	& \textcolor{red}{A12}\;\textcolor{black}{\rightarrow}\;\textcolor{red}{A8}\;\textcolor{red}{A9}\;\textcolor{red}{A7}\;\textcolor{blue}{[0.50]}\\
	& \textcolor{red}{A12}\;\textcolor{black}{\rightarrow}\;\textcolor{red}{A9}\;\textcolor{red}{A7}\;\textcolor{red}{A8}\;\textcolor{blue}{[0.50]}\\
	& \textcolor{red}{A7}\;\textcolor{black}{\rightarrow}\;\textcolor{teal}{3}\;\textcolor{teal}{1}\;\textcolor{blue}{[0.50]}\\
	& \textcolor{red}{A7}\;\textcolor{black}{\rightarrow}\;\textcolor{teal}{1}\;\textcolor{teal}{2}\;\textcolor{teal}{3}\;\textcolor{blue}{[0.50]}\\
	& \textcolor{red}{A8}\;\textcolor{black}{\rightarrow}\;\textcolor{teal}{6}\;\textcolor{teal}{5}\;\textcolor{blue}{[0.50]}\\
	& \textcolor{red}{A8}\;\textcolor{black}{\rightarrow}\;\textcolor{teal}{6}\;\textcolor{teal}{4}\;\textcolor{teal}{5}\;\textcolor{blue}{[0.50]}\\
	& \textcolor{red}{A9}\;\textcolor{black}{\rightarrow}\;\textcolor{teal}{9}\;\textcolor{teal}{8}\;\textcolor{teal}{7}\;\textcolor{blue}{[0.50]}\\
	& \textcolor{red}{A9}\;\textcolor{black}{\rightarrow}\;\textcolor{teal}{8}\;\textcolor{teal}{7}\;\textcolor{blue}{[0.50]}\\
\end{align*}
\end{minipage}%
\begin{minipage}{0.5\textwidth}
\centering
\begin{align*}
	& \textcolor{red}{S}\;\textcolor{black}{\rightarrow}\;\textcolor{red}{S5}\;\textcolor{blue}{[1]}\\
	& \textcolor{red}{S5}\;\textcolor{black}{\rightarrow}\;\textcolor{red}{B4}\;\textcolor{red}{C1_1}\;\textcolor{red}{E4}\;\textcolor{red}{T1_1}\;\textcolor{blue}{[0.25]}\\
	& \textcolor{red}{S5}\;\textcolor{black}{\rightarrow}\;\textcolor{red}{B4}\;\textcolor{red}{C1_2}\;\textcolor{red}{E4}\;\textcolor{red}{T1_2}\;\textcolor{blue}{[0.25]}\\
	& \textcolor{red}{S5}\;\textcolor{black}{\rightarrow}\;\textcolor{red}{B4}\;\textcolor{red}{C1_3}\;\textcolor{red}{E4}\;\textcolor{red}{T1_3}\;\textcolor{blue}{[0.25]}\\
	& \textcolor{red}{S5}\;\textcolor{black}{\rightarrow}\;\textcolor{red}{B4}\;\textcolor{red}{C1_4}\;\textcolor{red}{E4}\;\textcolor{red}{T1_4}\;\textcolor{blue}{[0.25]}\\
	& \textcolor{red}{B4}\;\textcolor{black}{\rightarrow}\;\textcolor{red}{B3}\;\textcolor{blue}{[0.3333]}\\
	& \textcolor{red}{B4}\;\textcolor{black}{\rightarrow}\;\textcolor{red}{B3}\;\textcolor{red}{B3}\;\textcolor{red}{B3}\;\textcolor{blue}{[0.3333]}\\
	& \textcolor{red}{B4}\;\textcolor{black}{\rightarrow}\;\textcolor{red}{B3}\;\textcolor{red}{B3}\;\textcolor{blue}{[0.3333]}\\
	& \textcolor{red}{B3}\;\textcolor{black}{\rightarrow}\;\textcolor{red}{B2}\;\textcolor{blue}{[0.3333]}\\
	& \textcolor{red}{B3}\;\textcolor{black}{\rightarrow}\;\textcolor{red}{B2}\;\textcolor{blue}{[0.3333]}\\
	& \textcolor{red}{B3}\;\textcolor{black}{\rightarrow}\;\textcolor{red}{B2}\;\textcolor{red}{B2}\;\textcolor{blue}{[0.3333]}\\
	& \textcolor{red}{B2}\;\textcolor{black}{\rightarrow}\;\textcolor{red}{B1}\;\textcolor{blue}{[0.3333]}\\
	& \textcolor{red}{B2}\;\textcolor{black}{\rightarrow}\;\textcolor{red}{B1}\;\textcolor{blue}{[0.3333]}\\
	& \textcolor{red}{B2}\;\textcolor{black}{\rightarrow}\;\textcolor{red}{B1}\;\textcolor{red}{B1}\;\textcolor{red}{B1}\;\textcolor{blue}{[0.3333]}\\
	& \textcolor{red}{B1}\;\textcolor{black}{\rightarrow}\;\textcolor{teal}{2}\;\textcolor{teal}{9}\;\textcolor{teal}{3}\;\textcolor{blue}{[0.3333]}\\
	& \textcolor{red}{B1}\;\textcolor{black}{\rightarrow}\;\textcolor{teal}{9}\;\textcolor{teal}{6}\;\textcolor{teal}{1}\;\textcolor{blue}{[0.3333]}\\
	& \textcolor{red}{B1}\;\textcolor{black}{\rightarrow}\;\textcolor{teal}{1}\;\textcolor{teal}{8}\;\textcolor{teal}{6}\;\textcolor{teal}{2}\;\textcolor{blue}{[0.3333]}\\
	& \textcolor{red}{E4}\;\textcolor{black}{\rightarrow}\;\textcolor{red}{E3}\;\textcolor{blue}{[0.3333]}\\
	& \textcolor{red}{E4}\;\textcolor{black}{\rightarrow}\;\textcolor{red}{E3}\;\textcolor{red}{E3}\;\textcolor{blue}{[0.3333]}\\
	& \textcolor{red}{E4}\;\textcolor{black}{\rightarrow}\;\textcolor{red}{E3}\;\textcolor{red}{E3}\;\textcolor{red}{E3}\;\textcolor{blue}{[0.3333]}\\
	& \textcolor{red}{E3}\;\textcolor{black}{\rightarrow}\;\textcolor{red}{E2}\;\textcolor{blue}{[0.3333]}\\
	& \textcolor{red}{E3}\;\textcolor{black}{\rightarrow}\;\textcolor{red}{E2}\;\textcolor{red}{E2}\;\textcolor{blue}{[0.3333]}\\
	& \textcolor{red}{E3}\;\textcolor{black}{\rightarrow}\;\textcolor{red}{E2}\;\textcolor{blue}{[0.3333]}\\
	& \textcolor{red}{E2}\;\textcolor{black}{\rightarrow}\;\textcolor{red}{E1}\;\textcolor{red}{E1}\;\textcolor{blue}{[0.3333]}\\
	& \textcolor{red}{E2}\;\textcolor{black}{\rightarrow}\;\textcolor{red}{E1}\;\textcolor{blue}{[0.3333]}\\
	& \textcolor{red}{E2}\;\textcolor{black}{\rightarrow}\;\textcolor{red}{E1}\;\textcolor{red}{E1}\;\textcolor{red}{E1}\;\textcolor{blue}{[0.3333]}\\
	& \textcolor{red}{E1}\;\textcolor{black}{\rightarrow}\;\textcolor{teal}{5}\;\textcolor{teal}{6}\;\textcolor{blue}{[0.3333]}\\
	& \textcolor{red}{E1}\;\textcolor{black}{\rightarrow}\;\textcolor{teal}{1}\;\textcolor{teal}{8}\;\textcolor{teal}{6}\;\textcolor{teal}{6}\;\textcolor{blue}{[0.3333]}\\
	& \textcolor{red}{E1}\;\textcolor{black}{\rightarrow}\;\textcolor{teal}{1}\;\textcolor{teal}{5}\;\textcolor{teal}{1}\;\textcolor{teal}{5}\;\textcolor{teal}{5}\;\textcolor{teal}{9}\;\textcolor{blue}{[0.3333]}\\
	& \textcolor{red}{T1_1}\;\textcolor{black}{\rightarrow}\;\textcolor{teal}{1}\;\textcolor{blue}{[1]}\\
	& \textcolor{red}{T1_2}\;\textcolor{black}{\rightarrow}\;\textcolor{teal}{2}\;\textcolor{blue}{[1]}\\
	& \textcolor{red}{T1_3}\;\textcolor{black}{\rightarrow}\;\textcolor{teal}{3}\;\textcolor{blue}{[1]}\\
	& \textcolor{red}{T1_4}\;\textcolor{black}{\rightarrow}\;\textcolor{teal}{4}\;\textcolor{blue}{[1]}\\
	& \textcolor{red}{C1_1}\;\textcolor{black}{\rightarrow}\;\textcolor{teal}{5}\;\textcolor{blue}{[1]}\\
	& \textcolor{red}{C1_2}\;\textcolor{black}{\rightarrow}\;\textcolor{teal}{6}\;\textcolor{blue}{[1]}\\
	& \textcolor{red}{C1_3}\;\textcolor{black}{\rightarrow}\;\textcolor{teal}{7}\;\textcolor{blue}{[1]}\\
	& \textcolor{red}{C1_4}\;\textcolor{black}{\rightarrow}\;\textcolor{teal}{8}\;\textcolor{blue}{[1]}\\
	& \textcolor{red}{C1_5}\;\textcolor{black}{\rightarrow}\;\textcolor{teal}{9}\;\textcolor{blue}{[1]}\\
\end{align*}
\end{minipage}%

\caption{Production rules of ${G}_5$ (left) and ${G}_6$ (right). These grammars are adapted from $ {G}_1 $ and $ {G}_3 $ respectively, by allowing non-uniform lengths of tokens in the lowest level production rules.}

\label{fig:grammar_g5_g6}
\end{figure*}

\begin{figure*}
\begin{minipage}{0.5\textwidth}
\centering

\begin{align*}
	& \textcolor{red}{S}\;\textcolor{black}{\rightarrow}\;\textcolor{red}{A16}\;\textcolor{blue}{[1]}\\
	& \textcolor{red}{A16}\;\textcolor{black}{\rightarrow}\;\textcolor{red}{A15}\;\textcolor{red}{A13}\;\textcolor{blue}{[0.50]}\\
	& \textcolor{red}{A16}\;\textcolor{black}{\rightarrow}\;\textcolor{red}{A13}\;\textcolor{red}{A15}\;\textcolor{red}{A14}\;\textcolor{blue}{[0.50]}\\
	& \textcolor{red}{A13}\;\textcolor{black}{\rightarrow}\;\textcolor{red}{A11}\;\textcolor{red}{A12}\;\textcolor{blue}{[0.50]}\\
	& \textcolor{red}{A13}\;\textcolor{black}{\rightarrow}\;\textcolor{red}{A12}\;\textcolor{red}{A11}\;\textcolor{blue}{[0.50]}\\
	& \textcolor{red}{A14}\;\textcolor{black}{\rightarrow}\;\textcolor{red}{A11}\;\textcolor{red}{A10}\;\textcolor{red}{A12}\;\textcolor{blue}{[0.50]}\\
	& \textcolor{red}{A14}\;\textcolor{black}{\rightarrow}\;\textcolor{red}{A10}\;\textcolor{red}{A11}\;\textcolor{red}{A12}\;\textcolor{blue}{[0.50]}\\
	& \textcolor{red}{A15}\;\textcolor{black}{\rightarrow}\;\textcolor{red}{A12}\;\textcolor{red}{A11}\;\textcolor{red}{A10}\;\textcolor{blue}{[0.50]}\\
	& \textcolor{red}{A15}\;\textcolor{black}{\rightarrow}\;\textcolor{red}{A11}\;\textcolor{red}{A12}\;\textcolor{red}{A10}\;\textcolor{blue}{[0.50]}\\
	& \textcolor{red}{A10}\;\textcolor{black}{\rightarrow}\;\textcolor{red}{A7}\;\textcolor{red}{A9}\;\textcolor{red}{A8}\;\textcolor{blue}{[0.50]}\\
	& \textcolor{red}{A10}\;\textcolor{black}{\rightarrow}\;\textcolor{red}{A9}\;\textcolor{red}{A8}\;\textcolor{red}{A7}\;\textcolor{blue}{[0.50]}\\
	& \textcolor{red}{A11}\;\textcolor{black}{\rightarrow}\;\textcolor{red}{A8}\;\textcolor{red}{A7}\;\textcolor{red}{A9}\;\textcolor{blue}{[0.50]}\\
	& \textcolor{red}{A11}\;\textcolor{black}{\rightarrow}\;\textcolor{red}{A7}\;\textcolor{red}{A8}\;\textcolor{red}{A9}\;\textcolor{blue}{[0.50]}\\
	& \textcolor{red}{A12}\;\textcolor{black}{\rightarrow}\;\textcolor{red}{A8}\;\textcolor{red}{A9}\;\textcolor{red}{A7}\;\textcolor{blue}{[0.50]}\\
	& \textcolor{red}{A12}\;\textcolor{black}{\rightarrow}\;\textcolor{red}{A9}\;\textcolor{red}{A7}\;\textcolor{red}{A8}\;\textcolor{blue}{[0.50]}\\
	& \textcolor{red}{A7}\;\textcolor{black}{\rightarrow}\;\textcolor{teal}{c}\;\textcolor{teal}{a}\;\textcolor{blue}{[0.50]}\\
	& \textcolor{red}{A7}\;\textcolor{black}{\rightarrow}\;\textcolor{teal}{a}\;\textcolor{teal}{b}\;\textcolor{teal}{c}\;\textcolor{blue}{[0.50]}\\
	& \textcolor{red}{A8}\;\textcolor{black}{\rightarrow}\;\textcolor{teal}{f}\;\textcolor{teal}{e}\;\textcolor{blue}{[0.50]}\\
	& \textcolor{red}{A8}\;\textcolor{black}{\rightarrow}\;\textcolor{teal}{f}\;\textcolor{teal}{d}\;\textcolor{teal}{e}\;\textcolor{blue}{[0.50]}\\
	& \textcolor{red}{A9}\;\textcolor{black}{\rightarrow}\;\textcolor{teal}{i}\;\textcolor{teal}{h}\;\textcolor{teal}{g}\;\textcolor{blue}{[0.50]}\\
	& \textcolor{red}{A9}\;\textcolor{black}{\rightarrow}\;\textcolor{teal}{h}\;\textcolor{teal}{g}\;\textcolor{blue}{[0.50]}\\
\end{align*}

\end{minipage}%
\begin{minipage}{0.5\textwidth}
\centering

\begin{align*}
	& \textcolor{red}{S}\;\textcolor{black}{\rightarrow}\;\textcolor{red}{S5}\;\textcolor{blue}{[1]}\\
	& \textcolor{red}{S5}\;\textcolor{black}{\rightarrow}\;\textcolor{red}{B4}\;\textcolor{red}{C1_1}\;\textcolor{red}{E4}\;\textcolor{red}{T1_1}\;\textcolor{blue}{[0.25]}\\
	& \textcolor{red}{S5}\;\textcolor{black}{\rightarrow}\;\textcolor{red}{B4}\;\textcolor{red}{C1_2}\;\textcolor{red}{E4}\;\textcolor{red}{T1_2}\;\textcolor{blue}{[0.25]}\\
	& \textcolor{red}{S5}\;\textcolor{black}{\rightarrow}\;\textcolor{red}{B4}\;\textcolor{red}{C1_3}\;\textcolor{red}{E4}\;\textcolor{red}{T1_3}\;\textcolor{blue}{[0.25]}\\
	& \textcolor{red}{S5}\;\textcolor{black}{\rightarrow}\;\textcolor{red}{B4}\;\textcolor{red}{C1_4}\;\textcolor{red}{E4}\;\textcolor{red}{T1_4}\;\textcolor{blue}{[0.25]}\\
	& \textcolor{red}{B4}\;\textcolor{black}{\rightarrow}\;\textcolor{red}{B3}\;\textcolor{blue}{[0.3333]}\\
	& \textcolor{red}{B4}\;\textcolor{black}{\rightarrow}\;\textcolor{red}{B3}\;\textcolor{red}{B3}\;\textcolor{red}{B3}\;\textcolor{blue}{[0.3333]}\\
	& \textcolor{red}{B4}\;\textcolor{black}{\rightarrow}\;\textcolor{red}{B3}\;\textcolor{red}{B3}\;\textcolor{blue}{[0.3333]}\\
	& \textcolor{red}{B3}\;\textcolor{black}{\rightarrow}\;\textcolor{red}{B2}\;\textcolor{blue}{[0.3333]}\\
	& \textcolor{red}{B3}\;\textcolor{black}{\rightarrow}\;\textcolor{red}{B2}\;\textcolor{blue}{[0.3333]}\\
	& \textcolor{red}{B3}\;\textcolor{black}{\rightarrow}\;\textcolor{red}{B2}\;\textcolor{red}{B2}\;\textcolor{blue}{[0.3333]}\\
	& \textcolor{red}{B2}\;\textcolor{black}{\rightarrow}\;\textcolor{red}{B1}\;\textcolor{blue}{[0.3333]}\\
	& \textcolor{red}{B2}\;\textcolor{black}{\rightarrow}\;\textcolor{red}{B1}\;\textcolor{blue}{[0.3333]}\\
	& \textcolor{red}{B2}\;\textcolor{black}{\rightarrow}\;\textcolor{red}{B1}\;\textcolor{red}{B1}\;\textcolor{red}{B1}\;\textcolor{blue}{[0.3333]}\\
	& \textcolor{red}{B1}\;\textcolor{black}{\rightarrow}\;\textcolor{teal}{b}\;\textcolor{teal}{i}\;\textcolor{teal}{c}\;\textcolor{blue}{[0.3333]}\\
	& \textcolor{red}{B1}\;\textcolor{black}{\rightarrow}\;\textcolor{teal}{i}\;\textcolor{teal}{f}\;\textcolor{teal}{a}\;\textcolor{blue}{[0.3333]}\\
	& \textcolor{red}{B1}\;\textcolor{black}{\rightarrow}\;\textcolor{teal}{a}\;\textcolor{teal}{h}\;\textcolor{teal}{f}\;\textcolor{teal}{b}\;\textcolor{blue}{[0.3333]}\\
	& \textcolor{red}{E4}\;\textcolor{black}{\rightarrow}\;\textcolor{red}{E3}\;\textcolor{blue}{[0.3333]}\\
	& \textcolor{red}{E4}\;\textcolor{black}{\rightarrow}\;\textcolor{red}{E3}\;\textcolor{red}{E3}\;\textcolor{blue}{[0.3333]}\\
	& \textcolor{red}{E4}\;\textcolor{black}{\rightarrow}\;\textcolor{red}{E3}\;\textcolor{red}{E3}\;\textcolor{red}{E3}\;\textcolor{blue}{[0.3333]}\\
	& \textcolor{red}{E3}\;\textcolor{black}{\rightarrow}\;\textcolor{red}{E2}\;\textcolor{blue}{[0.3333]}\\
	& \textcolor{red}{E3}\;\textcolor{black}{\rightarrow}\;\textcolor{red}{E2}\;\textcolor{red}{E2}\;\textcolor{blue}{[0.3333]}\\
	& \textcolor{red}{E3}\;\textcolor{black}{\rightarrow}\;\textcolor{red}{E2}\;\textcolor{blue}{[0.3333]}\\
	& \textcolor{red}{E2}\;\textcolor{black}{\rightarrow}\;\textcolor{red}{E1}\;\textcolor{red}{E1}\;\textcolor{blue}{[0.3333]}\\
	& \textcolor{red}{E2}\;\textcolor{black}{\rightarrow}\;\textcolor{red}{E1}\;\textcolor{blue}{[0.3333]}\\
	& \textcolor{red}{E2}\;\textcolor{black}{\rightarrow}\;\textcolor{red}{E1}\;\textcolor{red}{E1}\;\textcolor{red}{E1}\;\textcolor{blue}{[0.3333]}\\
	& \textcolor{red}{E1}\;\textcolor{black}{\rightarrow}\;\textcolor{teal}{e}\;\textcolor{teal}{f}\;\textcolor{blue}{[0.3333]}\\
	& \textcolor{red}{E1}\;\textcolor{black}{\rightarrow}\;\textcolor{teal}{a}\;\textcolor{teal}{h}\;\textcolor{teal}{f}\;\textcolor{teal}{f}\;\textcolor{blue}{[0.3333]}\\
	& \textcolor{red}{E1}\;\textcolor{black}{\rightarrow}\;\textcolor{teal}{a}\;\textcolor{teal}{e}\;\textcolor{teal}{a}\;\textcolor{teal}{e}\;\textcolor{teal}{e}\;\textcolor{teal}{i}\;\textcolor{blue}{[0.3333]}\\
	& \textcolor{red}{T1_1}\;\textcolor{black}{\rightarrow}\;\textcolor{teal}{a}\;\textcolor{blue}{[1]}\\
	& \textcolor{red}{T1_2}\;\textcolor{black}{\rightarrow}\;\textcolor{teal}{b}\;\textcolor{blue}{[1]}\\
	& \textcolor{red}{T1_3}\;\textcolor{black}{\rightarrow}\;\textcolor{teal}{c}\;\textcolor{blue}{[1]}\\
	& \textcolor{red}{T1_4}\;\textcolor{black}{\rightarrow}\;\textcolor{teal}{d}\;\textcolor{blue}{[1]}\\
	& \textcolor{red}{C1_1}\;\textcolor{black}{\rightarrow}\;\textcolor{teal}{e}\;\textcolor{blue}{[1]}\\
	& \textcolor{red}{C1_2}\;\textcolor{black}{\rightarrow}\;\textcolor{teal}{f}\;\textcolor{blue}{[1]}\\
	& \textcolor{red}{C1_3}\;\textcolor{black}{\rightarrow}\;\textcolor{teal}{g}\;\textcolor{blue}{[1]}\\
	& \textcolor{red}{C1_4}\;\textcolor{black}{\rightarrow}\;\textcolor{teal}{h}\;\textcolor{blue}{[1]}\\
	& \textcolor{red}{C1_5}\;\textcolor{black}{\rightarrow}\;\textcolor{teal}{i}\;\textcolor{blue}{[1]}\\
\end{align*}

\end{minipage}%

\caption{Production rules of ${G}_7$ (left) and ${G}_8$ (right). These grammars are adapted from $ {G}_5 $ and $ {G}_6 $ respectively, by replacing numerical tokens with Latin character tokens.}

\label{fig:grammar_g6_g7}
\end{figure*}

\begin{figure*}
    \begin{align*}
        & \textcolor{red}{S}\;\textcolor{black}{\rightarrow}\;\textcolor{red}{A13}\;\textcolor{blue}{[1]}\\
        & \textcolor{red}{A13}\;\textcolor{black}{\rightarrow}\;\textcolor{red}{A10}\;\textcolor{red}{A12}\;\textcolor{blue}{[0.8413447461]}\\
        & \textcolor{red}{A13}\;\textcolor{black}{\rightarrow}\;\textcolor{red}{A9}\;\textcolor{red}{A10}\;\textcolor{blue}{[0.1586552539]}\\
        & \textcolor{red}{A12}\;\textcolor{black}{\rightarrow}\;\textcolor{red}{A5}\;\textcolor{red}{A7}\;\textcolor{red}{A5}\;\textcolor{red}{A8}\;\textcolor{blue}{[0.8413447461]}\\
        & \textcolor{red}{A12}\;\textcolor{black}{\rightarrow}\;\textcolor{red}{A6}\;\textcolor{red}{A6}\;\textcolor{blue}{[0.1586552539]}\\
        & \textcolor{red}{A11}\;\textcolor{black}{\rightarrow}\;\textcolor{red}{A5}\;\textcolor{red}{A7}\;\textcolor{red}{A8}\;\textcolor{blue}{[0.8413447461]}\\
        & \textcolor{red}{A11}\;\textcolor{black}{\rightarrow}\;\textcolor{red}{A6}\;\textcolor{blue}{[0.1586552539]}\\
        & \textcolor{red}{A10}\;\textcolor{black}{\rightarrow}\;\textcolor{red}{A6}\;\textcolor{red}{A7}\;\textcolor{red}{A5}\;\textcolor{red}{A7}\;\textcolor{blue}{[0.8413447461]}\\
        & \textcolor{red}{A10}\;\textcolor{black}{\rightarrow}\;\textcolor{red}{A8}\;\textcolor{red}{A6}\;\textcolor{blue}{[0.1586552539]}\\
        & \textcolor{red}{A9}\;\textcolor{black}{\rightarrow}\;\textcolor{red}{A6}\;\textcolor{blue}{[0.8413447461]}\\
        & \textcolor{red}{A9}\;\textcolor{black}{\rightarrow}\;\textcolor{red}{A8}\;\textcolor{red}{A5}\;\textcolor{red}{A8}\;\textcolor{blue}{[0.1586552539]}\\
        & \textcolor{red}{A8}\;\textcolor{black}{\rightarrow}\;\textcolor{red}{A1}\;\textcolor{blue}{[0.8413447461]}\\
        & \textcolor{red}{A8}\;\textcolor{black}{\rightarrow}\;\textcolor{red}{A3}\;\textcolor{blue}{[0.1586552539]}\\
        & \textcolor{red}{A7}\;\textcolor{black}{\rightarrow}\;\textcolor{red}{A1}\;\textcolor{red}{A2}\;\textcolor{blue}{[0.8413447461]}\\
        & \textcolor{red}{A7}\;\textcolor{black}{\rightarrow}\;\textcolor{red}{A4}\;\textcolor{red}{A4}\;\textcolor{blue}{[0.1586552539]}\\
        & \textcolor{red}{A6}\;\textcolor{black}{\rightarrow}\;\textcolor{red}{A2}\;\textcolor{red}{A1}\;\textcolor{red}{A3}\;\textcolor{red}{A4}\;\textcolor{blue}{[0.8413447461]}\\
        & \textcolor{red}{A6}\;\textcolor{black}{\rightarrow}\;\textcolor{red}{A3}\;\textcolor{red}{A4}\;\textcolor{red}{A1}\;\textcolor{blue}{[0.1586552539]}\\
        & \textcolor{red}{A5}\;\textcolor{black}{\rightarrow}\;\textcolor{red}{A1}\;\textcolor{red}{A4}\;\textcolor{blue}{[0.8413447461]}\\
        & \textcolor{red}{A5}\;\textcolor{black}{\rightarrow}\;\textcolor{red}{A3}\;\textcolor{blue}{[0.1586552539]}\\
        & \textcolor{red}{A4}\;\textcolor{black}{\rightarrow}\;\textcolor{teal}{4}\;\textcolor{teal}{3}\;\textcolor{blue}{[0.8413447461]}\\
        & \textcolor{red}{A4}\;\textcolor{black}{\rightarrow}\;\textcolor{teal}{3}\;\textcolor{teal}{0}\;\textcolor{teal}{3}\;\textcolor{blue}{[0.1586552539]}\\
        & \textcolor{red}{A3}\;\textcolor{black}{\rightarrow}\;\textcolor{teal}{2}\;\textcolor{teal}{4}\;\textcolor{teal}{1}\;\textcolor{teal}{0}\;\textcolor{blue}{[0.8413447461]}\\
        & \textcolor{red}{A3}\;\textcolor{black}{\rightarrow}\;\textcolor{teal}{0}\;\textcolor{teal}{4}\;\textcolor{teal}{3}\;\textcolor{blue}{[0.1586552539]}\\
        & \textcolor{red}{A2}\;\textcolor{black}{\rightarrow}\;\textcolor{teal}{0}\;\textcolor{teal}{4}\;\textcolor{teal}{3}\;\textcolor{blue}{[0.8413447461]}\\
        & \textcolor{red}{A2}\;\textcolor{black}{\rightarrow}\;\textcolor{teal}{4}\;\textcolor{teal}{3}\;\textcolor{teal}{0}\;\textcolor{teal}{2}\;\textcolor{blue}{[0.1586552539]}\\
        & \textcolor{red}{A1}\;\textcolor{black}{\rightarrow}\;\textcolor{teal}{1}\;\textcolor{blue}{[0.8413447461]}\\
        & \textcolor{red}{A1}\;\textcolor{black}{\rightarrow}\;\textcolor{teal}{2}\;\textcolor{blue}{[0.1586552539]}\\
    \end{align*}

    \caption{Production rules of ${G}_9$. The hierarchical grammar has a maximum depth $ 4 $, maximum breadth $ 4 $, rules per non-terminals $ 2 $, and $ 5 $ numerical tokens.}
\end{figure*}

\begin{figure}
    \resizebox{\columnwidth}{!}{
    \begin{minipage}{0.1\textwidth}
        \begin{align*}
            & \textcolor{red}{S}\;\textcolor{black}{\rightarrow}\;\textcolor{red}{A19}\;\textcolor{blue}{[1]}\\
            & \textcolor{red}{A19}\;\textcolor{black}{\rightarrow}\;\textcolor{red}{A12}\;\textcolor{red}{A18}\;\textcolor{red}{A14}\;\textcolor{red}{A17}\;\textcolor{red}{A11}\;\textcolor{red}{A17}\;\textcolor{red}{A15}\;\textcolor{red}{A10}\;\textcolor{blue}{[0.16335]}\\
            & \textcolor{red}{A19}\;\textcolor{black}{\rightarrow}\;\textcolor{red}{A15}\;\textcolor{red}{A13}\;\textcolor{blue}{[0.2032909216]}\\
            & \textcolor{red}{A19}\;\textcolor{black}{\rightarrow}\;\textcolor{red}{A10}\;\textcolor{red}{A12}\;\textcolor{blue}{[0.2376710982]}\\
            & \textcolor{red}{A19}\;\textcolor{black}{\rightarrow}\;\textcolor{red}{A18}\;\textcolor{blue}{[0.2594097969]}\\
            & \textcolor{red}{A19}\;\textcolor{black}{\rightarrow}\;\textcolor{red}{A16}\;\textcolor{red}{A15}\;\textcolor{red}{A15}\;\textcolor{blue}{[0.1346583572]}\\
            & \textcolor{red}{A19}\;\textcolor{black}{\rightarrow}\;\textcolor{red}{A11}\;\textcolor{red}{A18}\;\textcolor{red}{A13}\;\textcolor{red}{A10}\;\textcolor{blue}{[0.0016209134]}\\
            & \textcolor{red}{A19}\;\textcolor{black}{\rightarrow}\;\textcolor{red}{A12}\;\textcolor{red}{A16}\;\textcolor{red}{A18}\;\textcolor{red}{A17}\;\textcolor{red}{A12}\;\textcolor{blue}{[0.0000000681]}\\
            & \textcolor{red}{A19}\;\textcolor{black}{\rightarrow}\;\textcolor{red}{A13}\;\textcolor{red}{A18}\;\textcolor{red}{A14}\;\textcolor{red}{A12}\;\textcolor{red}{A14}\;\textcolor{blue}{[0.0000000000]}\\
            & \textcolor{red}{A19}\;\textcolor{black}{\rightarrow}\;\textcolor{red}{A13}\;\textcolor{red}{A13}\;\textcolor{red}{A15}\;\textcolor{red}{A14}\;\textcolor{blue}{[0.0000000000]}\\
            & \textcolor{red}{A18}\;\textcolor{black}{\rightarrow}\;\textcolor{red}{A6}\;\textcolor{red}{A5}\;\textcolor{red}{A3}\;\textcolor{red}{A1}\;\textcolor{red}{A8}\;\textcolor{red}{A2}\;\textcolor{red}{A9}\;\textcolor{blue}{[0.1633488445]}\\
            & \textcolor{red}{A18}\;\textcolor{black}{\rightarrow}\;\textcolor{red}{A5}\;\textcolor{red}{A8}\;\textcolor{red}{A4}\;\textcolor{red}{A6}\;\textcolor{red}{A5}\;\textcolor{blue}{[0.2032909216]}\\
            & \textcolor{red}{A18}\;\textcolor{black}{\rightarrow}\;\textcolor{red}{A1}\;\textcolor{red}{A6}\;\textcolor{red}{A8}\;\textcolor{red}{A1}\;\textcolor{red}{A3}\;\textcolor{red}{A5}\;\textcolor{blue}{[0.2376710982]}\\
            & \textcolor{red}{A18}\;\textcolor{black}{\rightarrow}\;\textcolor{red}{A9}\;\textcolor{red}{A7}\;\textcolor{red}{A9}\;\textcolor{red}{A3}\;\textcolor{red}{A7}\;\textcolor{blue}{[0.2594097969]}\\
            & \textcolor{red}{A18}\;\textcolor{black}{\rightarrow}\;\textcolor{red}{A2}\;\textcolor{red}{A4}\;\textcolor{red}{A5}\;\textcolor{red}{A7}\;\textcolor{red}{A2}\;\textcolor{red}{A1}\;\textcolor{blue}{[0.1346583572]}\\
            & \textcolor{red}{A18}\;\textcolor{black}{\rightarrow}\;\textcolor{red}{A1}\;\textcolor{red}{A8}\;\textcolor{red}{A7}\;\textcolor{red}{A8}\;\textcolor{red}{A7}\;\textcolor{red}{A1}\;\textcolor{red}{A3}\;\textcolor{red}{A8}\;\textcolor{red}{A5}\;\textcolor{blue}{[0.0016209134]}\\
            & \textcolor{red}{A18}\;\textcolor{black}{\rightarrow}\;\textcolor{red}{A4}\;\textcolor{red}{A9}\;\textcolor{red}{A1}\;\textcolor{red}{A5}\;\textcolor{red}{A5}\;\textcolor{blue}{[0.0000000681]}\\
            & \textcolor{red}{A18}\;\textcolor{black}{\rightarrow}\;\textcolor{red}{A8}\;\textcolor{red}{A9}\;\textcolor{red}{A4}\;\textcolor{red}{A4}\;\textcolor{red}{A9}\;\textcolor{blue}{[0.0000000000]}\\
            & \textcolor{red}{A18}\;\textcolor{black}{\rightarrow}\;\textcolor{red}{A1}\;\textcolor{blue}{[0.0000000000]}\\
            & \textcolor{red}{A17}\;\textcolor{black}{\rightarrow}\;\textcolor{red}{A2}\;\textcolor{red}{A8}\;\textcolor{red}{A7}\;\textcolor{blue}{[0.1633488445]}\\
            & \textcolor{red}{A17}\;\textcolor{black}{\rightarrow}\;\textcolor{red}{A9}\;\textcolor{red}{A7}\;\textcolor{blue}{[0.2032909216]}\\
            & \textcolor{red}{A17}\;\textcolor{black}{\rightarrow}\;\textcolor{red}{A4}\;\textcolor{red}{A4}\;\textcolor{blue}{[0.2376710982]}\\
            & \textcolor{red}{A17}\;\textcolor{black}{\rightarrow}\;\textcolor{red}{A4}\;\textcolor{blue}{[0.2594097969]}\\
            & \textcolor{red}{A17}\;\textcolor{black}{\rightarrow}\;\textcolor{red}{A5}\;\textcolor{blue}{[0.1346583572]}\\
            & \textcolor{red}{A17}\;\textcolor{black}{\rightarrow}\;\textcolor{red}{A4}\;\textcolor{red}{A8}\;\textcolor{red}{A5}\;\textcolor{red}{A3}\;\textcolor{red}{A9}\;\textcolor{red}{A5}\;\textcolor{red}{A7}\;\textcolor{red}{A3}\;\textcolor{blue}{[0.0016209134]}\\
            & \textcolor{red}{A17}\;\textcolor{black}{\rightarrow}\;\textcolor{red}{A8}\;\textcolor{red}{A8}\;\textcolor{red}{A9}\;\textcolor{blue}{[0.0000000681]}\\
            & \textcolor{red}{A17}\;\textcolor{black}{\rightarrow}\;\textcolor{red}{A4}\;\textcolor{red}{A1}\;\textcolor{red}{A2}\;\textcolor{red}{A6}\;\textcolor{red}{A6}\;\textcolor{red}{A2}\;\textcolor{red}{A7}\;\textcolor{red}{A1}\;\textcolor{blue}{[0.0000000000]}\\
            & \textcolor{red}{A17}\;\textcolor{black}{\rightarrow}\;\textcolor{red}{A5}\;\textcolor{red}{A7}\;\textcolor{blue}{[0.0000000000]}\\
            & \textcolor{red}{A16}\;\textcolor{black}{\rightarrow}\;\textcolor{red}{A3}\;\textcolor{red}{A6}\;\textcolor{red}{A6}\;\textcolor{red}{A8}\;\textcolor{red}{A8}\;\textcolor{red}{A1}\;\textcolor{red}{A2}\;\textcolor{red}{A2}\;\textcolor{red}{A3}\;\textcolor{blue}{[0.1633488445]}\\
            & \textcolor{red}{A16}\;\textcolor{black}{\rightarrow}\;\textcolor{red}{A4}\;\textcolor{red}{A6}\;\textcolor{red}{A6}\;\textcolor{red}{A5}\;\textcolor{red}{A3}\;\textcolor{blue}{[0.2032909216]}\\
            & \textcolor{red}{A16}\;\textcolor{black}{\rightarrow}\;\textcolor{red}{A6}\;\textcolor{red}{A6}\;\textcolor{red}{A6}\;\textcolor{red}{A1}\;\textcolor{red}{A9}\;\textcolor{red}{A6}\;\textcolor{red}{A9}\;\textcolor{blue}{[0.2376710982]}\\
            & \textcolor{red}{A16}\;\textcolor{black}{\rightarrow}\;\textcolor{red}{A5}\;\textcolor{blue}{[0.2594097969]}\\
            & \textcolor{red}{A16}\;\textcolor{black}{\rightarrow}\;\textcolor{red}{A9}\;\textcolor{red}{A3}\;\textcolor{red}{A4}\;\textcolor{red}{A2}\;\textcolor{red}{A4}\;\textcolor{blue}{[0.1346583572]}\\
            & \textcolor{red}{A16}\;\textcolor{black}{\rightarrow}\;\textcolor{red}{A2}\;\textcolor{red}{A6}\;\textcolor{blue}{[0.0016209134]}\\
            & \textcolor{red}{A16}\;\textcolor{black}{\rightarrow}\;\textcolor{red}{A2}\;\textcolor{red}{A1}\;\textcolor{red}{A7}\;\textcolor{red}{A6}\;\textcolor{red}{A1}\;\textcolor{red}{A6}\;\textcolor{red}{A5}\;\textcolor{red}{A9}\;\textcolor{blue}{[0.0000000681]}\\
            & \textcolor{red}{A16}\;\textcolor{black}{\rightarrow}\;\textcolor{red}{A2}\;\textcolor{red}{A1}\;\textcolor{blue}{[0.0000000000]}\\
            & \textcolor{red}{A16}\;\textcolor{black}{\rightarrow}\;\textcolor{red}{A6}\;\textcolor{red}{A4}\;\textcolor{red}{A2}\;\textcolor{red}{A9}\;\textcolor{red}{A6}\;\textcolor{blue}{[0.0000000000]}\\
            & \textcolor{red}{A15}\;\textcolor{black}{\rightarrow}\;\textcolor{red}{A7}\;\textcolor{red}{A8}\;\textcolor{red}{A5}\;\textcolor{blue}{[0.1633488445]}\\
            & \textcolor{red}{A15}\;\textcolor{black}{\rightarrow}\;\textcolor{red}{A3}\;\textcolor{red}{A3}\;\textcolor{red}{A4}\;\textcolor{red}{A1}\;\textcolor{red}{A8}\;\textcolor{red}{A4}\;\textcolor{red}{A2}\;\textcolor{blue}{[0.2032909216]}\\
            & \textcolor{red}{A15}\;\textcolor{black}{\rightarrow}\;\textcolor{red}{A5}\;\textcolor{red}{A1}\;\textcolor{red}{A3}\;\textcolor{red}{A6}\;\textcolor{red}{A6}\;\textcolor{red}{A9}\;\textcolor{red}{A8}\;\textcolor{red}{A8}\;\textcolor{red}{A3}\;\textcolor{blue}{[0.2376710982]}\\
            & \textcolor{red}{A15}\;\textcolor{black}{\rightarrow}\;\textcolor{red}{A7}\;\textcolor{red}{A1}\;\textcolor{red}{A5}\;\textcolor{red}{A6}\;\textcolor{red}{A4}\;\textcolor{red}{A6}\;\textcolor{red}{A1}\;\textcolor{red}{A4}\;\textcolor{red}{A5}\;\textcolor{blue}{[0.2594097969]}\\
            & \textcolor{red}{A15}\;\textcolor{black}{\rightarrow}\;\textcolor{red}{A4}\;\textcolor{red}{A5}\;\textcolor{red}{A3}\;\textcolor{red}{A6}\;\textcolor{blue}{[0.1346583572]}\\
            & \textcolor{red}{A15}\;\textcolor{black}{\rightarrow}\;\textcolor{red}{A6}\;\textcolor{red}{A8}\;\textcolor{red}{A8}\;\textcolor{red}{A6}\;\textcolor{blue}{[0.0016209134]}\\
            & \textcolor{red}{A15}\;\textcolor{black}{\rightarrow}\;\textcolor{red}{A4}\;\textcolor{red}{A5}\;\textcolor{red}{A4}\;\textcolor{red}{A8}\;\textcolor{red}{A3}\;\textcolor{blue}{[0.0000000681]}\\
            & \textcolor{red}{A15}\;\textcolor{black}{\rightarrow}\;\textcolor{red}{A6}\;\textcolor{red}{A9}\;\textcolor{red}{A2}\;\textcolor{blue}{[0.0000000000]}\\
            & \textcolor{red}{A15}\;\textcolor{black}{\rightarrow}\;\textcolor{red}{A7}\;\textcolor{blue}{[0.0000000000]}\\
            & \textcolor{red}{A14}\;\textcolor{black}{\rightarrow}\;\textcolor{red}{A4}\;\textcolor{red}{A3}\;\textcolor{red}{A3}\;\textcolor{red}{A2}\;\textcolor{red}{A6}\;\textcolor{red}{A2}\;\textcolor{blue}{[0.1633488445]}\\
            & \textcolor{red}{A14}\;\textcolor{black}{\rightarrow}\;\textcolor{red}{A4}\;\textcolor{blue}{[0.2032909216]}\\
            & \textcolor{red}{A14}\;\textcolor{black}{\rightarrow}\;\textcolor{red}{A5}\;\textcolor{red}{A4}\;\textcolor{red}{A3}\;\textcolor{red}{A2}\;\textcolor{red}{A5}\;\textcolor{red}{A2}\;\textcolor{blue}{[0.2376710982]}\\
            & \textcolor{red}{A14}\;\textcolor{black}{\rightarrow}\;\textcolor{red}{A6}\;\textcolor{red}{A7}\;\textcolor{red}{A6}\;\textcolor{red}{A9}\;\textcolor{red}{A6}\;\textcolor{blue}{[0.2594097969]}\\
            & \textcolor{red}{A14}\;\textcolor{black}{\rightarrow}\;\textcolor{red}{A3}\;\textcolor{red}{A2}\;\textcolor{red}{A7}\;\textcolor{red}{A4}\;\textcolor{red}{A8}\;\textcolor{red}{A7}\;\textcolor{blue}{[0.1346583572]}\\
            & \textcolor{red}{A14}\;\textcolor{black}{\rightarrow}\;\textcolor{red}{A9}\;\textcolor{red}{A6}\;\textcolor{red}{A1}\;\textcolor{red}{A4}\;\textcolor{red}{A8}\;\textcolor{red}{A4}\;\textcolor{red}{A5}\;\textcolor{red}{A7}\;\textcolor{red}{A6}\;\textcolor{blue}{[0.0016209134]}\\
            & \textcolor{red}{A14}\;\textcolor{black}{\rightarrow}\;\textcolor{red}{A6}\;\textcolor{red}{A8}\;\textcolor{red}{A4}\;\textcolor{blue}{[0.0000000681]}\\
            & \textcolor{red}{A14}\;\textcolor{black}{\rightarrow}\;\textcolor{red}{A1}\;\textcolor{red}{A6}\;\textcolor{red}{A3}\;\textcolor{blue}{[0.0000000000]}\\
            & \textcolor{red}{A14}\;\textcolor{black}{\rightarrow}\;\textcolor{red}{A6}\;\textcolor{red}{A9}\;\textcolor{red}{A9}\;\textcolor{red}{A6}\;\textcolor{red}{A2}\;\textcolor{red}{A2}\;\textcolor{red}{A3}\;\textcolor{red}{A5}\;\textcolor{blue}{[0.0000000000]}\\
            & \textcolor{red}{A13}\;\textcolor{black}{\rightarrow}\;\textcolor{red}{A7}\;\textcolor{red}{A9}\;\textcolor{red}{A7}\;\textcolor{blue}{[0.1633488445]}\\
            & \textcolor{red}{A13}\;\textcolor{black}{\rightarrow}\;\textcolor{red}{A4}\;\textcolor{red}{A9}\;\textcolor{red}{A6}\;\textcolor{blue}{[0.2032909216]}\\
            & \textcolor{red}{A13}\;\textcolor{black}{\rightarrow}\;\textcolor{red}{A3}\;\textcolor{red}{A7}\;\textcolor{red}{A4}\;\textcolor{red}{A3}\;\textcolor{red}{A4}\;\textcolor{blue}{[0.2376710982]}\\
            & \textcolor{red}{A13}\;\textcolor{black}{\rightarrow}\;\textcolor{red}{A3}\;\textcolor{red}{A4}\;\textcolor{red}{A5}\;\textcolor{red}{A4}\;\textcolor{red}{A5}\;\textcolor{red}{A1}\;\textcolor{red}{A1}\;\textcolor{red}{A7}\;\textcolor{blue}{[0.2594097969]}\\
        \end{align*}
    \end{minipage}%
    \begin{minipage}{0.1\textwidth}
        \begin{align*}
            & \textcolor{red}{A13}\;\textcolor{black}{\rightarrow}\;\textcolor{red}{A3}\;\textcolor{red}{A5}\;\textcolor{red}{A9}\;\textcolor{red}{A9}\;\textcolor{red}{A6}\;\textcolor{red}{A8}\;\textcolor{blue}{[0.1346583572]}\\
            & \textcolor{red}{A13}\;\textcolor{black}{\rightarrow}\;\textcolor{red}{A7}\;\textcolor{red}{A2}\;\textcolor{red}{A1}\;\textcolor{red}{A4}\;\textcolor{red}{A7}\;\textcolor{red}{A1}\;\textcolor{red}{A5}\;\textcolor{red}{A3}\;\textcolor{red}{A9}\;\textcolor{blue}{[0.0016209134]}\\
            & \textcolor{red}{A13}\;\textcolor{black}{\rightarrow}\;\textcolor{red}{A7}\;\textcolor{blue}{[0.0000000681]}\\
            & \textcolor{red}{A13}\;\textcolor{black}{\rightarrow}\;\textcolor{red}{A7}\;\textcolor{red}{A8}\;\textcolor{red}{A8}\;\textcolor{blue}{[0.0000000000]}\\
            & \textcolor{red}{A13}\;\textcolor{black}{\rightarrow}\;\textcolor{red}{A7}\;\textcolor{red}{A4}\;\textcolor{red}{A2}\;\textcolor{red}{A7}\;\textcolor{red}{A8}\;\textcolor{red}{A4}\;\textcolor{red}{A8}\;\textcolor{blue}{[0.0000000000]}\\
            & \textcolor{red}{A12}\;\textcolor{black}{\rightarrow}\;\textcolor{red}{A5}\;\textcolor{red}{A9}\;\textcolor{red}{A6}\;\textcolor{blue}{[0.1633488445]}\\
            & \textcolor{red}{A12}\;\textcolor{black}{\rightarrow}\;\textcolor{red}{A3}\;\textcolor{red}{A6}\;\textcolor{blue}{[0.2032909216]}\\
            & \textcolor{red}{A12}\;\textcolor{black}{\rightarrow}\;\textcolor{red}{A5}\;\textcolor{red}{A9}\;\textcolor{red}{A7}\;\textcolor{red}{A6}\;\textcolor{red}{A9}\;\textcolor{blue}{[0.2376710982]}\\
            & \textcolor{red}{A12}\;\textcolor{black}{\rightarrow}\;\textcolor{red}{A5}\;\textcolor{blue}{[0.2594097969]}\\
            & \textcolor{red}{A12}\;\textcolor{black}{\rightarrow}\;\textcolor{red}{A9}\;\textcolor{red}{A6}\;\textcolor{red}{A1}\;\textcolor{red}{A8}\;\textcolor{blue}{[0.1346583572]}\\
            & \textcolor{red}{A12}\;\textcolor{black}{\rightarrow}\;\textcolor{red}{A7}\;\textcolor{red}{A1}\;\textcolor{red}{A7}\;\textcolor{red}{A2}\;\textcolor{red}{A5}\;\textcolor{red}{A4}\;\textcolor{red}{A4}\;\textcolor{blue}{[0.0016209134]}\\
            & \textcolor{red}{A12}\;\textcolor{black}{\rightarrow}\;\textcolor{red}{A5}\;\textcolor{red}{A1}\;\textcolor{red}{A6}\;\textcolor{red}{A5}\;\textcolor{red}{A3}\;\textcolor{red}{A2}\;\textcolor{red}{A9}\;\textcolor{blue}{[0.0000000681]}\\
            & \textcolor{red}{A12}\;\textcolor{black}{\rightarrow}\;\textcolor{red}{A4}\;\textcolor{red}{A1}\;\textcolor{red}{A6}\;\textcolor{red}{A7}\;\textcolor{red}{A5}\;\textcolor{blue}{[0.0000000000]}\\
            & \textcolor{red}{A12}\;\textcolor{black}{\rightarrow}\;\textcolor{red}{A4}\;\textcolor{red}{A5}\;\textcolor{red}{A8}\;\textcolor{red}{A7}\;\textcolor{red}{A6}\;\textcolor{blue}{[0.0000000000]}\\
            & \textcolor{red}{A11}\;\textcolor{black}{\rightarrow}\;\textcolor{red}{A2}\;\textcolor{red}{A7}\;\textcolor{red}{A9}\;\textcolor{red}{A7}\;\textcolor{blue}{[0.1633488445]}\\
            & \textcolor{red}{A11}\;\textcolor{black}{\rightarrow}\;\textcolor{red}{A5}\;\textcolor{red}{A4}\;\textcolor{red}{A4}\;\textcolor{red}{A7}\;\textcolor{red}{A3}\;\textcolor{red}{A2}\;\textcolor{red}{A1}\;\textcolor{red}{A6}\;\textcolor{blue}{[0.2032909216]}\\
            & \textcolor{red}{A11}\;\textcolor{black}{\rightarrow}\;\textcolor{red}{A2}\;\textcolor{blue}{[0.2376710982]}\\
            & \textcolor{red}{A11}\;\textcolor{black}{\rightarrow}\;\textcolor{red}{A9}\;\textcolor{red}{A3}\;\textcolor{red}{A7}\;\textcolor{red}{A1}\;\textcolor{red}{A5}\;\textcolor{red}{A1}\;\textcolor{red}{A3}\;\textcolor{blue}{[0.2594097969]}\\
            & \textcolor{red}{A11}\;\textcolor{black}{\rightarrow}\;\textcolor{red}{A1}\;\textcolor{blue}{[0.1346583572]}\\
            & \textcolor{red}{A11}\;\textcolor{black}{\rightarrow}\;\textcolor{red}{A3}\;\textcolor{red}{A3}\;\textcolor{red}{A4}\;\textcolor{red}{A2}\;\textcolor{red}{A8}\;\textcolor{red}{A1}\;\textcolor{red}{A6}\;\textcolor{red}{A8}\;\textcolor{blue}{[0.0016209134]}\\
            & \textcolor{red}{A11}\;\textcolor{black}{\rightarrow}\;\textcolor{red}{A6}\;\textcolor{red}{A4}\;\textcolor{red}{A9}\;\textcolor{red}{A5}\;\textcolor{red}{A5}\;\textcolor{red}{A7}\;\textcolor{red}{A5}\;\textcolor{red}{A4}\;\textcolor{blue}{[0.0000000681]}\\
            & \textcolor{red}{A11}\;\textcolor{black}{\rightarrow}\;\textcolor{red}{A9}\;\textcolor{red}{A6}\;\textcolor{red}{A7}\;\textcolor{red}{A7}\;\textcolor{red}{A3}\;\textcolor{red}{A6}\;\textcolor{red}{A5}\;\textcolor{red}{A3}\;\textcolor{blue}{[0.0000000000]}\\
            & \textcolor{red}{A11}\;\textcolor{black}{\rightarrow}\;\textcolor{red}{A3}\;\textcolor{red}{A5}\;\textcolor{red}{A8}\;\textcolor{red}{A4}\;\textcolor{red}{A2}\;\textcolor{red}{A4}\;\textcolor{red}{A5}\;\textcolor{red}{A2}\;\textcolor{red}{A9}\;\textcolor{blue}{[0.0000000000]}\\
            & \textcolor{red}{A10}\;\textcolor{black}{\rightarrow}\;\textcolor{red}{A5}\;\textcolor{blue}{[0.1633488445]}\\
            & \textcolor{red}{A10}\;\textcolor{black}{\rightarrow}\;\textcolor{red}{A8}\;\textcolor{red}{A5}\;\textcolor{blue}{[0.2032909216]}\\
            & \textcolor{red}{A10}\;\textcolor{black}{\rightarrow}\;\textcolor{red}{A2}\;\textcolor{red}{A2}\;\textcolor{blue}{[0.2376710982]}\\
            & \textcolor{red}{A10}\;\textcolor{black}{\rightarrow}\;\textcolor{red}{A5}\;\textcolor{red}{A8}\;\textcolor{red}{A2}\;\textcolor{red}{A8}\;\textcolor{red}{A4}\;\textcolor{blue}{[0.2594097969]}\\
            & \textcolor{red}{A10}\;\textcolor{black}{\rightarrow}\;\textcolor{red}{A4}\;\textcolor{red}{A1}\;\textcolor{red}{A7}\;\textcolor{red}{A5}\;\textcolor{red}{A7}\;\textcolor{red}{A8}\;\textcolor{red}{A1}\;\textcolor{blue}{[0.1346583572]}\\
            & \textcolor{red}{A10}\;\textcolor{black}{\rightarrow}\;\textcolor{red}{A3}\;\textcolor{red}{A3}\;\textcolor{red}{A3}\;\textcolor{red}{A1}\;\textcolor{blue}{[0.0016209134]}\\
            & \textcolor{red}{A10}\;\textcolor{black}{\rightarrow}\;\textcolor{red}{A2}\;\textcolor{red}{A6}\;\textcolor{red}{A5}\;\textcolor{red}{A8}\;\textcolor{red}{A2}\;\textcolor{red}{A1}\;\textcolor{red}{A2}\;\textcolor{red}{A2}\;\textcolor{red}{A4}\;\textcolor{blue}{[0.0000000681]}\\
            & \textcolor{red}{A10}\;\textcolor{black}{\rightarrow}\;\textcolor{red}{A7}\;\textcolor{red}{A8}\;\textcolor{red}{A4}\;\textcolor{red}{A9}\;\textcolor{red}{A9}\;\textcolor{red}{A5}\;\textcolor{red}{A7}\;\textcolor{red}{A8}\;\textcolor{blue}{[0.0000000000]}\\
            & \textcolor{red}{A10}\;\textcolor{black}{\rightarrow}\;\textcolor{red}{A8}\;\textcolor{red}{A2}\;\textcolor{red}{A5}\;\textcolor{blue}{[0.0000000000]}\\
            & \textcolor{red}{A9}\;\textcolor{black}{\rightarrow}\;\textcolor{teal}{8}\;\textcolor{teal}{3}\;\textcolor{teal}{2}\;\textcolor{teal}{12}\;\textcolor{teal}{10}\;\textcolor{teal}{10}\;\textcolor{teal}{13}\;\textcolor{teal}{14}\;\textcolor{blue}{[0.1633488445]}\\
            & \textcolor{red}{A9}\;\textcolor{black}{\rightarrow}\;\textcolor{teal}{2}\;\textcolor{teal}{19}\;\textcolor{blue}{[0.2032909216]}\\
            & \textcolor{red}{A9}\;\textcolor{black}{\rightarrow}\;\textcolor{teal}{13}\;\textcolor{teal}{16}\;\textcolor{teal}{18}\;\textcolor{teal}{14}\;\textcolor{blue}{[0.2376710982]}\\
            & \textcolor{red}{A9}\;\textcolor{black}{\rightarrow}\;\textcolor{teal}{0}\;\textcolor{teal}{0}\;\textcolor{teal}{7}\;\textcolor{blue}{[0.2594097969]}\\
            & \textcolor{red}{A9}\;\textcolor{black}{\rightarrow}\;\textcolor{teal}{18}\;\textcolor{teal}{14}\;\textcolor{blue}{[0.1346583572]}\\
            & \textcolor{red}{A9}\;\textcolor{black}{\rightarrow}\;\textcolor{teal}{7}\;\textcolor{blue}{[0.0016209134]}\\
            & \textcolor{red}{A9}\;\textcolor{black}{\rightarrow}\;\textcolor{teal}{7}\;\textcolor{teal}{17}\;\textcolor{teal}{19}\;\textcolor{teal}{14}\;\textcolor{teal}{6}\;\textcolor{teal}{1}\;\textcolor{teal}{1}\;\textcolor{teal}{6}\;\textcolor{teal}{10}\;\textcolor{blue}{[0.0000000681]}\\
            & \textcolor{red}{A9}\;\textcolor{black}{\rightarrow}\;\textcolor{teal}{11}\;\textcolor{teal}{3}\;\textcolor{teal}{14}\;\textcolor{teal}{19}\;\textcolor{teal}{9}\;\textcolor{teal}{7}\;\textcolor{teal}{19}\;\textcolor{teal}{15}\;\textcolor{blue}{[0.0000000000]}\\
            & \textcolor{red}{A9}\;\textcolor{black}{\rightarrow}\;\textcolor{teal}{6}\;\textcolor{teal}{13}\;\textcolor{teal}{3}\;\textcolor{blue}{[0.0000000000]}\\
            & \textcolor{red}{A8}\;\textcolor{black}{\rightarrow}\;\textcolor{teal}{0}\;\textcolor{teal}{12}\;\textcolor{teal}{9}\;\textcolor{blue}{[0.1633488445]}\\
            & \textcolor{red}{A8}\;\textcolor{black}{\rightarrow}\;\textcolor{teal}{3}\;\textcolor{teal}{0}\;\textcolor{teal}{10}\;\textcolor{teal}{14}\;\textcolor{teal}{9}\;\textcolor{teal}{9}\;\textcolor{teal}{9}\;\textcolor{blue}{[0.2032909216]}\\
            & \textcolor{red}{A8}\;\textcolor{black}{\rightarrow}\;\textcolor{teal}{8}\;\textcolor{teal}{10}\;\textcolor{teal}{14}\;\textcolor{teal}{12}\;\textcolor{teal}{12}\;\textcolor{teal}{12}\;\textcolor{teal}{10}\;\textcolor{teal}{4}\;\textcolor{teal}{6}\;\textcolor{blue}{[0.2376710982]}\\
            & \textcolor{red}{A8}\;\textcolor{black}{\rightarrow}\;\textcolor{teal}{6}\;\textcolor{teal}{15}\;\textcolor{teal}{10}\;\textcolor{teal}{5}\;\textcolor{teal}{12}\;\textcolor{teal}{6}\;\textcolor{teal}{0}\;\textcolor{teal}{5}\;\textcolor{teal}{18}\;\textcolor{blue}{[0.2594097969]}\\
            & \textcolor{red}{A8}\;\textcolor{black}{\rightarrow}\;\textcolor{teal}{1}\;\textcolor{teal}{0}\;\textcolor{teal}{5}\;\textcolor{blue}{[0.1346583572]}\\
            & \textcolor{red}{A8}\;\textcolor{black}{\rightarrow}\;\textcolor{teal}{6}\;\textcolor{teal}{0}\;\textcolor{teal}{12}\;\textcolor{teal}{9}\;\textcolor{teal}{14}\;\textcolor{teal}{7}\;\textcolor{teal}{1}\;\textcolor{teal}{5}\;\textcolor{blue}{[0.0016209134]}\\
            & \textcolor{red}{A8}\;\textcolor{black}{\rightarrow}\;\textcolor{teal}{14}\;\textcolor{teal}{3}\;\textcolor{teal}{8}\;\textcolor{teal}{0}\;\textcolor{teal}{3}\;\textcolor{teal}{14}\;\textcolor{teal}{11}\;\textcolor{blue}{[0.0000000681]}\\
            & \textcolor{red}{A8}\;\textcolor{black}{\rightarrow}\;\textcolor{teal}{2}\;\textcolor{teal}{7}\;\textcolor{teal}{0}\;\textcolor{blue}{[0.0000000000]}\\
            & \textcolor{red}{A8}\;\textcolor{black}{\rightarrow}\;\textcolor{teal}{13}\;\textcolor{blue}{[0.0000000000]}\\
            & \textcolor{red}{A7}\;\textcolor{black}{\rightarrow}\;\textcolor{teal}{12}\;\textcolor{teal}{11}\;\textcolor{teal}{8}\;\textcolor{teal}{11}\;\textcolor{teal}{0}\;\textcolor{teal}{0}\;\textcolor{teal}{0}\;\textcolor{blue}{[0.1633488445]}\\
            & \textcolor{red}{A7}\;\textcolor{black}{\rightarrow}\;\textcolor{teal}{2}\;\textcolor{teal}{19}\;\textcolor{teal}{4}\;\textcolor{teal}{18}\;\textcolor{blue}{[0.2032909216]}\\
            & \textcolor{red}{A7}\;\textcolor{black}{\rightarrow}\;\textcolor{teal}{6}\;\textcolor{teal}{18}\;\textcolor{teal}{14}\;\textcolor{teal}{7}\;\textcolor{teal}{12}\;\textcolor{teal}{19}\;\textcolor{teal}{3}\;\textcolor{blue}{[0.2376710982]}\\
            & \textcolor{red}{A7}\;\textcolor{black}{\rightarrow}\;\textcolor{teal}{0}\;\textcolor{blue}{[0.2594097969]}\\
            & \textcolor{red}{A7}\;\textcolor{black}{\rightarrow}\;\textcolor{teal}{1}\;\textcolor{teal}{17}\;\textcolor{teal}{12}\;\textcolor{teal}{11}\;\textcolor{teal}{0}\;\textcolor{teal}{8}\;\textcolor{blue}{[0.1346583572]}\\
            & \textcolor{red}{A7}\;\textcolor{black}{\rightarrow}\;\textcolor{teal}{13}\;\textcolor{teal}{1}\;\textcolor{teal}{12}\;\textcolor{teal}{9}\;\textcolor{teal}{11}\;\textcolor{teal}{7}\;\textcolor{teal}{4}\;\textcolor{blue}{[0.0016209134]}\\
            & \textcolor{red}{A7}\;\textcolor{black}{\rightarrow}\;\textcolor{teal}{12}\;\textcolor{teal}{17}\;\textcolor{teal}{6}\;\textcolor{teal}{18}\;\textcolor{teal}{12}\;\textcolor{teal}{12}\;\textcolor{teal}{10}\;\textcolor{teal}{0}\;\textcolor{teal}{3}\;\textcolor{blue}{[0.0000000681]}\\
            & \textcolor{red}{A7}\;\textcolor{black}{\rightarrow}\;\textcolor{teal}{19}\;\textcolor{teal}{7}\;\textcolor{teal}{16}\;\textcolor{teal}{4}\;\textcolor{teal}{8}\;\textcolor{teal}{12}\;\textcolor{teal}{14}\;\textcolor{blue}{[0.0000000000]}\\
            & \textcolor{red}{A7}\;\textcolor{black}{\rightarrow}\;\textcolor{teal}{17}\;\textcolor{teal}{3}\;\textcolor{teal}{17}\;\textcolor{teal}{11}\;\textcolor{teal}{8}\;\textcolor{teal}{10}\;\textcolor{blue}{[0.0000000000]}\\
        \end{align*}
    \end{minipage}%
    \begin{minipage}{0.33\textwidth}
        \begin{align*}
            & \textcolor{red}{A6}\;\textcolor{black}{\rightarrow}\;\textcolor{teal}{9}\;\textcolor{teal}{3}\;\textcolor{teal}{16}\;\textcolor{teal}{19}\;\textcolor{blue}{[0.1633488445]}\\
            & \textcolor{red}{A6}\;\textcolor{black}{\rightarrow}\;\textcolor{teal}{18}\;\textcolor{teal}{0}\;\textcolor{teal}{11}\;\textcolor{teal}{8}\;\textcolor{teal}{6}\;\textcolor{blue}{[0.2032909216]}\\
            & \textcolor{red}{A6}\;\textcolor{black}{\rightarrow}\;\textcolor{teal}{1}\;\textcolor{teal}{11}\;\textcolor{teal}{16}\;\textcolor{blue}{[0.2376710982]}\\
            & \textcolor{red}{A6}\;\textcolor{black}{\rightarrow}\;\textcolor{teal}{11}\;\textcolor{teal}{19}\;\textcolor{teal}{12}\;\textcolor{teal}{3}\;\textcolor{teal}{19}\;\textcolor{teal}{4}\;\textcolor{blue}{[0.2594097969]}\\
            & \textcolor{red}{A6}\;\textcolor{black}{\rightarrow}\;\textcolor{teal}{12}\;\textcolor{teal}{0}\;\textcolor{teal}{17}\;\textcolor{teal}{12}\;\textcolor{teal}{3}\;\textcolor{teal}{8}\;\textcolor{teal}{18}\;\textcolor{blue}{[0.1346583572]}\\
            & \textcolor{red}{A6}\;\textcolor{black}{\rightarrow}\;\textcolor{teal}{19}\;\textcolor{teal}{3}\;\textcolor{teal}{16}\;\textcolor{teal}{16}\;\textcolor{teal}{6}\;\textcolor{teal}{11}\;\textcolor{teal}{9}\;\textcolor{teal}{3}\;\textcolor{blue}{[0.0016209134]}\\
            & \textcolor{red}{A6}\;\textcolor{black}{\rightarrow}\;\textcolor{teal}{10}\;\textcolor{teal}{6}\;\textcolor{teal}{7}\;\textcolor{teal}{11}\;\textcolor{teal}{6}\;\textcolor{teal}{2}\;\textcolor{teal}{17}\;\textcolor{teal}{14}\;\textcolor{teal}{12}\;\textcolor{blue}{[0.0000000681]}\\
            & \textcolor{red}{A6}\;\textcolor{black}{\rightarrow}\;\textcolor{teal}{14}\;\textcolor{blue}{[0.0000000000]}\\
            & \textcolor{red}{A6}\;\textcolor{black}{\rightarrow}\;\textcolor{teal}{12}\;\textcolor{teal}{3}\;\textcolor{teal}{3}\;\textcolor{teal}{12}\;\textcolor{teal}{17}\;\textcolor{teal}{19}\;\textcolor{blue}{[0.0000000000]}\\
            & \textcolor{red}{A5}\;\textcolor{black}{\rightarrow}\;\textcolor{teal}{17}\;\textcolor{teal}{13}\;\textcolor{teal}{11}\;\textcolor{teal}{0}\;\textcolor{teal}{7}\;\textcolor{teal}{18}\;\textcolor{blue}{[0.1633488445]}\\
            & \textcolor{red}{A5}\;\textcolor{black}{\rightarrow}\;\textcolor{teal}{18}\;\textcolor{teal}{16}\;\textcolor{teal}{11}\;\textcolor{teal}{7}\;\textcolor{teal}{6}\;\textcolor{teal}{17}\;\textcolor{teal}{3}\;\textcolor{teal}{18}\;\textcolor{blue}{[0.2032909216]}\\
            & \textcolor{red}{A5}\;\textcolor{black}{\rightarrow}\;\textcolor{teal}{19}\;\textcolor{teal}{3}\;\textcolor{blue}{[0.2376710982]}\\
            & \textcolor{red}{A5}\;\textcolor{black}{\rightarrow}\;\textcolor{teal}{4}\;\textcolor{blue}{[0.2594097969]}\\
            & \textcolor{red}{A5}\;\textcolor{black}{\rightarrow}\;\textcolor{teal}{12}\;\textcolor{teal}{4}\;\textcolor{teal}{6}\;\textcolor{teal}{11}\;\textcolor{teal}{16}\;\textcolor{blue}{[0.1346583572]}\\
            & \textcolor{red}{A5}\;\textcolor{black}{\rightarrow}\;\textcolor{teal}{0}\;\textcolor{teal}{6}\;\textcolor{teal}{6}\;\textcolor{teal}{14}\;\textcolor{teal}{12}\;\textcolor{blue}{[0.0016209134]}\\
            & \textcolor{red}{A5}\;\textcolor{black}{\rightarrow}\;\textcolor{teal}{11}\;\textcolor{teal}{14}\;\textcolor{blue}{[0.0000000681]}\\
            & \textcolor{red}{A5}\;\textcolor{black}{\rightarrow}\;\textcolor{teal}{16}\;\textcolor{teal}{8}\;\textcolor{teal}{15}\;\textcolor{teal}{19}\;\textcolor{teal}{19}\;\textcolor{teal}{14}\;\textcolor{blue}{[0.0000000000]}\\
            & \textcolor{red}{A5}\;\textcolor{black}{\rightarrow}\;\textcolor{teal}{9}\;\textcolor{teal}{8}\;\textcolor{teal}{9}\;\textcolor{teal}{9}\;\textcolor{teal}{13}\;\textcolor{blue}{[0.0000000000]}\\
            & \textcolor{red}{A4}\;\textcolor{black}{\rightarrow}\;\textcolor{teal}{10}\;\textcolor{teal}{10}\;\textcolor{teal}{5}\;\textcolor{teal}{8}\;\textcolor{teal}{6}\;\textcolor{blue}{[0.1633488445]}\\
            & \textcolor{red}{A4}\;\textcolor{black}{\rightarrow}\;\textcolor{teal}{7}\;\textcolor{teal}{8}\;\textcolor{teal}{3}\;\textcolor{teal}{14}\;\textcolor{teal}{9}\;\textcolor{blue}{[0.2032909216]}\\
            & \textcolor{red}{A4}\;\textcolor{black}{\rightarrow}\;\textcolor{teal}{1}\;\textcolor{teal}{8}\;\textcolor{teal}{1}\;\textcolor{teal}{1}\;\textcolor{blue}{[0.2376710982]}\\
            & \textcolor{red}{A4}\;\textcolor{black}{\rightarrow}\;\textcolor{teal}{8}\;\textcolor{teal}{8}\;\textcolor{teal}{7}\;\textcolor{teal}{16}\;\textcolor{blue}{[0.2594097969]}\\
            & \textcolor{red}{A4}\;\textcolor{black}{\rightarrow}\;\textcolor{teal}{12}\;\textcolor{teal}{12}\;\textcolor{blue}{[0.1346583572]}\\
            & \textcolor{red}{A4}\;\textcolor{black}{\rightarrow}\;\textcolor{teal}{17}\;\textcolor{teal}{15}\;\textcolor{teal}{16}\;\textcolor{teal}{14}\;\textcolor{teal}{8}\;\textcolor{teal}{14}\;\textcolor{teal}{4}\;\textcolor{teal}{11}\;\textcolor{blue}{[0.0016209134]}\\
            & \textcolor{red}{A4}\;\textcolor{black}{\rightarrow}\;\textcolor{teal}{13}\;\textcolor{teal}{6}\;\textcolor{blue}{[0.0000000681]}\\
            & \textcolor{red}{A4}\;\textcolor{black}{\rightarrow}\;\textcolor{teal}{2}\;\textcolor{teal}{11}\;\textcolor{teal}{18}\;\textcolor{teal}{3}\;\textcolor{teal}{6}\;\textcolor{teal}{1}\;\textcolor{teal}{3}\;\textcolor{teal}{4}\;\textcolor{blue}{[0.0000000000]}\\
            & \textcolor{red}{A4}\;\textcolor{black}{\rightarrow}\;\textcolor{teal}{19}\;\textcolor{teal}{5}\;\textcolor{blue}{[0.0000000000]}\\
            & \textcolor{red}{A3}\;\textcolor{black}{\rightarrow}\;\textcolor{teal}{14}\;\textcolor{teal}{17}\;\textcolor{teal}{0}\;\textcolor{teal}{18}\;\textcolor{blue}{[0.1633488445]}\\
            & \textcolor{red}{A3}\;\textcolor{black}{\rightarrow}\;\textcolor{teal}{19}\;\textcolor{teal}{8}\;\textcolor{blue}{[0.2032909216]}\\
            & \textcolor{red}{A3}\;\textcolor{black}{\rightarrow}\;\textcolor{teal}{18}\;\textcolor{teal}{3}\;\textcolor{teal}{4}\;\textcolor{teal}{12}\;\textcolor{teal}{13}\;\textcolor{teal}{12}\;\textcolor{teal}{11}\;\textcolor{teal}{16}\;\textcolor{teal}{9}\;\textcolor{blue}{[0.2376710982]}\\
            & \textcolor{red}{A3}\;\textcolor{black}{\rightarrow}\;\textcolor{teal}{15}\;\textcolor{teal}{18}\;\textcolor{teal}{6}\;\textcolor{teal}{5}\;\textcolor{teal}{15}\;\textcolor{teal}{9}\;\textcolor{teal}{6}\;\textcolor{blue}{[0.2594097969]}\\
            & \textcolor{red}{A3}\;\textcolor{black}{\rightarrow}\;\textcolor{teal}{10}\;\textcolor{teal}{1}\;\textcolor{teal}{1}\;\textcolor{teal}{2}\;\textcolor{teal}{9}\;\textcolor{teal}{19}\;\textcolor{teal}{9}\;\textcolor{teal}{9}\;\textcolor{blue}{[0.1346583572]}\\
            & \textcolor{red}{A3}\;\textcolor{black}{\rightarrow}\;\textcolor{teal}{17}\;\textcolor{blue}{[0.0016209134]}\\
            & \textcolor{red}{A3}\;\textcolor{black}{\rightarrow}\;\textcolor{teal}{11}\;\textcolor{teal}{1}\;\textcolor{teal}{19}\;\textcolor{blue}{[0.0000000681]}\\
            & \textcolor{red}{A3}\;\textcolor{black}{\rightarrow}\;\textcolor{teal}{1}\;\textcolor{teal}{9}\;\textcolor{teal}{9}\;\textcolor{teal}{8}\;\textcolor{teal}{8}\;\textcolor{teal}{11}\;\textcolor{teal}{10}\;\textcolor{teal}{9}\;\textcolor{blue}{[0.0000000000]}\\
            & \textcolor{red}{A3}\;\textcolor{black}{\rightarrow}\;\textcolor{teal}{14}\;\textcolor{teal}{19}\;\textcolor{teal}{5}\;\textcolor{teal}{9}\;\textcolor{blue}{[0.0000000000]}\\
            & \textcolor{red}{A2}\;\textcolor{black}{\rightarrow}\;\textcolor{teal}{7}\;\textcolor{teal}{17}\;\textcolor{teal}{13}\;\textcolor{blue}{[0.1633488445]}\\
            & \textcolor{red}{A2}\;\textcolor{black}{\rightarrow}\;\textcolor{teal}{4}\;\textcolor{teal}{10}\;\textcolor{teal}{0}\;\textcolor{teal}{17}\;\textcolor{teal}{4}\;\textcolor{teal}{5}\;\textcolor{teal}{8}\;\textcolor{blue}{[0.2032909216]}\\
            & \textcolor{red}{A2}\;\textcolor{black}{\rightarrow}\;\textcolor{teal}{15}\;\textcolor{teal}{9}\;\textcolor{teal}{6}\;\textcolor{blue}{[0.2376710982]}\\
            & \textcolor{red}{A2}\;\textcolor{black}{\rightarrow}\;\textcolor{teal}{8}\;\textcolor{teal}{13}\;\textcolor{teal}{2}\;\textcolor{teal}{10}\;\textcolor{teal}{17}\;\textcolor{teal}{5}\;\textcolor{teal}{6}\;\textcolor{blue}{[0.2594097969]}\\
            & \textcolor{red}{A2}\;\textcolor{black}{\rightarrow}\;\textcolor{teal}{19}\;\textcolor{teal}{3}\;\textcolor{teal}{6}\;\textcolor{blue}{[0.1346583572]}\\
            & \textcolor{red}{A2}\;\textcolor{black}{\rightarrow}\;\textcolor{teal}{2}\;\textcolor{teal}{15}\;\textcolor{teal}{1}\;\textcolor{blue}{[0.0016209134]}\\
            & \textcolor{red}{A2}\;\textcolor{black}{\rightarrow}\;\textcolor{teal}{7}\;\textcolor{teal}{18}\;\textcolor{teal}{18}\;\textcolor{blue}{[0.0000000681]}\\
            & \textcolor{red}{A2}\;\textcolor{black}{\rightarrow}\;\textcolor{teal}{10}\;\textcolor{teal}{17}\;\textcolor{blue}{[0.0000000000]}\\
            & \textcolor{red}{A2}\;\textcolor{black}{\rightarrow}\;\textcolor{teal}{16}\;\textcolor{teal}{3}\;\textcolor{teal}{4}\;\textcolor{teal}{8}\;\textcolor{teal}{0}\;\textcolor{teal}{9}\;\textcolor{blue}{[0.0000000000]}\\
            & \textcolor{red}{A1}\;\textcolor{black}{\rightarrow}\;\textcolor{teal}{6}\;\textcolor{teal}{10}\;\textcolor{teal}{8}\;\textcolor{teal}{1}\;\textcolor{blue}{[0.1633488445]}\\
            & \textcolor{red}{A1}\;\textcolor{black}{\rightarrow}\;\textcolor{teal}{3}\;\textcolor{teal}{18}\;\textcolor{teal}{10}\;\textcolor{blue}{[0.2032909216]}\\
            & \textcolor{red}{A1}\;\textcolor{black}{\rightarrow}\;\textcolor{teal}{19}\;\textcolor{teal}{12}\;\textcolor{teal}{0}\;\textcolor{teal}{0}\;\textcolor{blue}{[0.2376710982]}\\
            & \textcolor{red}{A1}\;\textcolor{black}{\rightarrow}\;\textcolor{teal}{9}\;\textcolor{teal}{13}\;\textcolor{teal}{13}\;\textcolor{teal}{2}\;\textcolor{teal}{15}\;\textcolor{blue}{[0.2594097969]}\\
            & \textcolor{red}{A1}\;\textcolor{black}{\rightarrow}\;\textcolor{teal}{12}\;\textcolor{teal}{7}\;\textcolor{teal}{5}\;\textcolor{blue}{[0.1346583572]}\\
            & \textcolor{red}{A1}\;\textcolor{black}{\rightarrow}\;\textcolor{teal}{8}\;\textcolor{blue}{[0.0016209134]}\\
            & \textcolor{red}{A1}\;\textcolor{black}{\rightarrow}\;\textcolor{teal}{16}\;\textcolor{teal}{13}\;\textcolor{teal}{3}\;\textcolor{teal}{10}\;\textcolor{teal}{6}\;\textcolor{teal}{4}\;\textcolor{teal}{19}\;\textcolor{teal}{19}\;\textcolor{blue}{[0.0000000681]}\\
            & \textcolor{red}{A1}\;\textcolor{black}{\rightarrow}\;\textcolor{teal}{4}\;\textcolor{blue}{[0.0000000000]}\\
            & \textcolor{red}{A1}\;\textcolor{black}{\rightarrow}\;\textcolor{teal}{8}\;\textcolor{blue}{[0.0000000000]}\\
        \end{align*}
    \end{minipage}
    }

    \caption{
        For simulating pre-training, we consider a large skewed grammar $ G_{10} $ with maximum depth $ 3 $, maximum breadth $ 10 $, rules per non-terminals $ 9 $, and $ 20 $ numerical tokens.
    }
    
\end{figure}

In each fine-tuning, we provide the LLM with strings sampled from a probabilistic formal language, with the learning task of generating unseen strings from the same language via syntactic pattern recognition. Underneath, a probabilistic formal language is represented by a \emph{probabilistic formal grammars}, or simply \emph{grammars}~\citep{collins2013probabilistic}.
Specifically, a grammar consists of two sets of symbols called the \emph{non-terminals} and \emph{terminals}, a set of rules to rewrite strings over these symbols that contain at least one nonterminal, also called the \emph{production rules}, and a probability distribution over the production rules. Formally, a probabilistic formal grammar, is defined as a quintuple.  
\begin{align*}
    G = (\textcolor{red}{\mathbf{N}}, \textcolor{teal}{\mathbf{T}}, \textcolor{black}{\mathbf{R}}, \textcolor{red}{S}, \textcolor{blue}{\mathbf{P}})
\end{align*}

where $\textcolor{red}{\mathbf{N}}$ is the set of non-terminals, $\textcolor{teal}{\mathbf{T}}$ is the set of terminals (equivalently, tokens), $\textcolor{black}{\mathbf{R}}$ is the set of production rules, $\textcolor{red}{S} \in \textcolor{red}{\mathbf{N}} $ is the begin non-terminal, and $\textcolor{blue}{\mathbf{P}}$ is the set of probabilities on production rules.

Formal languages are divided into well-known classes based on the \emph{complexity} of the language membership problem, i.e., the \emph{complexity} of the grammars needed to generate them~\citep{chomsky1956three}. In this paper, we use one class of grammars, namely, hierarchical probabilistic context-free grammars (HPCFGs)~\citep{allen2023physics}. Specifically, our experiments are based on teaching LLMs languages represented by HPCFGs. We use HPCFGs because they are simple syntactically and can represent languages that are structurally similar to natural languages~\citep{allen2023physics,shi2022learning}.

\paragraph{Description of Grammars and Identified Languages.} In our experiments, we consider two generic structures for the considered grammars, one adapted from~\citet{allen2023physics}, namely $ G_1, G_2, G_5, G_7, G_9, G_{10} $, and another is proposed by~\citet{ghosh2026fine-tuning}, namely $ G_3, G_4, G_6, G_8 $.

In the first generic structure, such as $ G_1 $, the grammar has $ \mathbf{N} = \{S, A7, A8, \dots, A16\}$ and $\mathbf{T} = \{1, 2, 3, \dots, 9\}$. The grammar has four levels of hierarchy: the non-terminals from top to bottom levels are $\{A16\}$, $\{A13, A14, A15\}$, $\{A10, A11, A12\}$, and $\{A7, A8, A9\}$, followed by terminals $\{1, 2, 3, \dots, 9\}$. Each non-terminal (except the start non-terminal) has two expansion rules, consisting of non-terminals from the immediate lower level. Further, the expansion rules are probabilistic, where the sum of probabilities of all expansion rules from a given non-terminal is $1$.

The second generic structure is inspired by bridging two HPCFGs together, and simulating a long range dependencies within the generated strings. Specifically, the sub-grammar at $ B4 $ and the sub-grammar at $ E4 $ are connected by non-terminal $C1_i$; and $E4$ ends with $T1_j$. Long range dependencies are communicated through $C1_i$ and $T1_j$, by enforcing $ i = j $ at each expansion of $ S5 $.

In all cases, ${G}_i$ produces a probabilistic context free language $L_i$. Figure~\ref{fig:length_distribution} denotes the length distribution of different languages, and Figure~\ref{fig:representaive_string} demonstrates how hierarchical non-terminals are applied in different positions in the representative strings.

{
\color{black}

\paragraph{Sampling Strings from a Formal Language.}
Given a language $\lang$ generated by a HPCFG, we first need to obtain \emph{training} samples, i.e., set of i.i.d.\ samples of strings from $\lang$.
To \emph{sample a string from the language},
we start from a special string in the grammar containing a single, distinguished nonterminal called the ``start'' or ``root'' symbol, and apply the production rules to rewrite the string repeatedly. 
If several rules can be used to rewrite the string at any stage, we sample one such rule from the probability distribution over the rules and apply it. 
We stop when we obtain a string containing terminal tokens only. This string is a sample drawn from the language.
We can repeat this process to draw any number of i.i.d.\ samples from the language.

In our experiments, we aim to split the sampled strings into training and test sets, which are disjoint but have a similar probability distribution over string occurrences. To realize this goal, we first sample a finite number of strings from the language. Then, we remove a random string from the finite set along with all its occurrences and put them into the training set. In the next iteration, we put the next random string with all its occurrences into the test set. The process repeats until the initial finite set is exhausted. Thus, we empirically try to obtain a similar probability distribution over string occurrences in the training and test sets.

\begin{figure}
    \centering

    \subfloat[$L_1$ (also $L_2$)]{
    \includegraphics[scale=0.55]{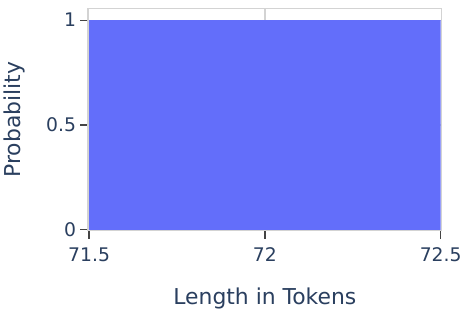}
    }
    \subfloat[$L_3$ (also $L_4$)]{
    \includegraphics[scale=0.55]{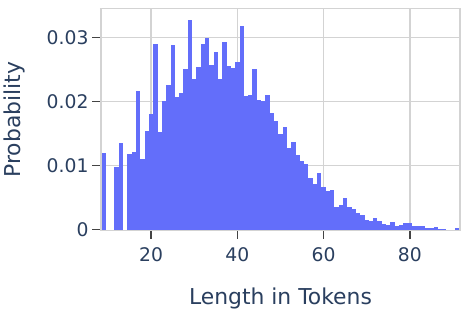}
    }
    \subfloat[$L_5$ (also $L_7$)]{
    \includegraphics[scale=0.55]{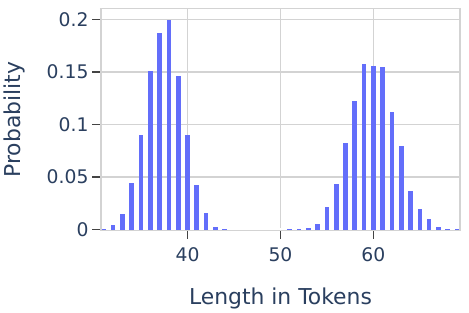}
    }

    \subfloat[$L_6$ (also $L_8$)]{
    \includegraphics[scale=0.55]{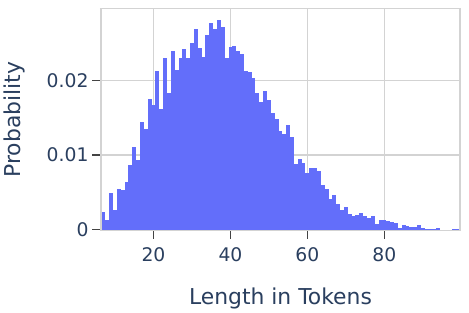}
    }
    \subfloat[$L_9$]{
    \includegraphics[scale=0.55]{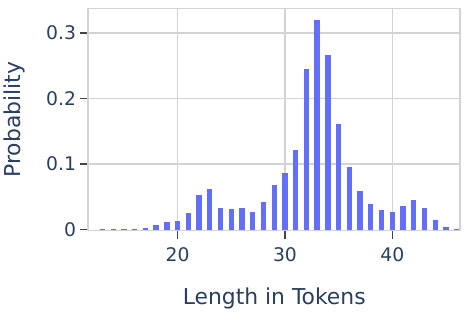}
    }
    \subfloat[$L_{10}$]{
    \includegraphics[scale=0.55]{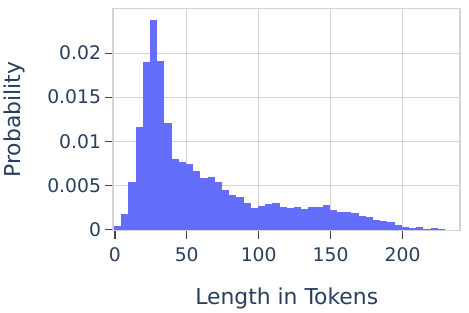}
    }

    \caption{Length distribution of strings sampled from multiple formal languages. The length distribution is based on $10000$ sampled strings per language.}
    \label{fig:length_distribution}
\end{figure}

\begin{figure*}
    \centering

    \subfloat[$ L_1 $]{
    \includegraphics[trim={0 0 0 2.5cm},clip,scale=0.5]{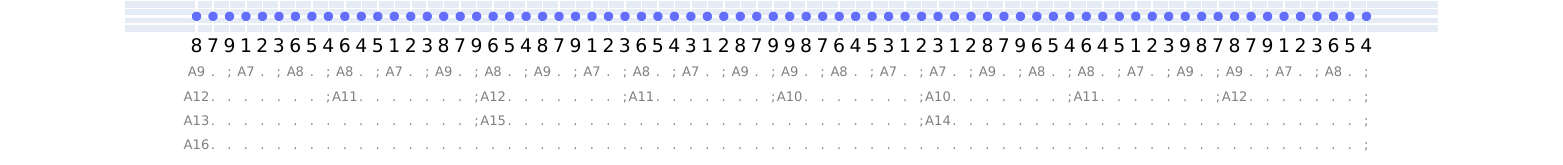}
    }

    \subfloat[$ L_2 $]{
    \includegraphics[trim={0 0 0 2.5cm},clip,scale=0.5]{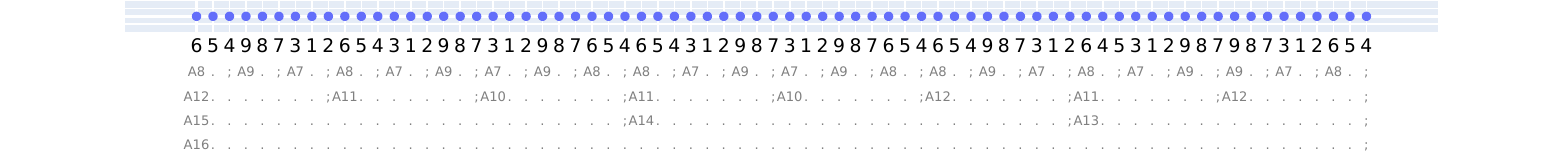}
    }

    \subfloat[$ L_3 $]{
    \includegraphics[trim={0 0 0 2.5cm},clip,scale=0.5]{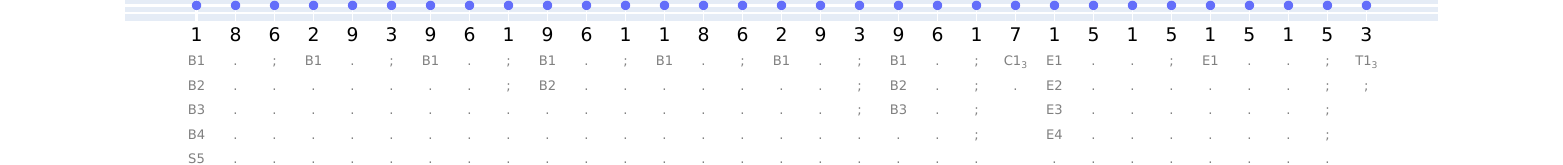}
    }

    \subfloat[$ L_4 $]{
    \includegraphics[trim={0 0 0 2.5cm},clip,scale=0.5]{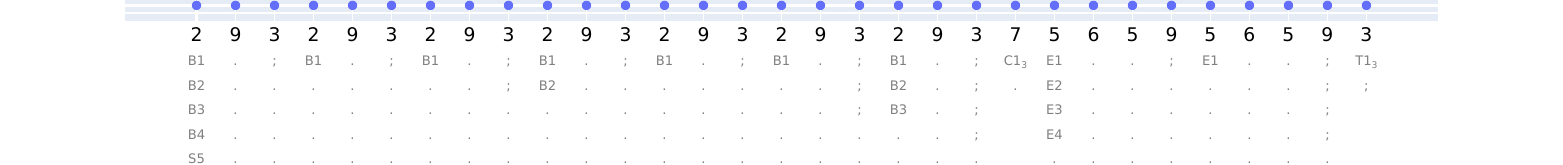}
    }

    \subfloat[$ L_5 $]{
    \includegraphics[trim={0 0 0 2.5cm},clip,scale=0.5]{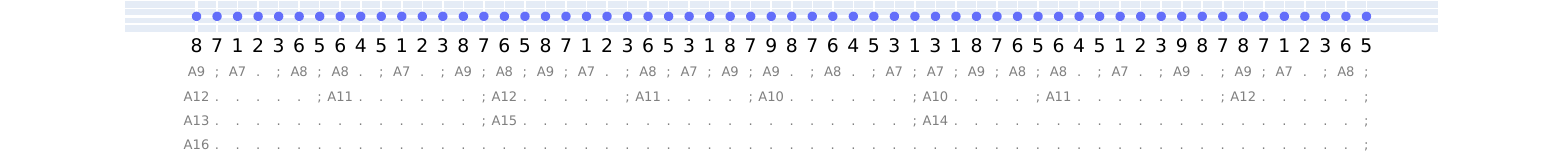}
    }

    \caption{Representative strings from different languages, annotated with non-terminals applied in different positions by the respective hierarchical grammar.}

    \label{fig:representaive_string}
    
\end{figure*}

\begin{figure*}
    \centering

    \subfloat[$ L_6 $]{
    \includegraphics[trim={0 0 0 2.5cm},clip,scale=0.5]{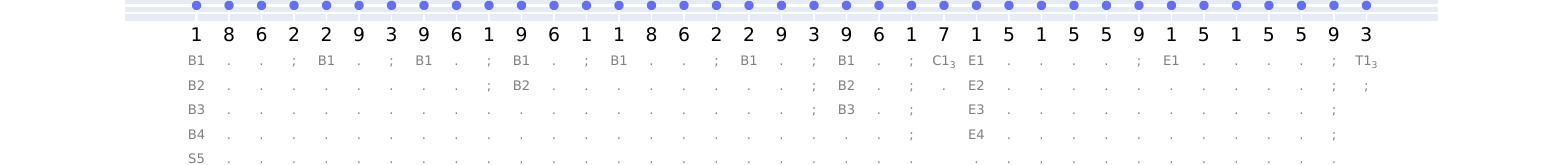}
    }

    \subfloat[$ L_7 $]{
    \includegraphics[trim={0 0 0 2.5cm},clip,scale=0.5]{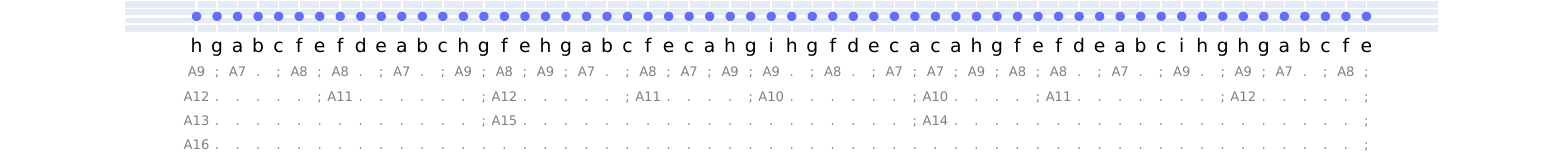}
    }

    \subfloat[$ L_8 $]{
    \includegraphics[trim={0 0 0 2.5cm},clip,scale=0.5]{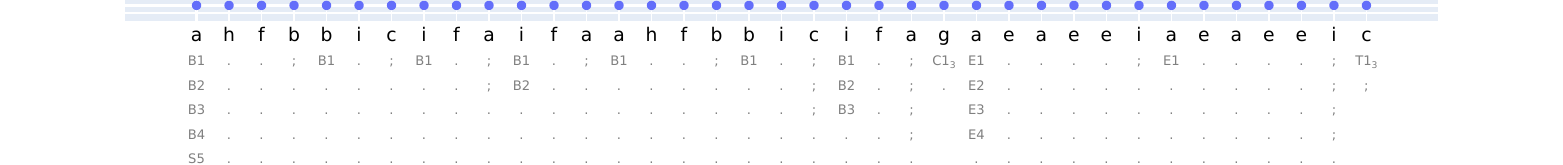}
    }

    \subfloat[$ L_9 $]{
    \includegraphics[trim={0 0 0 2.5cm},clip,scale=0.5]{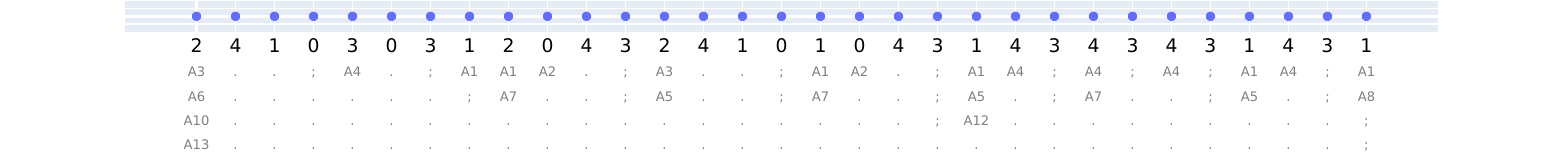}
    }

    \subfloat[$ L_{10} $]{
    \includegraphics[trim={0 0 0 2.5cm},clip,scale=0.5]{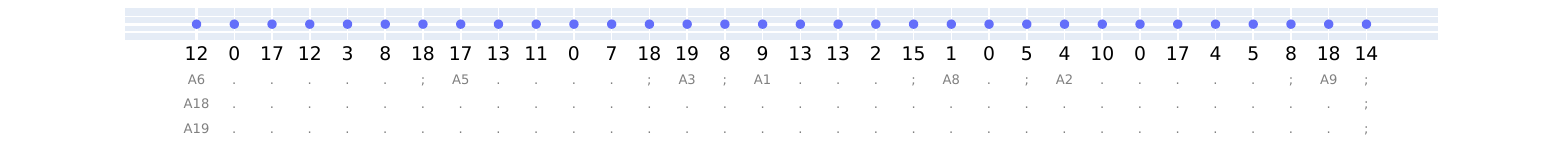}
    }

    \caption{Representative strings from different languages, annotated with non-terminals applied in different positions by the respective hierarchical grammar.}

    \label{fig:representaive_string_2}
    
\end{figure*}

\clearpage
\section{Approximation of Contextual Loss}
\label{sec:validation_ctx}

We validate the frequency-matched approximation of the lowest contextual loss (Section~\ref{sec:definitions}) across three LLMs (Mistral-$7$B, Pythia-$1$B, Qwen-$2.5$-$1.5$B), three formal languages ($L_2$, $L_4$, $L_9$), and four training sizes ($16$, $64$, $256$, $1024$). For each combination, we sample three strings from frequency groups (high, medium, low) and repeat over three seeds of random training data.

\begin{figure}[h]
  \centering
  \subfloat[]{\includegraphics[scale=0.5]{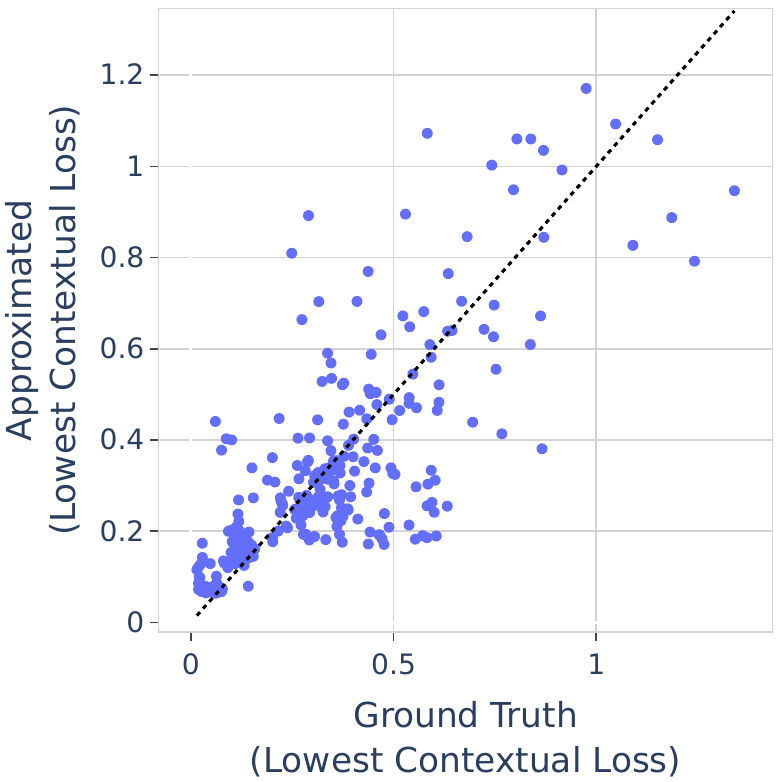}
  \label{fig:scatter_lowest_contextual_loss}}
  \subfloat[]{\includegraphics[scale=0.5]{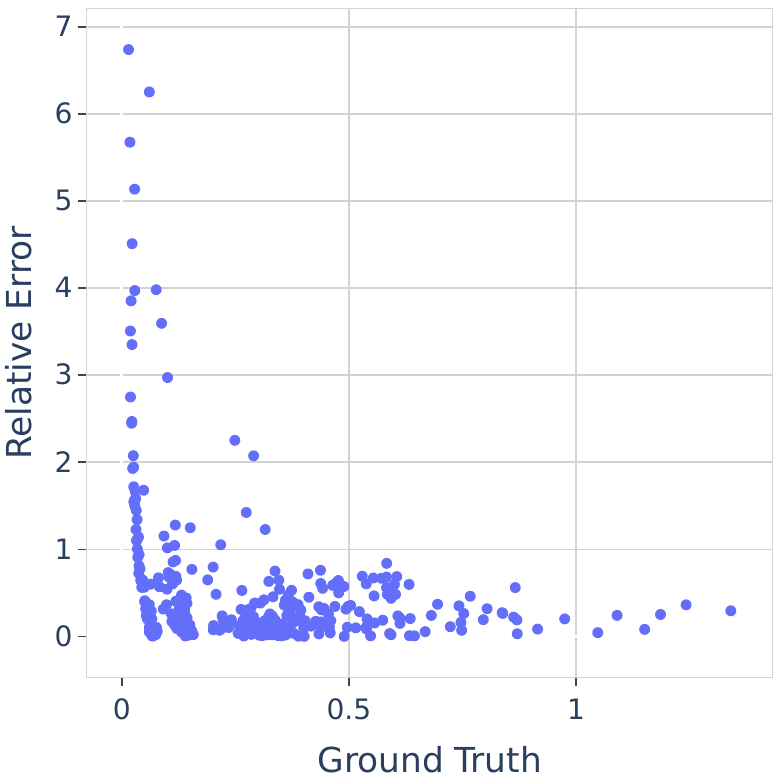}
  \label{fig:relative_error_vs_ground_truth}}

  \subfloat[]{\includegraphics[scale=0.5]{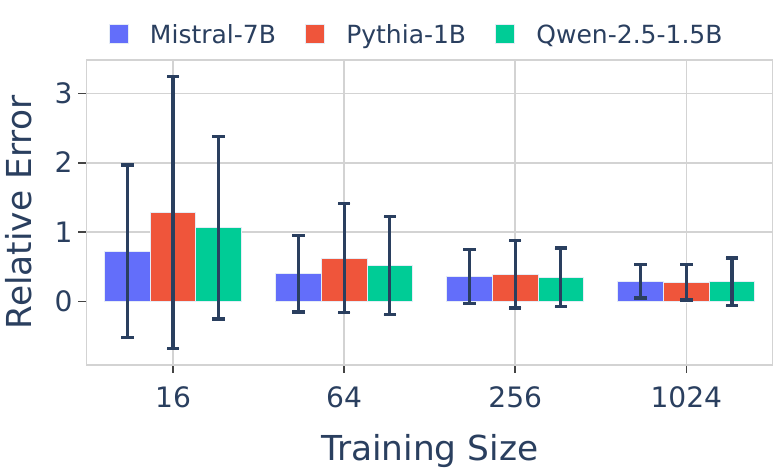}
  \label{fig:relative_error_by_training_size}}

  \caption{\textbf{Validation of the Approximation of the Lowest Contextual Loss.} (a) Ground truth vs.\ approximated lowest contextual loss (Figure~\ref{fig:scatter_lowest_contextual_loss}): points cluster tightly around the diagonal, indicating strong agreement between the two. (b) Relative error (ratio of absolute error to ground truth) vs.\ ground truth (Figure~\ref{fig:relative_error_vs_ground_truth}): relative error is higher when the ground-truth loss is small, a numerical artifact of dividing a similar absolute error by a smaller denominator. (c) Relative error by training size, grouped by model (Figure~\ref{fig:relative_error_by_training_size}): the approximation error decreases as training size increases, and this trend is consistent across models.
  }
\end{figure}

\clearpage
\section{Order and Score of Memorization}
\label{sec:app_experiments}

In Figure~\ref{fig:degree_of_memorization}, we demonstrate the memorization scores of strings, corresponding to Figure~\ref{fig:begin_of_memorization}, across multiple memorization measures. In all measures, the  memorization score usually increases with epochs. As we theoretically demonstrate, contextual memorization score provides a lower bound of counterfactual memorization score.

\begin{figure*}[h]
    \centering
    
    \subfloat[Recollection ($ 0.2 $)]{
        \includegraphics[scale=0.4]{memorization_figures/counterfactual_memorization/rec_mistral-7b_pcfg_cfg3b_eq_len_skewed_prob_counterfactual_16_loss_cf_mem_truncated.pdf}
    }\hfil
    \subfloat[Counterfactual]{
        \includegraphics[scale=0.4]{memorization_figures/counterfactual_memorization/cf_mistral-7b_pcfg_cfg3b_eq_len_skewed_prob_counterfactual_16_loss_cf_mem_truncated.pdf}
    }\hfil
    \subfloat[Contextual]{
        \includegraphics[scale=0.4]{memorization_figures/counterfactual_memorization/ctx_mistral-7b_pcfg_cfg3b_eq_len_skewed_prob_counterfactual_16_loss_cf_mem_truncated.pdf}
    }

    \caption{Start of memorization of selected strings in Language $ L_2 $.}
    
    \subfloat[Recollection ($ 0.2 $)]{
        \includegraphics[scale=0.4]{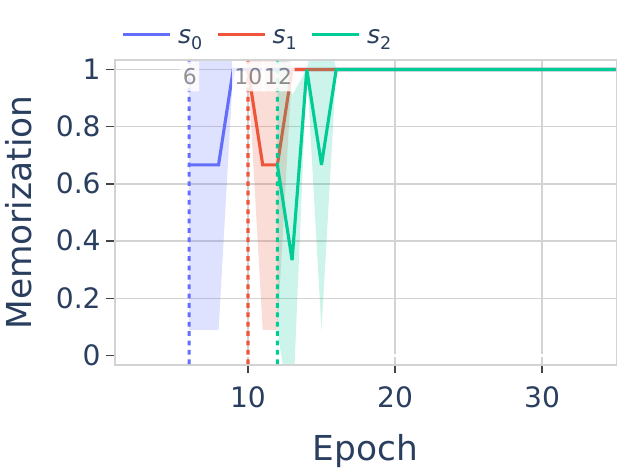}
        \label{fig:recollection_based_memorization_degree}
        
    }\hfil
    \subfloat[Counterfactual]{
        \includegraphics[scale=0.4]{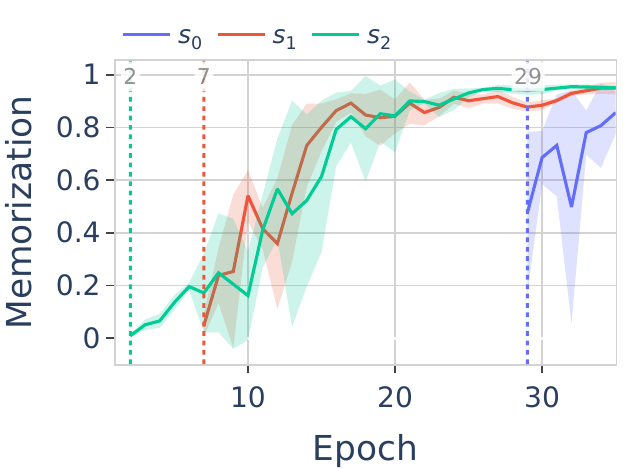}
        \label{fig:counterfactual_memorization_degree}
    }\hfil 
    \subfloat[Contextual]{
        \includegraphics[scale=0.4]{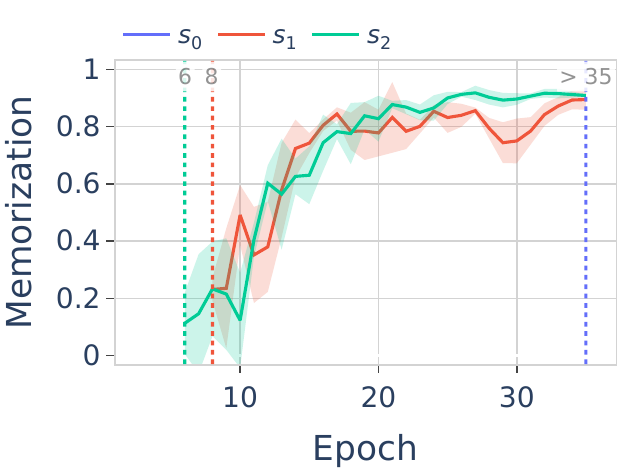}
        \label{fig:contextual_memorization_degree}
    }

    \caption{Memorization score of strings in language $ L_2 $, respective to Figure~\ref{fig:begin_of_memorization}. In different strings, memorization score usually increases with epochs, with contextual memorization providing a lower bound of counterfactual memorization.
}
    \label{fig:degree_of_memorization}
\end{figure*}

\begin{figure*}
    \centering
    
    \subfloat[Recollection ($ 0.2 $)]{
        \includegraphics[scale=0.4]{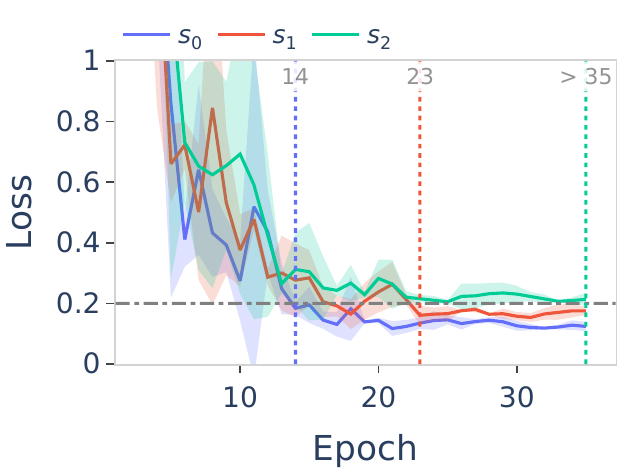}
    }\hfil
    \subfloat[Counterfactual]{
        \includegraphics[scale=0.4]{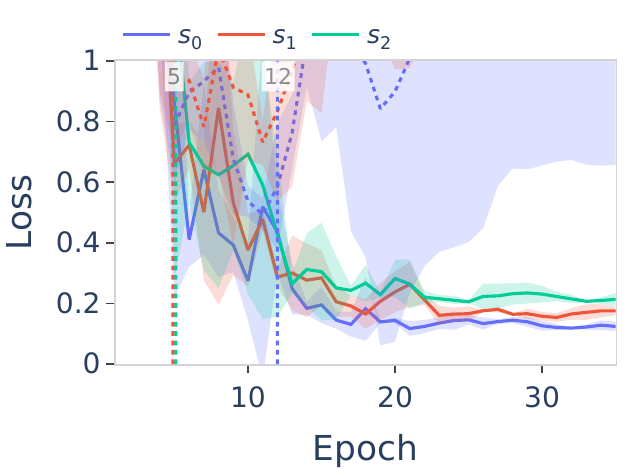}
    }\hfil
    \subfloat[Contextual]{
        \includegraphics[scale=0.4]{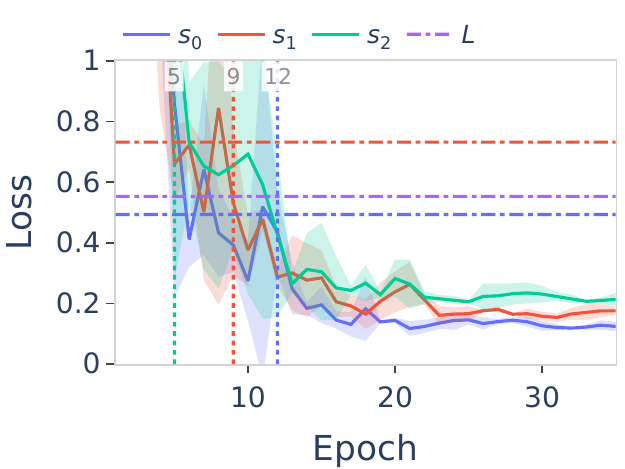}
    }

    \caption{Start of memorization of selected strings in language $ L_4 $ (specifically, a modified version of $ L_4 $ as explained below). The observation is consistent with language $ L_2 $, as shown in Figure~\ref{fig:begin_of_memorization}, where frequency of strings correlates with the beginning of recollection-based memorization. Similarly, frequency often inversely correlates with counterfactual and contextual memorization, with an exception that both $ s_1 $ and $ s_2 $ are memorized at the same epoch in the counterfactual memorization. Thus, regardless of whether correlation or inverse correlation exists \textit{strongly} between string frequency and the order of memorization, a more consistent observation is that memorization measures disagree with each other when applied to the same training dynamic on identical strings.\\~\\
    In this experiment, to better differentiate the strings $ s_0, s_1, s_2 $ based on frequency, we modify $ L_4 $ to be even more skewed. We apply high probability to one random production rule in each non-terminal in all levels, beyond the lowest level non-terminals in $ L_4 $, as shown in Figure~\ref{fig:grammar_g3_g4}.}
    \label{fig:begin_of_memorization_l_4}

    \subfloat[Recollection ($ 0.2 $)]{
        \includegraphics[scale=0.4]{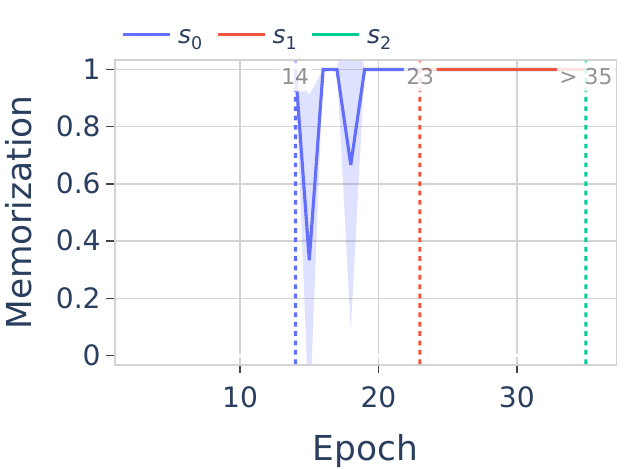}
        
    }\hfil
    \subfloat[Counterfactual]{
        \includegraphics[scale=0.4]{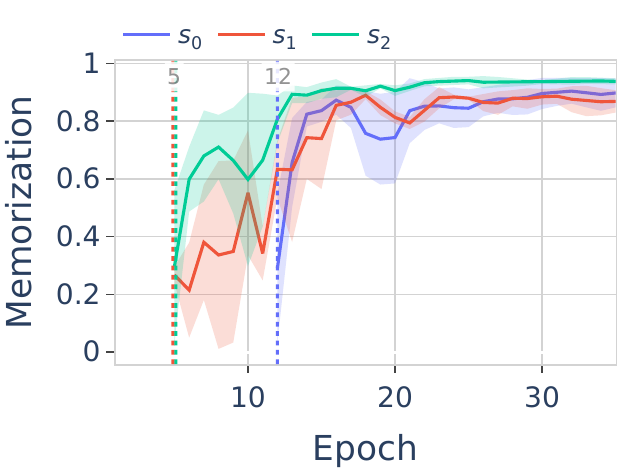}
    
    }\hfil
    \subfloat[Contextual]{
        \includegraphics[scale=0.4]{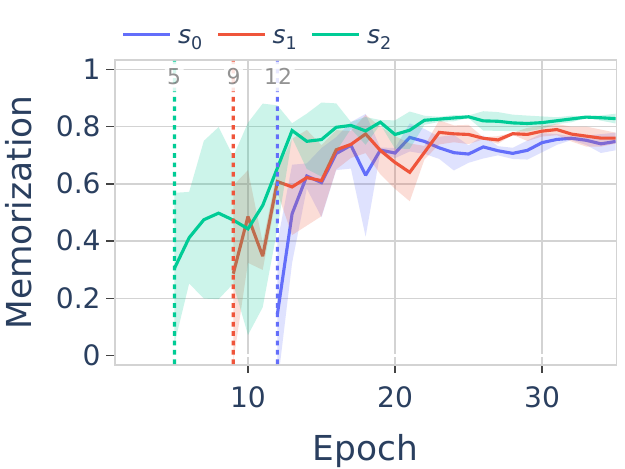}
        
    }

    \caption{Memorization score of strings in language $ L_4 $.
    }

\end{figure*}

\clearpage

\subsection*{Results on Natural Language Datasets}
\label{sec:nlp_results_appendix}

\begin{figure}[h]
    \centering
    
    \subfloat[Recollection ($ 0.2 $)]{
        \includegraphics[scale=0.4]{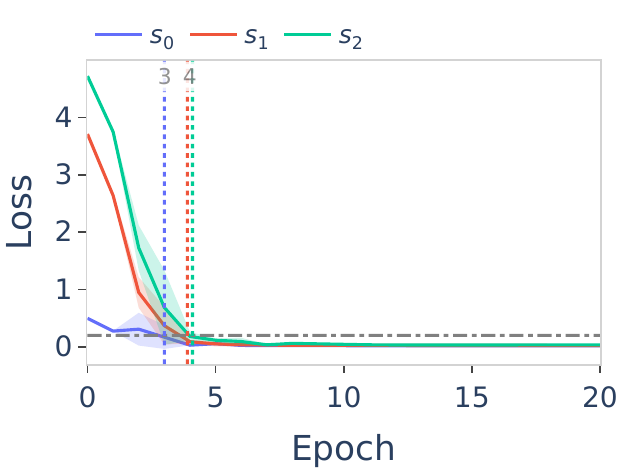}
        \label{fig:recollection_based_memorization_nlp}
    }\hfil
    \subfloat[Counterfactual]{
        \includegraphics[scale=0.4]{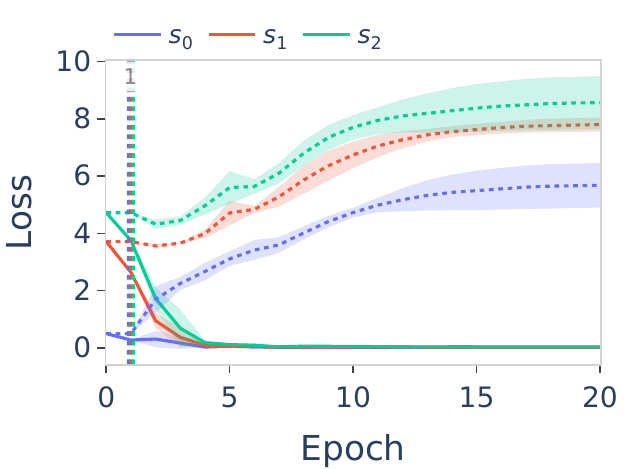}
        \label{fig:counterfactual_memorization_nlp}
    }\hfil
    \subfloat[Contextual]{
        \includegraphics[scale=0.4]{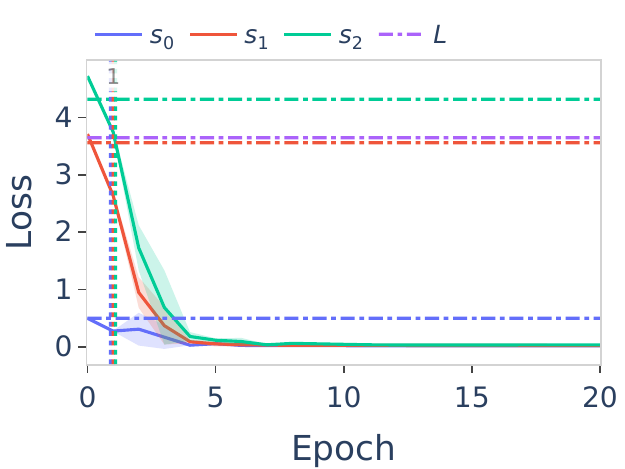}
        \label{fig:contextual_memorization_nlp}
    }

    \caption{
    \textbf{Disagreement Among Memorization Measures on Natural Language.}
    We select three training strings $s_0$, $s_1$, $s_2$ from the Tiny Shakespeare dataset~\cite{karpathy2015charrnn} with low, medium, and high loss under the base model (before any fine-tuning), respectively. Similar to our results in formal language, different memorization measures disagree on when a string in natural language begins to be memorized, shown using vertical lines. More importantly in Figure~\ref{fig:contextual_memorization_nlp}, \textbf{strings have different values of optimal contextual loss, making the case for contextual memorization.}
   }
   \label{fig:memorization_measures_nlp}

   \subfloat[Recollection ($ 0.2 $)]{
        \includegraphics[scale=0.4]{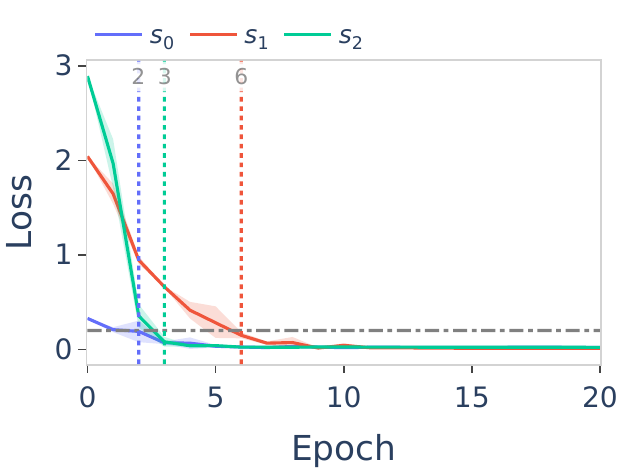}
    }\hfil
    \subfloat[Counterfactual]{
        \includegraphics[scale=0.4]{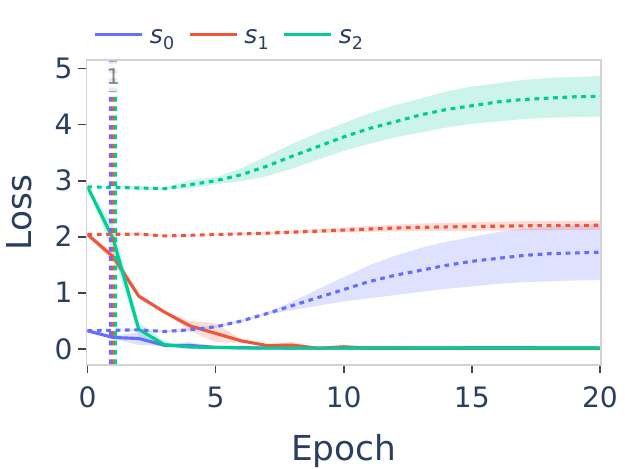}
    }\hfil
    \subfloat[Contextual]{
        \includegraphics[scale=0.4]{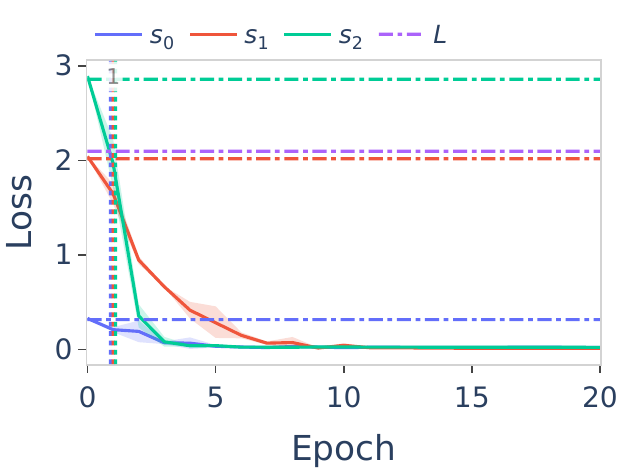}
    }

    \caption{\textbf{FineWeb-Edu}~\cite{lozhkov2024fineweb-edu} (English). Same results of Figure~\ref{fig:memorization_measures_nlp} as observed in the English language.}
    \label{fig:nlp_fineweb_edu}

    \subfloat[Recollection ($ 0.2 $)]{
        \includegraphics[scale=0.4]{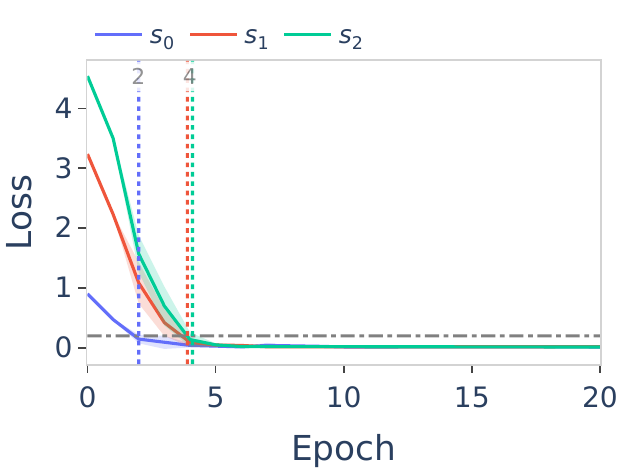}
    }\hfil
    \subfloat[Counterfactual]{
        \includegraphics[scale=0.4]{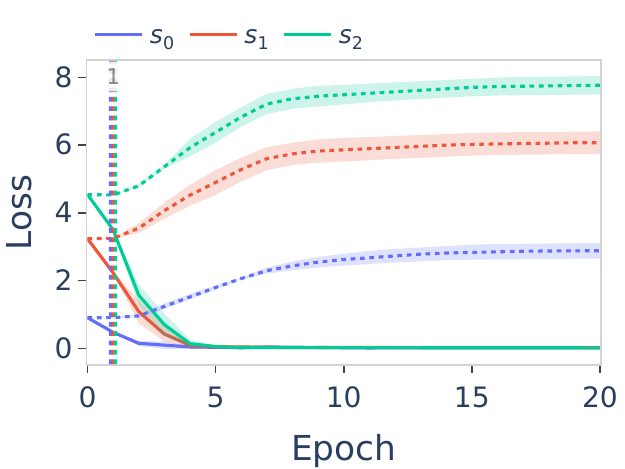}
    }\hfil
    \subfloat[Contextual]{
        \includegraphics[scale=0.4]{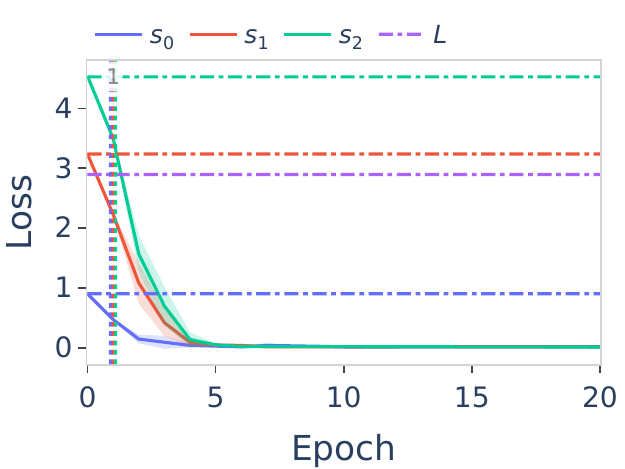}
    }

    \caption{\textbf{German}, drawn from FineWeb2~\cite{penedo2025fineweb2}. Same results of Figure~\ref{fig:memorization_measures_nlp} as observed in the German language.}
    \label{fig:nlp_german}

\end{figure}

\begin{figure}
    \centering

    \subfloat[Recollection ($ 0.2 $)]{
        \includegraphics[scale=0.4]{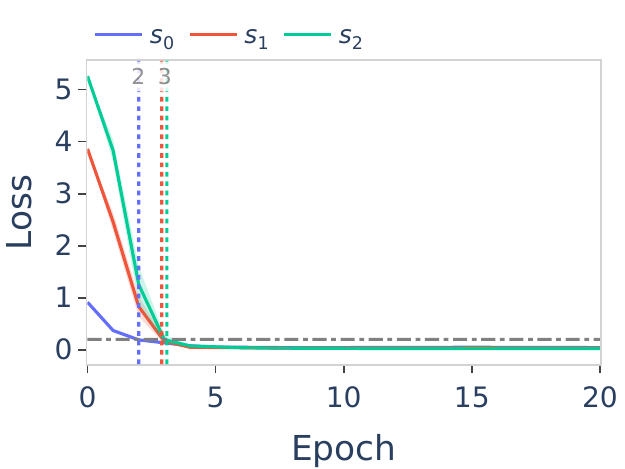}
    }\hfil
    \subfloat[Counterfactual]{
        \includegraphics[scale=0.4]{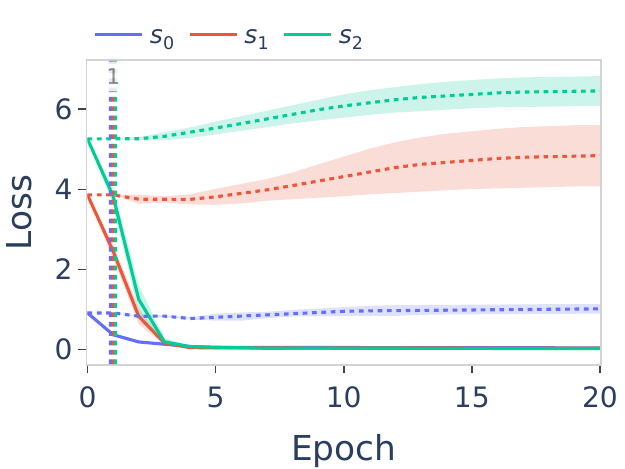}
    }\hfil
    \subfloat[Contextual]{
        \includegraphics[scale=0.4]{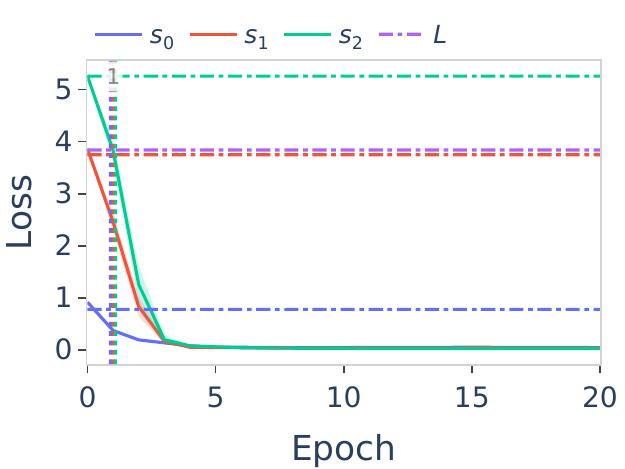}
    }

    \caption{\textbf{Chinese}, drawn from FineWeb2~\cite{penedo2025fineweb2}. Same results of Figure~\ref{fig:memorization_measures_nlp} as observed in the Chinese language.}
    \label{fig:nlp_chinese}

    \subfloat[Recollection ($ 0.2 $)]{
        \includegraphics[scale=0.4]{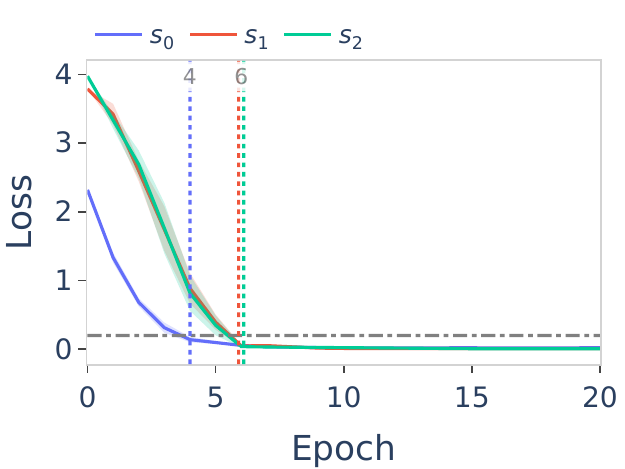}
    }\hfil
    \subfloat[Counterfactual]{
        \includegraphics[scale=0.4]{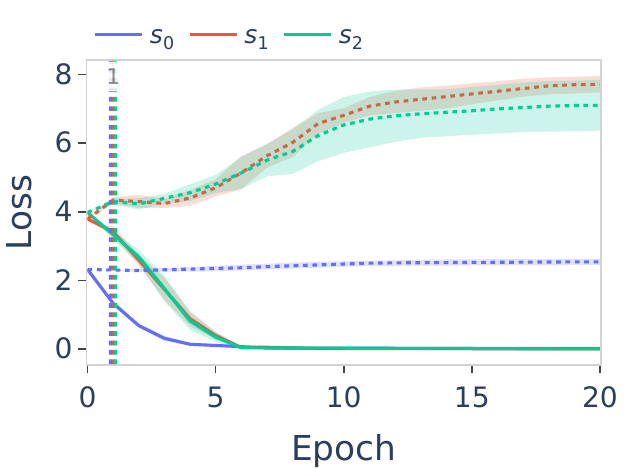}
    }\hfil
    \subfloat[Contextual]{
        \includegraphics[scale=0.4]{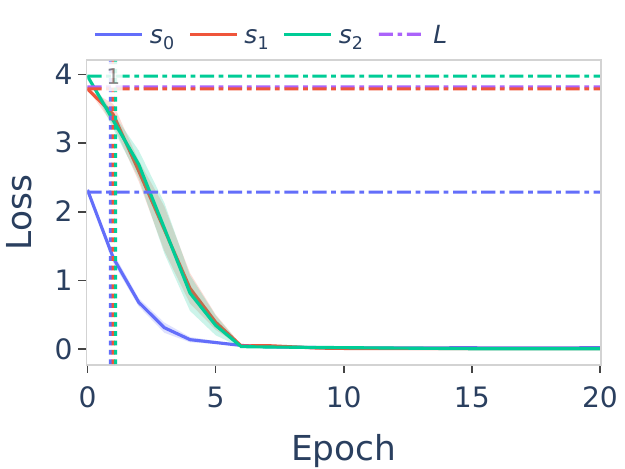}
    }

    \caption{\textbf{Swahili}, drawn from FineWeb2~\cite{penedo2025fineweb2}. Same results of Figure~\ref{fig:memorization_measures_nlp} as observed in the Swahili language.}
    \label{fig:nlp_swahili}

    \subfloat[Recollection ($ 0.2 $)]{
        \includegraphics[scale=0.4]{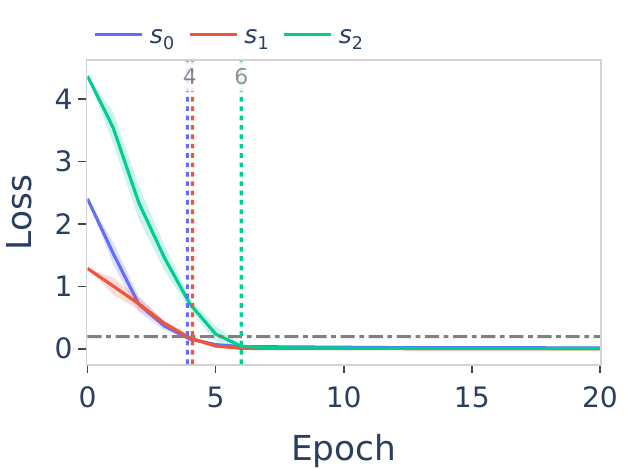}
    }\hfil
    \subfloat[Counterfactual]{
        \includegraphics[scale=0.4]{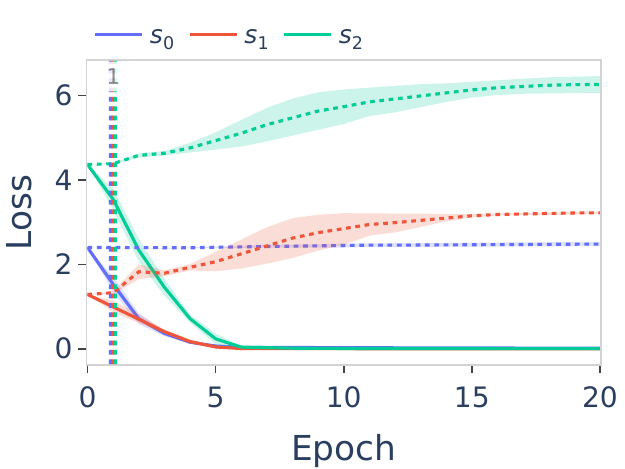}
    }\hfil
    \subfloat[Contextual]{
        \includegraphics[scale=0.4]{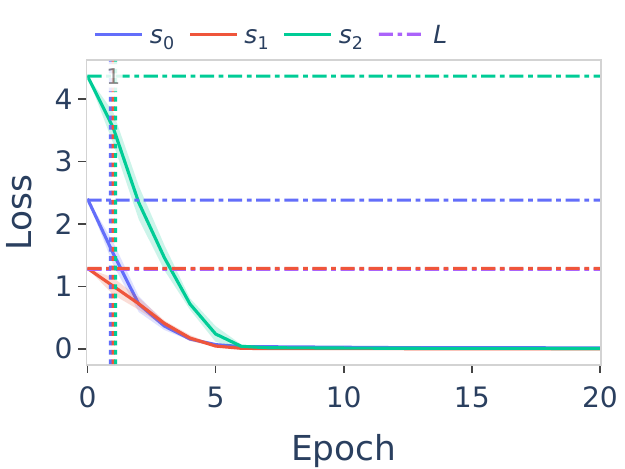}
    }

    \caption{\textbf{Bengali}, drawn from FineWeb2~\cite{penedo2025fineweb2}. Same results of Figure~\ref{fig:memorization_measures_nlp} as observed in the Bengali language.}
    \label{fig:nlp_bengali}

\end{figure}

\clearpage
\section{Learning and Memorization Trade-offs}
\label{sec:learning_memorization_tradeoffs_appendix}

\begin{figure*}[h]
    \centering
    \captionsetup[subfigure]{justification=centering}

    \subfloat[Mistral-$7$B]{
        \includegraphics[scale=0.4]{memorization_figures/operationalization/language_memorization_pcfg_cfg3b_eq_len_uniform_prob_mistral-7b_256_loss_memorization.pdf}
        
    }
    \subfloat[Pythia-$6.9$B]{
        \includegraphics[scale=0.4]{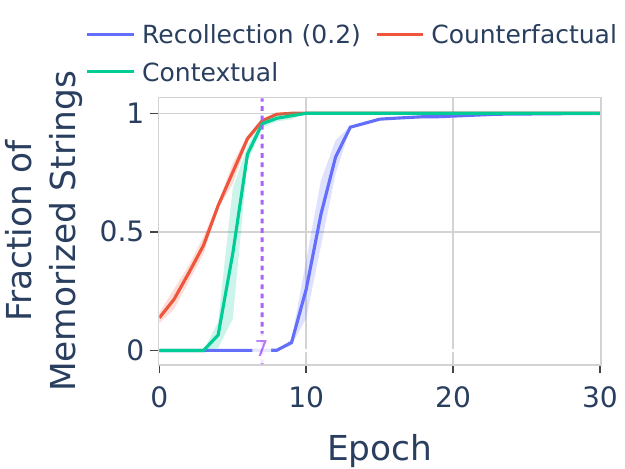}
        
    }
    \subfloat[Qwen-$2.5$B]{
        \includegraphics[scale=0.4]{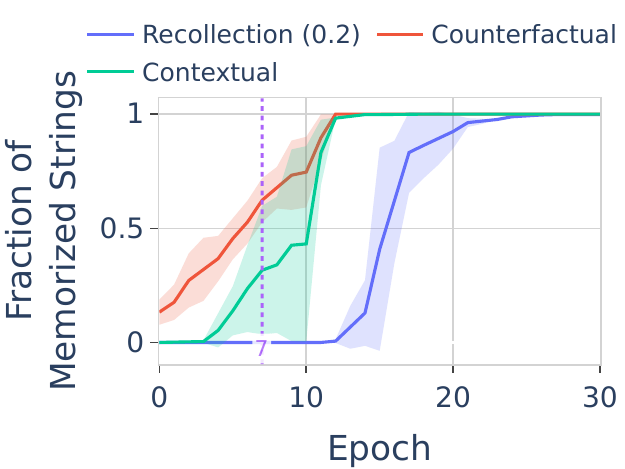}
        
    }

    \subfloat[Mistral-$7$B]{
        \includegraphics[scale=0.4]{memorization_figures/operationalization/language_memorization_weighted_pcfg_cfg3b_eq_len_uniform_prob_mistral-7b_256_loss_memorization.pdf}
        
    }
    \subfloat[Pythia-$6.9$B]{
        \includegraphics[scale=0.4]{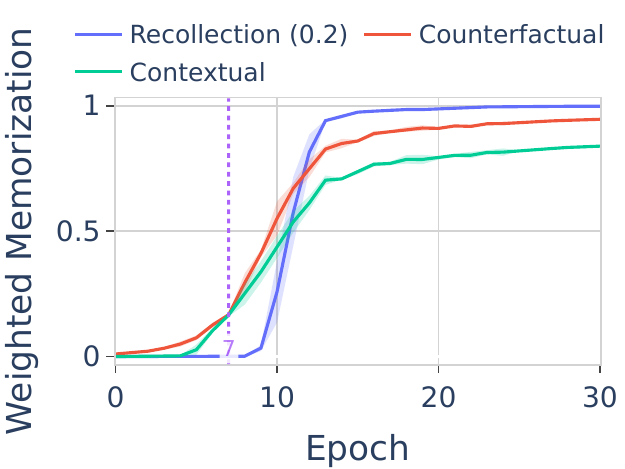}
        
    }
    \subfloat[Qwen-$2.5$B]{
        \includegraphics[scale=0.4]{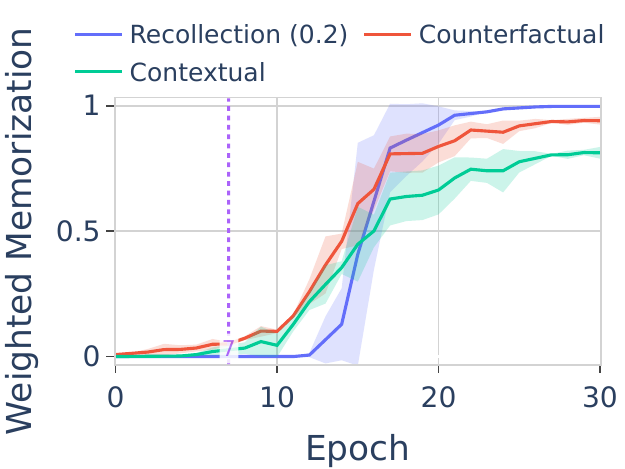}
        
    }

    \caption{Memorization of training strings in languages of different entropy across different memorization measures. Results are for language $ L_1 $, which is a high entropy language.
    }

    \subfloat[Mistral-$7$B]{
        \includegraphics[scale=0.4]{memorization_figures/operationalization/language_memorization_pcfg_cfg3b_eq_len_skewed_prob_mistral-7b_256_loss_memorization.pdf}
        
    }
    \subfloat[Pythia-$6.9$B]{
        \includegraphics[scale=0.4]{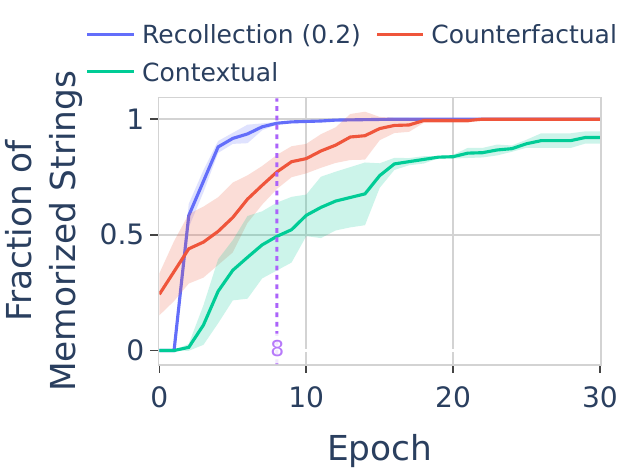}
        
    }
    \subfloat[Qwen-$2.5$B]{
        \includegraphics[scale=0.4]{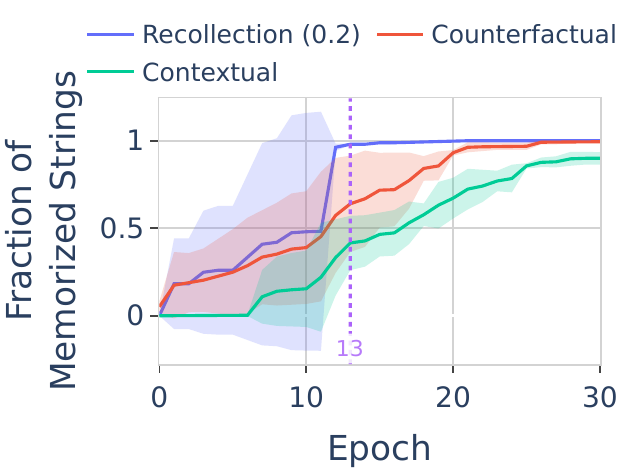}
        
    }

    \subfloat[Mistral-$7$B]{
        \includegraphics[scale=0.4]{memorization_figures/operationalization/language_memorization_weighted_pcfg_cfg3b_eq_len_skewed_prob_mistral-7b_256_loss_memorization.pdf}
        
    }
    \subfloat[Pythia-$6.9$B]{
        \includegraphics[scale=0.4]{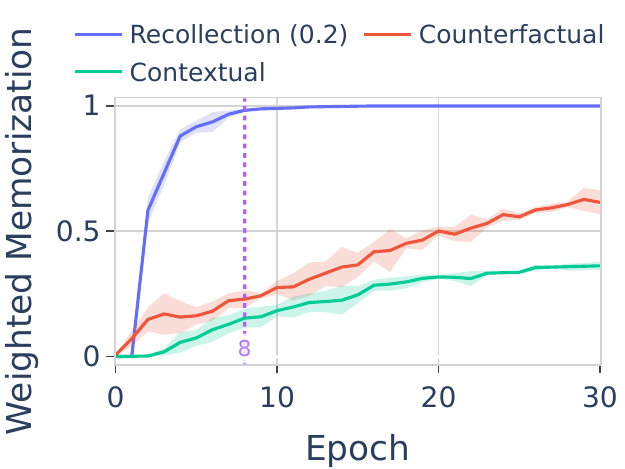}
        
    }
    \subfloat[Qwen-$2.5$B]{
        \includegraphics[scale=0.4]{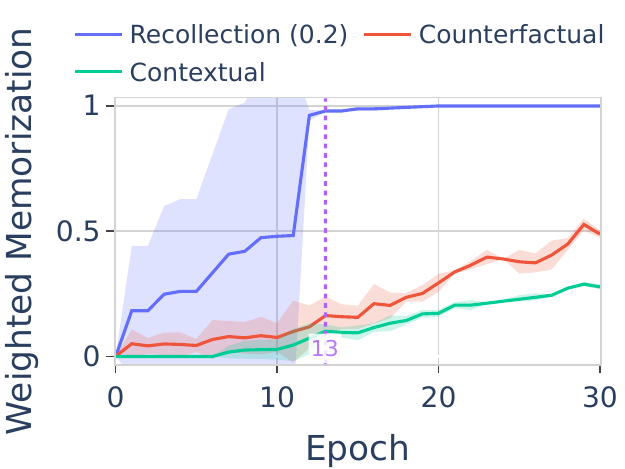}
        
    }

    \caption{Memorization of training strings in languages of different entropy across different memorization measures. Results are for language $ L_2 $, which is a low entropy language.
    }
\end{figure*}

\begin{figure*}[!t]
    \centering
    \captionsetup[subfigure]{justification=centering}

    \subfloat[Low Entropy]{
        \includegraphics[scale=0.4]{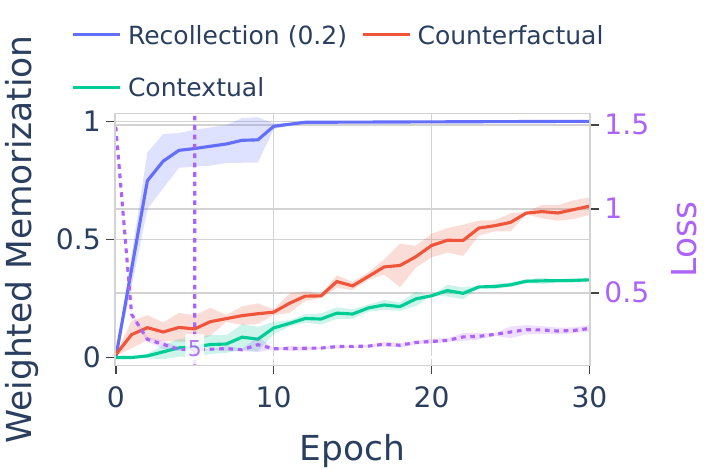}
    }
    \subfloat[High Entropy]{
        \includegraphics[scale=0.4]{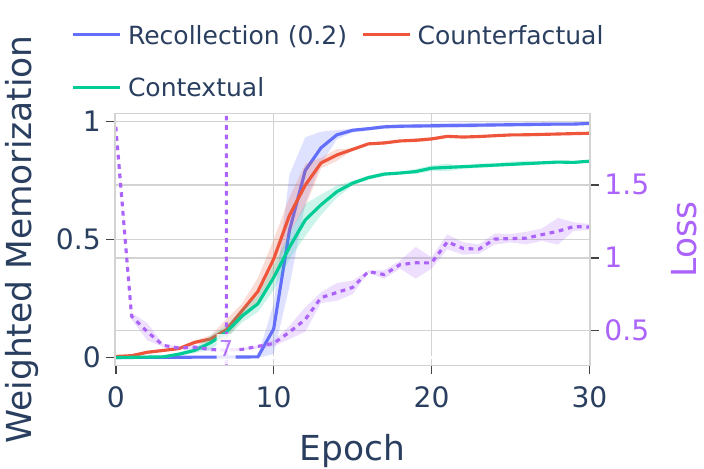}
    }

    \caption{Continuing Figure~\ref{fig:language_memorization}, we demonstrate associated loss with weighted memorization.}
    \label{fig:weighted_mem_vs_learning}
    
\end{figure*}

\begin{figure*}
    \centering
    \captionsetup[subfigure]{justification=centering}

    \subfloat[Recollection, $ n=64 $]{
        \includegraphics[scale=0.4]{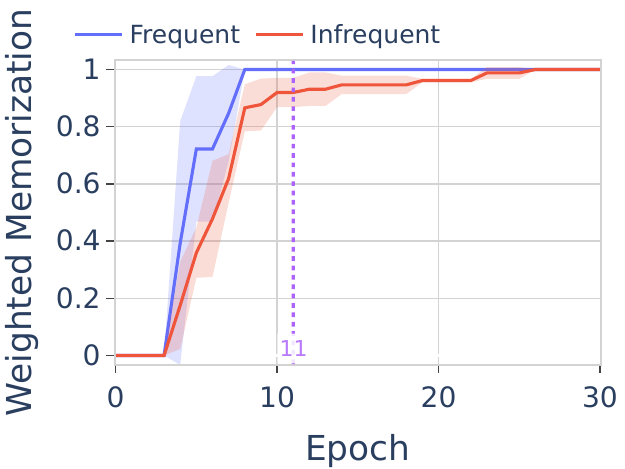}
    }
    \subfloat[Counterfactual, $ n=64 $]{
        \includegraphics[scale=0.4]{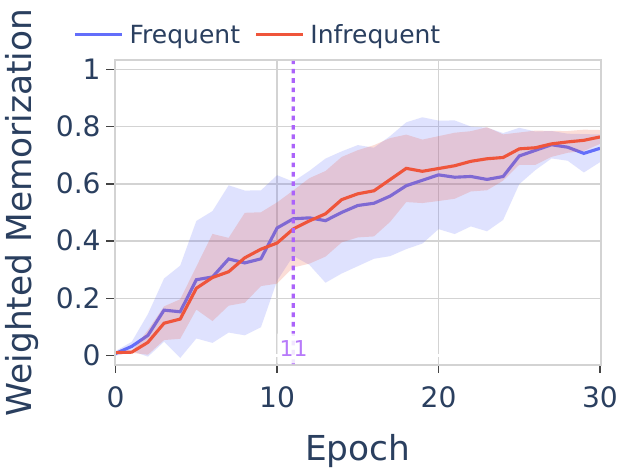}
    }
    \subfloat[Contextual, $ n=64 $]{
        \includegraphics[scale=0.4]{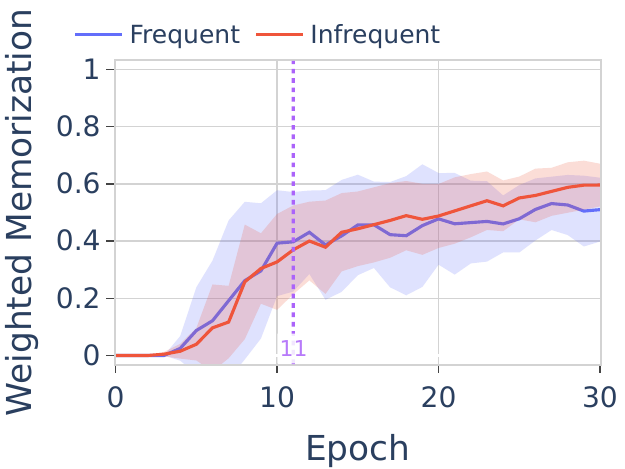}
    }

    \subfloat[Recollection, $ n=256 $]{
        \includegraphics[scale=0.4]{memorization_figures/operationalization/weighted_freq_split_pcfg_cfg3b_eq_len_skewed_prob_mistral-7b_256_recollection_memorization_loss_memorization.pdf}
    }
    \subfloat[Counterfactual, $ n=256 $]{
        \includegraphics[scale=0.4]{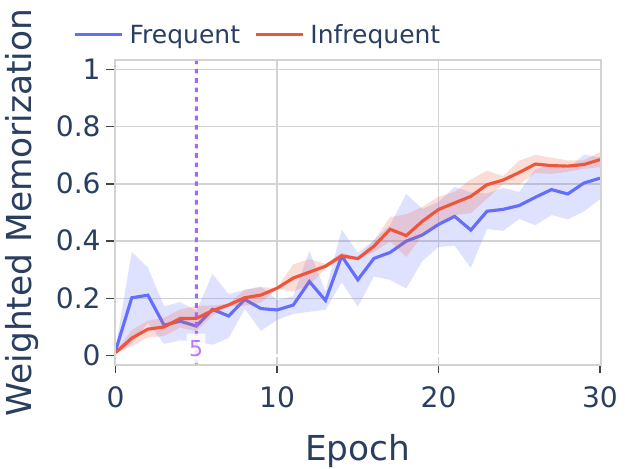}
    }
    \subfloat[Contextual, $ n=256 $]{
        \includegraphics[scale=0.4]{memorization_figures/operationalization/weighted_freq_split_pcfg_cfg3b_eq_len_skewed_prob_mistral-7b_256_contextual_memorization_loss_memorization.pdf}
    }

    \subfloat[Recollection, $ n=1024 $]{
        \includegraphics[scale=0.4]{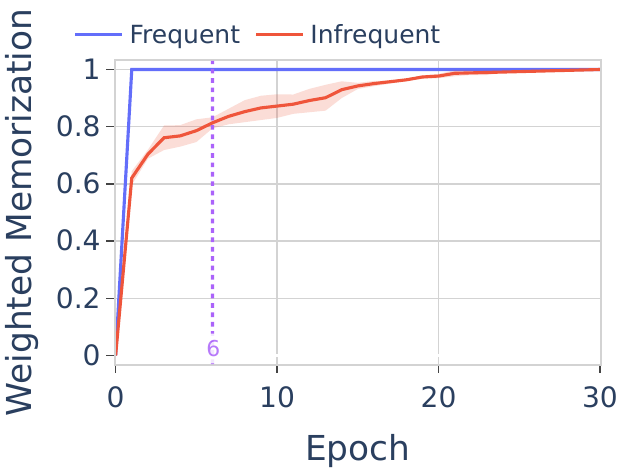}
    }
    \subfloat[Counterfactual, $ n=1024 $]{
        \includegraphics[scale=0.4]{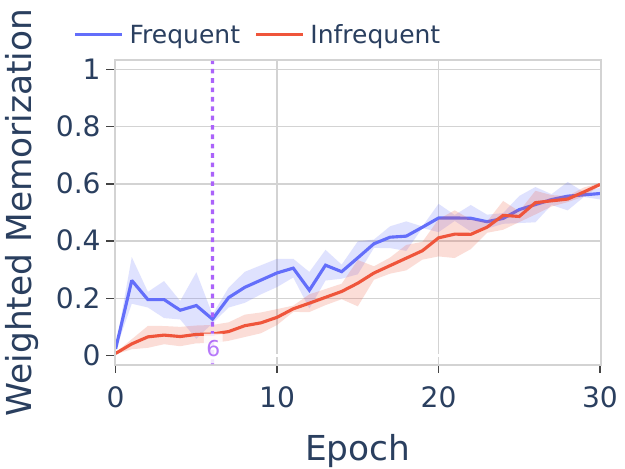}
    }
    \subfloat[Contextual, $ n=1024 $]{
        \includegraphics[scale=0.4]{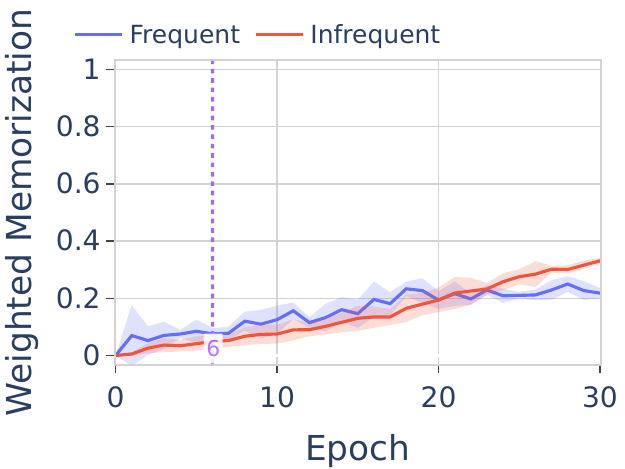}
    }

    \caption{Continuing Figure~\ref{fig:low_entropy_language_contradiction}, contradiction between recollection-based and contextual (or counterfactual) memorization on determining memorization of top $ 10\% $ frequent strings and bottom $ 10\% $ infrequent strings in a low entropy language. The results are for Mistral-$ 7 $B on language $ L_2 $, which is a low entropy language.}
    \label{fig:low_entropy_language_contradiction_detailed_fraction}

\end{figure*}

\begin{figure*}
    \centering
    \captionsetup[subfigure]{justification=centering}

    \subfloat[Recollection, $ n=64 $]{
        \includegraphics[scale=0.4]{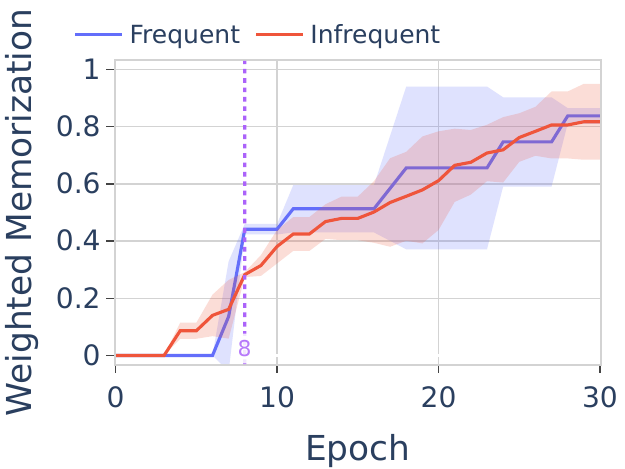}
    }
    \subfloat[Counterfactual, $ n=64 $]{
        \includegraphics[scale=0.4]{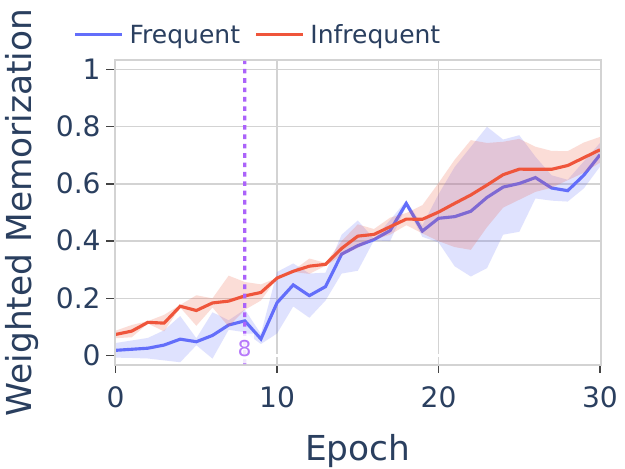}
    }
    \subfloat[Contextual, $ n=64 $]{
        \includegraphics[scale=0.4]{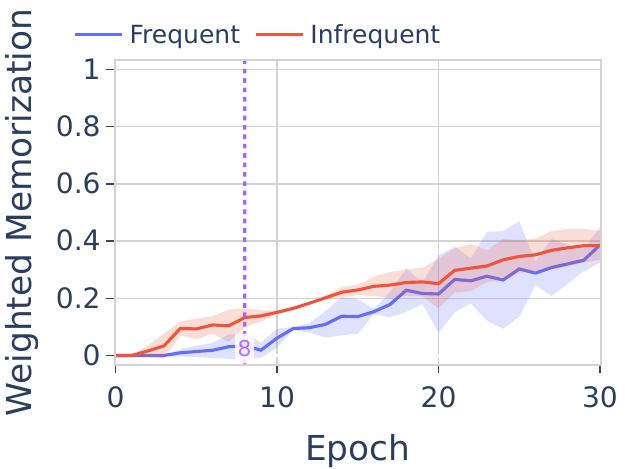}
    }

    \subfloat[Recollection, $ n=256 $]{
        \includegraphics[scale=0.4]{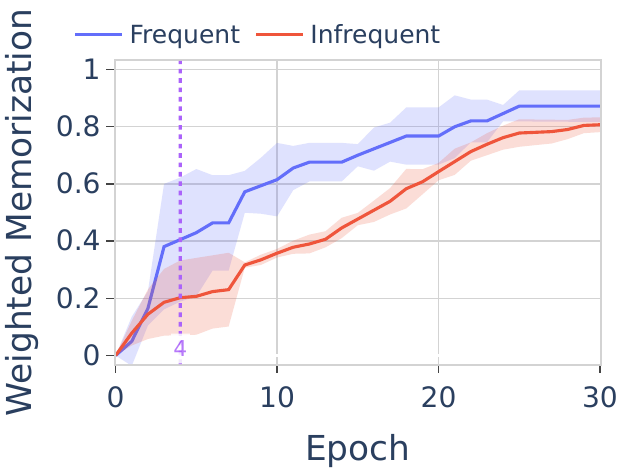}
    }
    \subfloat[Counterfactual, $ n=256 $]{
        \includegraphics[scale=0.4]{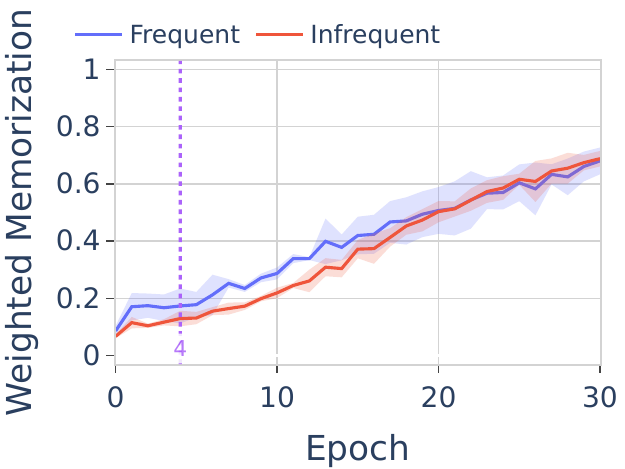}
    }
    \subfloat[Contextual, $ n=256 $]{
        \includegraphics[scale=0.4]{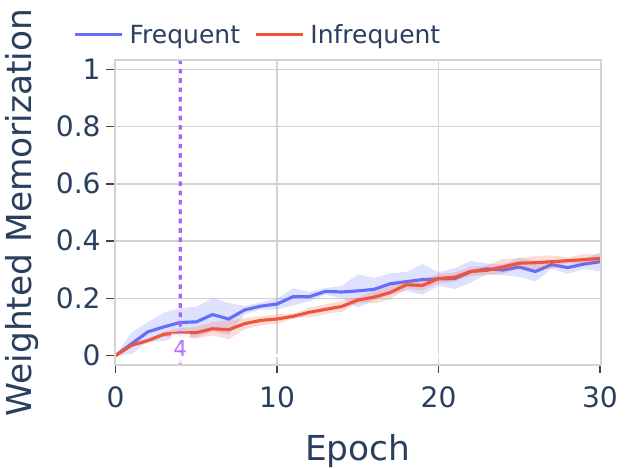}
    }

    \subfloat[Recollection, $ n=1024 $]{
        \includegraphics[scale=0.4]{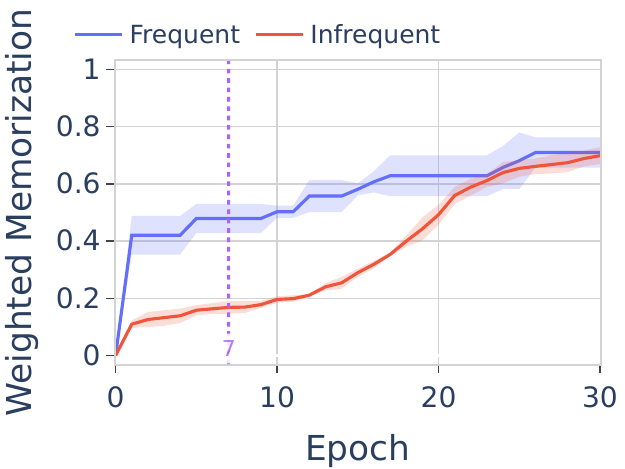}
    }
    \subfloat[Counterfactual, $ n=1024 $]{
        \includegraphics[scale=0.4]{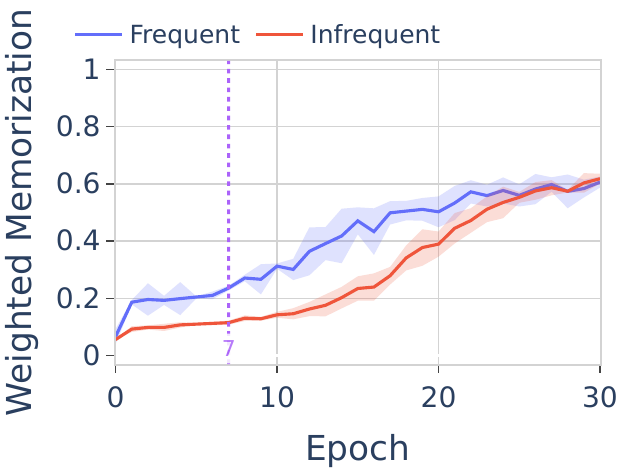}
    }
    \subfloat[Contextual, $ n=1024 $]{
        \includegraphics[scale=0.4]{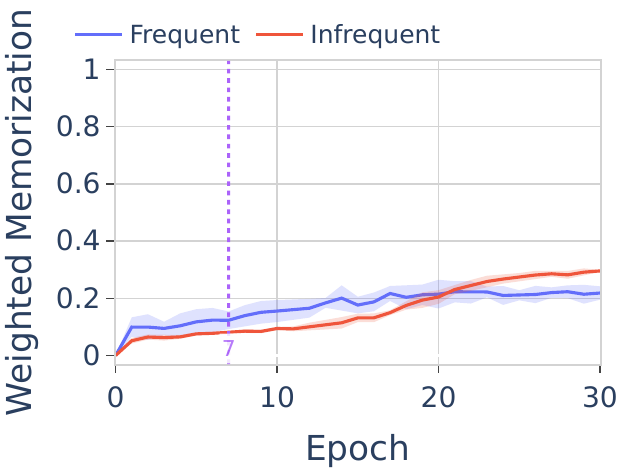}
    }

    \caption{Continuing Figure~\ref{fig:low_entropy_language_contradiction_detailed_fraction}, contradiction between recollection-based and contextual (or counterfactual) memorization on determining memorization of top $ 10\% $ frequent strings and bottom $ 10\% $ infrequent strings in a low entropy language. The results are for Mistral-$ 7 $B on language $ L_4 $, which is a low entropy language.}
    \label{fig:low_entropy_language_contradiction_detailed_fraction_l4}

\end{figure*}

\clearpage

\begin{figure}[!t]
    \centering
    \captionsetup[subfigure]{justification=centering}

    \subfloat[Mistral-$ 7 $B]{
        \includegraphics[scale=0.4]{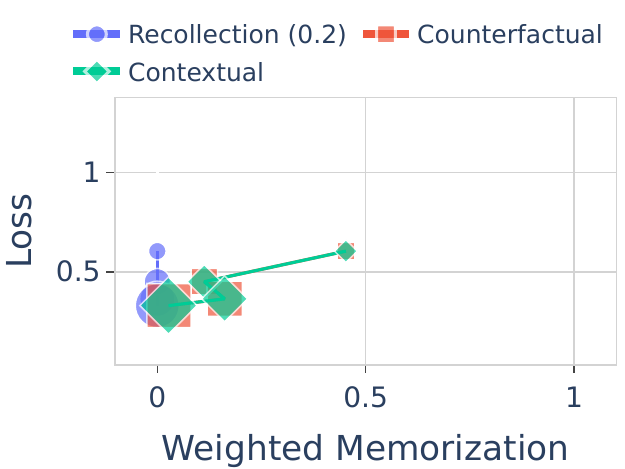}
    }
    \subfloat[Qwen-$ 2.5 $-$ 7 $B]{
        \includegraphics[scale=0.4]{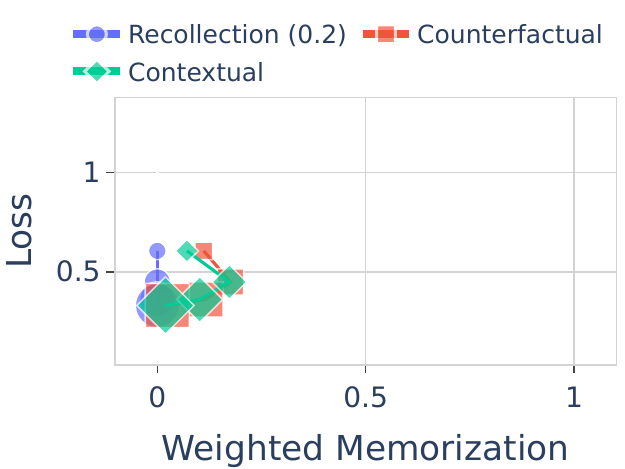}
    }

    \subfloat[Llama-$2$-$ 7 $B]{
        \includegraphics[scale=0.4]{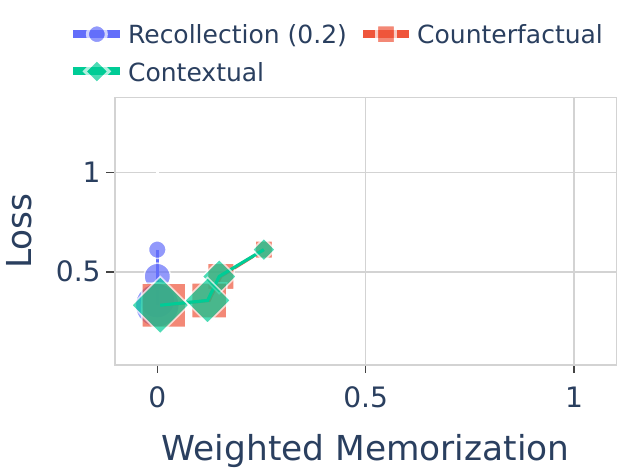}
    }
    \subfloat[Llama-$3.1$-$ * $B]{
        \includegraphics[scale=0.4]{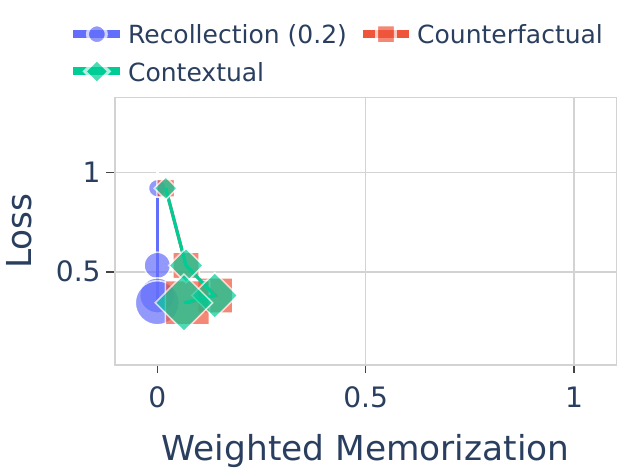}
    }

    \subfloat[Gemma-$ 2 $-$ 9 $B]{
        \includegraphics[scale=0.4]{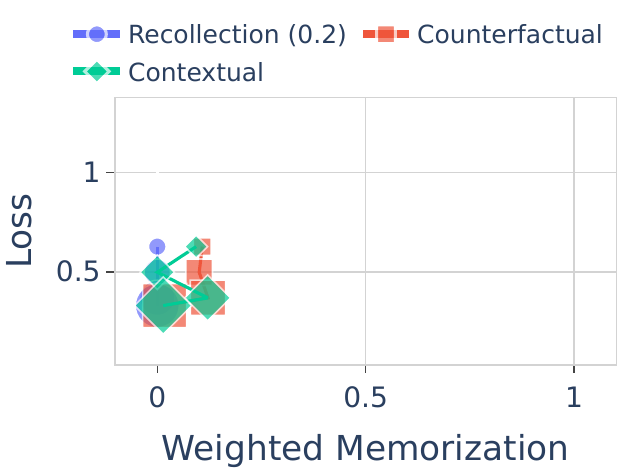}
    }
    \subfloat[Pythia-$ 6.9 $B]{
        \includegraphics[scale=0.4]{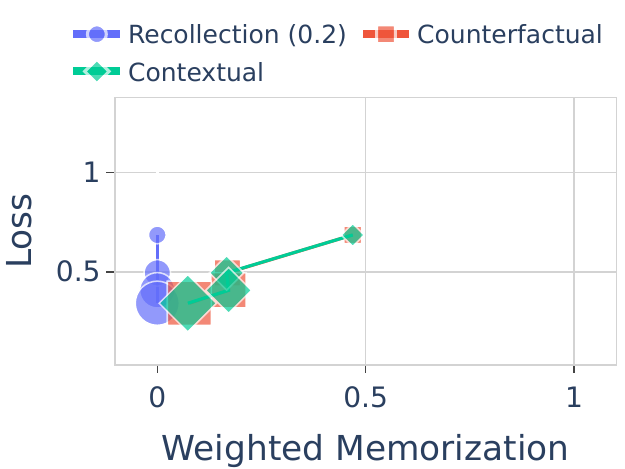}
    }

    \subfloat[Opt-$ 6.7 $B]{
        \includegraphics[scale=0.4]{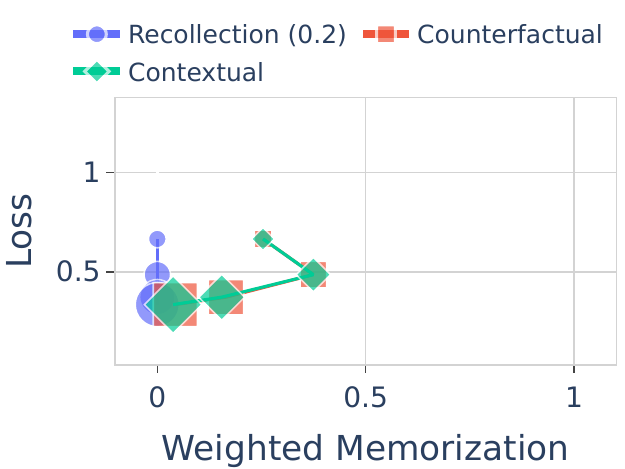}
    }

    \caption{Tradeoffs between optimal learning and memorization among comparable $ \approx 7 $B parameter size models on language $ L_1 $, which is a high entropy language.}

\end{figure}

\begin{figure}[!t]
    \centering
    \captionsetup[subfigure]{justification=centering}

    \subfloat[Mistral-$ 7 $B]{
        \includegraphics[scale=0.4]{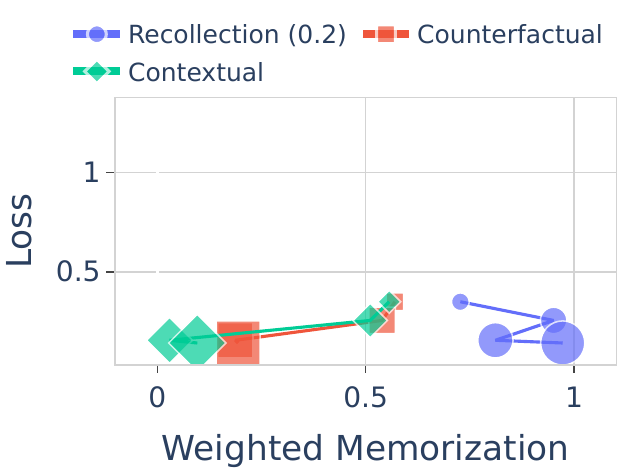}
    }
    \subfloat[Qwen-$ 2.5 $-$ 7 $B]{
        \includegraphics[scale=0.4]{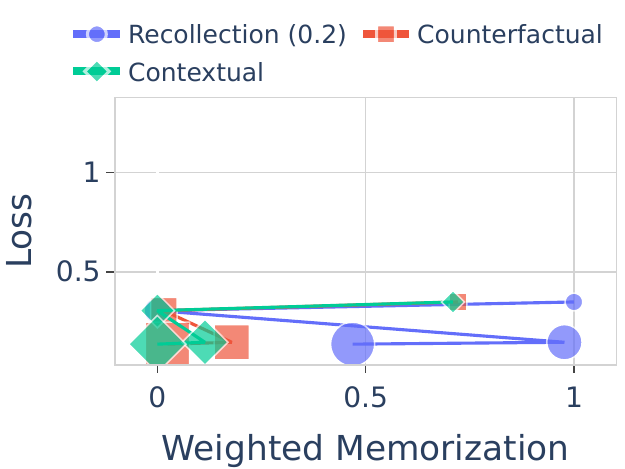}
    }

    \subfloat[Llama-$2$-$ 7 $B]{
        \includegraphics[scale=0.4]{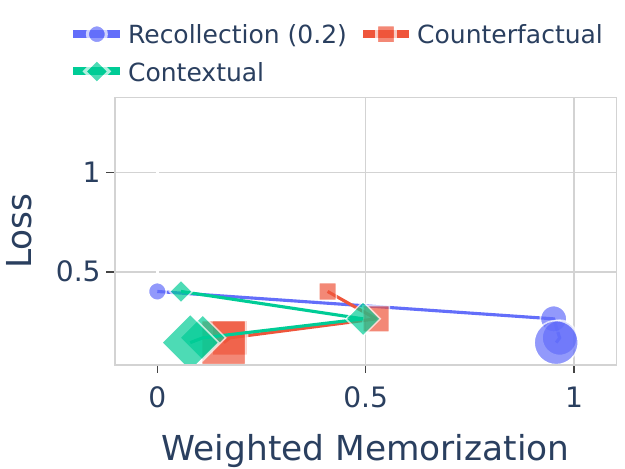}
    }
    \subfloat[Llama-$3.1$-$ * $B]{
        \includegraphics[scale=0.4]{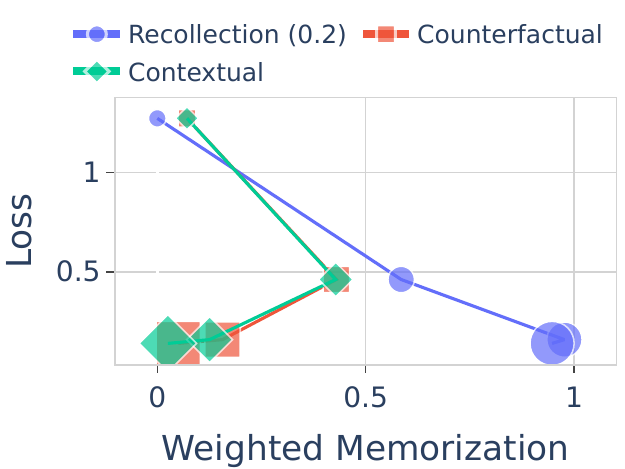}
    }

    \subfloat[Gemma-$ 2 $-$ 9 $B]{
        \includegraphics[scale=0.4]{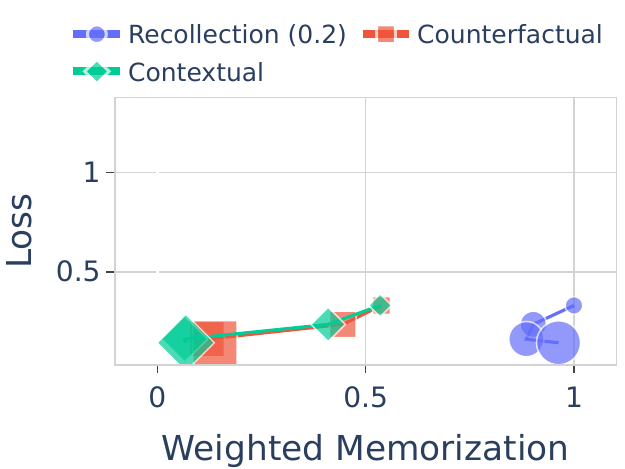}
    }
    \subfloat[Pythia-$ 6.9 $B]{
        \includegraphics[scale=0.4]{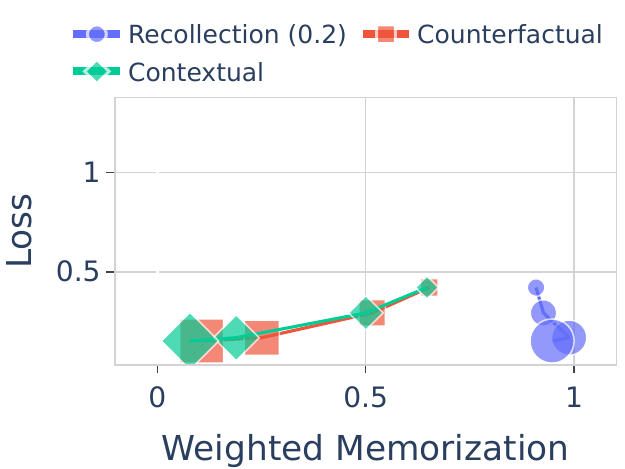}
    }

    \subfloat[Opt-$ 6.7 $B]{
        \includegraphics[scale=0.4]{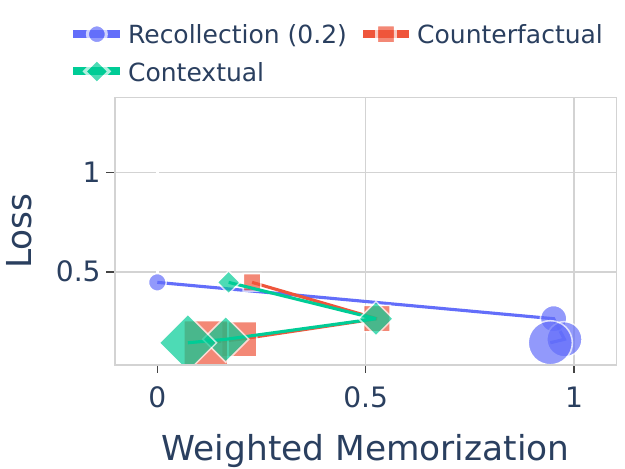}
    }

    \caption{Tradeoffs between optimal learning and memorization among comparable $ \approx 7 $B parameter size models on language $ L_2 $, which is a low entropy language.}

\end{figure}

\begin{figure}[!t]
    \centering
    \captionsetup[subfigure]{justification=centering}

    \subfloat[Mistral-$ 7 $B]{
        \includegraphics[scale=0.4]{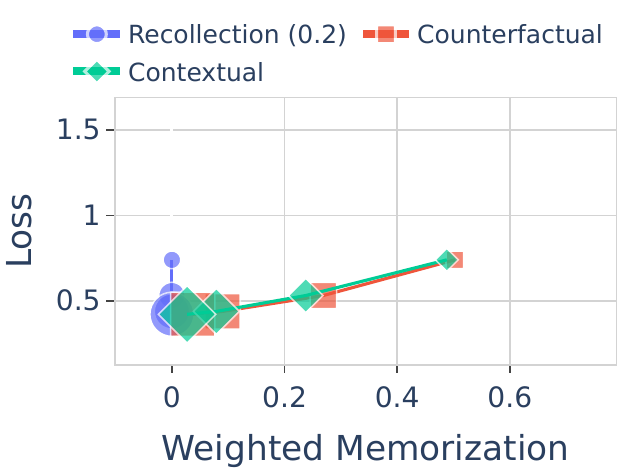}
    }
    \subfloat[Qwen-$ 2.5 $-$ 7 $B]{
        \includegraphics[scale=0.4]{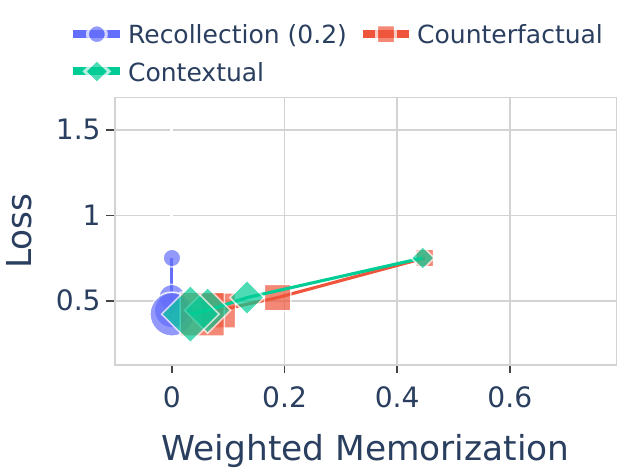}
    }

    \subfloat[Llama-$2$-$ 7 $B]{
        \includegraphics[scale=0.4]{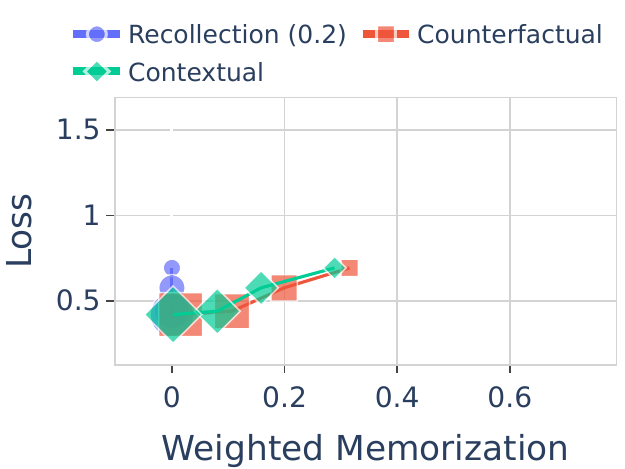}
    }
    \subfloat[Llama-$3.1$-$ * $B]{
        \includegraphics[scale=0.4]{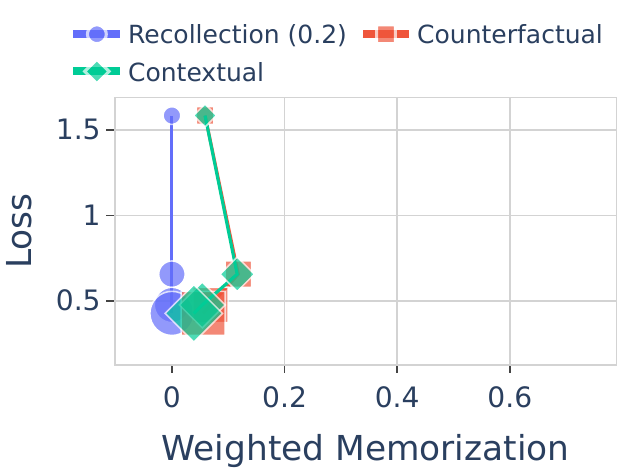}
    }

    \subfloat[Gemma-$ 2 $-$ 9 $B]{
        \includegraphics[scale=0.4]{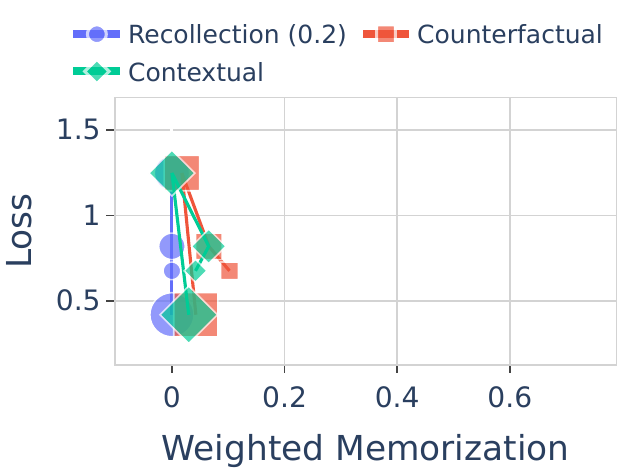}
    }
    \subfloat[Pythia-$ 6.9 $B]{
        \includegraphics[scale=0.4]{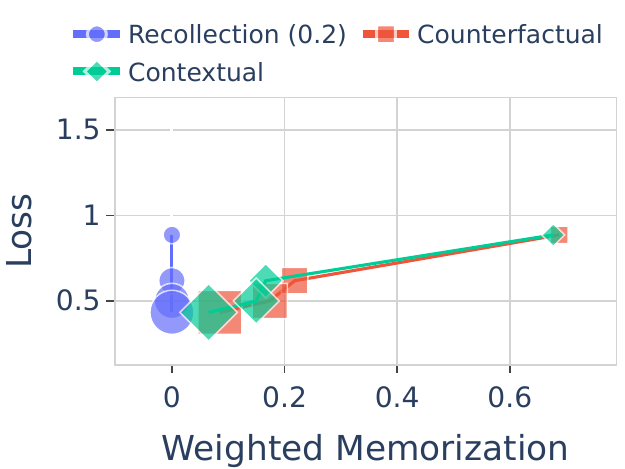}
    }

    \subfloat[Opt-$ 6.7 $B]{
        \includegraphics[scale=0.4]{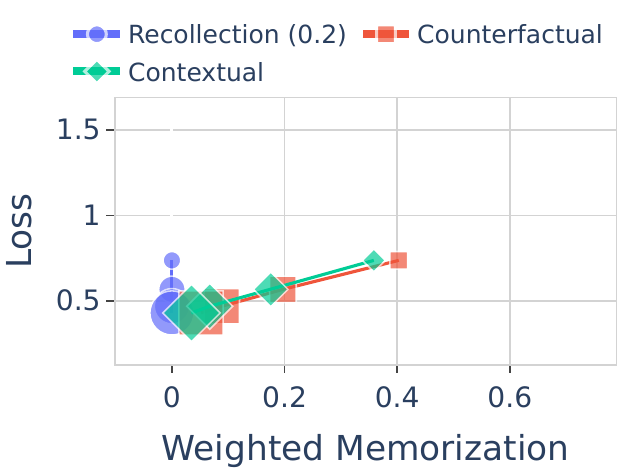}
    }

    \caption{Tradeoffs between optimal learning and memorization among comparable $ \approx 7 $B parameter size models on language $ L_3 $, which is a high entropy language.}

\end{figure}

\begin{figure}[!t]
    \centering
    \captionsetup[subfigure]{justification=centering}

    \subfloat[Mistral-$ 7 $B]{
        \includegraphics[scale=0.4]{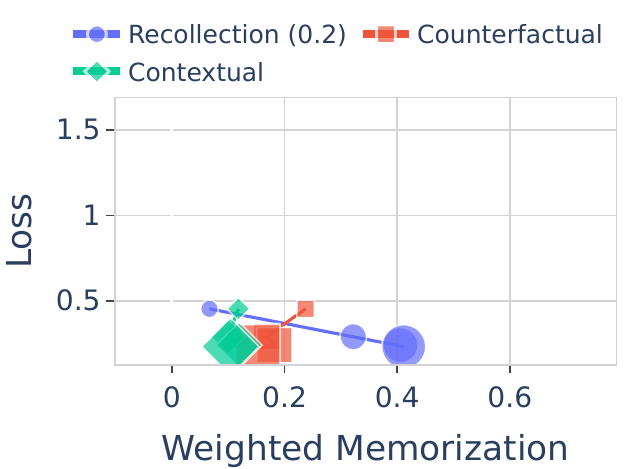}
    }
    \subfloat[Qwen-$ 2.5 $-$ 7 $B]{
        \includegraphics[scale=0.4]{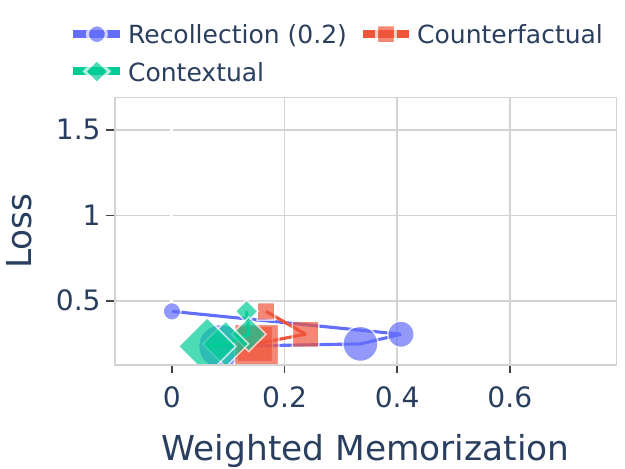}
    }

    \subfloat[Llama-$2$-$ 7 $B]{
        \includegraphics[scale=0.4]{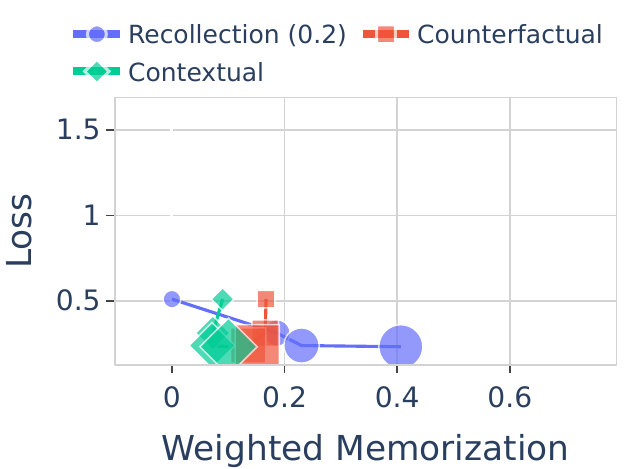}
    }
    \subfloat[Llama-$3.1$-$ * $B]{
        \includegraphics[scale=0.4]{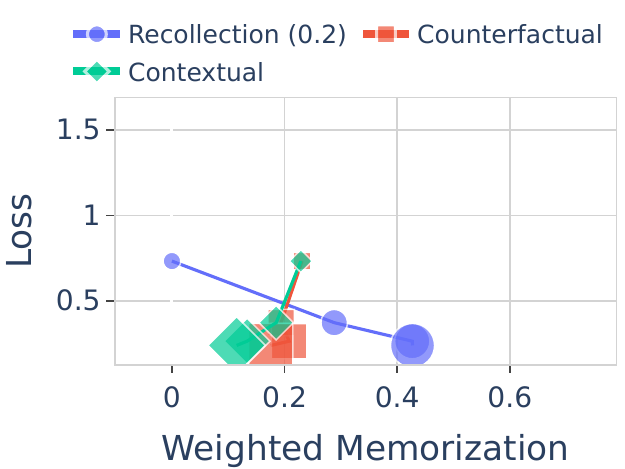}
    }

    \subfloat[Gemma-$ 2 $-$ 9 $B]{
        \includegraphics[scale=0.4]{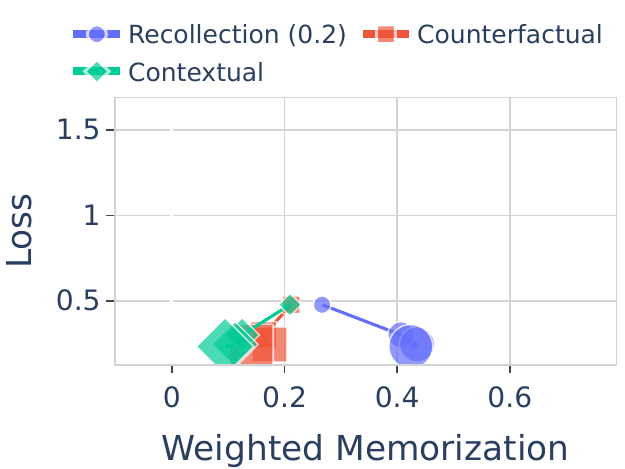}
    }
    \subfloat[Pythia-$ 6.9 $B]{
        \includegraphics[scale=0.4]{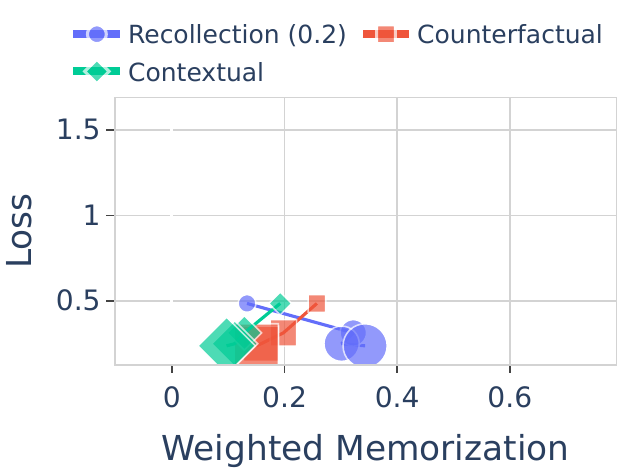}
    }

    \subfloat[Opt-$ 6.7 $B]{
        \includegraphics[scale=0.4]{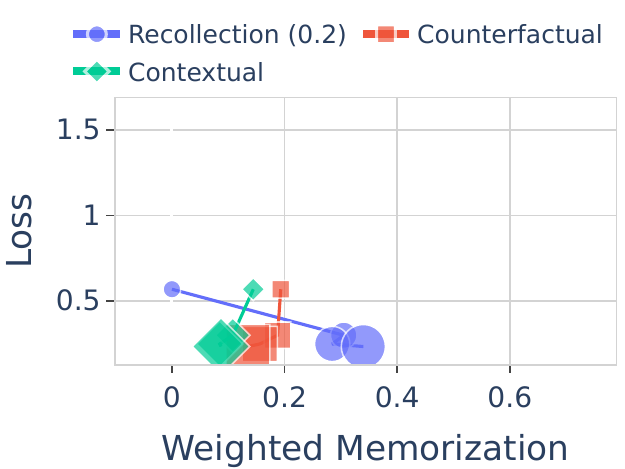}
    }

    \caption{Tradeoffs between optimal learning and memorization among comparable $ \approx 7 $B parameter size models on language $ L_4 $, which is a low entropy language.}

\end{figure}

\begin{figure}[!t]
    \centering
    \captionsetup[subfigure]{justification=centering}

    \subfloat[Mistral-$ 7 $B]{
        \includegraphics[scale=0.4]{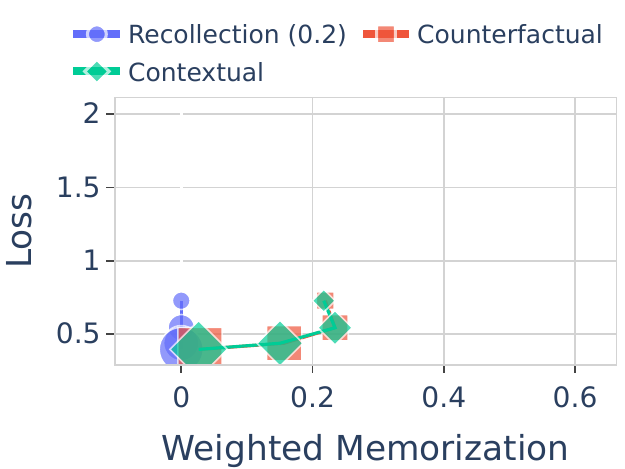}
    }
    \subfloat[Qwen-$ 2.5 $-$ 7 $B]{
        \includegraphics[scale=0.4]{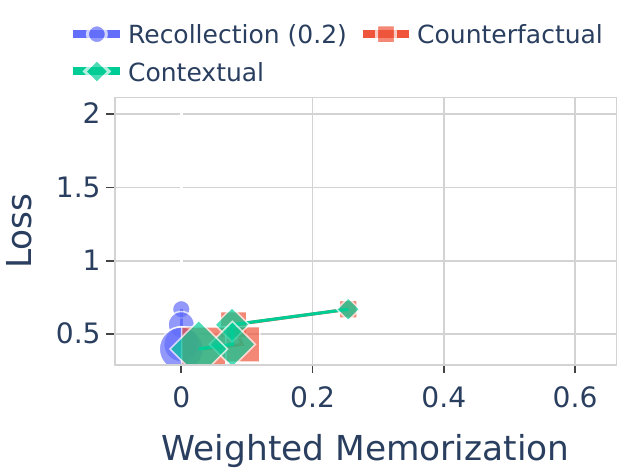}
    }

    \subfloat[Llama-$2$-$ 7 $B]{
        \includegraphics[scale=0.4]{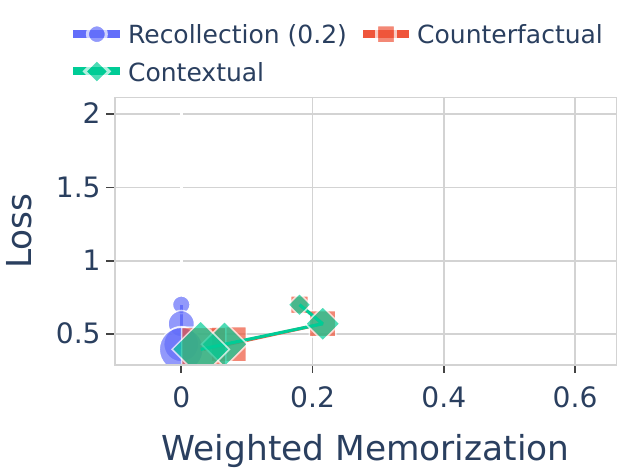}
    }
    \subfloat[Llama-$3.1$-$ * $B]{
        \includegraphics[scale=0.4]{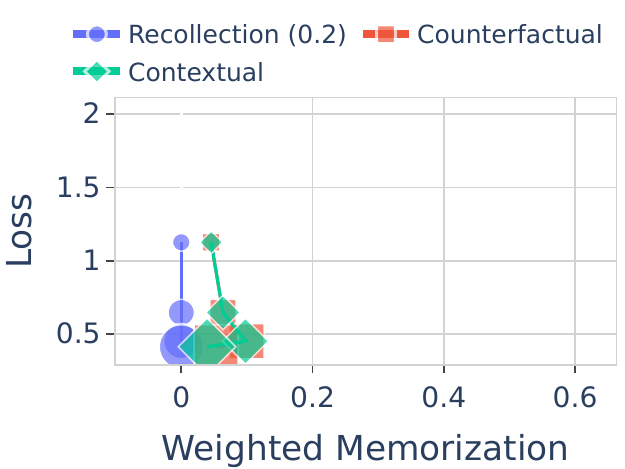}
    }

    \subfloat[Gemma-$ 2 $-$ 9 $B]{
        \includegraphics[scale=0.4]{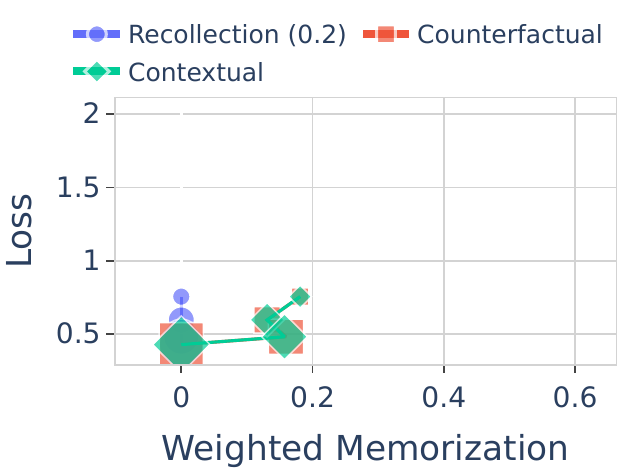}
    }
    \subfloat[Pythia-$ 6.9 $B]{
        \includegraphics[scale=0.4]{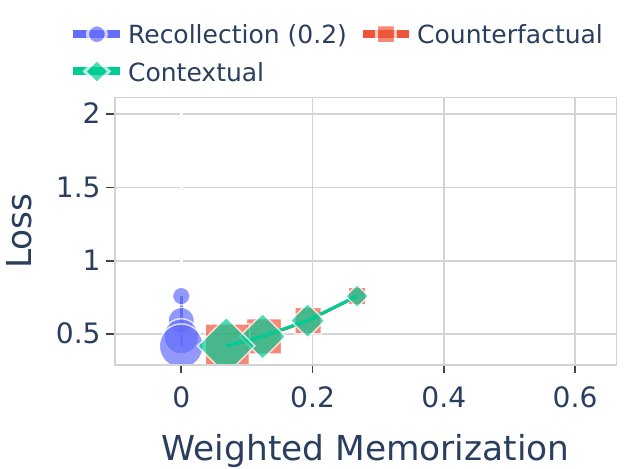}
    }

    \subfloat[Opt-$ 6.7 $B]{
        \includegraphics[scale=0.4]{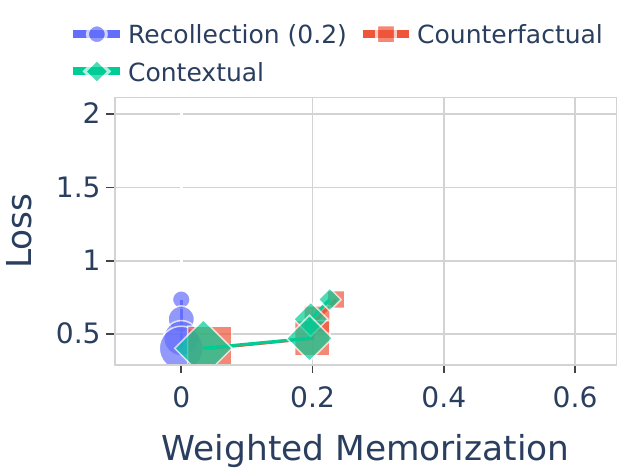}
    }

    \caption{Tradeoffs between optimal learning and memorization among comparable $ \approx 7 $B parameter size models on language $ L_5 $, which is a high entropy language.}

\end{figure}

\begin{figure}[!t]
    \centering
    \captionsetup[subfigure]{justification=centering}

    \subfloat[Mistral-$ 7 $B]{
        \includegraphics[scale=0.4]{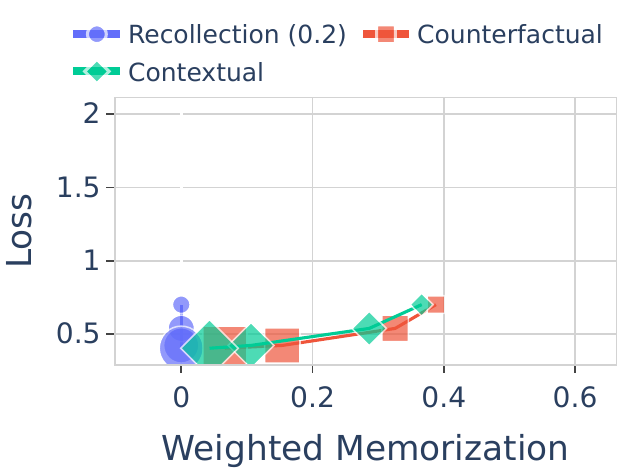}
    }
    \subfloat[Qwen-$ 2.5 $-$ 7 $B]{
        \includegraphics[scale=0.4]{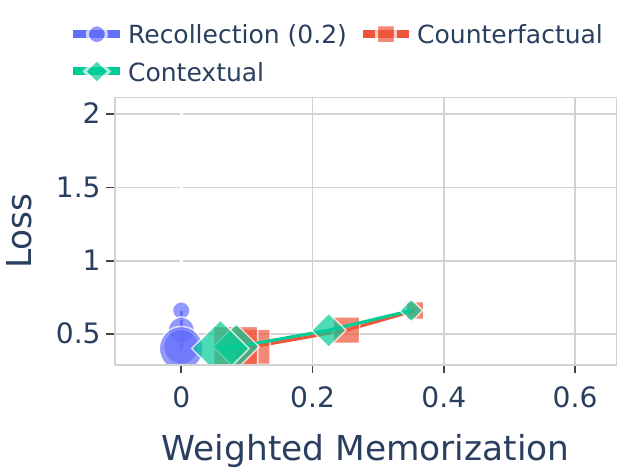}
    }

    \subfloat[Llama-$2$-$ 7 $B]{
        \includegraphics[scale=0.4]{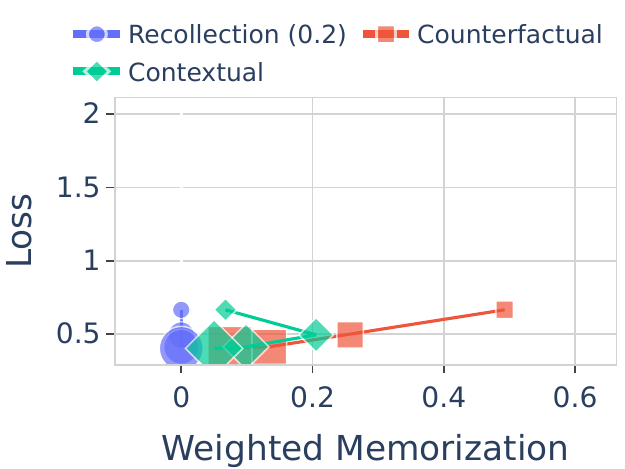}
    }
    \subfloat[Llama-$3.1$-$ * $B]{
        \includegraphics[scale=0.4]{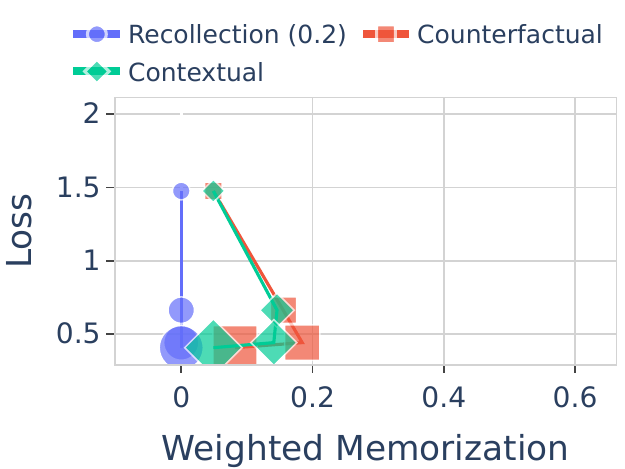}
    }

    \subfloat[Gemma-$ 2 $-$ 9 $B]{
        \includegraphics[scale=0.4]{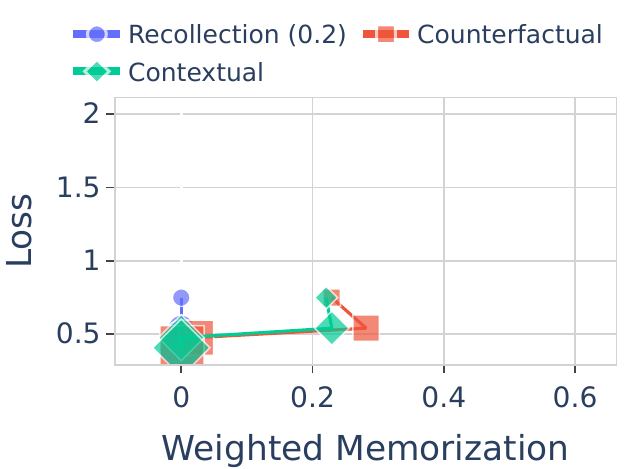}
    }
    \subfloat[Pythia-$ 6.9 $B]{
        \includegraphics[scale=0.4]{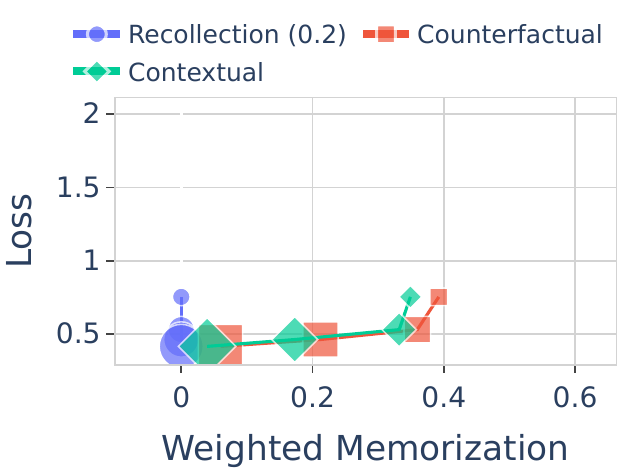}
    }

    \subfloat[Opt-$ 6.7 $B]{
        \includegraphics[scale=0.4]{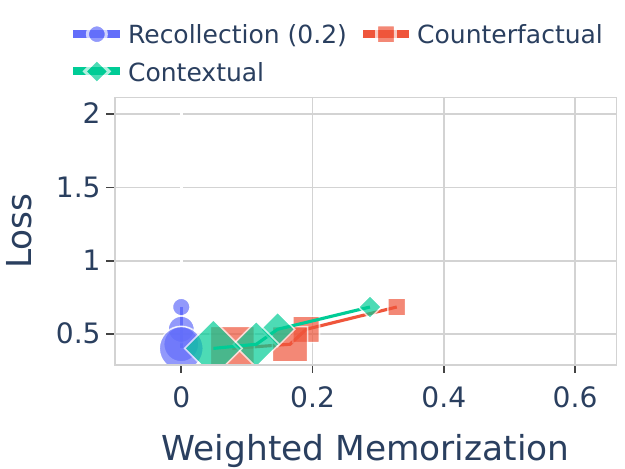}
    }

    \caption{Tradeoffs between optimal learning and memorization among comparable $ \approx 7 $B parameter size models on language $ L_6 $, which is a high entropy language.}

\end{figure}

\begin{figure}[!t]
    \centering
    \captionsetup[subfigure]{justification=centering}

    \subfloat[Mistral-$ 7 $B]{
        \includegraphics[scale=0.4]{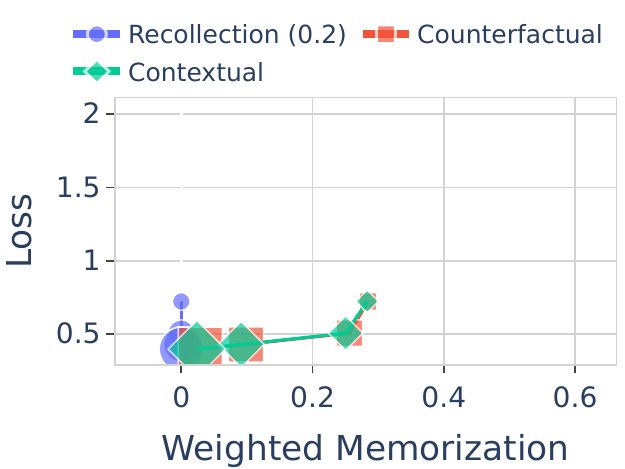}
    }
    \subfloat[Qwen-$ 2.5 $-$ 7 $B]{
        \includegraphics[scale=0.4]{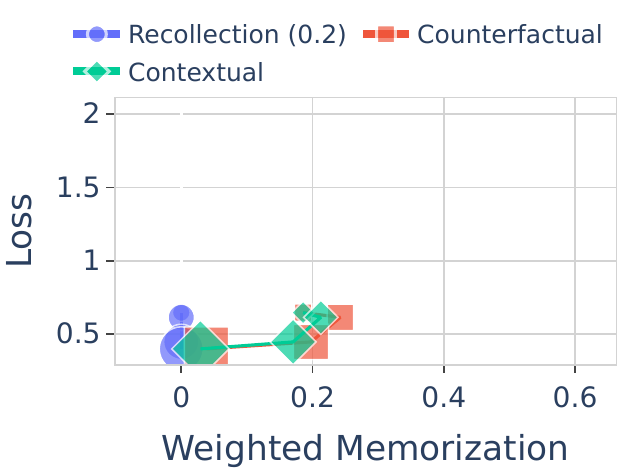}
    }

    \subfloat[Llama-$2$-$ 7 $B]{
        \includegraphics[scale=0.4]{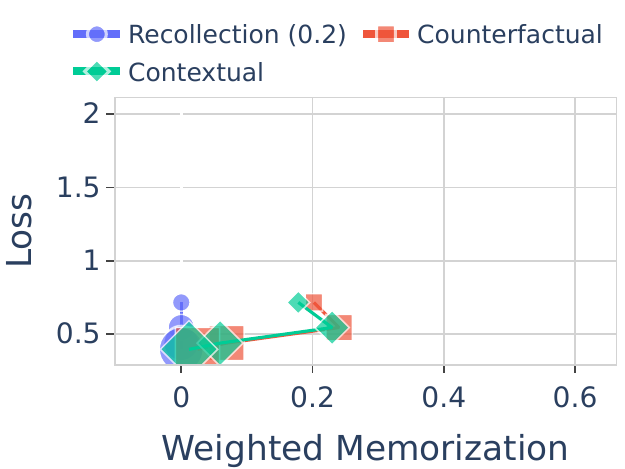}
    }

    \subfloat[Gemma-$ 2 $-$ 9 $B]{
        \includegraphics[scale=0.4]{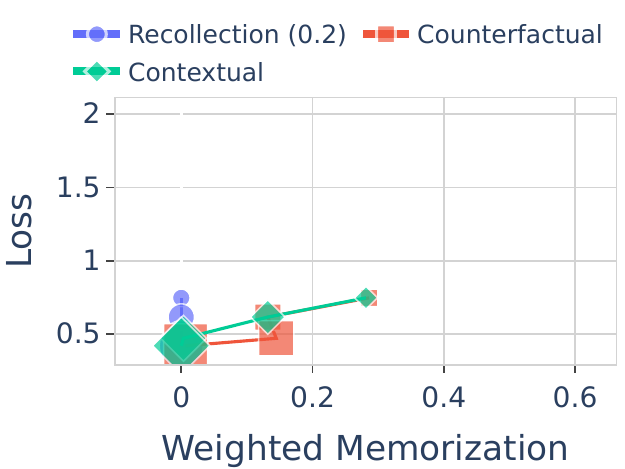}
    }
    \subfloat[Pythia-$ 6.9 $B]{
        \includegraphics[scale=0.4]{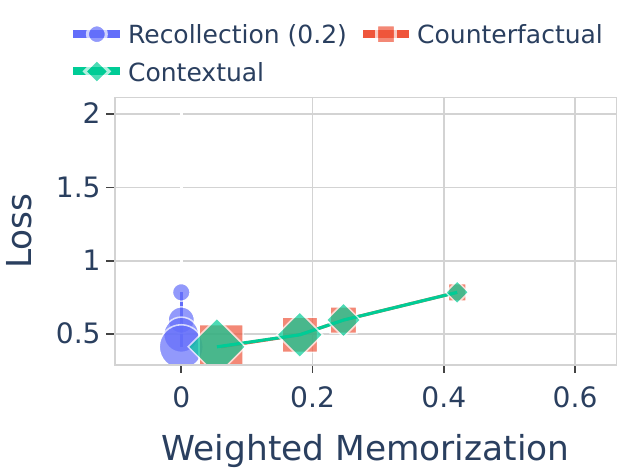}
    }

    \caption{Tradeoffs between optimal learning and memorization among comparable $ \approx 7 $B parameter size models on language $ L_7 $, which is a high entropy language.}

\end{figure}

\begin{figure}[!t]
    \centering
    \captionsetup[subfigure]{justification=centering}

    \subfloat[Mistral-$ 7 $B]{
        \includegraphics[scale=0.4]{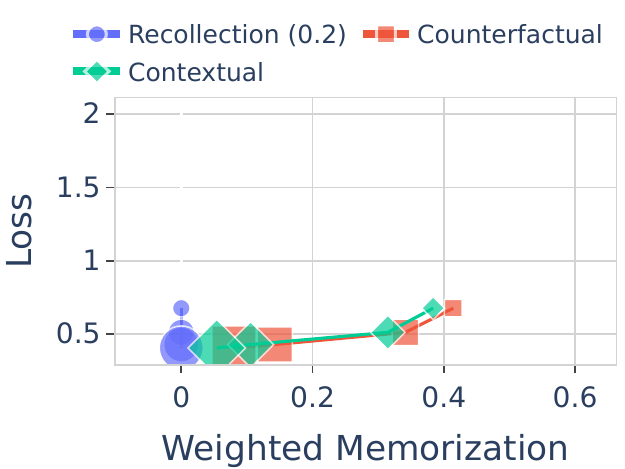}
    }
    \subfloat[Qwen-$ 2.5 $-$ 7 $B]{
        \includegraphics[scale=0.4]{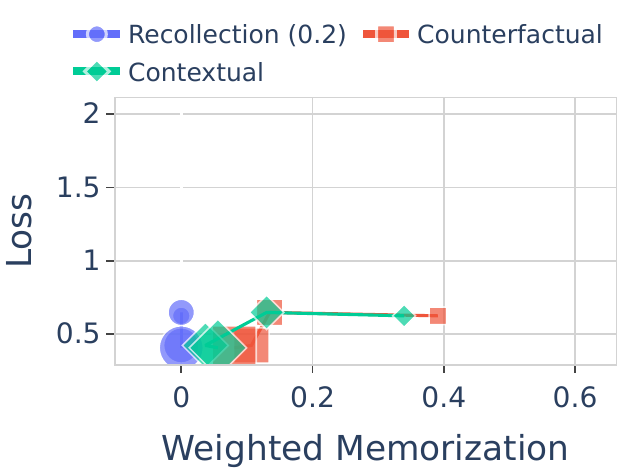}
    }

    \subfloat[Llama-$2$-$ 7 $B]{
        \includegraphics[scale=0.4]{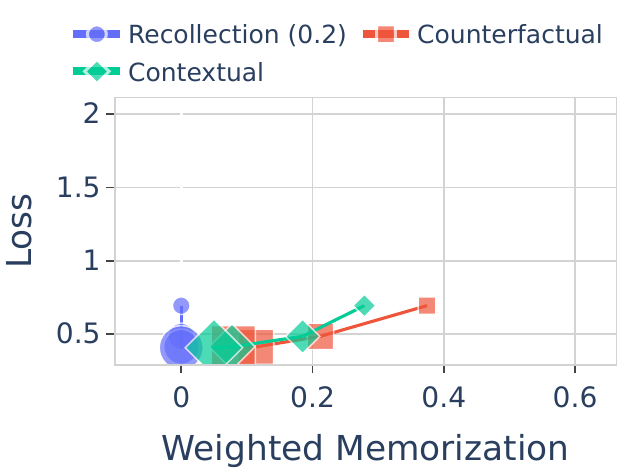}
    }
    \subfloat[Llama-$3.1$-$ * $B]{
        \includegraphics[scale=0.4]{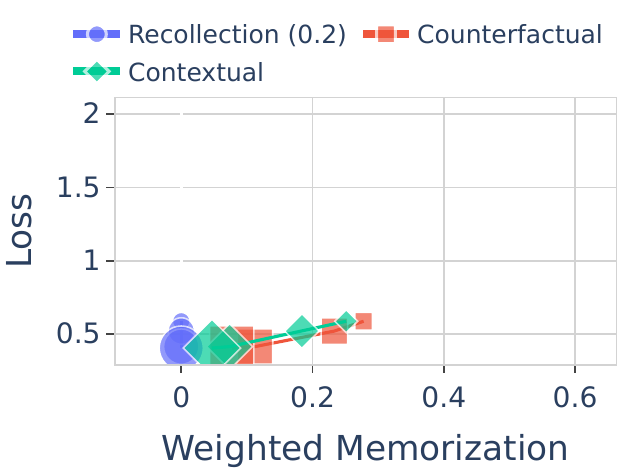}
    }

    \subfloat[Gemma-$ 2 $-$ 9 $B]{
        \includegraphics[scale=0.4]{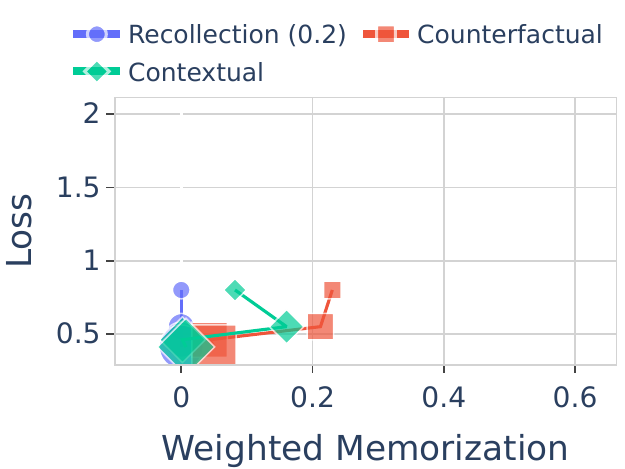}
    }
    \subfloat[Pythia-$ 6.9 $B]{
        \includegraphics[scale=0.4]{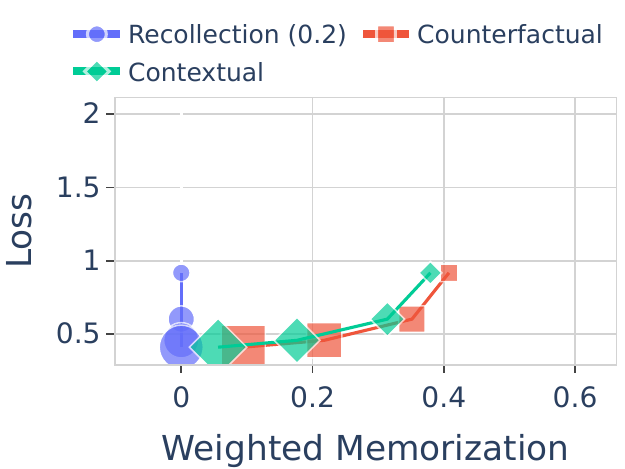}
    }

    \caption{Tradeoffs between optimal learning and memorization among comparable $ \approx 7 $B parameter size models on language $ L_8 $, which is a high entropy language.}

\end{figure}

\cleardoublepage

\begin{figure}
    \centering
    \captionsetup[subfigure]{justification=centering}

    \subfloat[Qwen-$ 2.5 $]{
        \includegraphics[scale=0.4]{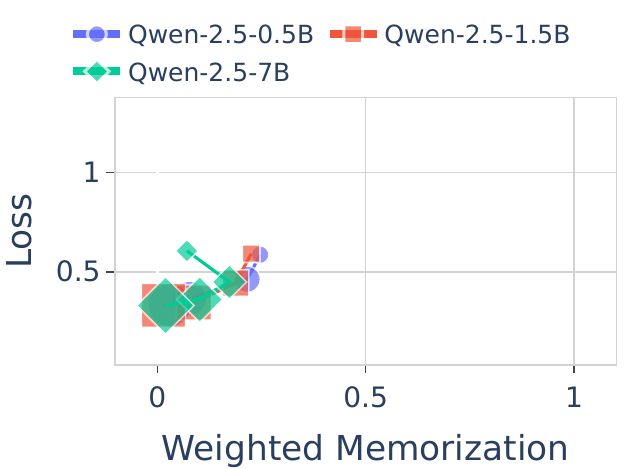}
    }
    \subfloat[Mistral]{
        \includegraphics[scale=0.4]{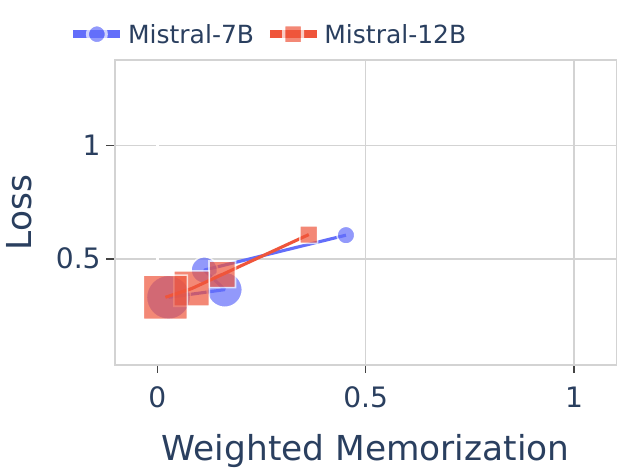}
    }
    
    \subfloat[Llama-$ 2 $]{
        \includegraphics[scale=0.4]{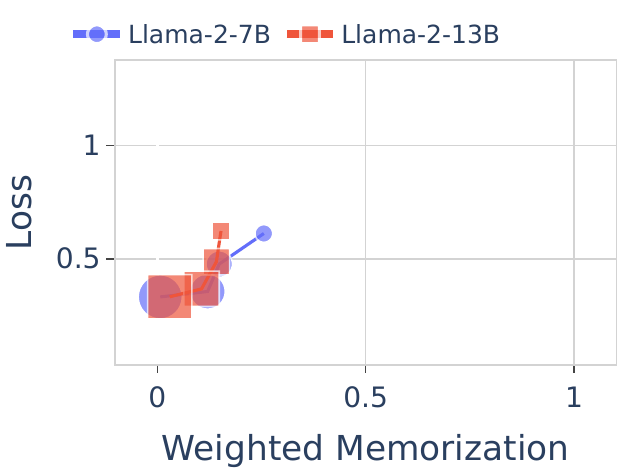}
    }
    \subfloat[Llama-$ 3 $]{
        \includegraphics[scale=0.4]{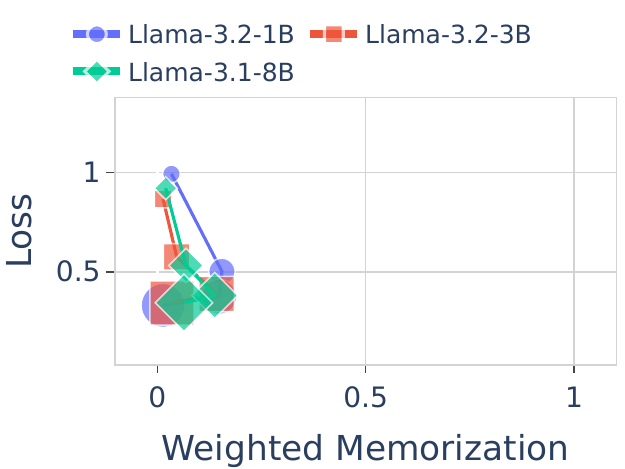}
    }
    
    \subfloat[Gemma-$ 2 $]{
        \includegraphics[scale=0.4]{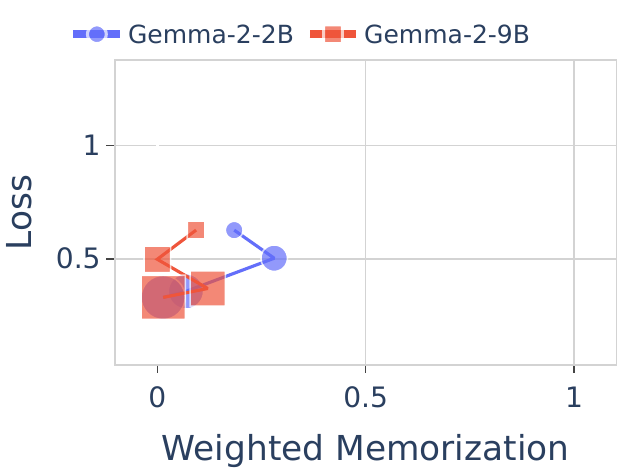}
    }
    \subfloat[Opt]{
        \includegraphics[scale=0.4]{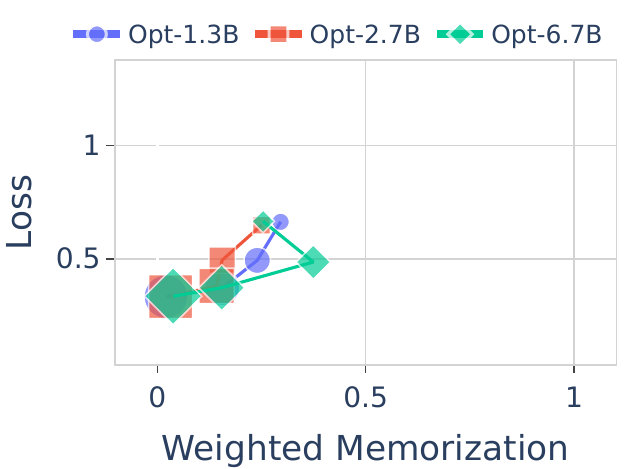}
    }
    
    \subfloat[Pythia]{
        \includegraphics[scale=0.4]{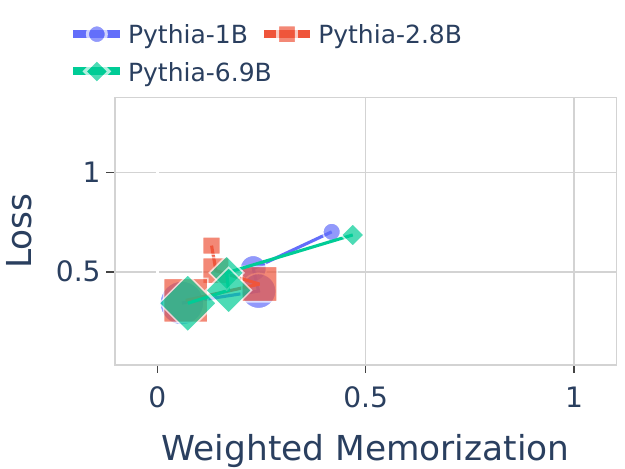}
    }

    \caption{Contextual memorization vs.\ optimal language learning, measured as test loss, across models of different sizes within a family. Results are on language $ L_1 $, which is a high entropy language.}
\end{figure}

\begin{figure}
    \centering
    \captionsetup[subfigure]{justification=centering}

    \subfloat[Qwen-$ 2.5 $]{
        \includegraphics[scale=0.4]{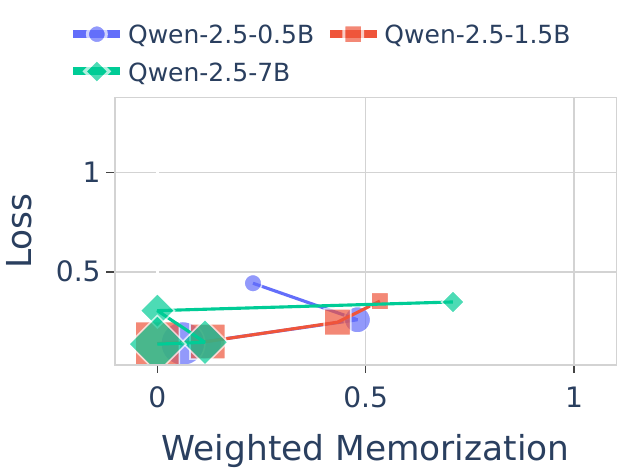}
    }
    \subfloat[Mistral]{
        \includegraphics[scale=0.4]{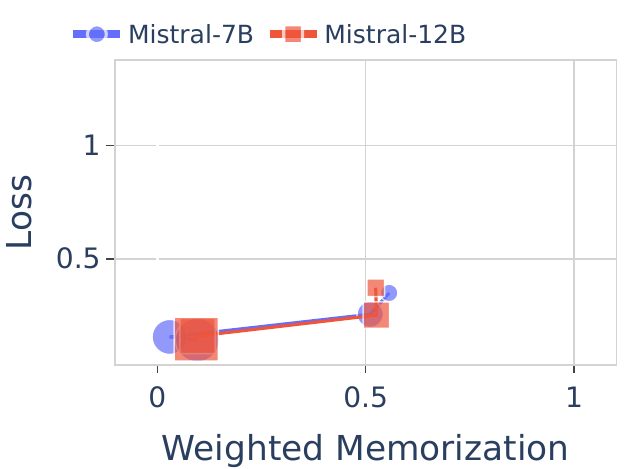}
    }
    
    \subfloat[Llama-$ 2 $]{
        \includegraphics[scale=0.4]{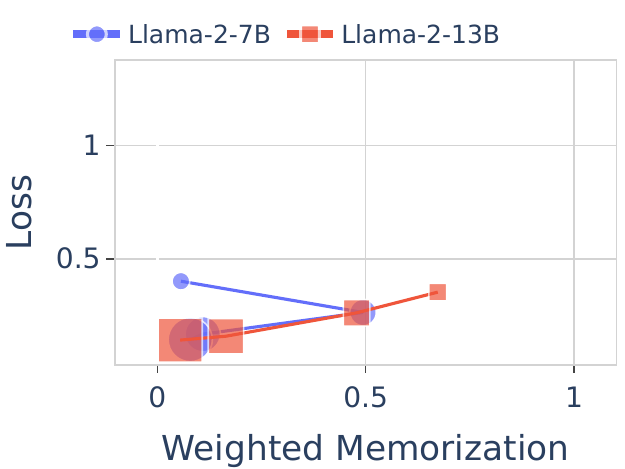}
    }
    \subfloat[Llama-$ 3 $]{
        \includegraphics[scale=0.4]{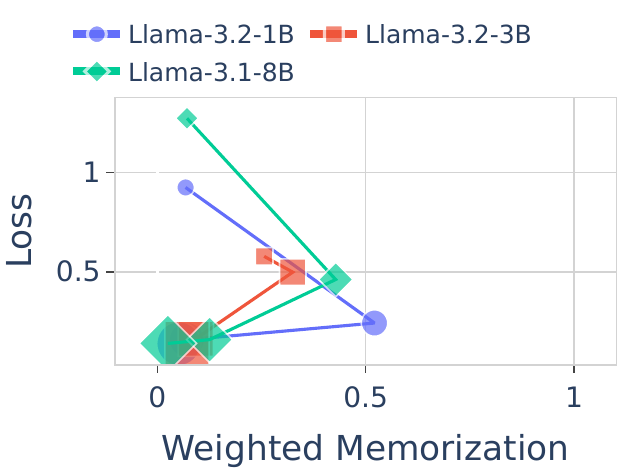}
    }
    
    \subfloat[Gemma-$ 2 $]{
        \includegraphics[scale=0.4]{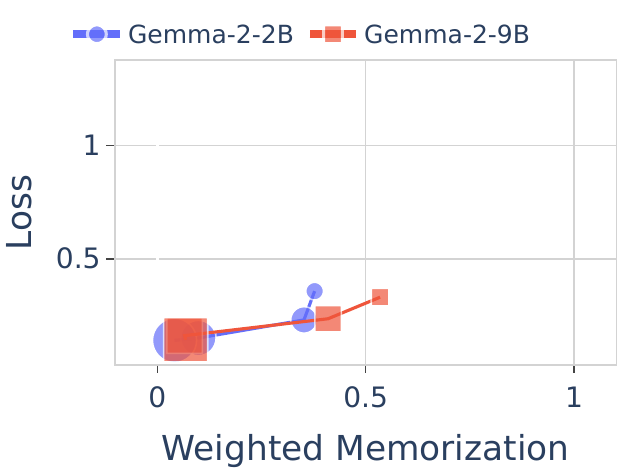}
    }
    \subfloat[Opt]{
        \includegraphics[scale=0.4]{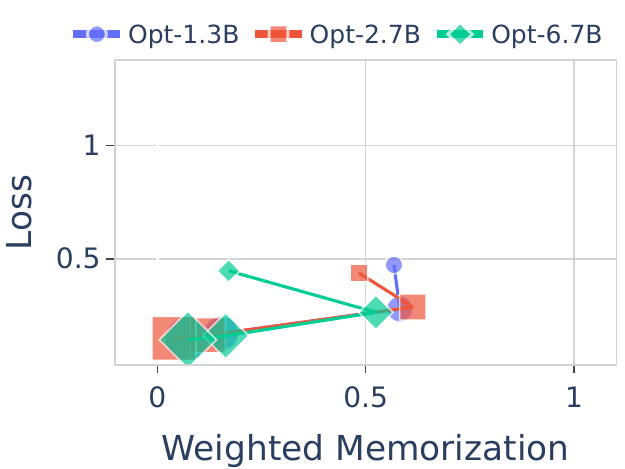}
    }
    
    \subfloat[Pythia]{
        \includegraphics[scale=0.4]{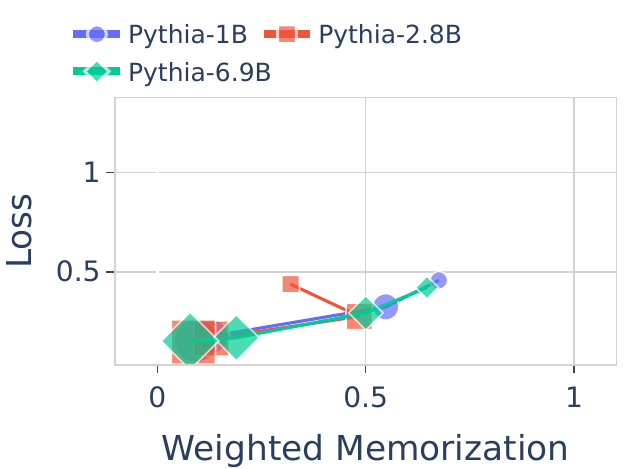}
    }

    \caption{Contextual memorization vs.\ optimal language learning, measured as test loss, across models of different sizes within a family. Results are on language $ L_2 $, which is a low entropy language.}
\end{figure}

\begin{figure}
    \centering
    \captionsetup[subfigure]{justification=centering}

    \subfloat[Qwen-$ 2.5 $]{
        \includegraphics[scale=0.4]{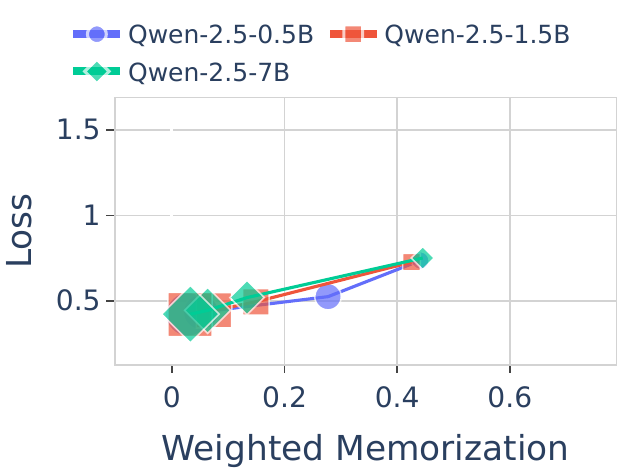}
    }
    \subfloat[Mistral]{
        \includegraphics[scale=0.4]{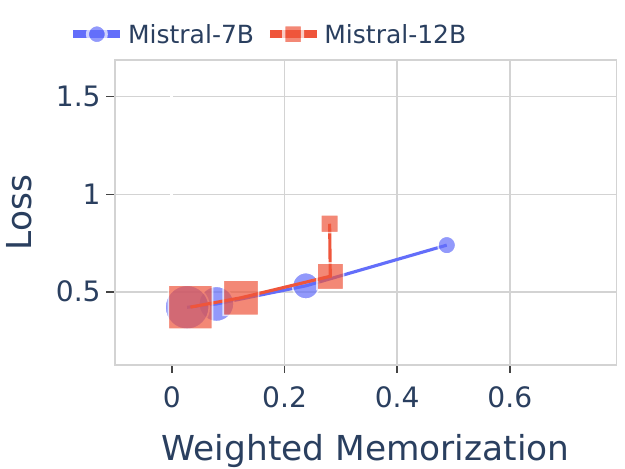}
    }
    
    \subfloat[Llama-$ 2 $]{
        \includegraphics[scale=0.4]{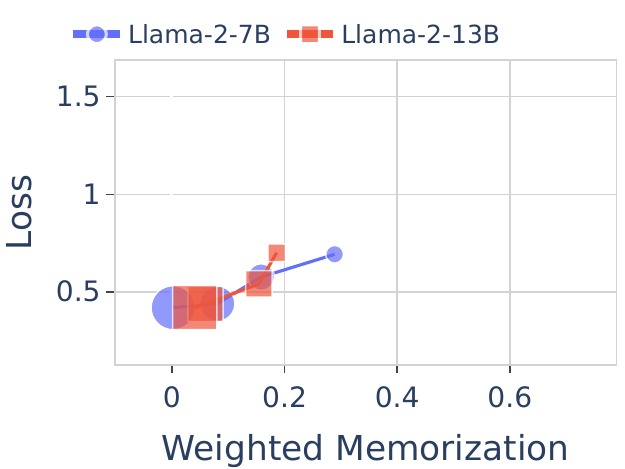}
    }
    \subfloat[Llama-$ 3 $]{
        \includegraphics[scale=0.4]{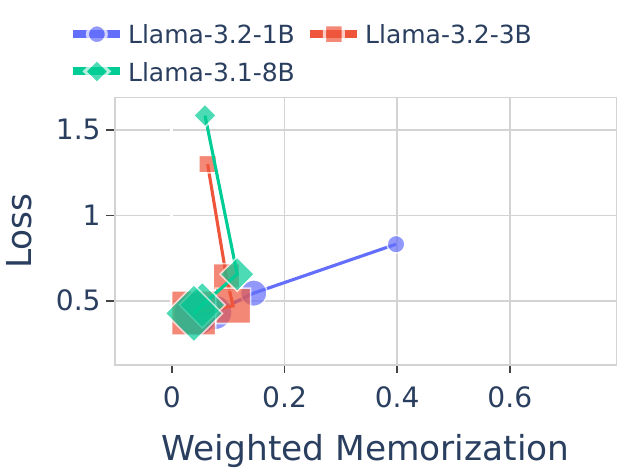}
    }
    
    \subfloat[Gemma-$ 2 $]{
        \includegraphics[scale=0.4]{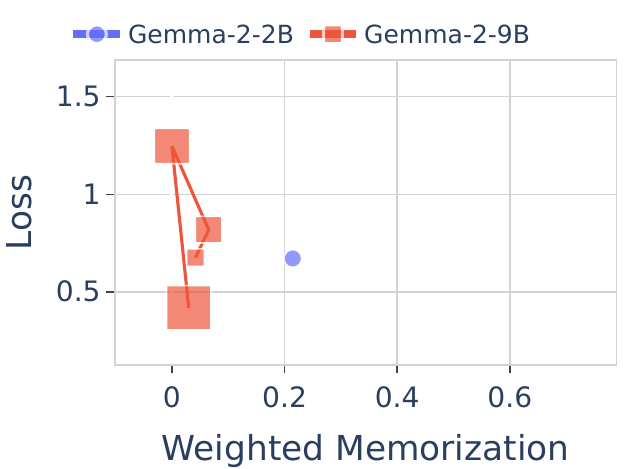}
    }
    \subfloat[Opt]{
        \includegraphics[scale=0.4]{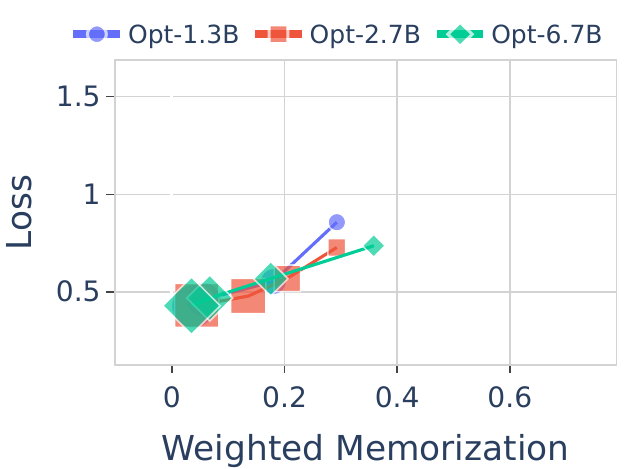}
    }
    
    \subfloat[Pythia]{
        \includegraphics[scale=0.4]{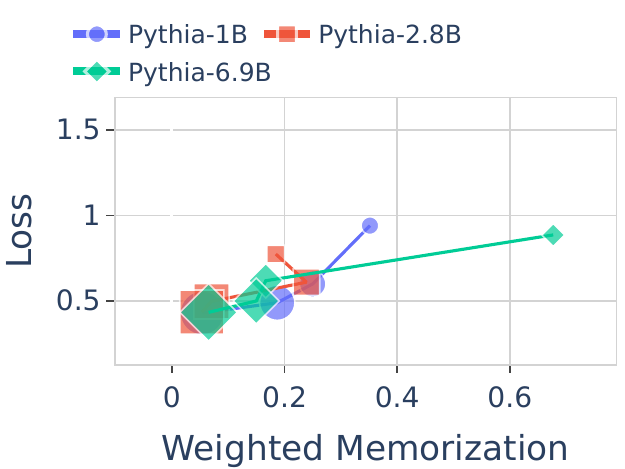}
    }

    \caption{Contextual memorization vs.\ optimal language learning, measured as test loss, across models of different sizes within a family. Results are on language $ L_3 $, which is a high entropy language.}
\end{figure}

\begin{figure}
    \centering
    \captionsetup[subfigure]{justification=centering}

    \subfloat[Qwen-$ 2.5 $]{
        \includegraphics[scale=0.4]{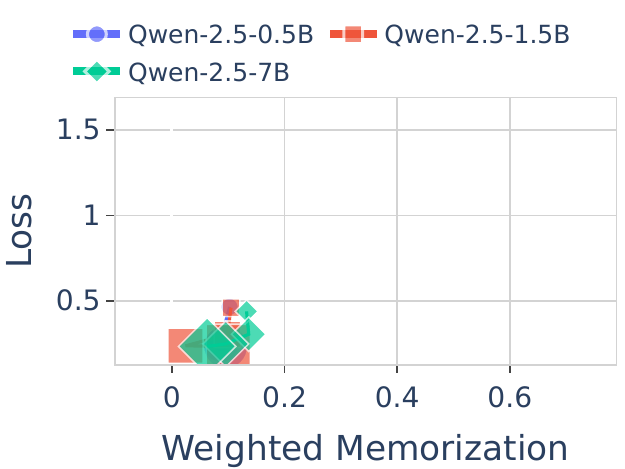}
    }
    \subfloat[Mistral]{
        \includegraphics[scale=0.4]{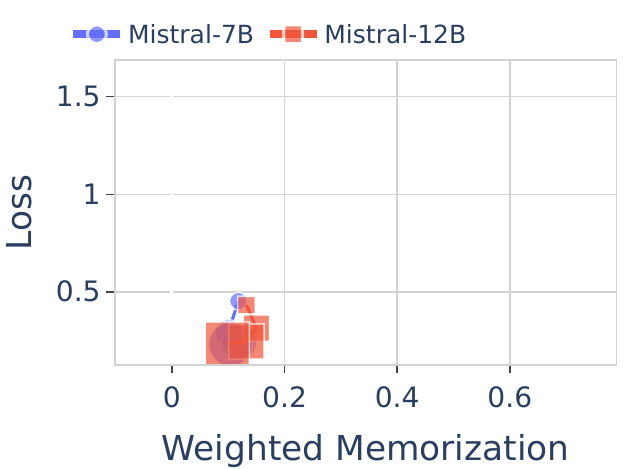}
    }
    
    \subfloat[Llama-$ 2 $]{
        \includegraphics[scale=0.4]{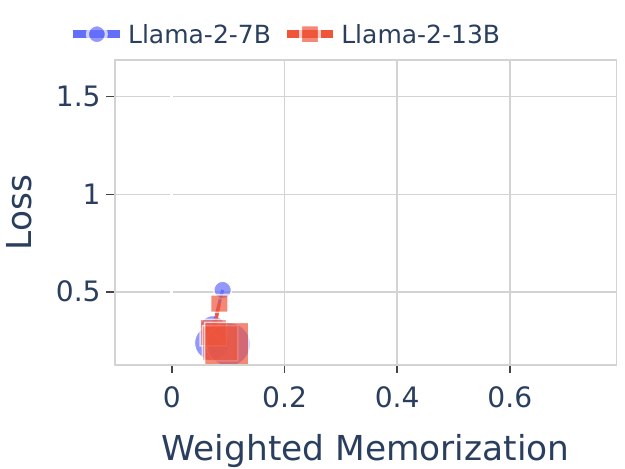}
    }
    \subfloat[Llama-$ 3 $]{
        \includegraphics[scale=0.4]{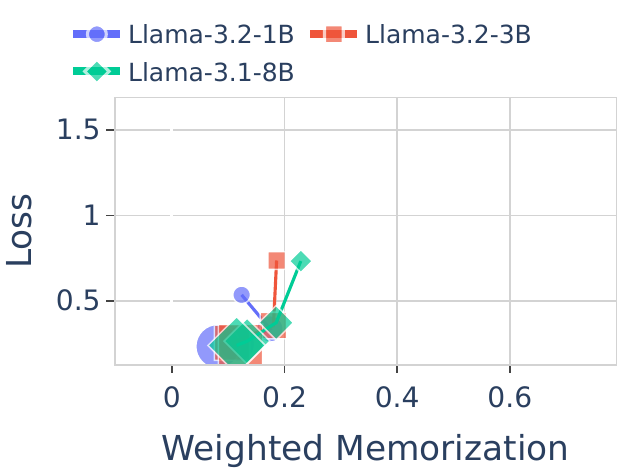}
    }
    
    \subfloat[Gemma-$ 2 $]{
        \includegraphics[scale=0.4]{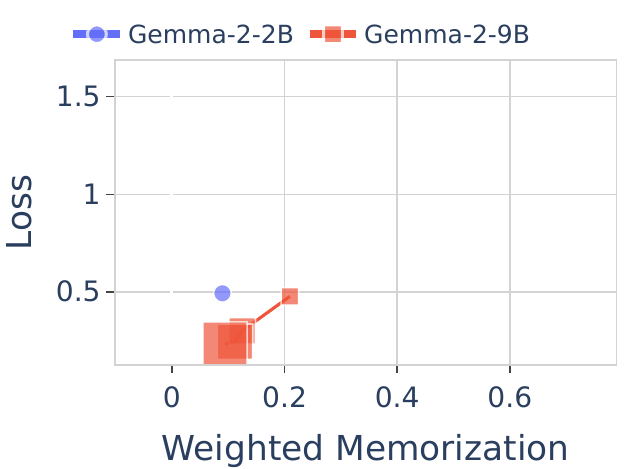}
    }
    \subfloat[Opt]{
        \includegraphics[scale=0.4]{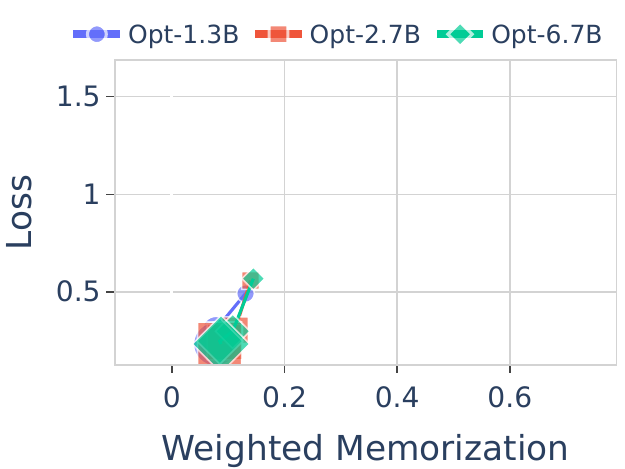}
    }
    
    \subfloat[Pythia]{
        \includegraphics[scale=0.4]{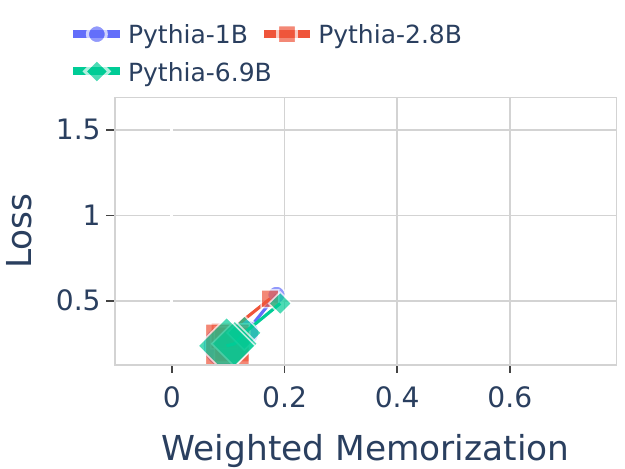}
    }

    \caption{Contextual memorization vs.\ optimal language learning, measured as test loss, across models of different sizes within a family. Results are on language $ L_4 $, which is a low entropy language.}
\end{figure}

\begin{figure}
    \centering
    \captionsetup[subfigure]{justification=centering}

    \subfloat[Qwen-$ 2.5 $]{
        \includegraphics[scale=0.4]{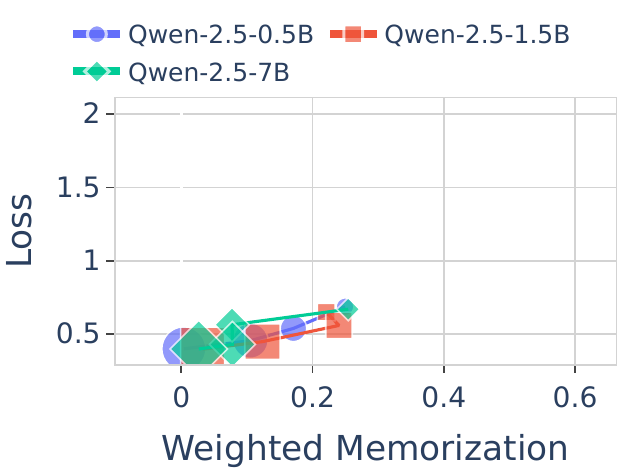}
    }
    \subfloat[Mistral]{
        \includegraphics[scale=0.4]{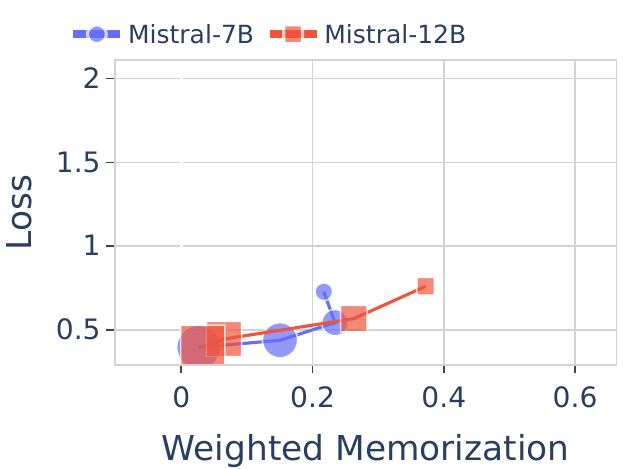}
    }
    
    \subfloat[Llama-$ 2 $]{
        \includegraphics[scale=0.4]{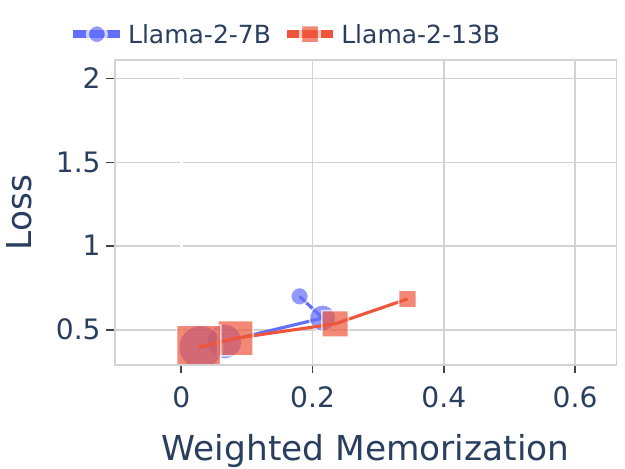}
    }
    \subfloat[Llama-$ 3 $]{
        \includegraphics[scale=0.4]{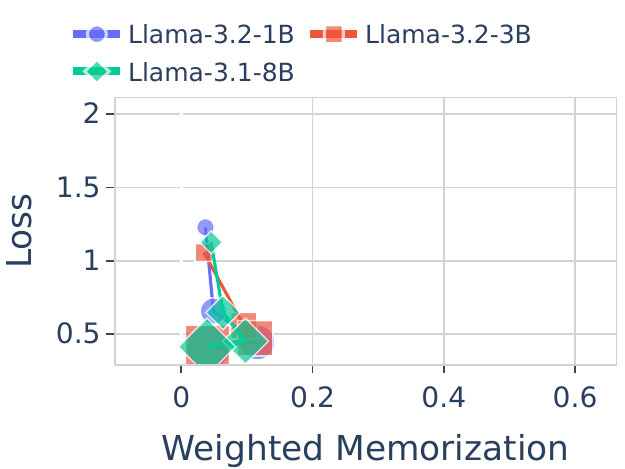}
    }
    
    \subfloat[Gemma-$ 2 $]{
        \includegraphics[scale=0.4]{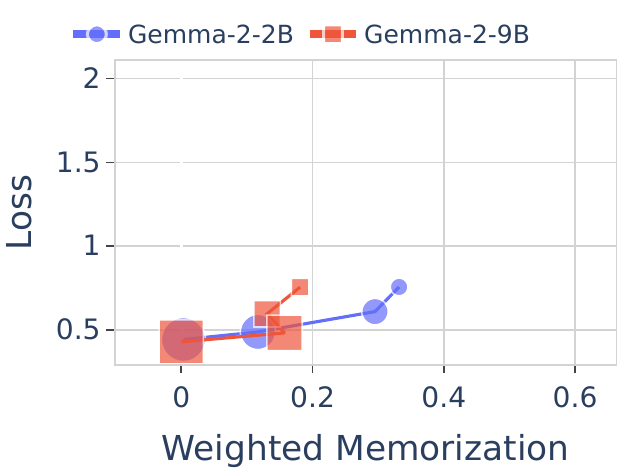}
    }
    \subfloat[Opt]{
        \includegraphics[scale=0.4]{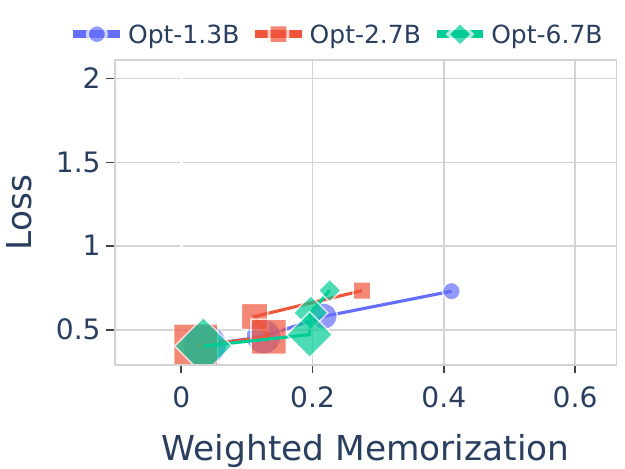}
    }
    
    \subfloat[Pythia]{
        \includegraphics[scale=0.4]{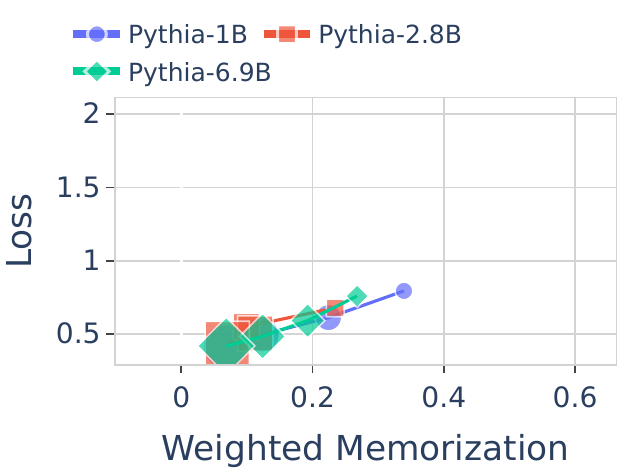}
    }

    \caption{Contextual memorization vs.\ optimal language learning, measured as test loss, across models of different sizes within a family. Results are on language $ L_5 $, which is a high entropy language.}
\end{figure}

\begin{figure}
    \centering
    \captionsetup[subfigure]{justification=centering}

    \subfloat[Qwen-$ 2.5 $]{
        \includegraphics[scale=0.4]{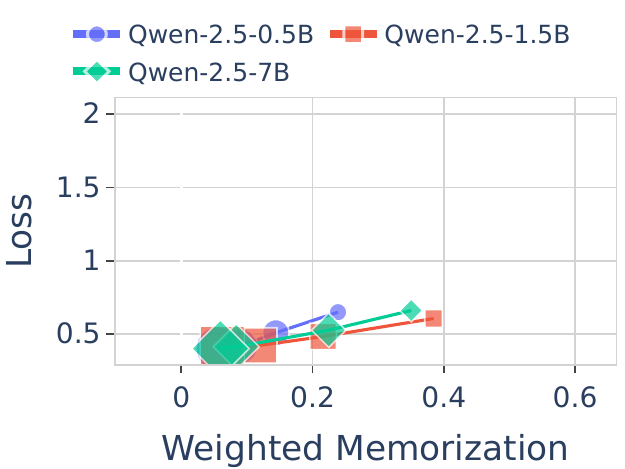}
    }
    \subfloat[Mistral]{
        \includegraphics[scale=0.4]{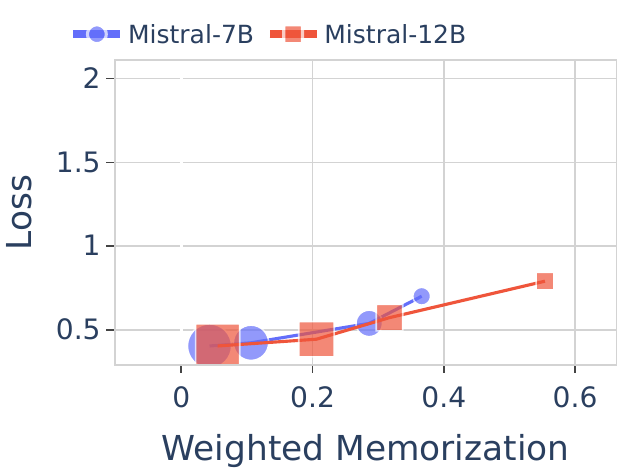}
    }
    
    \subfloat[Llama-$ 2 $]{
        \includegraphics[scale=0.4]{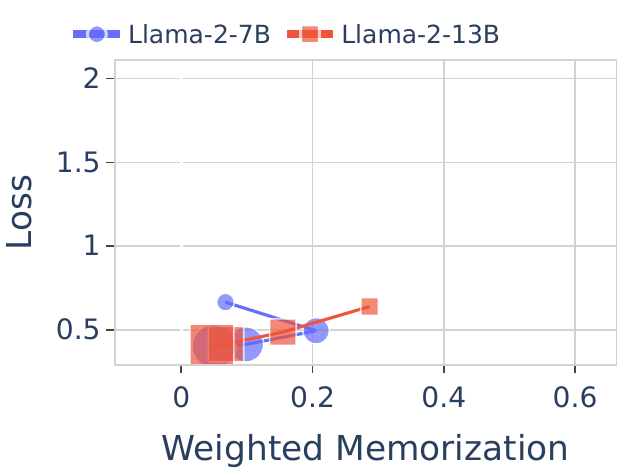}
    }
    \subfloat[Llama-$ 3 $]{
        \includegraphics[scale=0.4]{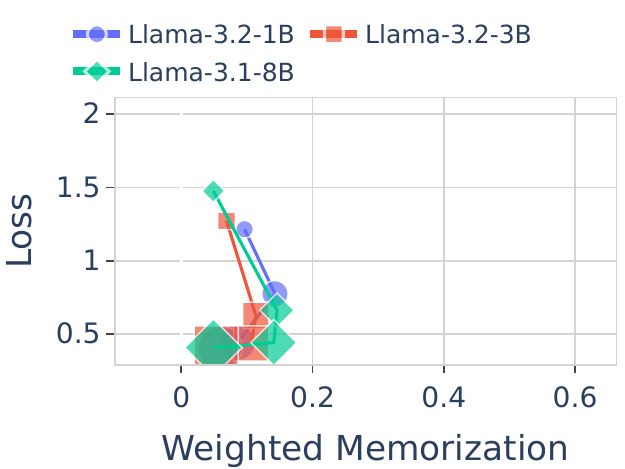}
    }
    
    \subfloat[Gemma-$ 2 $]{
        \includegraphics[scale=0.4]{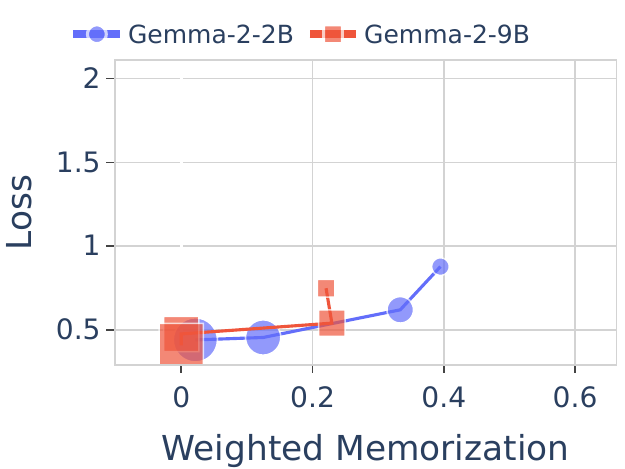}
    }
    \subfloat[Opt]{
        \includegraphics[scale=0.4]{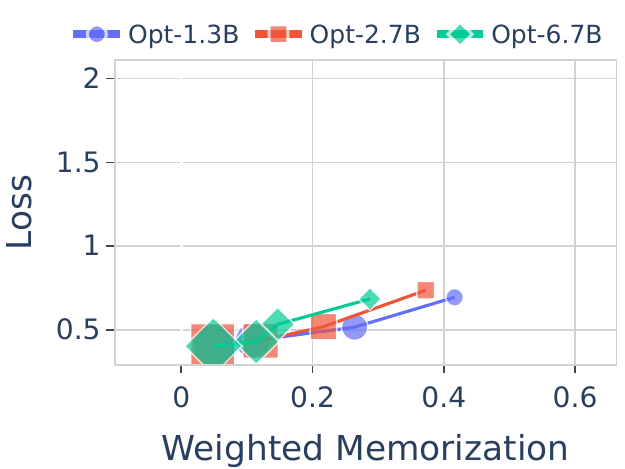}
    }
    
    \subfloat[Pythia]{
        \includegraphics[scale=0.4]{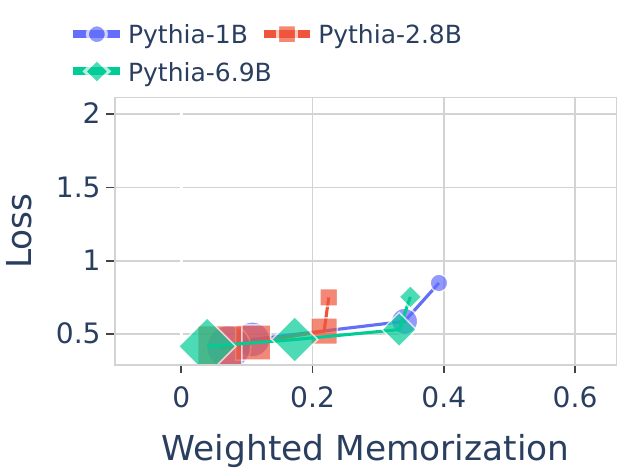}
    }

    \caption{Contextual memorization vs.\ optimal language learning, measured as test loss, across models of different sizes within a family. Results are on language $ L_6 $, which is a high entropy language.}
\end{figure}

\begin{figure}
    \centering
    \captionsetup[subfigure]{justification=centering}

    \subfloat[Qwen-$ 2.5 $]{
        \includegraphics[scale=0.4]{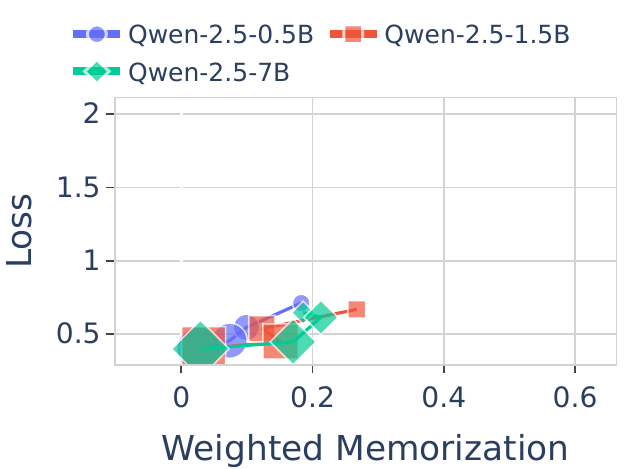}
    }
    \subfloat[Mistral]{
        \includegraphics[scale=0.4]{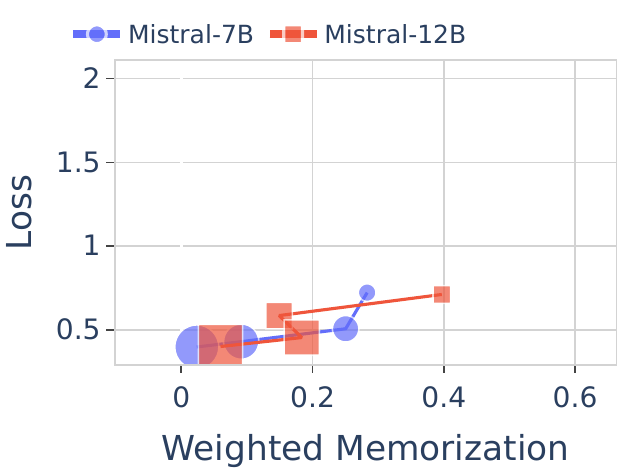}
    }
    
    \subfloat[Llama-$ 2 $]{
        \includegraphics[scale=0.4]{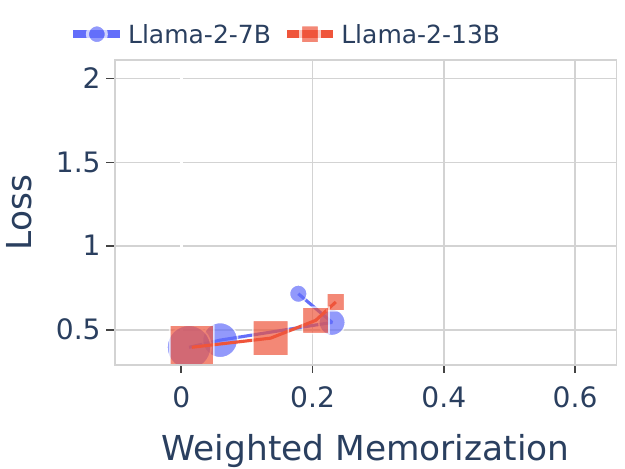}
    }
    \subfloat[Llama-$ 3 $]{
        \includegraphics[scale=0.4]{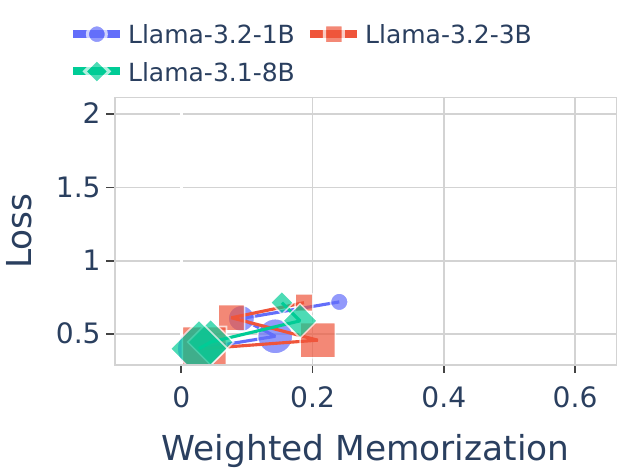}
    }
    
    \subfloat[Gemma-$ 2 $]{
        \includegraphics[scale=0.4]{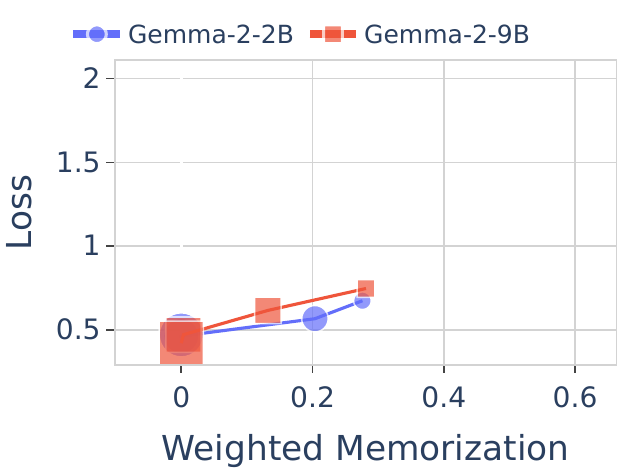}
    }
    \subfloat[Opt]{
        \includegraphics[scale=0.4]{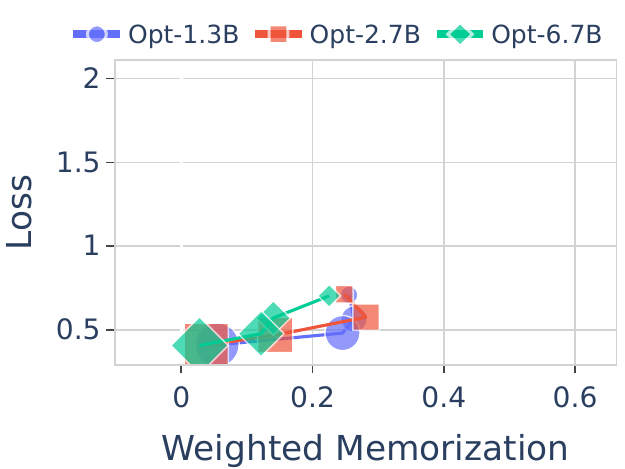}
    }
    
    \subfloat[Pythia]{
        \includegraphics[scale=0.4]{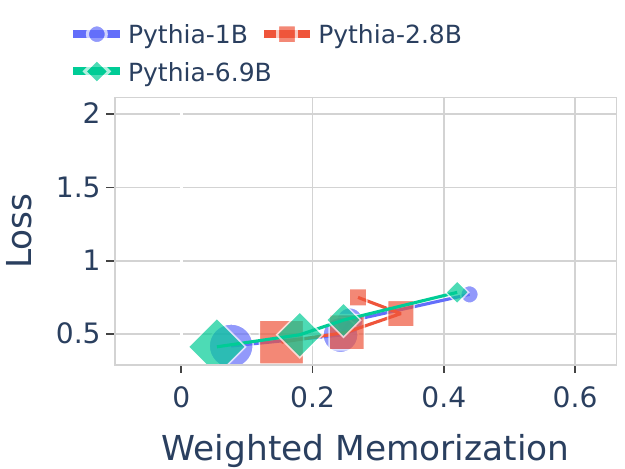}
    }

    \caption{Contextual memorization vs.\ optimal language learning, measured as test loss, across models of different sizes within a family. Results are on language $ L_7 $, which is a high entropy language and contains Latin characters as tokens.}
\end{figure}

\begin{figure}
    \centering
    \captionsetup[subfigure]{justification=centering}

    \subfloat[Qwen-$ 2.5 $]{
        \includegraphics[scale=0.4]{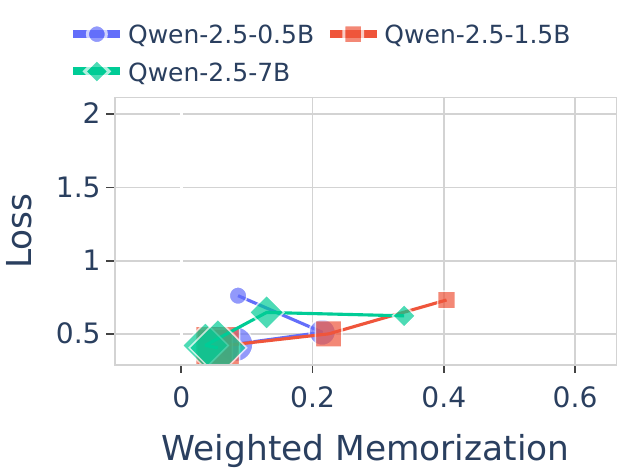}
    }
    \subfloat[Mistral]{
        \includegraphics[scale=0.4]{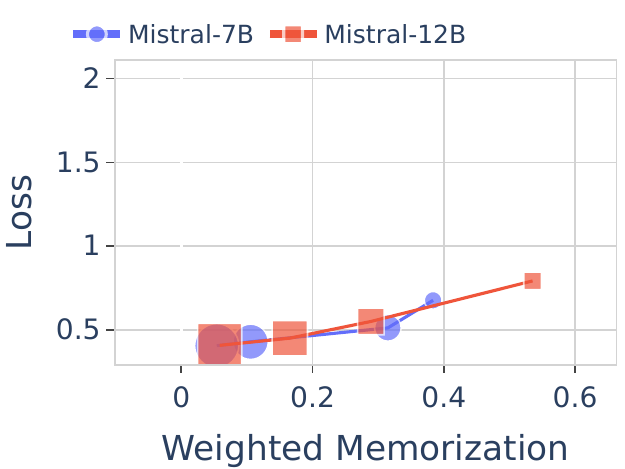}
    }
    
    \subfloat[Llama-$ 2 $]{
        \includegraphics[scale=0.4]{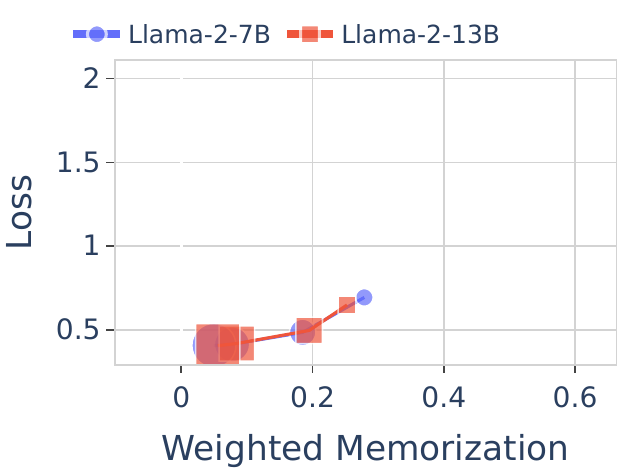}
    }
    \subfloat[Llama-$ 3 $]{
        \includegraphics[scale=0.4]{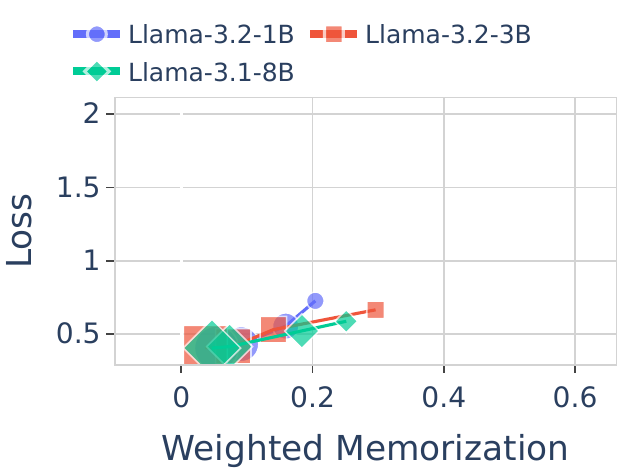}
    }
    
    \subfloat[Gemma-$ 2 $]{
        \includegraphics[scale=0.4]{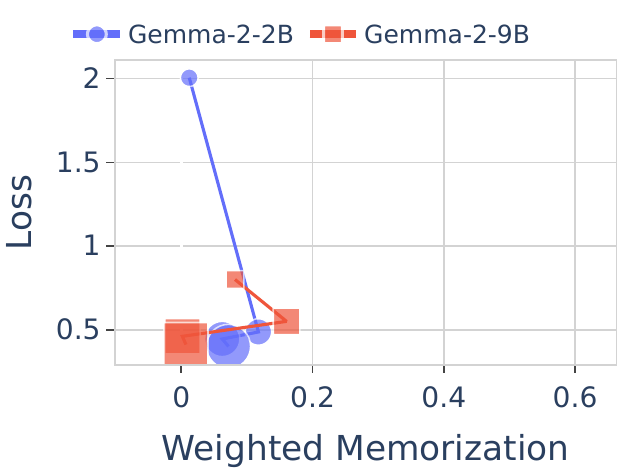}
    }
    \subfloat[Opt]{
        \includegraphics[scale=0.4]{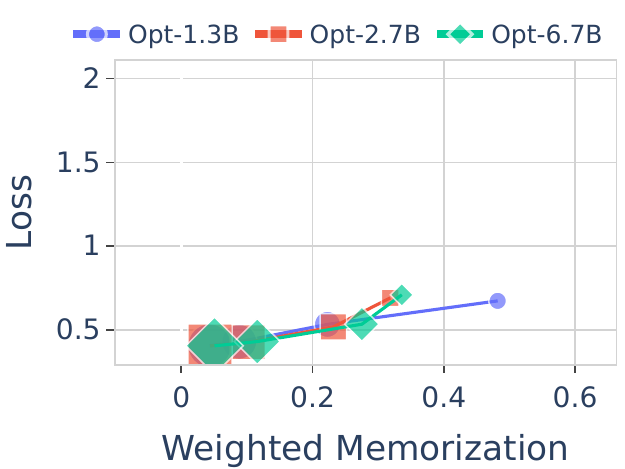}
    }
    
    \subfloat[Pythia]{
        \includegraphics[scale=0.4]{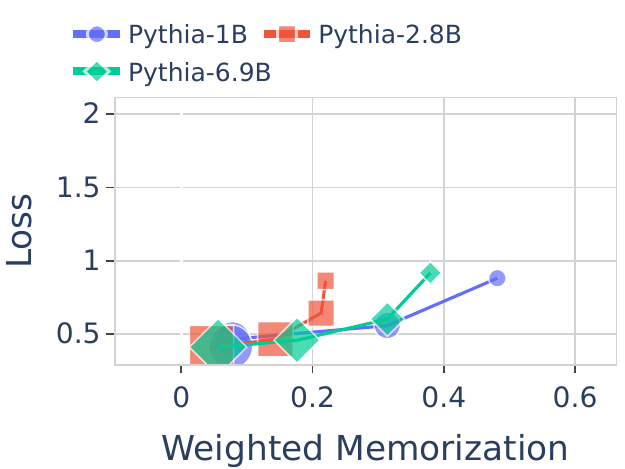}
    }

    \caption{Contextual memorization vs.\ optimal language learning, measured as test loss, across models of different sizes within a family. Results are on language $ L_8 $, which is a high entropy language and contains Latin characters as tokens.}
\end{figure}

\clearpage
\section{Memorization-aware Training}

We propose MAT and MAT*, two memorization-aware training schemes that improve the stability and uniformity of learning. In both schemes, training strings are down-sampled once they are learned, and re-added if subsequently forgotten. MAT uses a simple heuristic: down-sampling strings with the lowest training loss. MAT* down-samples a string when its loss falls below the lowest contextual loss, approximated using similarly-frequent test strings. In practice, the down-sampling fraction is determined by monitoring learning stability on the test set, or equivalently a held-out validation set. MAT down-samples frequent strings more, as they tend to have lower losses. MAT* is more effective at down-sampling strings across all frequencies, as studied in~\ref{itm:rq_mem_learn_2}. Consequently, MAT stabilizes learning (test loss does not rise past its local minimum) for higher-frequency strings, while MAT* stabilizes learning across all frequency groups. As a byproduct, both schemes reduce memorization and training cost.
\label{sec:mat_appendix}
\subsection{Performance on Fine-tuning}
\label{sec:mat_performance_finetuning}
\begin{figure}[h]
    \raggedleft

    \subfloat[Base Training]{
        \includegraphics[scale=0.35]{memorization_figures/deduplication/ft_train_all_pcfg_cfg3b_eq_len_skewed_prob_mistral-7b_256_0_base_epoch_mem_intervention.pdf}
    }
    \subfloat[Base Training]{
        \includegraphics[scale=0.35]{memorization_figures/deduplication/ft_test_all_pcfg_cfg3b_eq_len_skewed_prob_mistral-7b_256_0_base_epoch_mem_intervention.pdf}
    }
    \subfloat[Base Training]{
        \includegraphics[scale=0.35]{memorization_figures/deduplication/memorization_weighted_pcfg_cfg3b_eq_len_skewed_prob_mistral-7b_256_0_base_contextual_epoch_mem_intervention.pdf}
    }

    \subfloat[Dedup]{
        \includegraphics[scale=0.35]{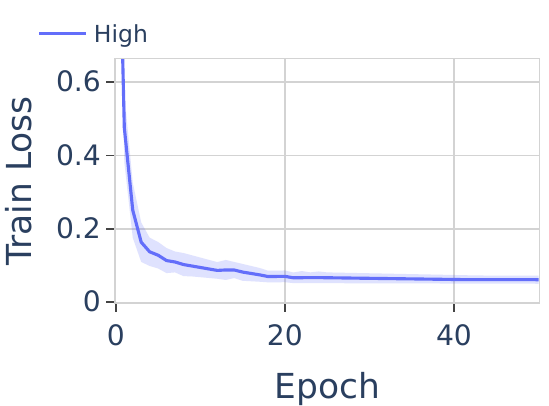}
    }
    \subfloat[Dedup]{
        \includegraphics[scale=0.35]{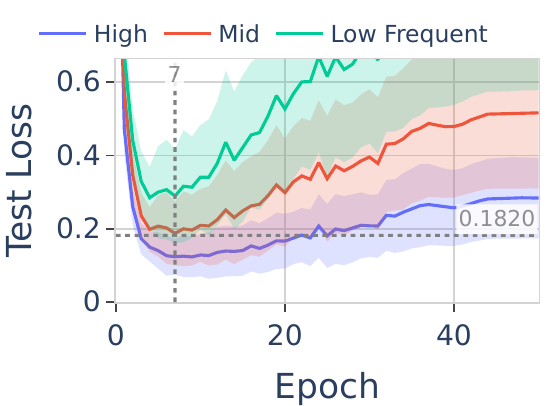}
        \label{fig:mat_training_dynamics_dedup_test_loss_appendix}
    }
    \subfloat[Dedup]{
        \includegraphics[scale=0.35]{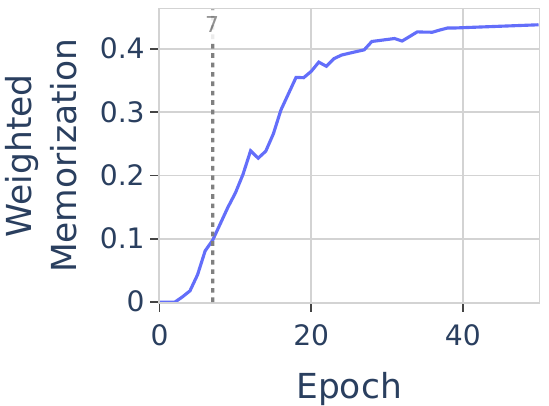}
    }

    \subfloat[MAT]{
        \includegraphics[scale=0.35]{memorization_figures/deduplication/training_pruning_group_ratio_pcfg_cfg3b_eq_len_skewed_prob_mistral-7b_256_mat_0_epoch_mem_intervention.pdf}
    }
    \subfloat[MAT]{
        \includegraphics[scale=0.35]{memorization_figures/deduplication/ft_train_all_pcfg_cfg3b_eq_len_skewed_prob_mistral-7b_256_0_mat_epoch_mem_intervention.pdf}
    }
    \subfloat[MAT]{
        \includegraphics[scale=0.35]{memorization_figures/deduplication/ft_test_all_pcfg_cfg3b_eq_len_skewed_prob_mistral-7b_256_0_mat_epoch_mem_intervention.pdf}
    }
    \subfloat[MAT]{
        \includegraphics[scale=0.35]{memorization_figures/deduplication/memorization_weighted_pcfg_cfg3b_eq_len_skewed_prob_mistral-7b_256_0_mat_contextual_epoch_mem_intervention.pdf}
    }

    \subfloat[MAT*]{
        \includegraphics[scale=0.35]{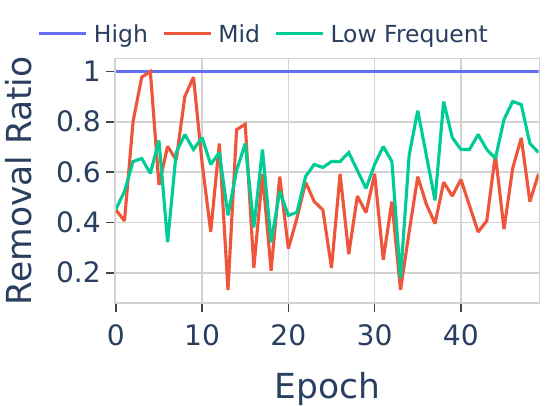}
    }
    \subfloat[MAT*]{
        \includegraphics[scale=0.35]{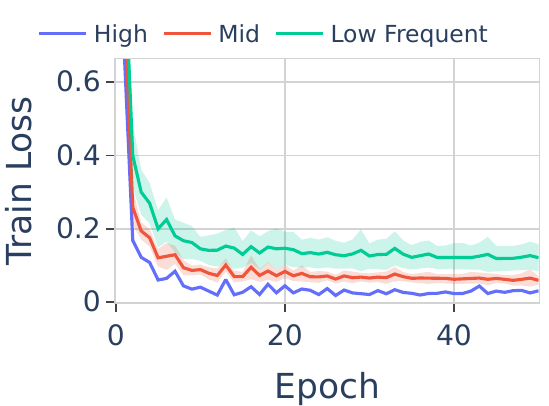}
    }
    \subfloat[MAT*]{
        \includegraphics[scale=0.35]{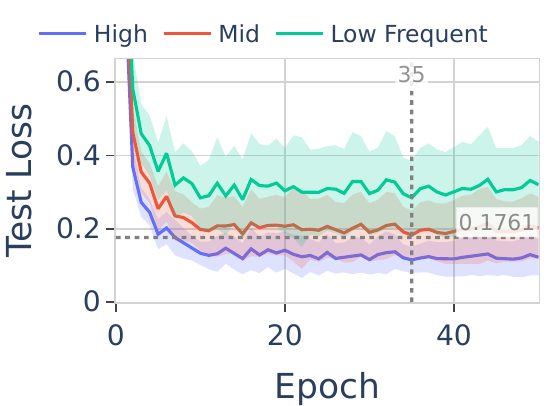}
    }
    \subfloat[MAT*]{
        \includegraphics[scale=0.35]{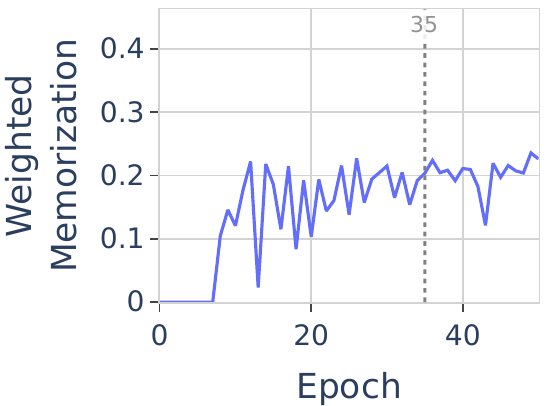}
    }

    \caption{Training dynamics under Base Training, Dedup, MAT, and MAT* on language $L_2$ using Mistral-$7$B (seed $0$). Each row corresponds to a training scheme; columns show train loss, test loss, and contextual memorization by frequency group across epochs. For MAT and MAT*, the fraction of training strings removed per group is shown leftmost. Vertical lines mark the optimal learning epoch. In the case of deduplication, all the training strings have the same frequency, hereby grouped as high-frequency.}

\end{figure}

\begin{figure}
    \raggedleft

    \subfloat[Base Training]{
        \includegraphics[scale=0.35]{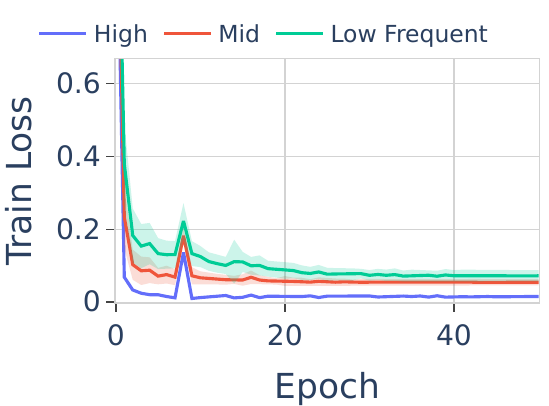}
    }
    \subfloat[Base Training]{
        \includegraphics[scale=0.35]{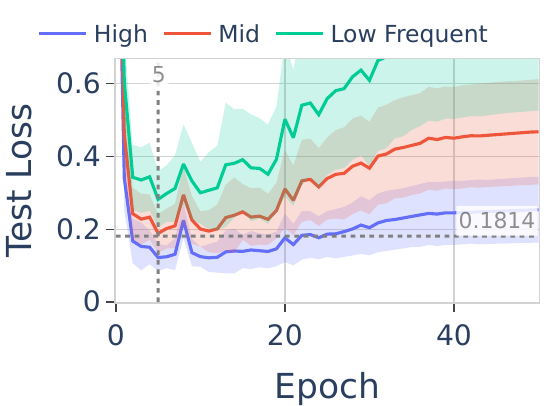}
    }
    \subfloat[Base Training]{
        \includegraphics[scale=0.35]{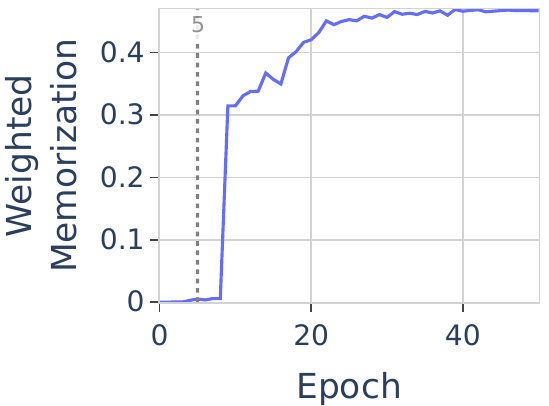}
    }

    \subfloat[Dedup]{
        \includegraphics[scale=0.35]{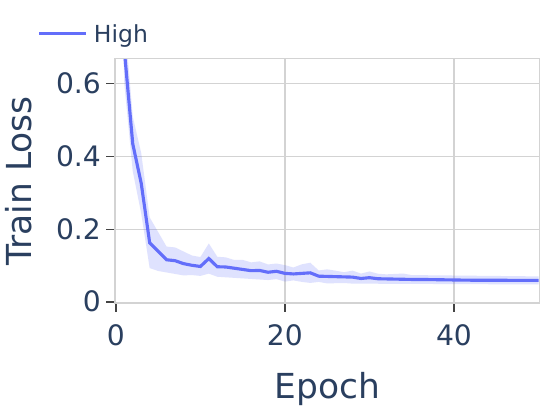}
    }
    \subfloat[Dedup]{
        \includegraphics[scale=0.35]{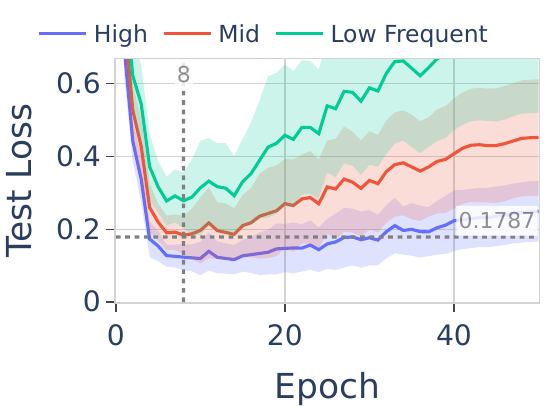}
    }
    \subfloat[Dedup]{
        \includegraphics[scale=0.35]{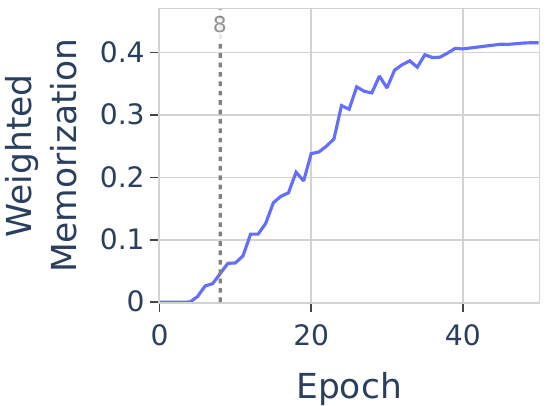}
    }

    \subfloat[MAT]{
        \includegraphics[scale=0.35]{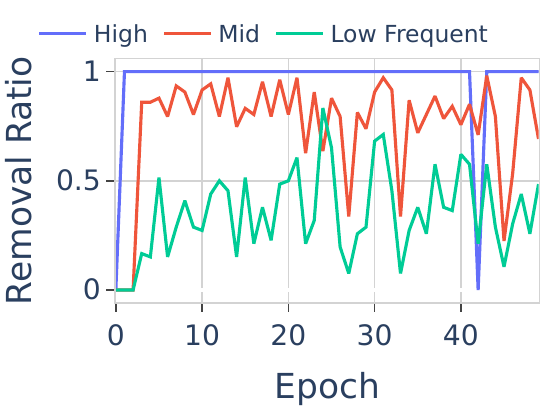}
    }
    \subfloat[MAT]{
        \includegraphics[scale=0.35]{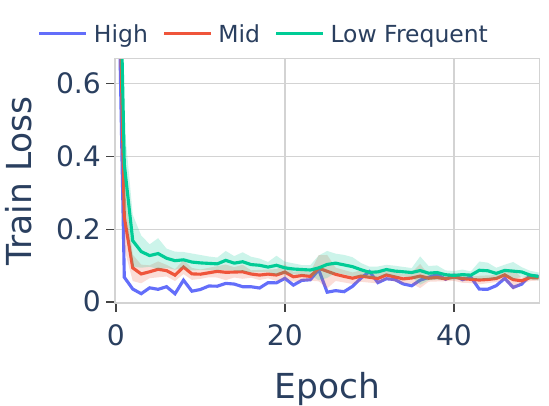}
    }
    \subfloat[MAT]{
        \includegraphics[scale=0.35]{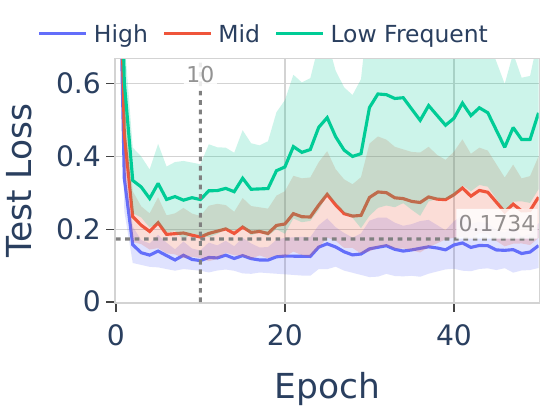}
    }
    \subfloat[MAT]{
        \includegraphics[scale=0.35]{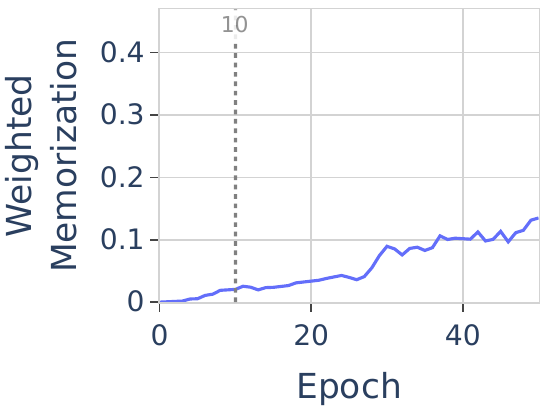}
    }

    \subfloat[MAT*]{
        \includegraphics[scale=0.35]{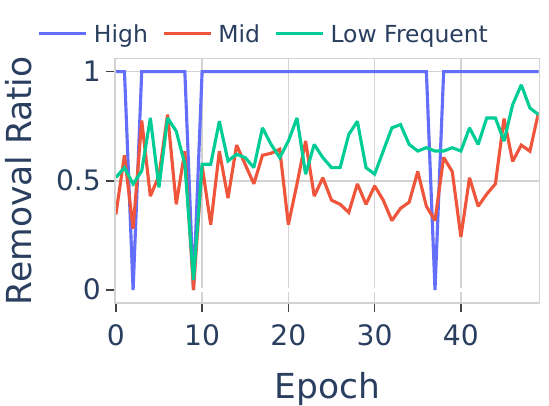}
    }
    \subfloat[MAT*]{
        \includegraphics[scale=0.35]{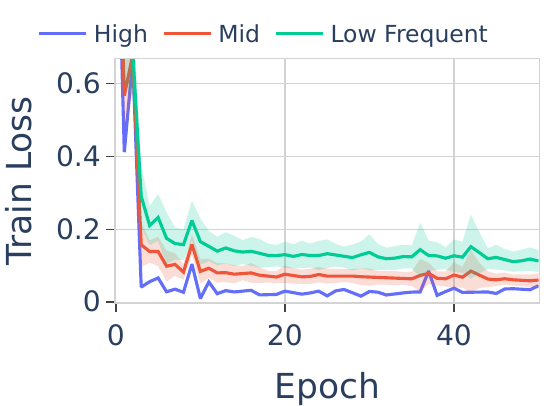}
    }
    \subfloat[MAT*]{
        \includegraphics[scale=0.35]{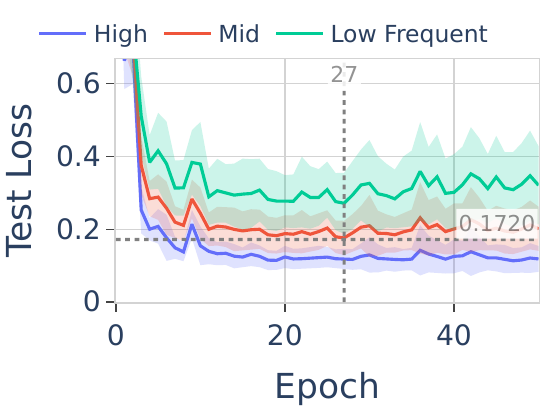}
    }
    \subfloat[MAT*]{
        \includegraphics[scale=0.35]{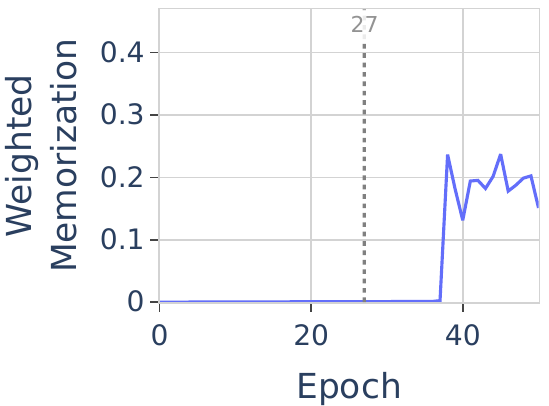}
    }

    \caption{Training dynamics under Base Training, Dedup, MAT, and MAT* on language $L_2$ using Mistral-$7$B (seed $1$). Each row corresponds to a training scheme; columns show train loss, test loss, and contextual memorization by frequency group across epochs. For MAT and MAT*, the fraction of training strings removed per group is shown leftmost. Vertical lines mark the optimal learning epoch. In the case of deduplication, all the training strings have the same frequency, hereby grouped as high-frequency.}

\end{figure}

\begin{figure}
    \raggedleft

    \subfloat[Base Training]{
        \includegraphics[scale=0.35]{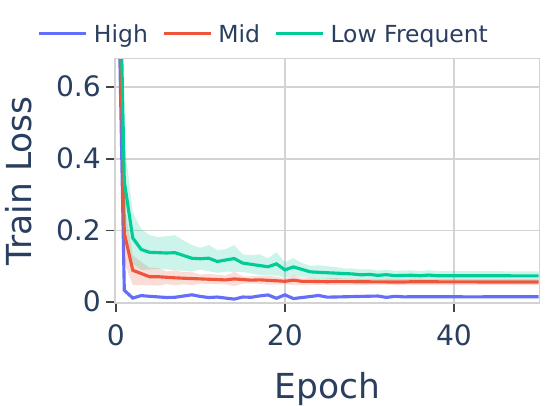}
    }
    \subfloat[Base Training]{
        \includegraphics[scale=0.35]{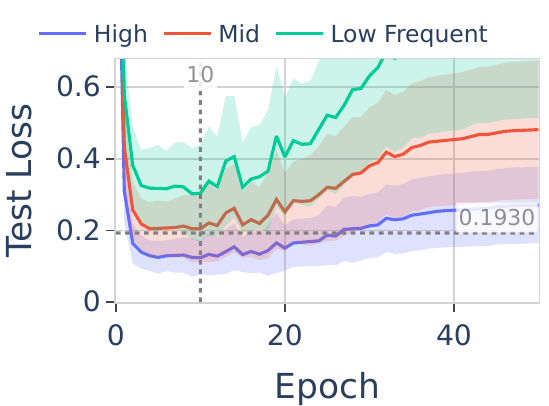}
    }
    \subfloat[Base Training]{
        \includegraphics[scale=0.35]{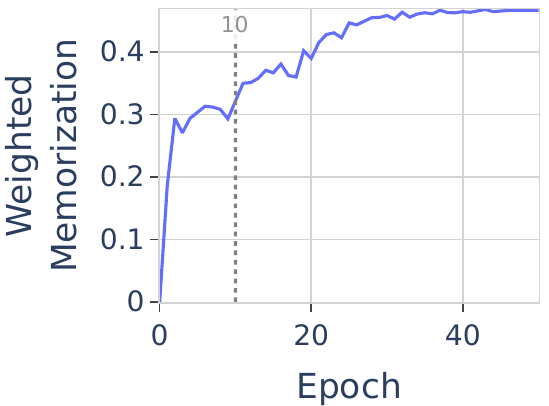}
    }

    \subfloat[Dedup]{
        \includegraphics[scale=0.35]{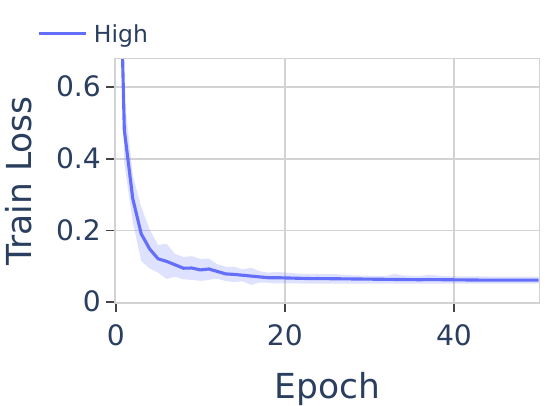}
    }
    \subfloat[Dedup]{
        \includegraphics[scale=0.35]{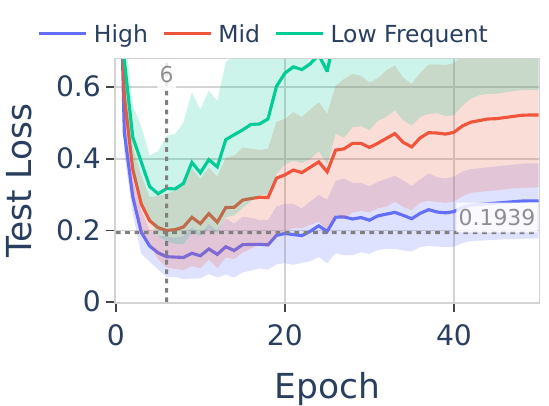}
    }
    \subfloat[Dedup]{
        \includegraphics[scale=0.35]{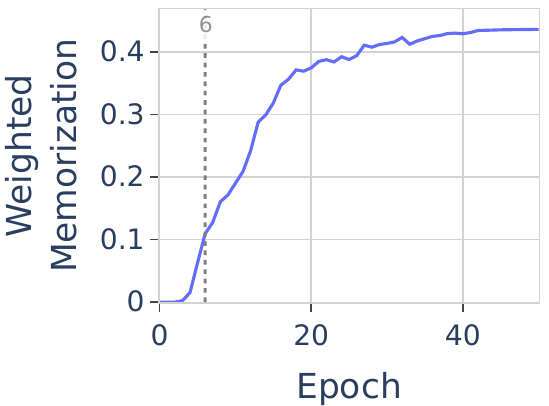}
    }

    \subfloat[MAT]{
        \includegraphics[scale=0.35]{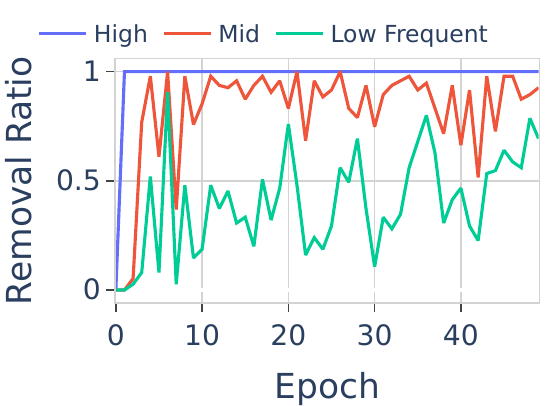}
    }
    \subfloat[MAT]{
        \includegraphics[scale=0.35]{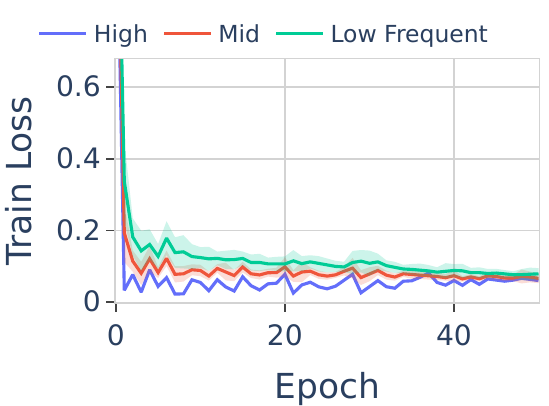}
    }
    \subfloat[MAT]{
        \includegraphics[scale=0.35]{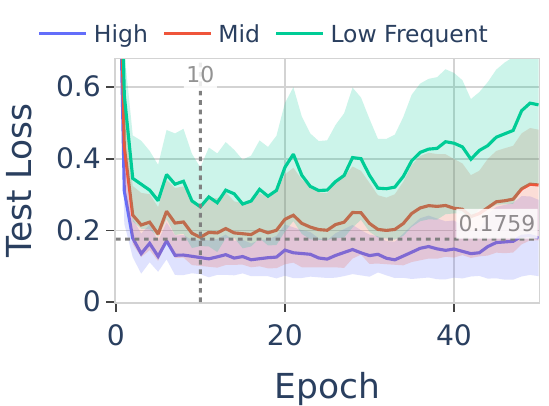}
    }
    \subfloat[MAT]{
        \includegraphics[scale=0.35]{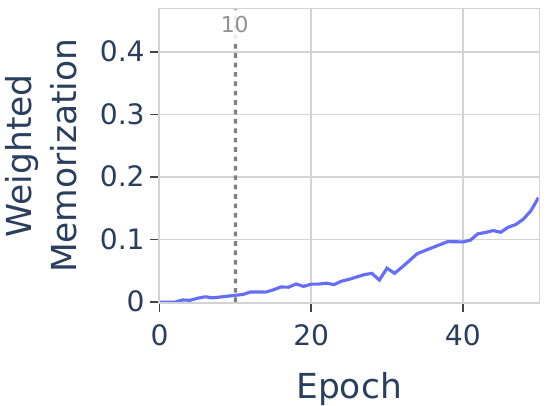}
    }

    \subfloat[MAT*]{
        \includegraphics[scale=0.35]{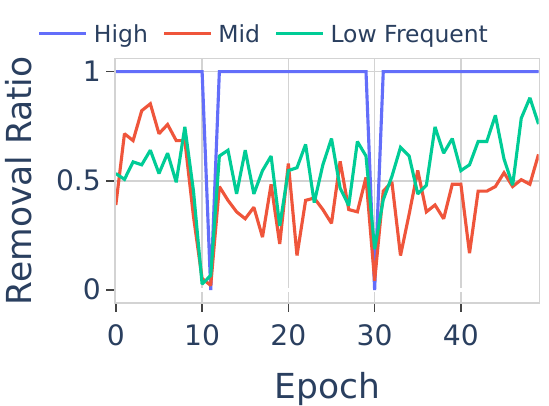}
    }
    \subfloat[MAT*]{
        \includegraphics[scale=0.35]{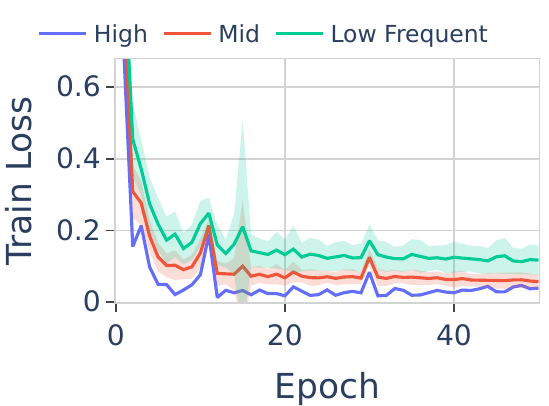}
    }
    \subfloat[MAT*]{
        \includegraphics[scale=0.35]{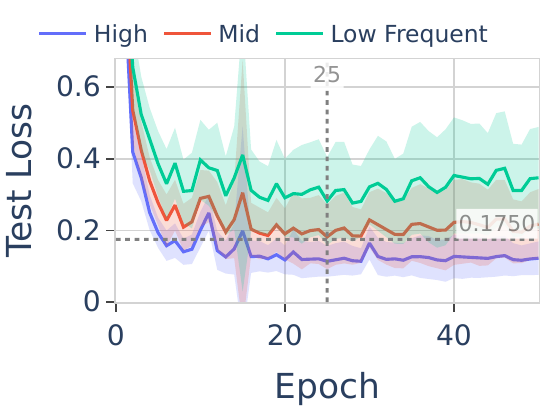}
    }
    \subfloat[MAT*]{
        \includegraphics[scale=0.35]{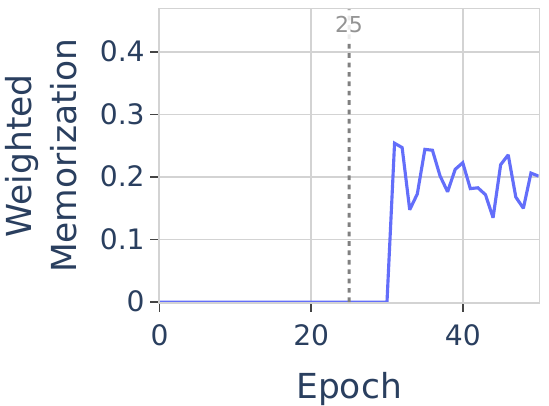}
    }

    \caption{Training dynamics under Base Training, Dedup, MAT, and MAT* on language $L_2$ using Mistral-$7$B (seed $2$). Each row corresponds to a training scheme; columns show train loss, test loss, and contextual memorization by frequency group across epochs. For MAT and MAT*, the fraction of training strings removed per group is shown leftmost. Vertical lines mark the optimal learning epoch. In the case of deduplication, all the training strings have the same frequency, hereby grouped as high-frequency.}

\end{figure}

\clearpage
\begin{figure}
    \raggedleft

    \subfloat[Base Training]{
        \includegraphics[scale=0.35]{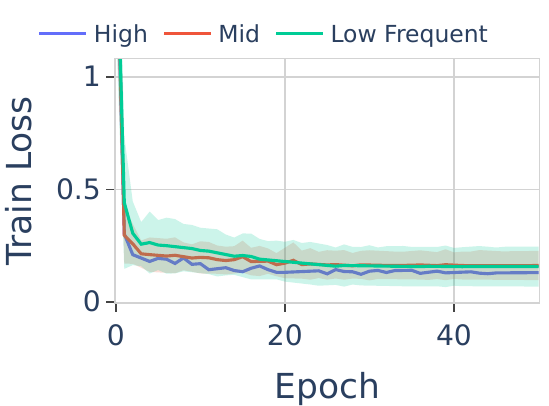}
    }
    \subfloat[Base Training]{
        \includegraphics[scale=0.35]{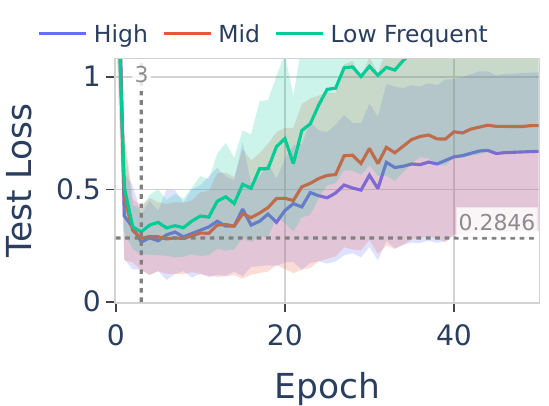}
    }
    \subfloat[Base Training]{
        \includegraphics[scale=0.35]{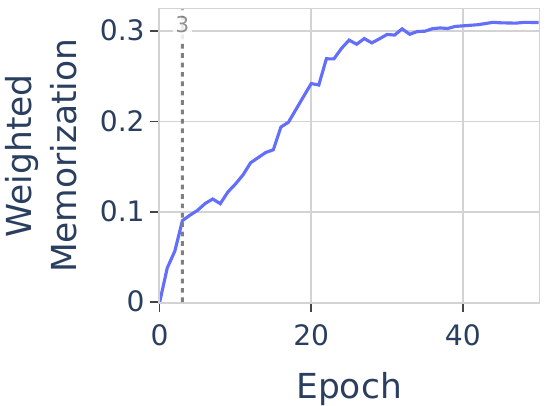}
    }

    \subfloat[Dedup]{
        \includegraphics[scale=0.35]{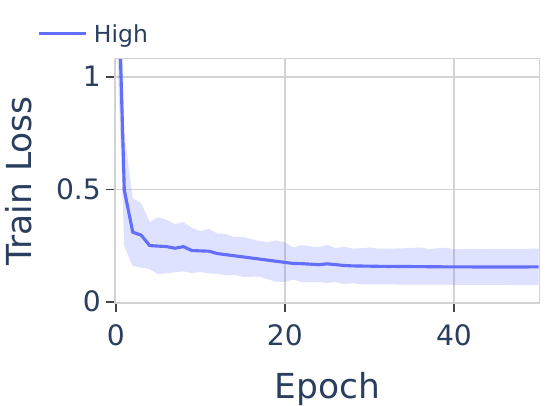}
    }
    \subfloat[Dedup]{
        \includegraphics[scale=0.35]{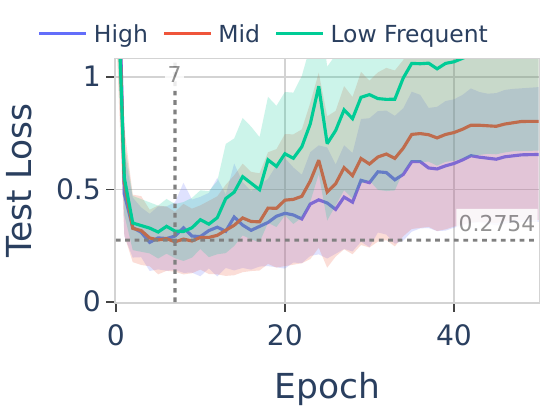}
    }
    \subfloat[Dedup]{
        \includegraphics[scale=0.35]{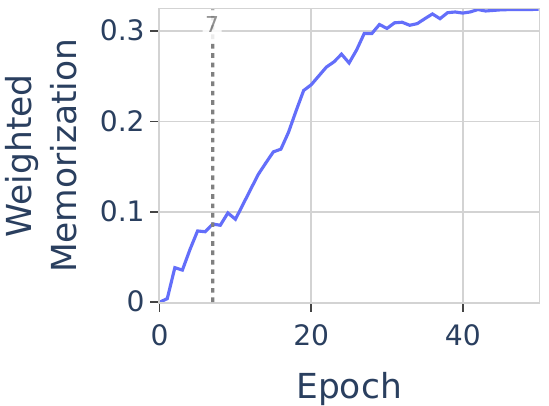}
    }

    \subfloat[MAT]{
        \includegraphics[scale=0.35]{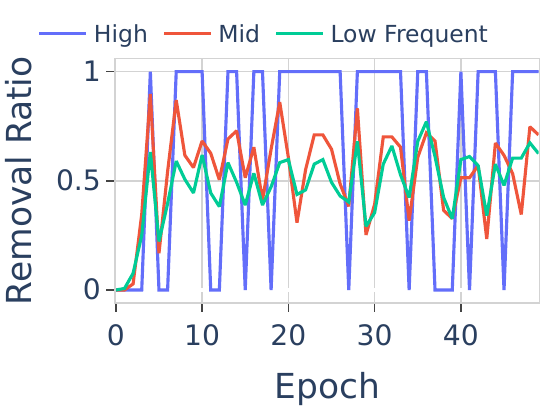}
    }
    \subfloat[MAT]{
        \includegraphics[scale=0.35]{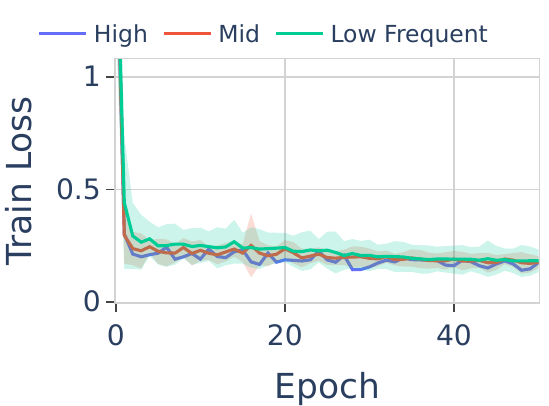}
    }
    \subfloat[MAT]{
        \includegraphics[scale=0.35]{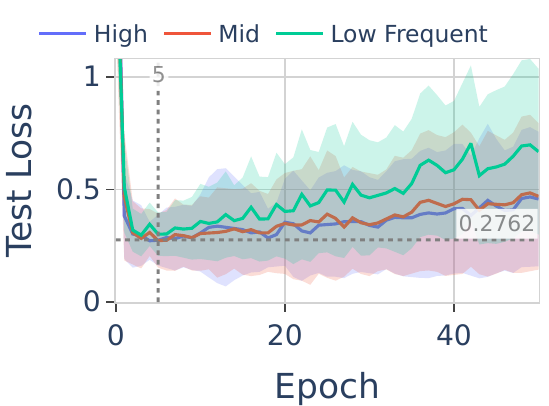}
    }
    \subfloat[MAT]{
        \includegraphics[scale=0.35]{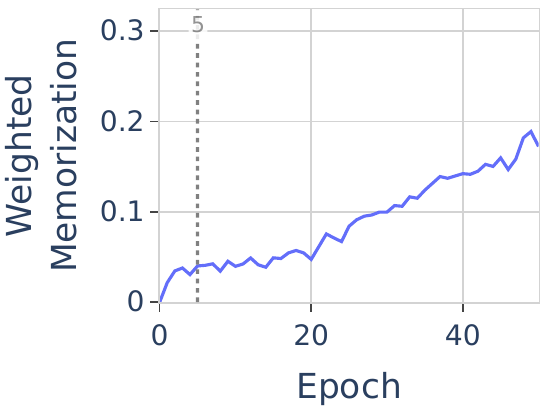}
    }

    \subfloat[MAT*]{
        \includegraphics[scale=0.35]{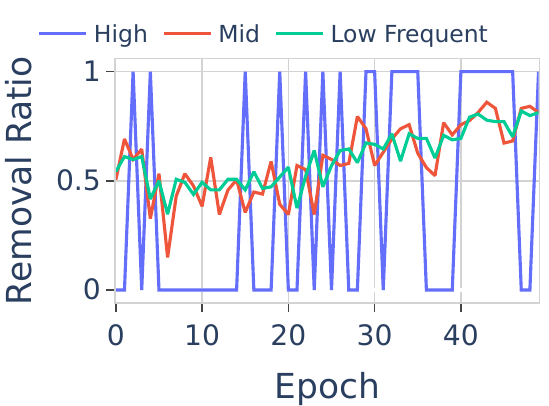}
    }
    \subfloat[MAT*]{
        \includegraphics[scale=0.35]{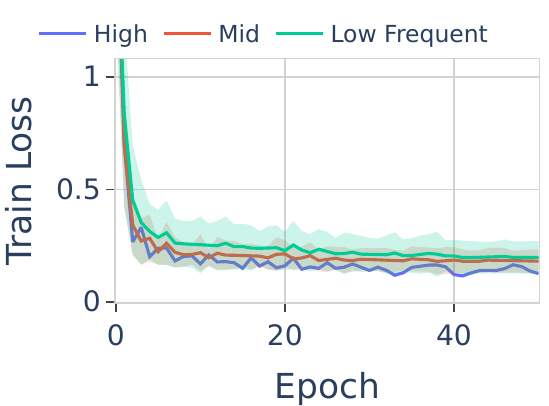}
    }
    \subfloat[MAT*]{
        \includegraphics[scale=0.35]{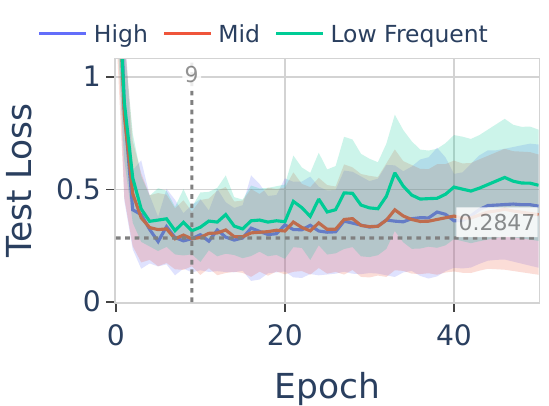}
    }
    \subfloat[MAT*]{
        \includegraphics[scale=0.35]{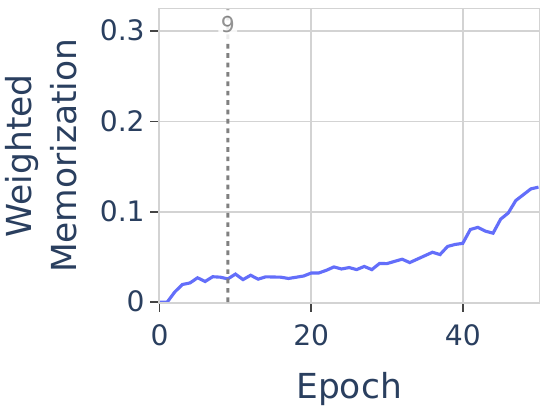}
    }

    \caption{Training dynamics under Base Training, Dedup, MAT, and MAT* on language $L_4$ using Mistral-$7$B (seed $0$). Each row corresponds to a training scheme; columns show train loss, test loss, and contextual memorization by frequency group across epochs. For MAT and MAT*, the fraction of training strings removed per group is shown leftmost. Vertical lines mark the optimal learning epoch. In the case of deduplication, all the training strings have the same frequency, hereby grouped as high-frequency.}

\end{figure}

\begin{figure}
    \raggedleft

    \subfloat[Base Training]{
        \includegraphics[scale=0.35]{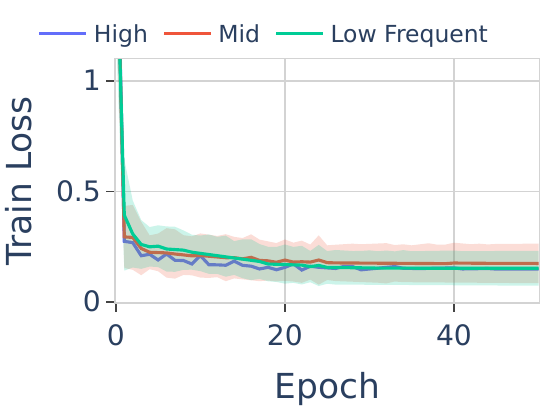}
    }
    \subfloat[Base Training]{
        \includegraphics[scale=0.35]{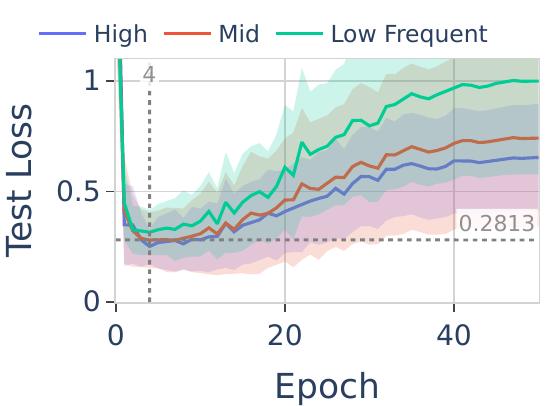}
    }
    \subfloat[Base Training]{
        \includegraphics[scale=0.35]{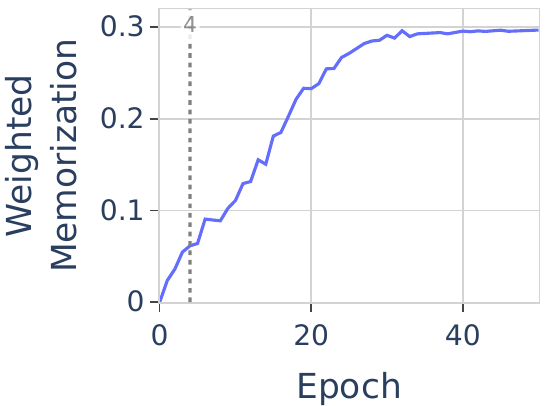}
    }

    \subfloat[Dedup]{
        \includegraphics[scale=0.35]{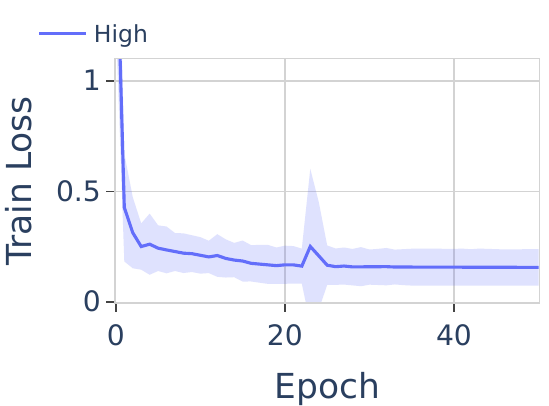}
    }
    \subfloat[Dedup]{
        \includegraphics[scale=0.35]{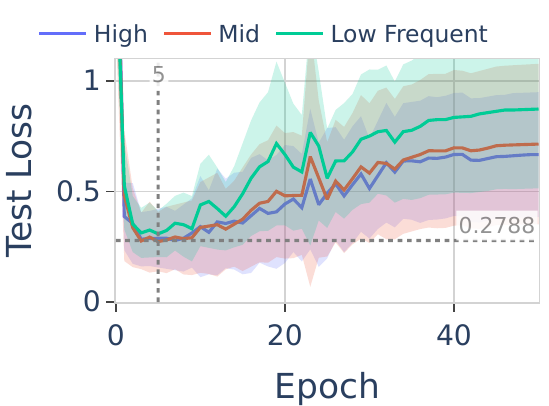}
    }
    \subfloat[Dedup]{
        \includegraphics[scale=0.35]{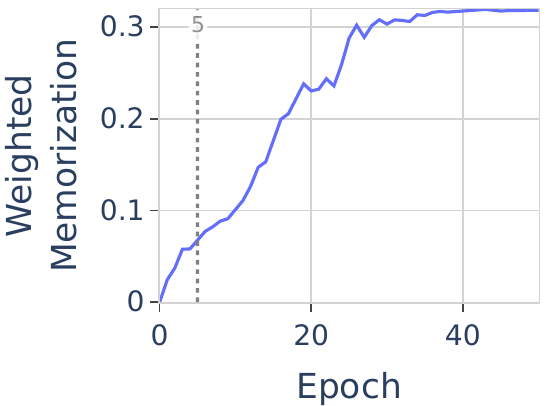}
    }

    \subfloat[MAT]{
        \includegraphics[scale=0.35]{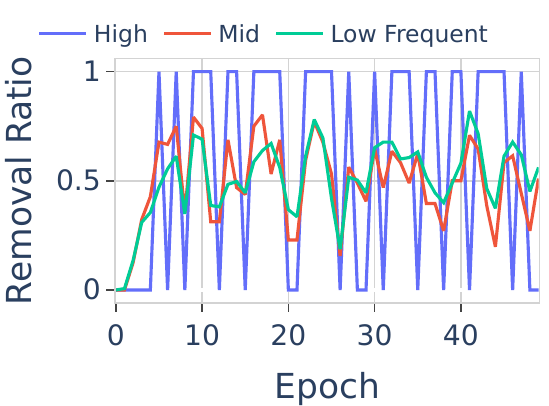}
    }
    \subfloat[MAT]{
        \includegraphics[scale=0.35]{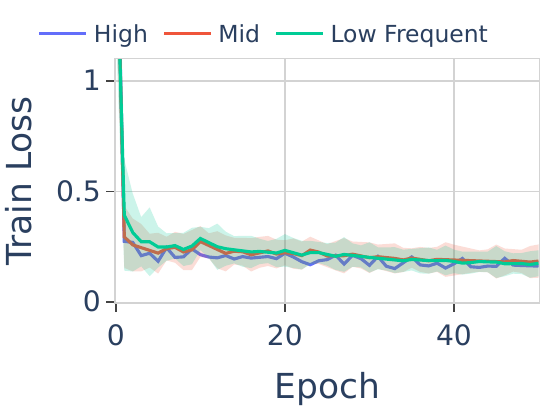}
    }
    \subfloat[MAT]{
        \includegraphics[scale=0.35]{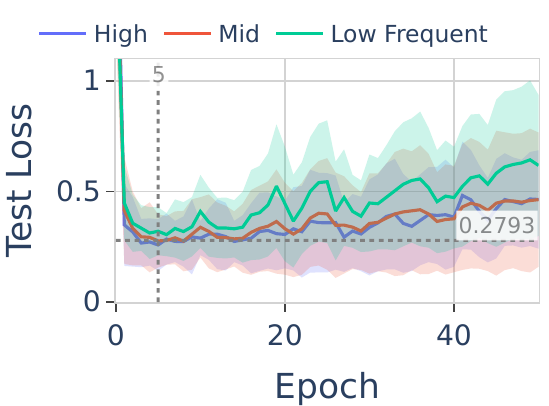}
    }
    \subfloat[MAT]{
        \includegraphics[scale=0.35]{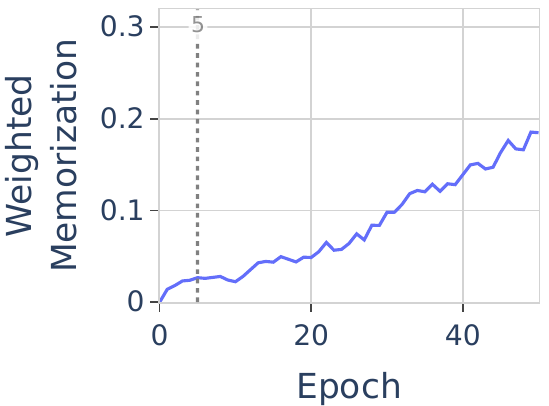}
    }

    \subfloat[MAT*]{
        \includegraphics[scale=0.35]{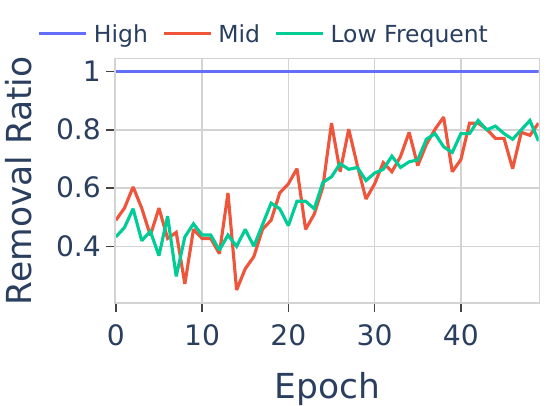}
    }
    \subfloat[MAT*]{
        \includegraphics[scale=0.35]{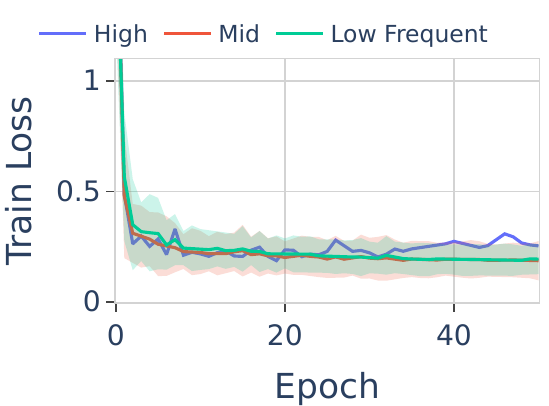}
    }
    \subfloat[MAT*]{
        \includegraphics[scale=0.35]{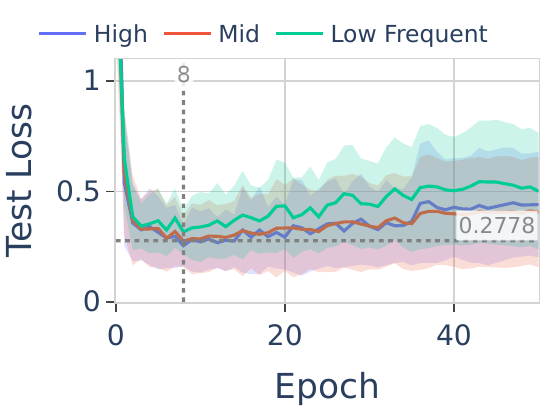}
    }
    \subfloat[MAT*]{
        \includegraphics[scale=0.35]{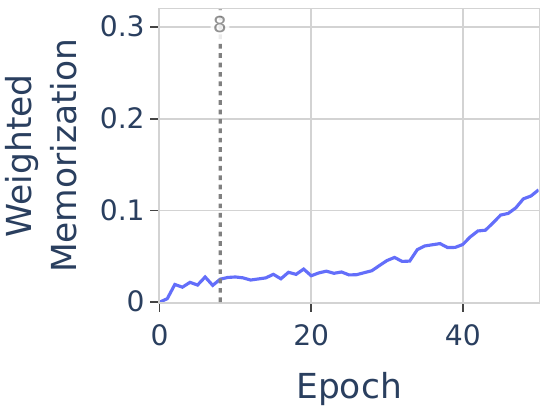}
    }

    \caption{Training dynamics under Base Training, Dedup, MAT, and MAT* on language $L_4$ using Mistral-$7$B (seed $1$). Each row corresponds to a training scheme; columns show train loss, test loss, and contextual memorization by frequency group across epochs. For MAT and MAT*, the fraction of training strings removed per group is shown leftmost. Vertical lines mark the optimal learning epoch. In the case of deduplication, all the training strings have the same frequency, hereby grouped as high-frequency.}

\end{figure}

\begin{figure}
    \raggedleft

    \subfloat[Base Training]{
        \includegraphics[scale=0.35]{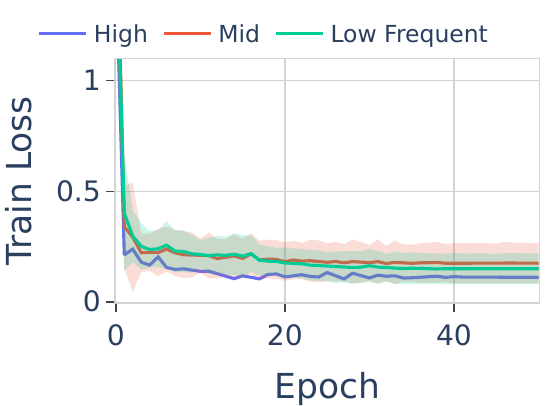}
    }
    \subfloat[Base Training]{
        \includegraphics[scale=0.35]{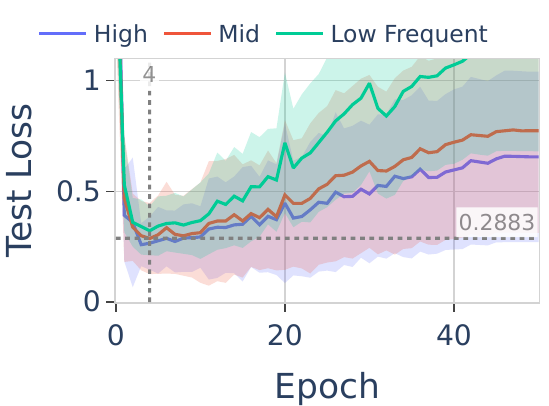}
    }
    \subfloat[Base Training]{
        \includegraphics[scale=0.35]{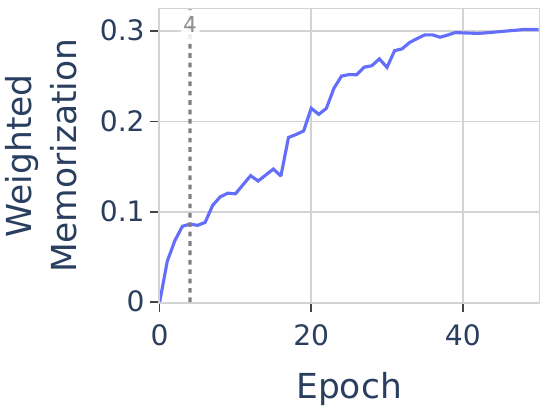}
    }

    \subfloat[Dedup]{
        \includegraphics[scale=0.35]{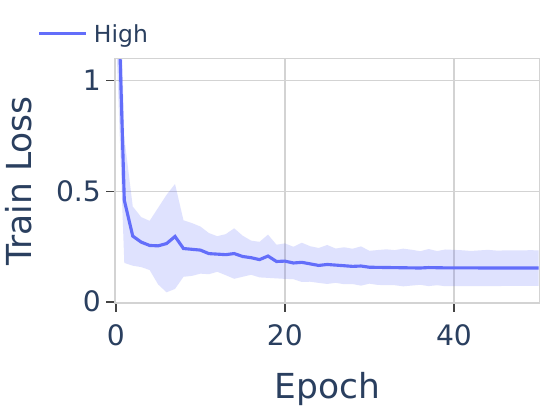}
    }
    \subfloat[Dedup]{
        \includegraphics[scale=0.35]{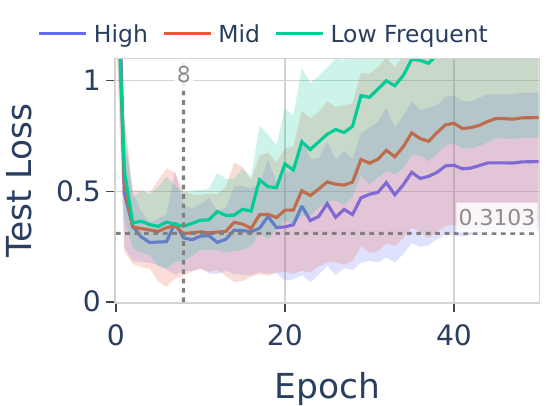}
    }
    \subfloat[Dedup]{
        \includegraphics[scale=0.35]{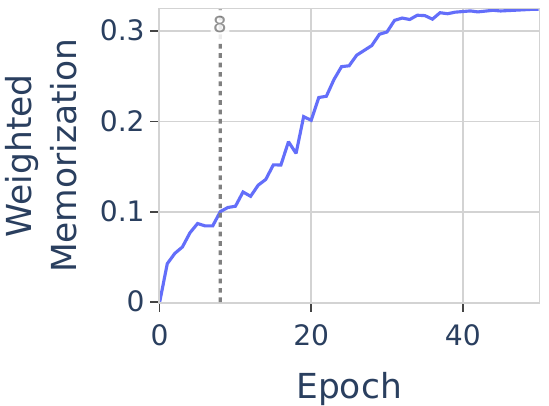}
    }

    \subfloat[MAT]{
        \includegraphics[scale=0.35]{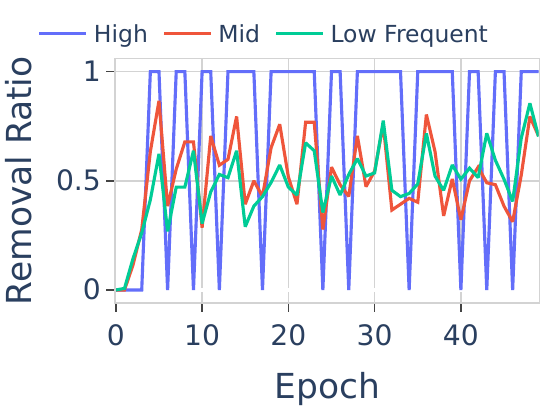}
    }
    \subfloat[MAT]{
        \includegraphics[scale=0.35]{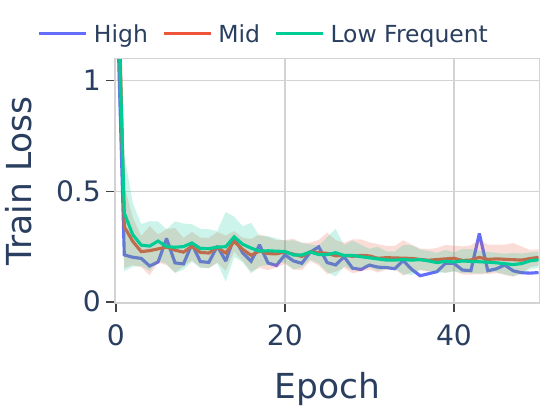}
    }
    \subfloat[MAT]{
        \includegraphics[scale=0.35]{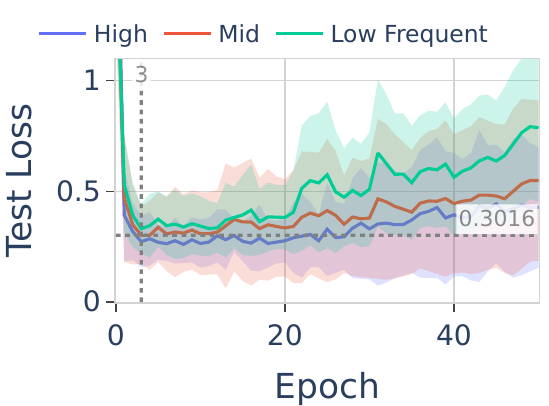}
    }
    \subfloat[MAT]{
        \includegraphics[scale=0.35]{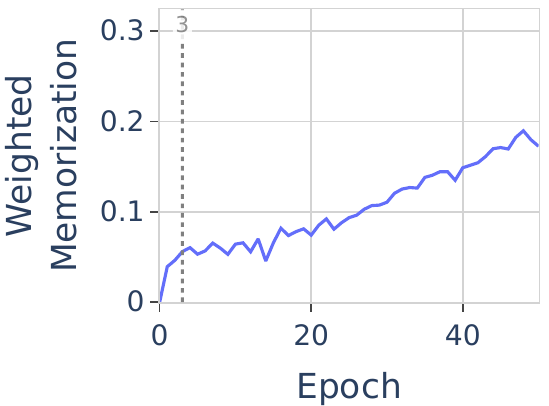}
    }

    \subfloat[MAT*]{
        \includegraphics[scale=0.35]{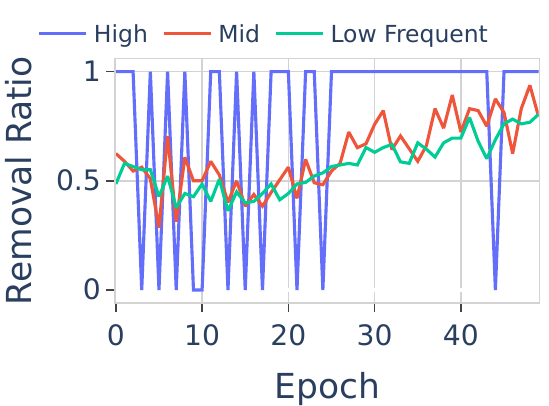}
    }
    \subfloat[MAT*]{
        \includegraphics[scale=0.35]{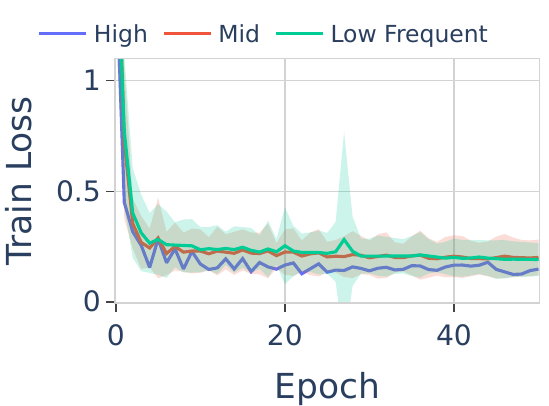}
    }
    \subfloat[MAT*]{
        \includegraphics[scale=0.35]{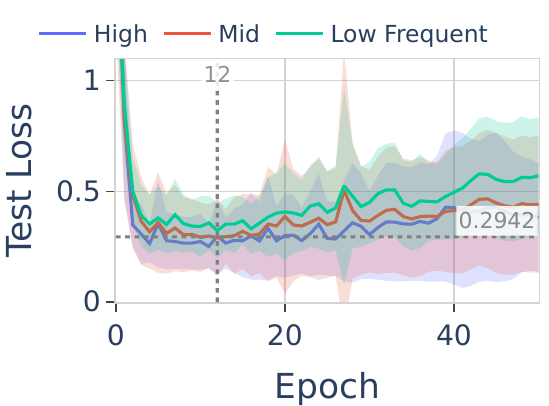}
    }
    \subfloat[MAT*]{
        \includegraphics[scale=0.35]{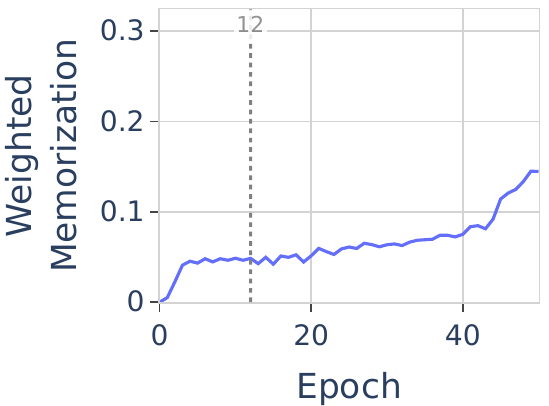}
    }

    \caption{Training dynamics under Base Training, Dedup, MAT, and MAT* on language $L_4$ using Mistral-$7$B (seed $2$). Each row corresponds to a training scheme; columns show train loss, test loss, and contextual memorization by frequency group across epochs. For MAT and MAT*, the fraction of training strings removed per group is shown leftmost. Vertical lines mark the optimal learning epoch. In the case of deduplication, all the training strings have the same frequency, hereby grouped as high-frequency.}

\end{figure}

\clearpage
\begin{figure}
    \raggedleft

    \subfloat[Base Training]{
        \includegraphics[scale=0.35]{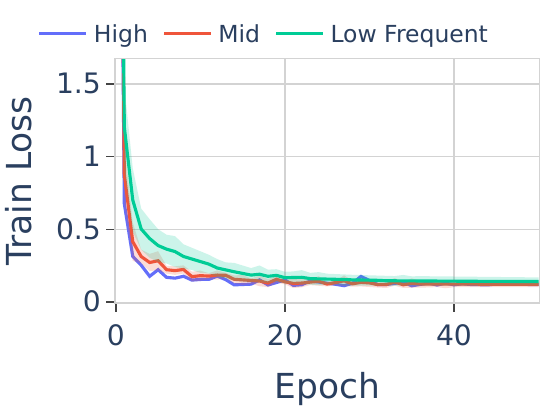}
    }
    \subfloat[Base Training]{
        \includegraphics[scale=0.35]{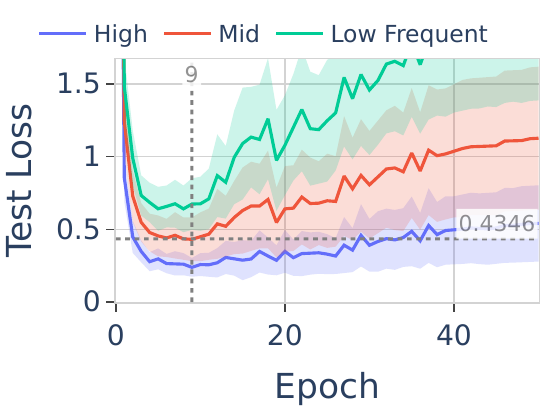}
    }
    \subfloat[Base Training]{
        \includegraphics[scale=0.35]{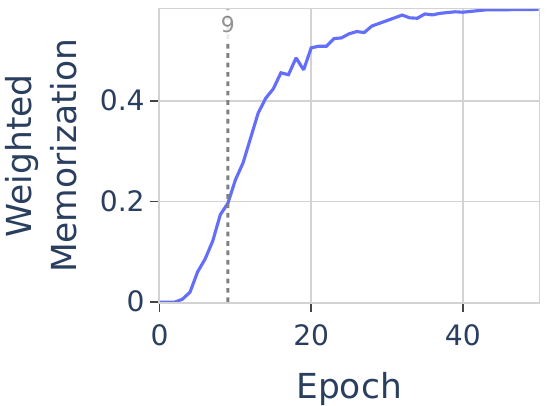}
    }

    \subfloat[Dedup]{
        \includegraphics[scale=0.35]{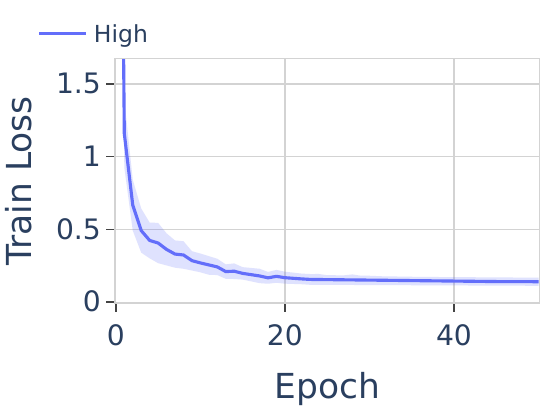}
    }
    \subfloat[Dedup]{
        \includegraphics[scale=0.35]{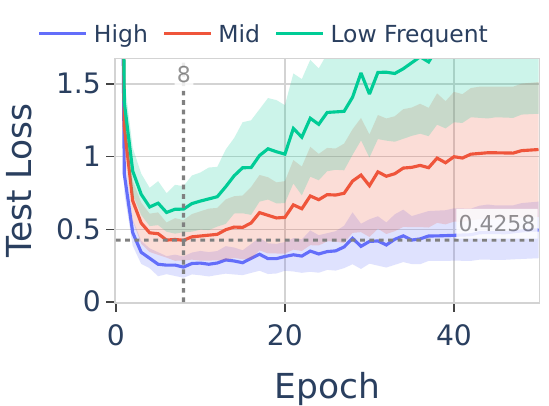}
    }
    \subfloat[Dedup]{
        \includegraphics[scale=0.35]{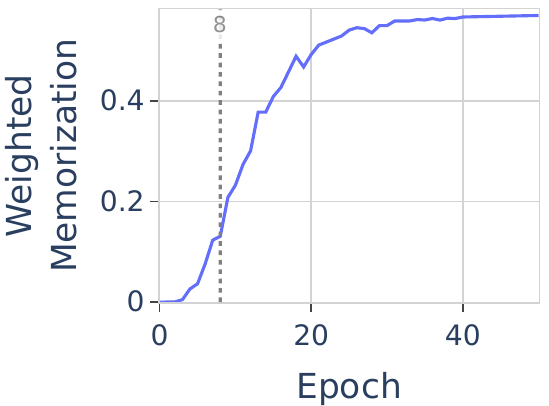}
    }

    \subfloat[MAT]{
        \includegraphics[scale=0.35]{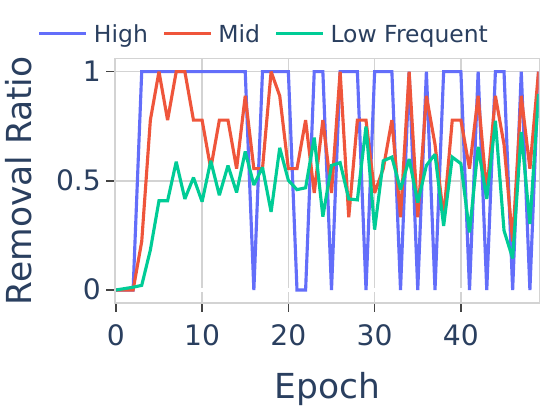}
    }
    \subfloat[MAT]{
        \includegraphics[scale=0.35]{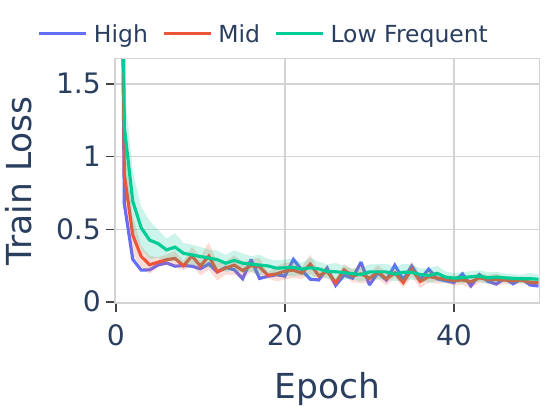}
    }
    \subfloat[MAT]{
        \includegraphics[scale=0.35]{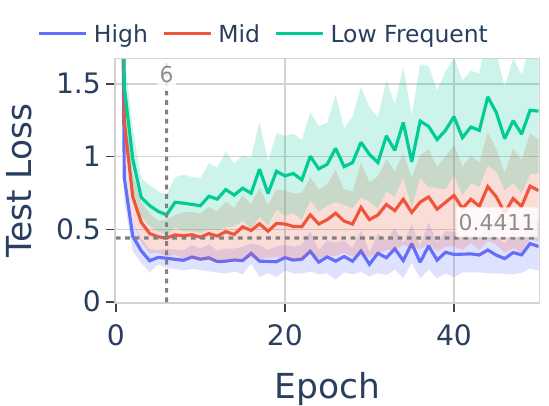}
    }
    \subfloat[MAT]{
        \includegraphics[scale=0.35]{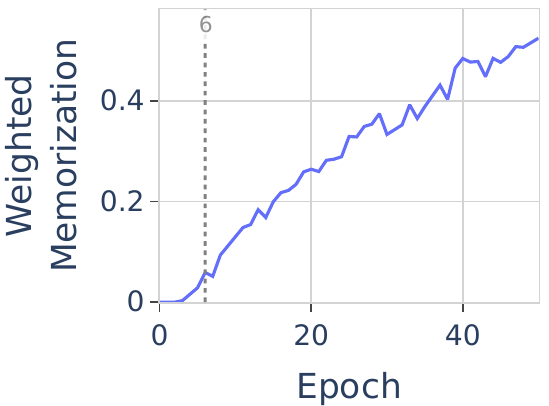}
    }

    \subfloat[MAT*]{
        \includegraphics[scale=0.35]{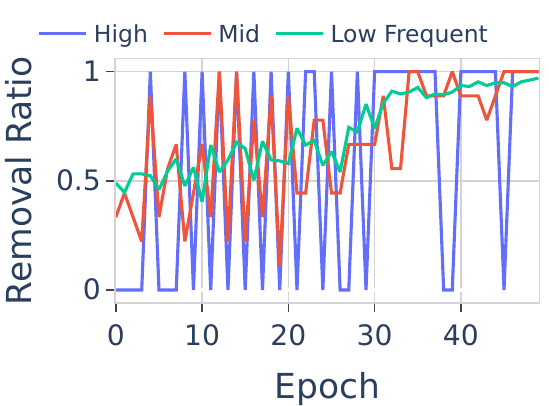}
    }
    \subfloat[MAT*]{
        \includegraphics[scale=0.35]{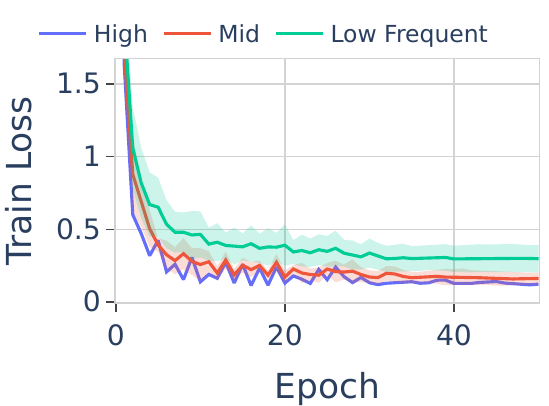}
    }
    \subfloat[MAT*]{
        \includegraphics[scale=0.35]{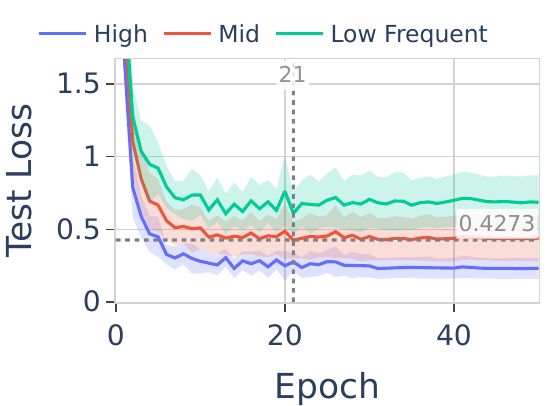}
    }
    \subfloat[MAT*]{
        \includegraphics[scale=0.35]{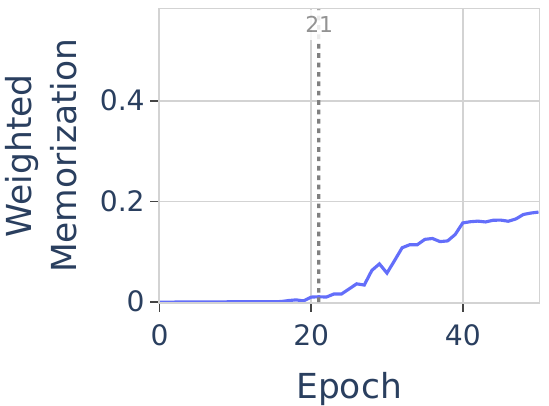}
    }

    \caption{Training dynamics under Base Training, Dedup, MAT, and MAT* on language $L_9$ using Pythia-$1$B (seed $0$). Each row corresponds to a training scheme; columns show train loss, test loss, and contextual memorization by frequency group across epochs. For MAT and MAT*, the fraction of training strings removed per group is shown leftmost. Vertical lines mark the optimal learning epoch. In the case of deduplication, all the training strings have the same frequency, hereby grouped as high-frequency.}

\end{figure}

\begin{figure}
    \raggedleft

    \subfloat[Base Training]{
        \includegraphics[scale=0.35]{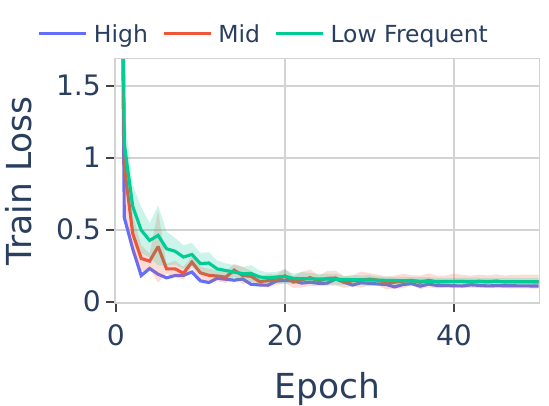}
    }
    \subfloat[Base Training]{
        \includegraphics[scale=0.35]{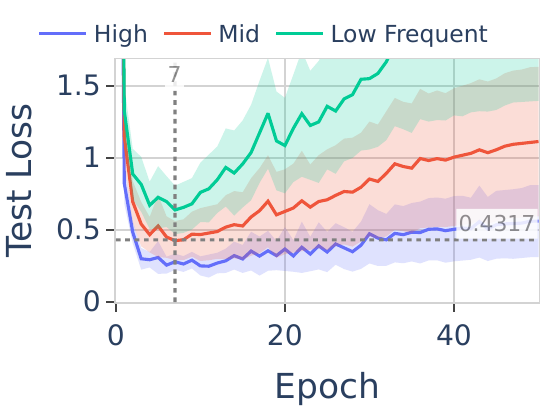}
    }
    \subfloat[Base Training]{
        \includegraphics[scale=0.35]{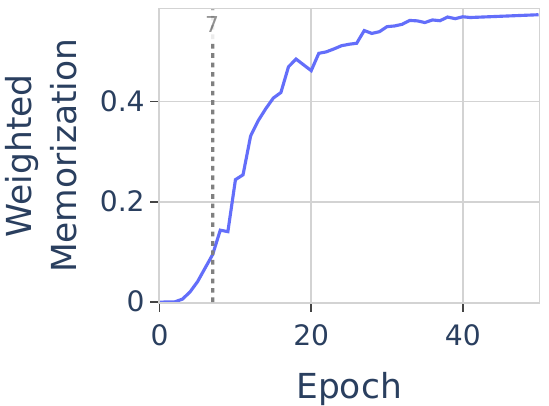}
    }

    \subfloat[Dedup]{
        \includegraphics[scale=0.35]{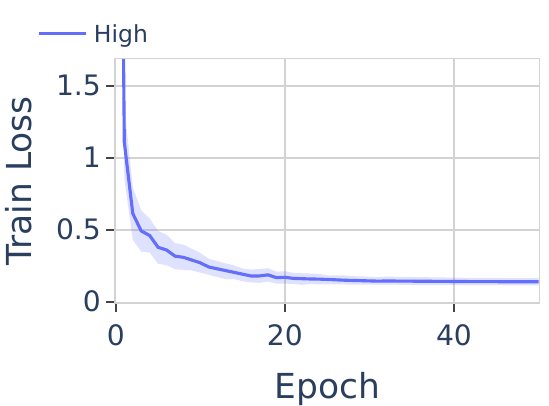}
    }
    \subfloat[Dedup]{
        \includegraphics[scale=0.35]{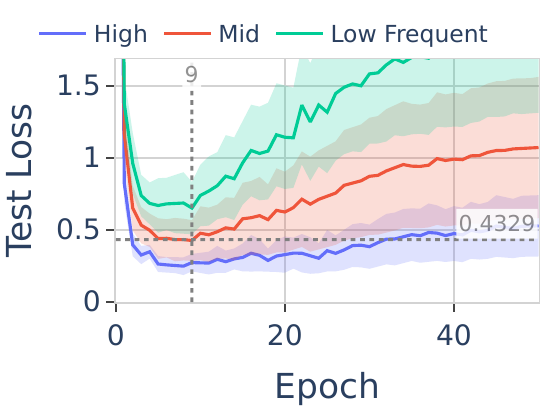}
    }
    \subfloat[Dedup]{
        \includegraphics[scale=0.35]{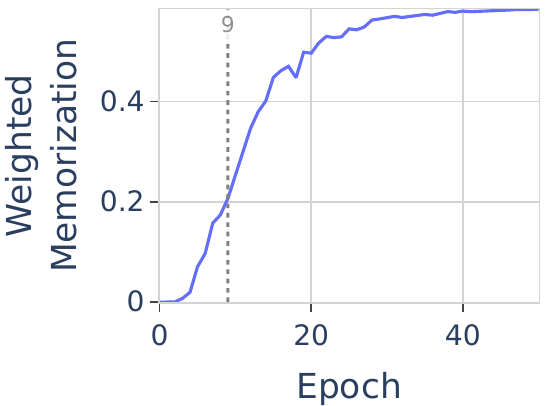}
    }

    \subfloat[MAT]{
        \includegraphics[scale=0.35]{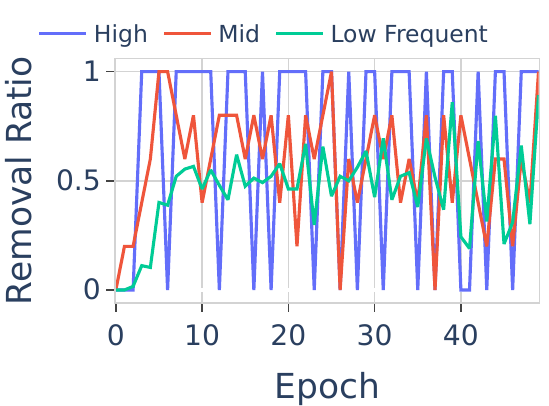}
    }
    \subfloat[MAT]{
        \includegraphics[scale=0.35]{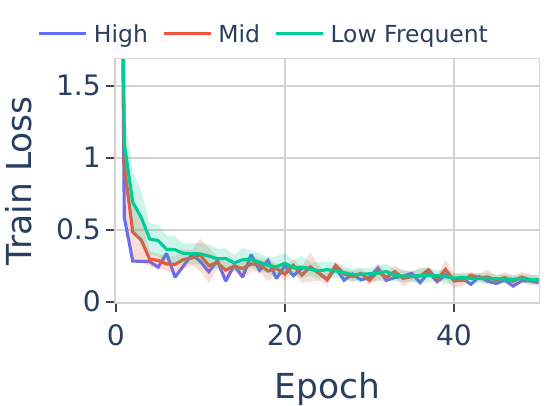}
    }
    \subfloat[MAT]{
        \includegraphics[scale=0.35]{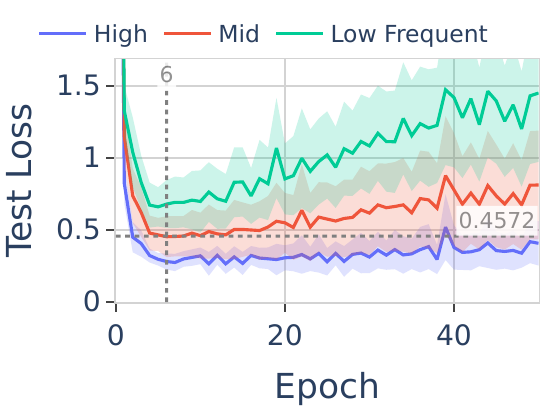}
    }
    \subfloat[MAT]{
        \includegraphics[scale=0.35]{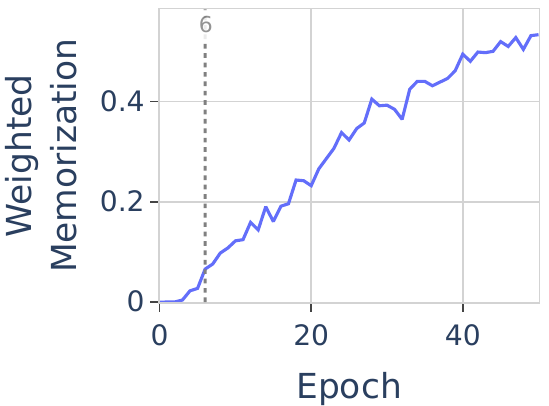}
    }

    \subfloat[MAT*]{
        \includegraphics[scale=0.35]{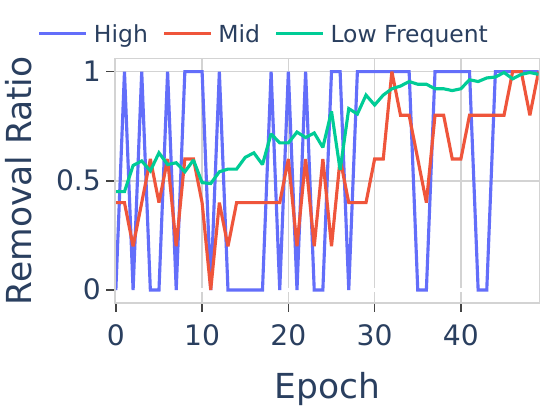}
    }
    \subfloat[MAT*]{
        \includegraphics[scale=0.35]{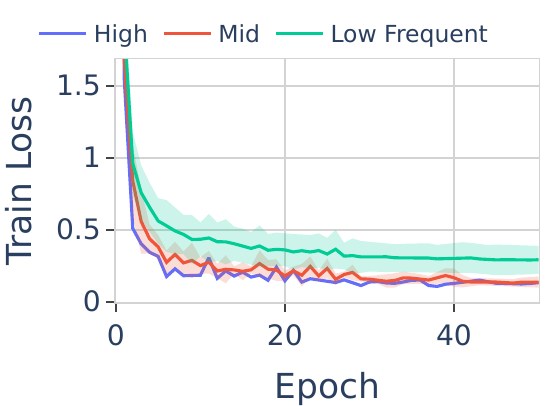}
    }
    \subfloat[MAT*]{
        \includegraphics[scale=0.35]{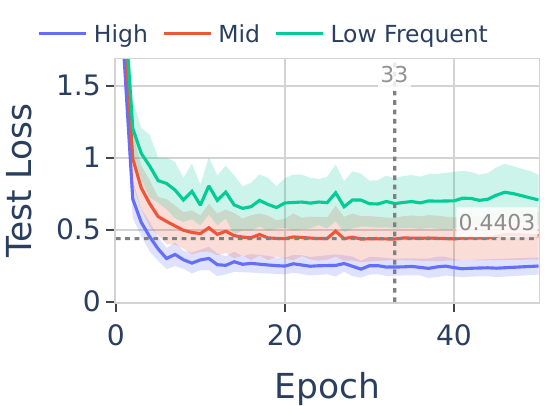}
    }
    \subfloat[MAT*]{
        \includegraphics[scale=0.35]{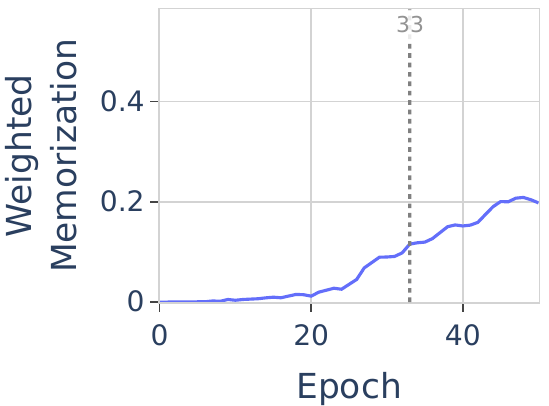}
    }

    \caption{Training dynamics under Base Training, Dedup, MAT, and MAT* on language $L_9$ using Pythia-$1$B (seed $1$). Each row corresponds to a training scheme; columns show train loss, test loss, and contextual memorization by frequency group across epochs. For MAT and MAT*, the fraction of training strings removed per group is shown leftmost. Vertical lines mark the optimal learning epoch. In the case of deduplication, all the training strings have the same frequency, hereby grouped as high-frequency.}

\end{figure}

\begin{figure}
    \raggedleft

    \subfloat[Base Training]{
        \includegraphics[scale=0.35]{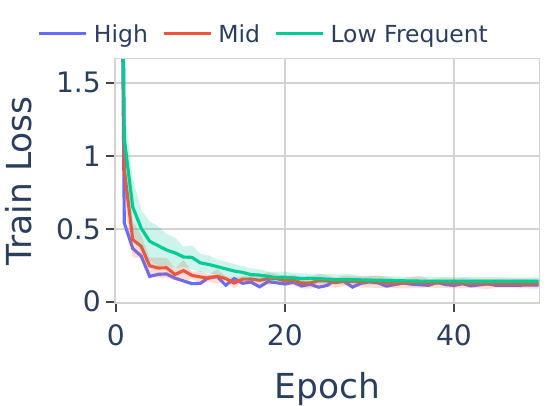}
    }
    \subfloat[Base Training]{
        \includegraphics[scale=0.35]{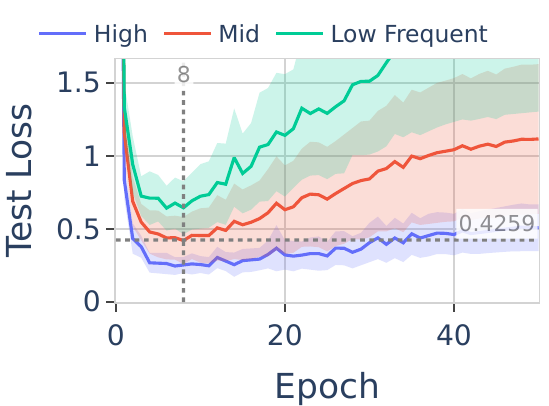}
    }
    \subfloat[Base Training]{
        \includegraphics[scale=0.35]{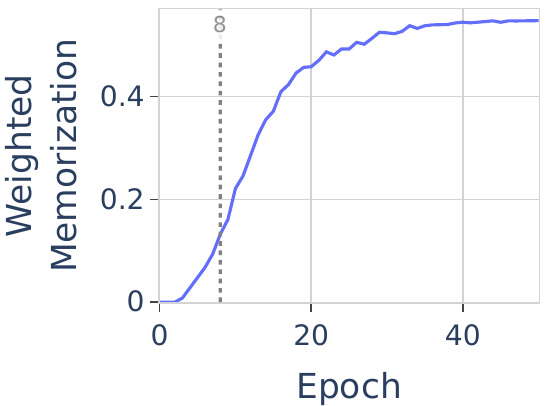}
    }

    \subfloat[Dedup]{
        \includegraphics[scale=0.35]{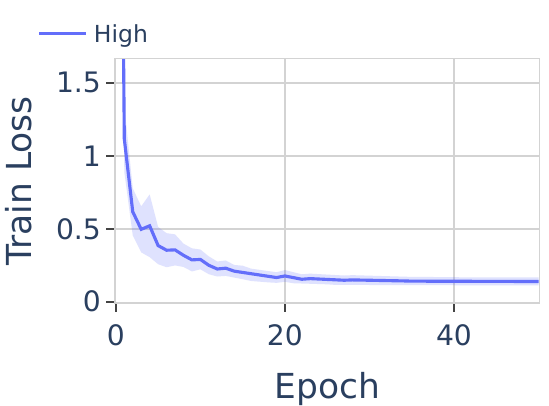}
    }
    \subfloat[Dedup]{
        \includegraphics[scale=0.35]{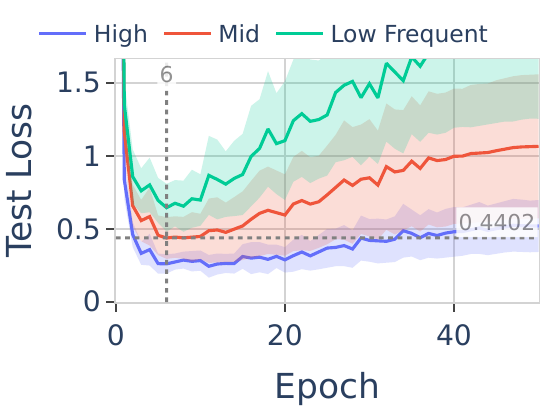}
    }
    \subfloat[Dedup]{
        \includegraphics[scale=0.35]{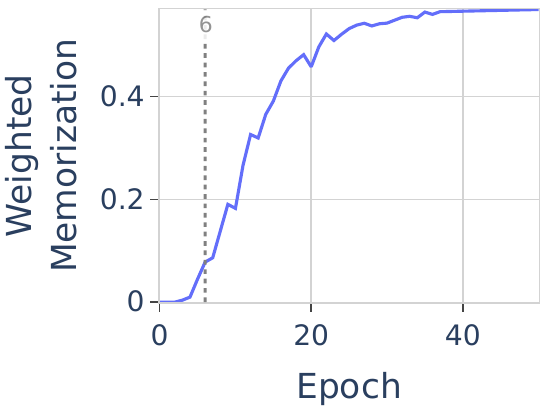}
    }

    \subfloat[MAT]{
        \includegraphics[scale=0.35]{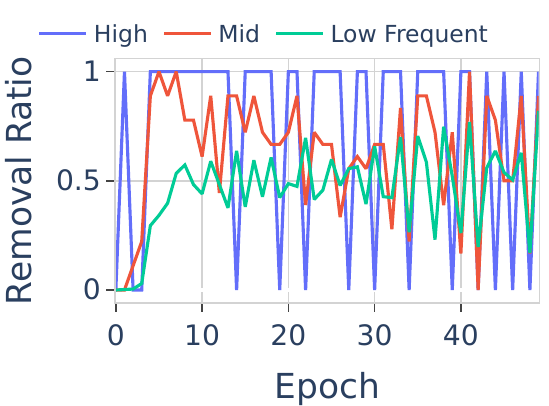}
    }
    \subfloat[MAT]{
        \includegraphics[scale=0.35]{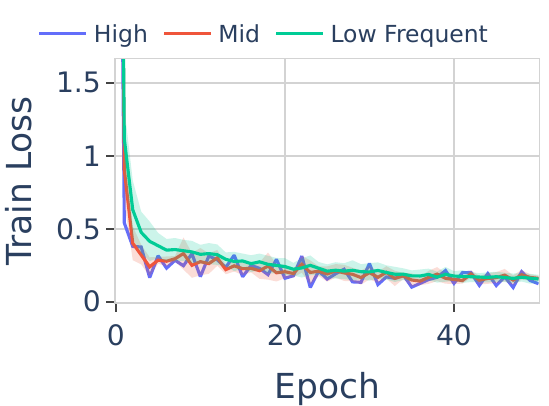}
    }
    \subfloat[MAT]{
        \includegraphics[scale=0.35]{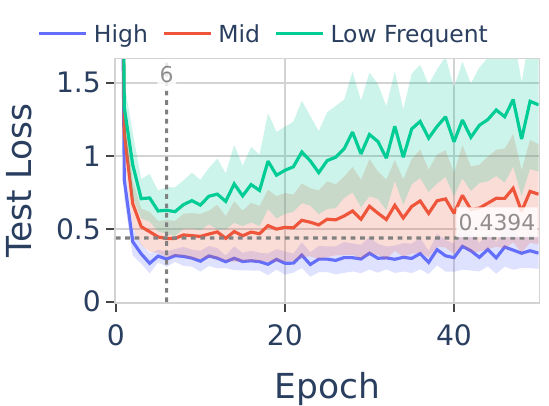}
    }
    \subfloat[MAT]{
        \includegraphics[scale=0.35]{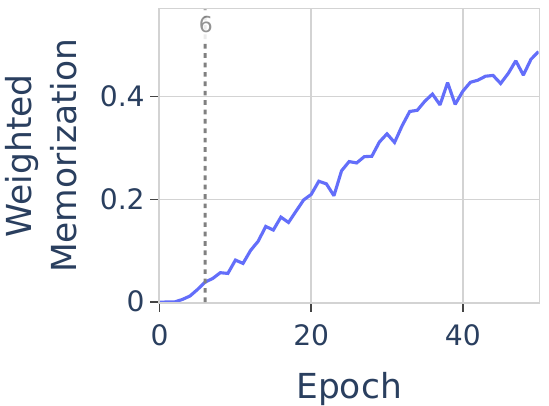}
    }

    \subfloat[MAT*]{
        \includegraphics[scale=0.35]{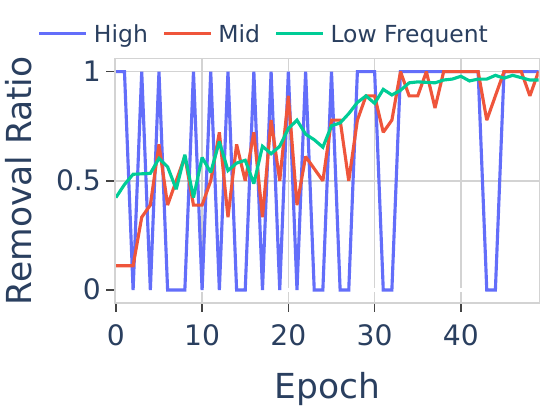}
    }
    \subfloat[MAT*]{
        \includegraphics[scale=0.35]{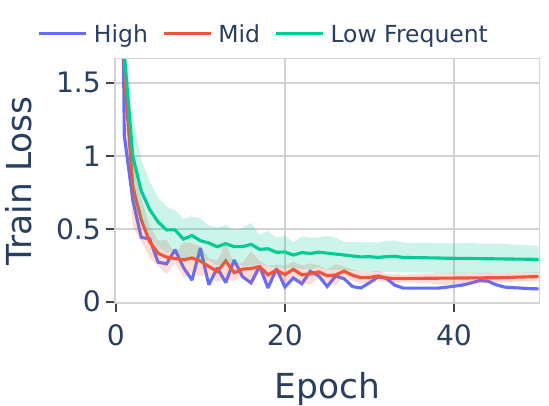}
    }
    \subfloat[MAT*]{
        \includegraphics[scale=0.35]{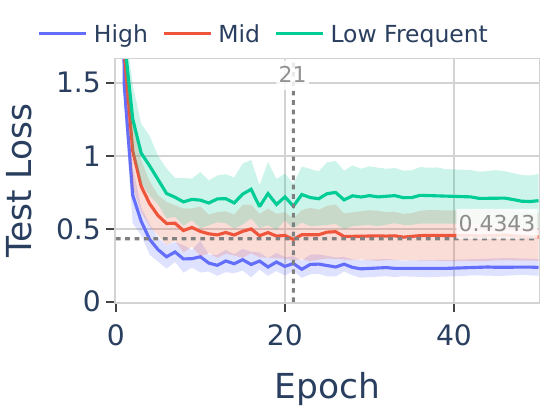}
    }
    \subfloat[MAT*]{
        \includegraphics[scale=0.35]{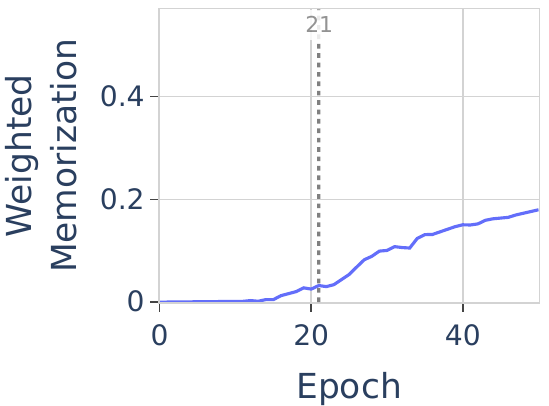}
    }

    \caption{Training dynamics under Base Training, Dedup, MAT, and MAT* on language $L_9$ using Pythia-$1$B (seed $2$). Each row corresponds to a training scheme; columns show train loss, test loss, and contextual memorization by frequency group across epochs. For MAT and MAT*, the fraction of training strings removed per group is shown leftmost. Vertical lines mark the optimal learning epoch. In the case of deduplication, all the training strings have the same frequency, hereby grouped as high-frequency.}

\end{figure}

\begin{figure*}
    \centering
    \subfloat[$ L_2 $]{\includegraphics[scale=0.45]{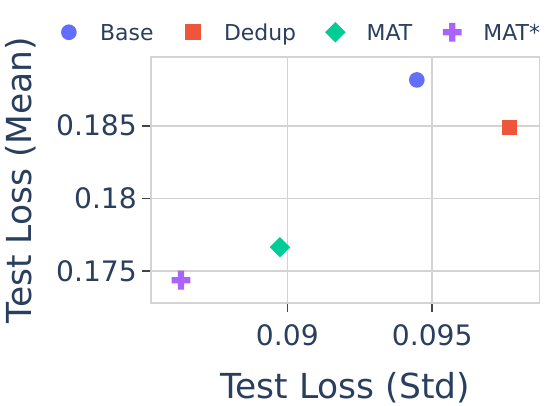}}
    \subfloat[$ L_2 $]{\includegraphics[scale=0.45]{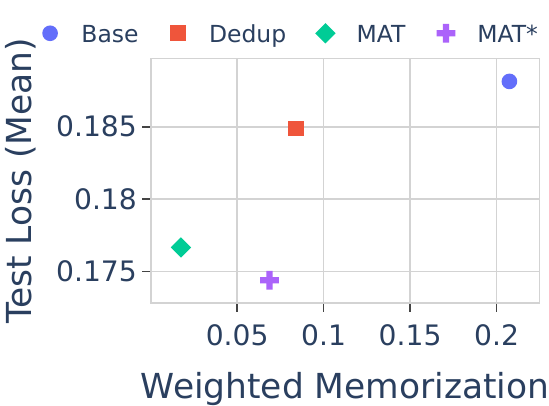}}
    \subfloat[$ L_2 $]{\includegraphics[scale=0.45]{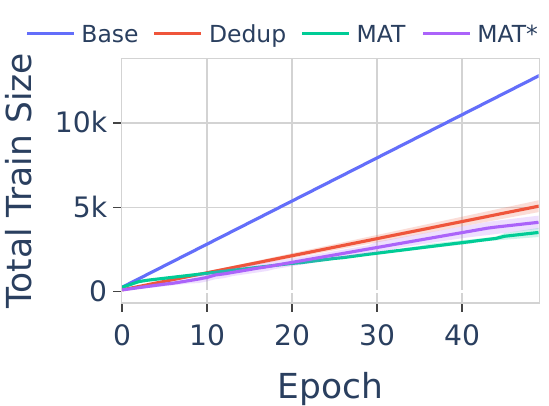}}

    \subfloat[$ L_4 $]{\includegraphics[scale=0.45]{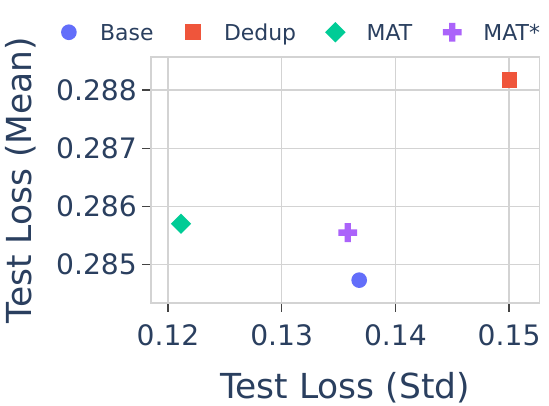}}
    \subfloat[$ L_4 $]{\includegraphics[scale=0.45]{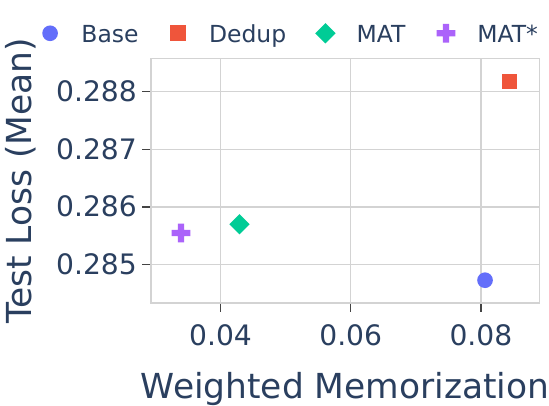}}
    \subfloat[$ L_4 $]{\includegraphics[scale=0.45]{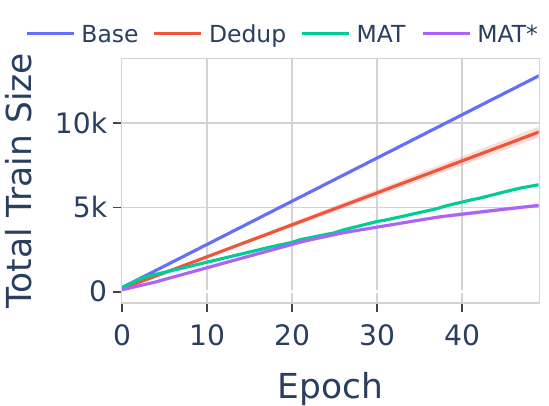}}

    \subfloat[$ L_9 $]{\includegraphics[scale=0.45]{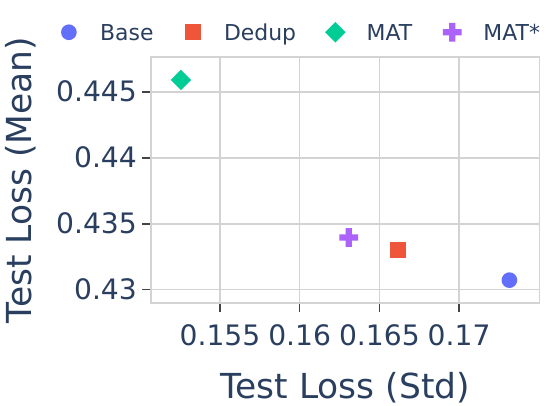}}
    \subfloat[$ L_9 $]{\includegraphics[scale=0.45]{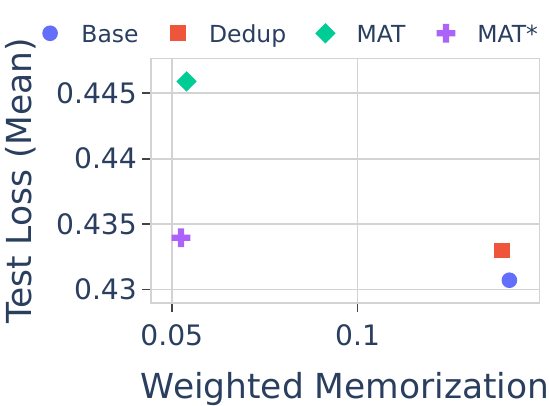}}
    \subfloat[$ L_9 $]{\includegraphics[scale=0.45]{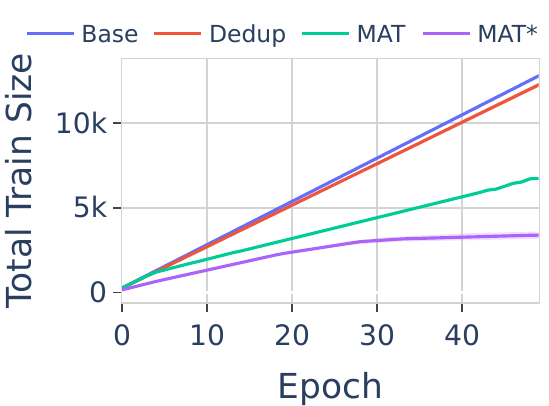}}

    \caption{Summary results on learning, memorization, and training cost efficiency on different formal languages, averaged across multiple runs. Lower values are better in all dimensions.}
\end{figure*}

\begin{figure*}
    \centering

    \subfloat[High frequency, $ L_2 $]{\includegraphics[scale=0.45]{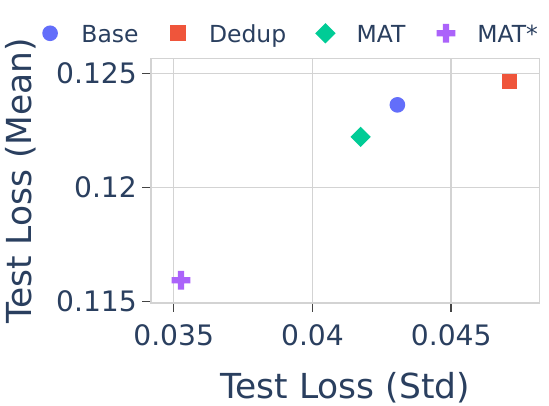}}
    \subfloat[Mid frequency, $ L_2 $]{\includegraphics[scale=0.45]{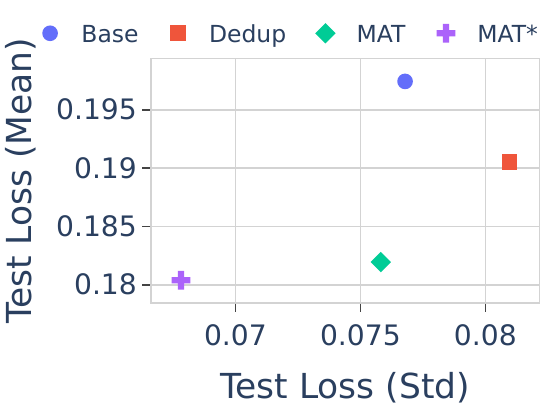}}
    \subfloat[Low frequency, $ L_2 $]{\includegraphics[scale=0.45]{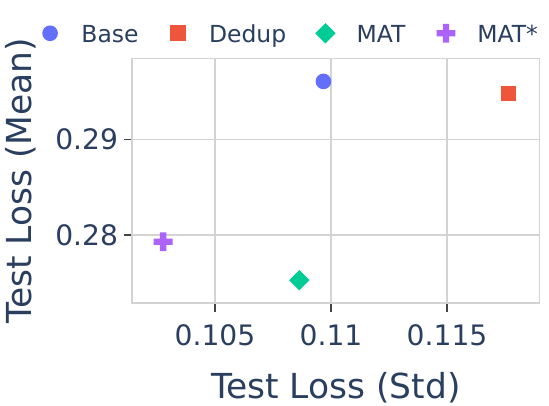}}

    \subfloat[High frequency, $ L_4 $]{\includegraphics[scale=0.45]{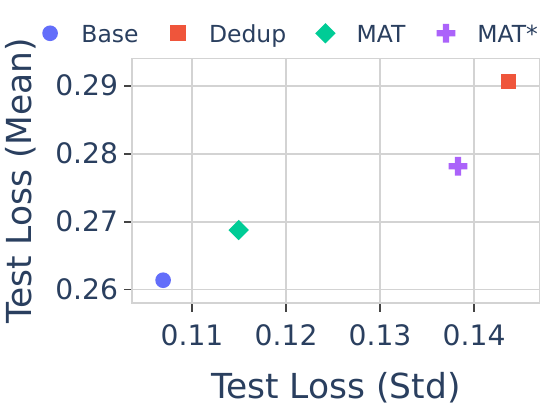}}
    \subfloat[Mid frequency, $ L_4 $]{\includegraphics[scale=0.45]{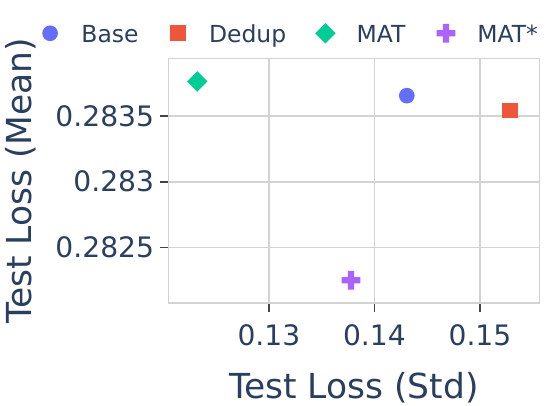}}
    \subfloat[Low frequency, $ L_4 $]{\includegraphics[scale=0.45]{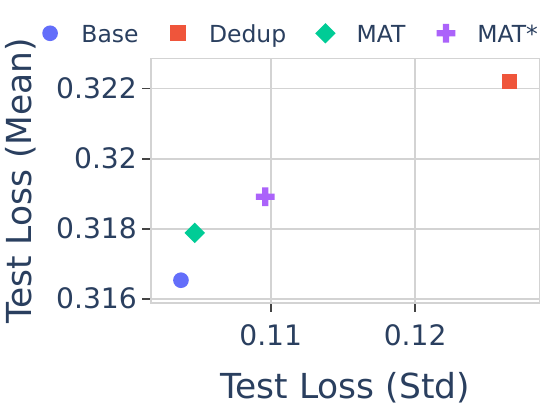}}

    \subfloat[High frequency, $ L_9 $]{\includegraphics[scale=0.45]{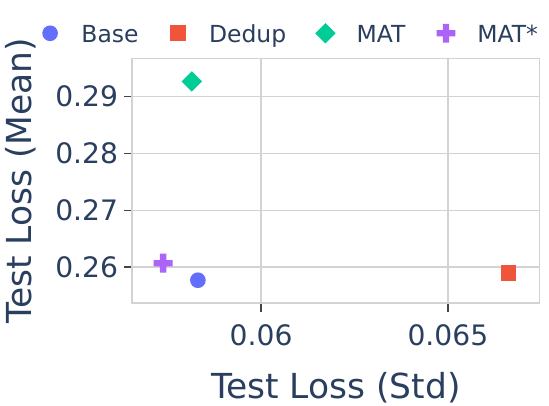}}
    \subfloat[Mid frequency, $ L_9 $]{\includegraphics[scale=0.45]{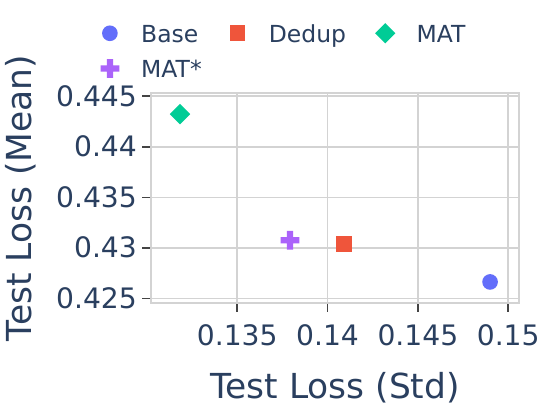}}
    \subfloat[Low frequency, $ L_9 $]{\includegraphics[scale=0.45]{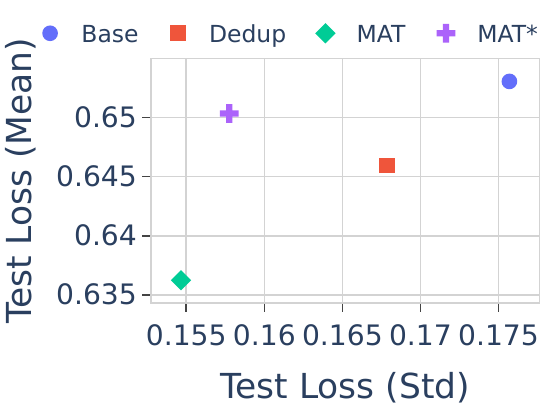}}

    \caption{Detailed results of the standard deviation and mean of test loss across test strings from different frequency groups. The benefit of MAT and MAT* is also pronounced in mid and low-frequency strings.}
    \label{fig:mat_learning_different_freq}
\end{figure*}

\clearpage
\subsection{Sensitivity of MAT to the Down-sampling Threshold}
\label{sec:mat_threshold_sensitivity}

MAT down-samples $x$ fraction of training strings (by lowest training loss), where $x$ is the fraction of test sequences whose loss increased relative to the previous epoch. Here we test how sensitive MAT is to this heuristic by scaling it to $0.5x$ (conservative) and $2x$ (aggressive), and comparing against the default $x$ and against Base training, on all three formal languages $L_2$, $L_4$, $L_9$ using Mistral-$7$B. Figures~\ref{fig:mat_threshold_dynamics_l2}--\ref{fig:mat_threshold_dynamics_l9} show the training dynamics (single seed); Figure~\ref{fig:mat_threshold_summary} summarizes memorization, learning stability, and training cost across all four schemes, averaged over three seeds.

\begin{figure}[h]
    \raggedleft

    \subfloat[Base]{
        \includegraphics[scale=0.35]{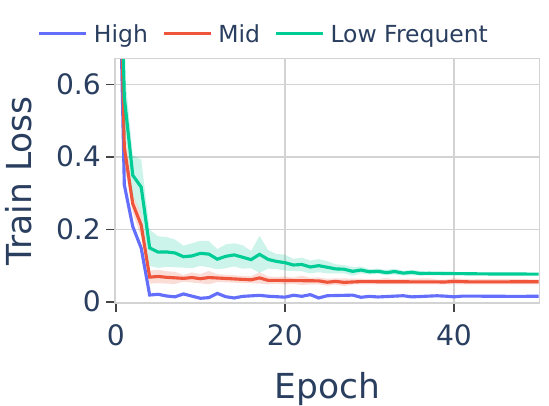}
    }
    \subfloat[Base]{
        \includegraphics[scale=0.35]{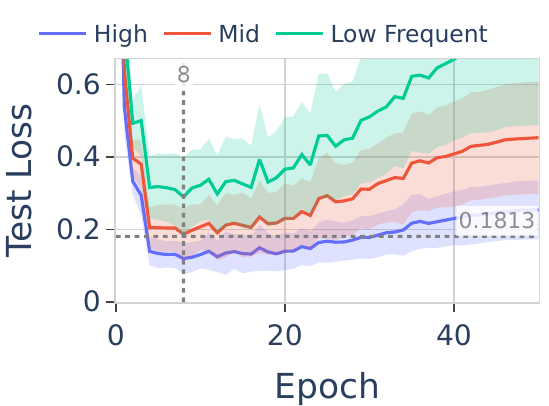}
    }
    \subfloat[Base]{
        \includegraphics[scale=0.35]{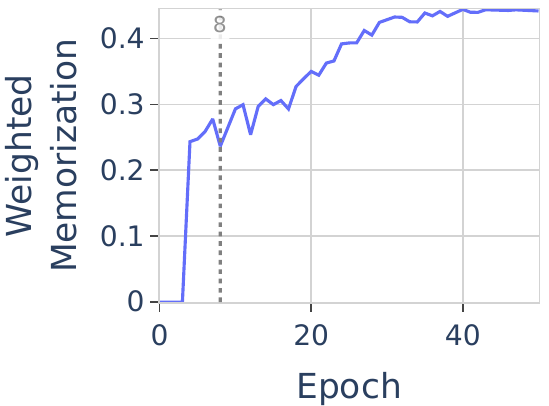}
    }

    \subfloat[$0.5x$]{
        \includegraphics[scale=0.35]{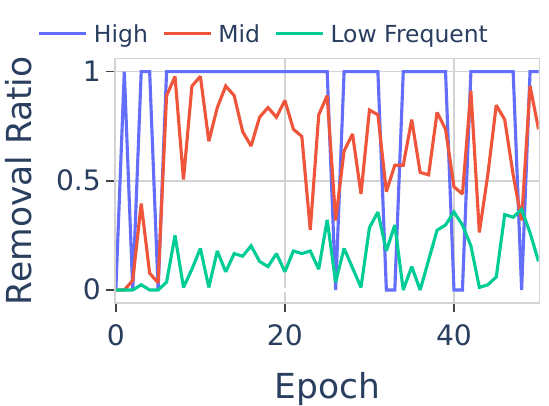}
    }
    \subfloat[$0.5x$]{
        \includegraphics[scale=0.35]{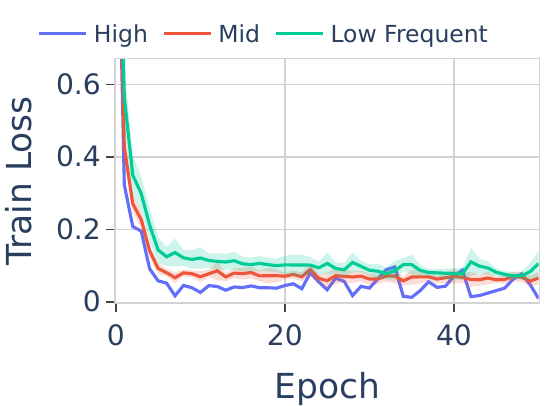}
    }
    \subfloat[$0.5x$]{
        \includegraphics[scale=0.35]{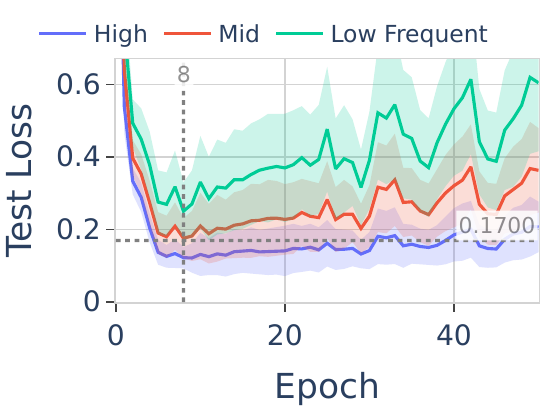}
    }
    \subfloat[$0.5x$]{
        \includegraphics[scale=0.35]{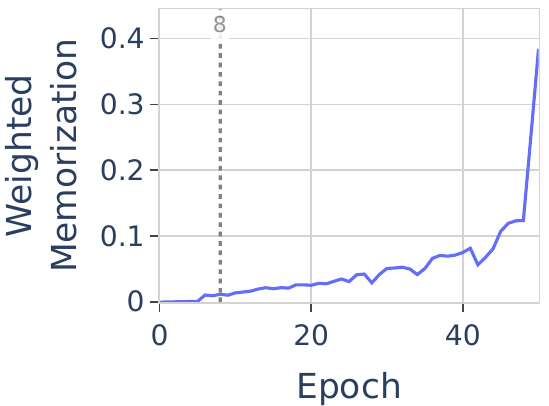}
    }

    \subfloat[$x$]{
        \includegraphics[scale=0.35]{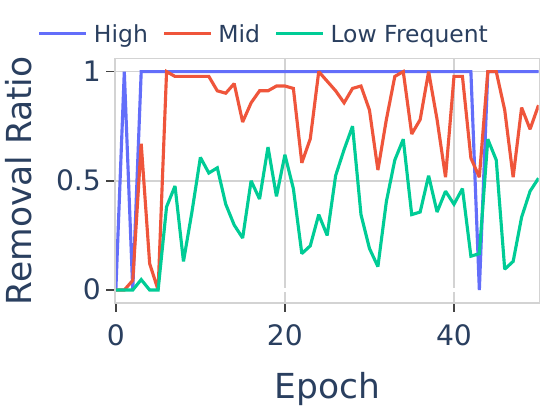}
    }
    \subfloat[$x$]{
        \includegraphics[scale=0.35]{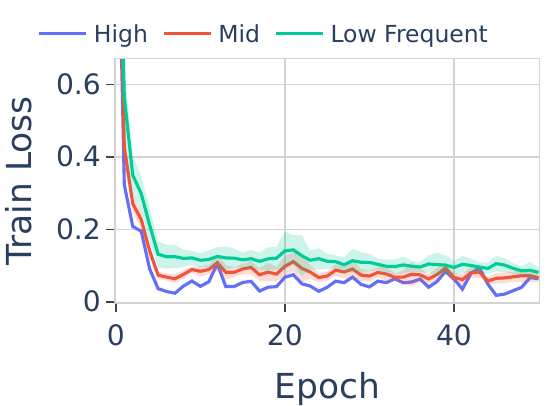}
    }
    \subfloat[$x$]{
        \includegraphics[scale=0.35]{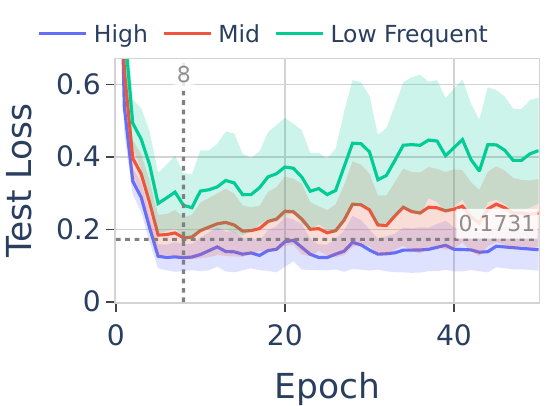}
    }
    \subfloat[$x$]{
        \includegraphics[scale=0.35]{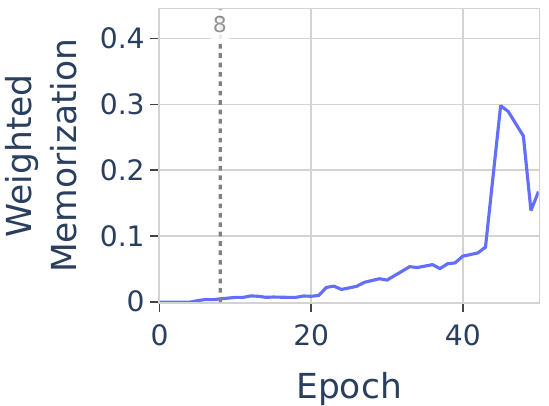}
    }

    \subfloat[$2x$]{
        \includegraphics[scale=0.35]{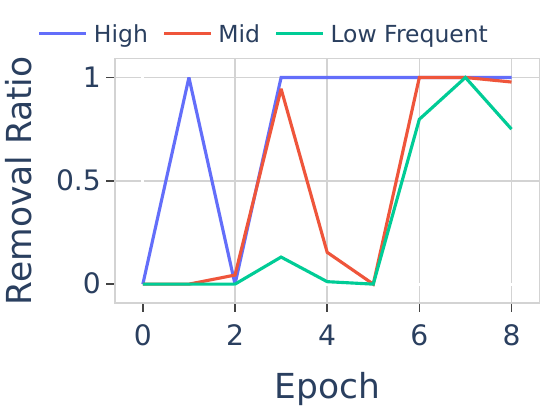}
    }
    \subfloat[$2x$]{
        \includegraphics[scale=0.35]{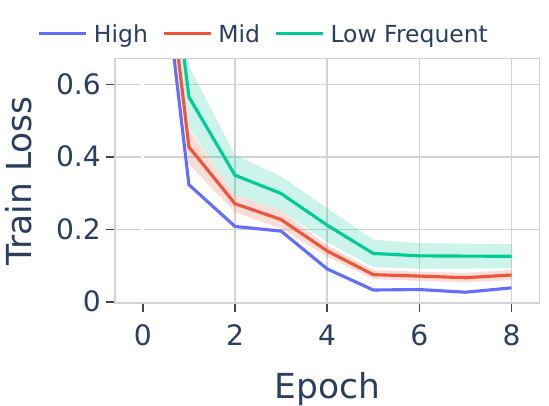}
    }
    \subfloat[$2x$]{
        \includegraphics[scale=0.35]{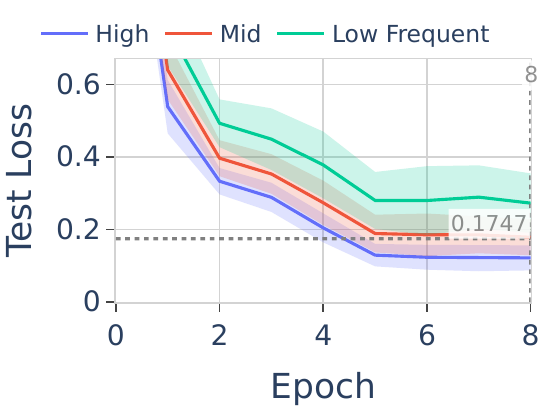}
    }
    \subfloat[$2x$]{
        \includegraphics[scale=0.35]{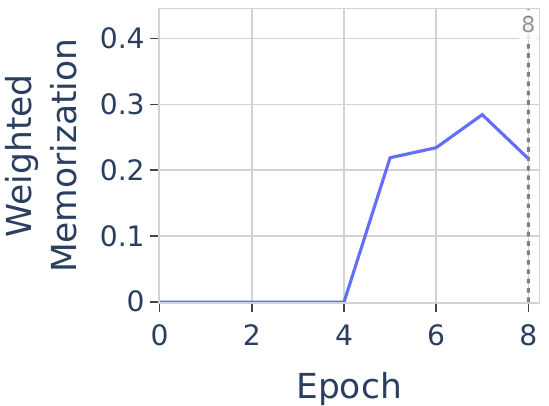}
    }

    \caption{Training dynamics under Base training and MAT with down-sampling rate $0.5x$, $x$, and $2x$ on language $L_2$ using Mistral-$7$B (seed $0$), where $x$ is the fraction of test sequences whose loss increases relative to the previous epoch. Each row corresponds to a training scheme; columns show train loss, test loss, and contextual memorization across epochs.}
    \label{fig:mat_threshold_dynamics_l2}

\end{figure}

\begin{figure}[h]
    \raggedleft

    \subfloat[Base]{
        \includegraphics[scale=0.35]{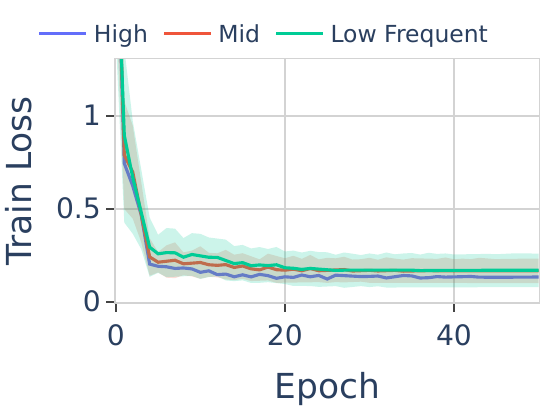}
    }
    \subfloat[Base]{
        \includegraphics[scale=0.35]{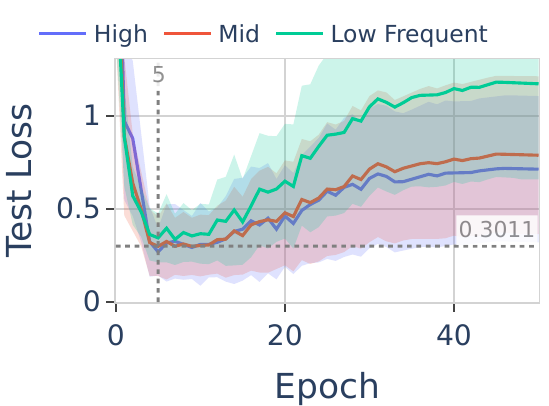}
    }
    \subfloat[Base]{
        \includegraphics[scale=0.35]{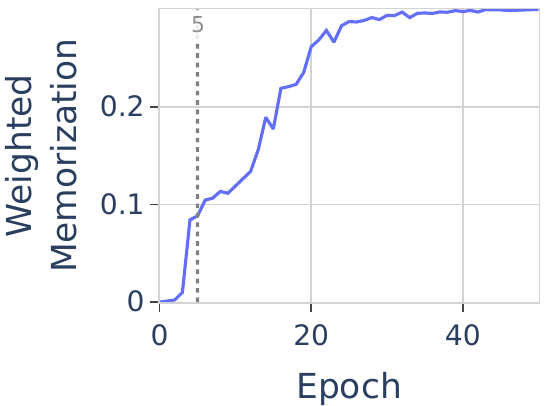}
    }

    \subfloat[$0.5x$]{
        \includegraphics[scale=0.35]{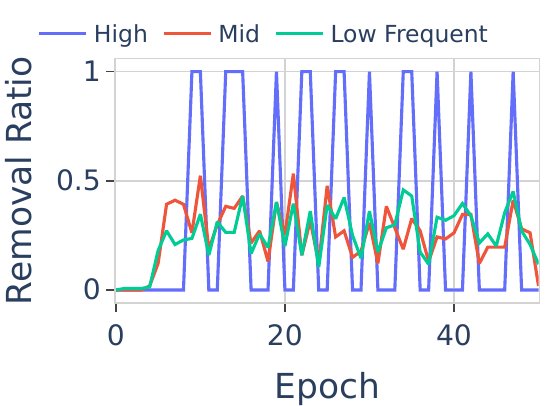}
    }
    \subfloat[$0.5x$]{
        \includegraphics[scale=0.35]{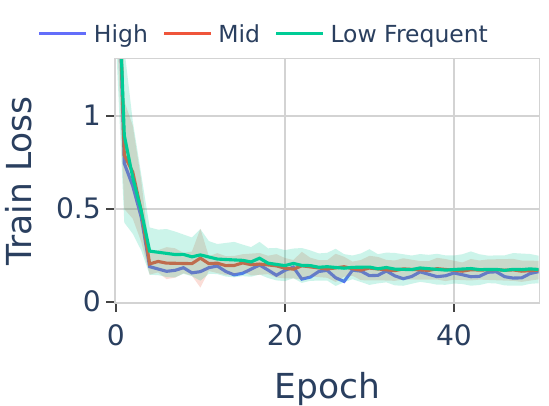}
    }
    \subfloat[$0.5x$]{
        \includegraphics[scale=0.35]{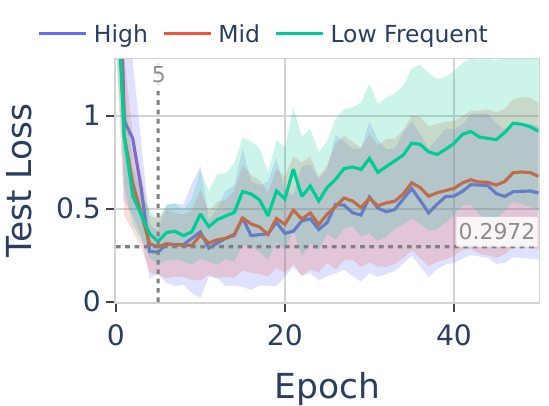}
    }
    \subfloat[$0.5x$]{
        \includegraphics[scale=0.35]{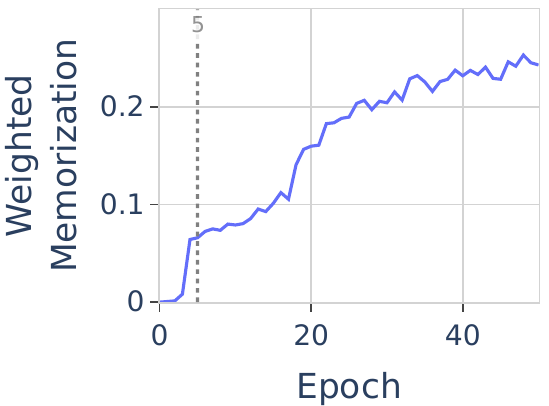}
    }

    \subfloat[$x$]{
        \includegraphics[scale=0.35]{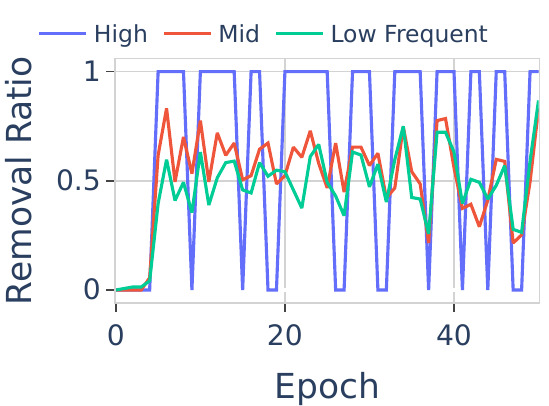}
    }
    \subfloat[$x$]{
        \includegraphics[scale=0.35]{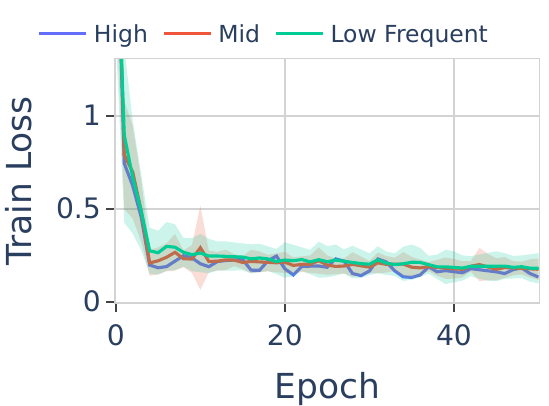}
    }
    \subfloat[$x$]{
        \includegraphics[scale=0.35]{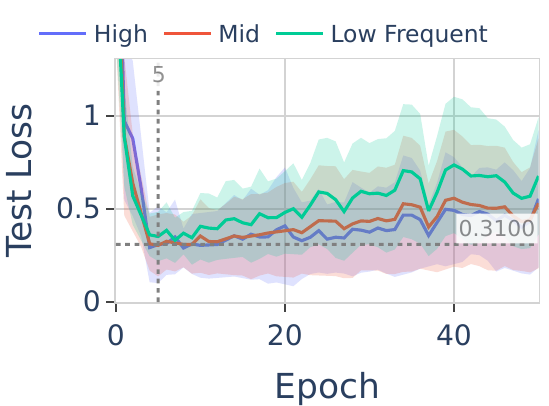}
    }
    \subfloat[$x$]{
        \includegraphics[scale=0.35]{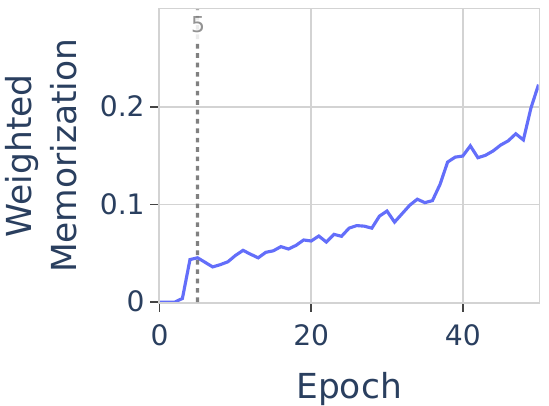}
    }

    \subfloat[$2x$]{
        \includegraphics[scale=0.35]{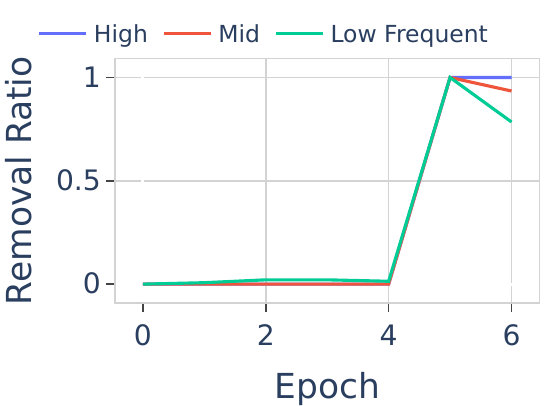}
    }
    \subfloat[$2x$]{
        \includegraphics[scale=0.35]{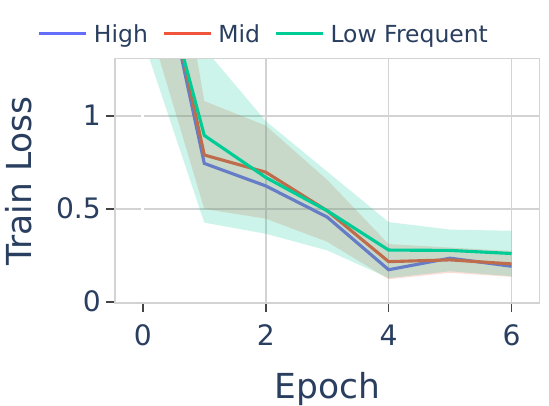}
    }
    \subfloat[$2x$]{
        \includegraphics[scale=0.35]{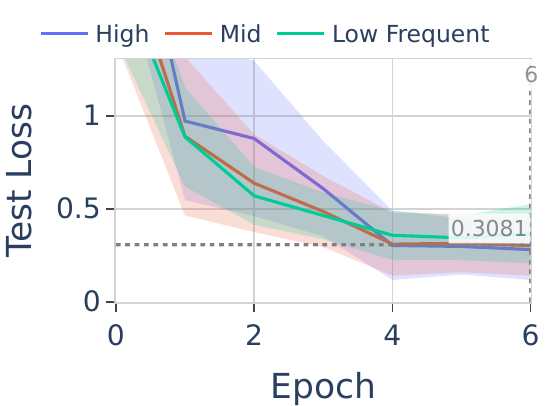}
    }
    \subfloat[$2x$]{
        \includegraphics[scale=0.35]{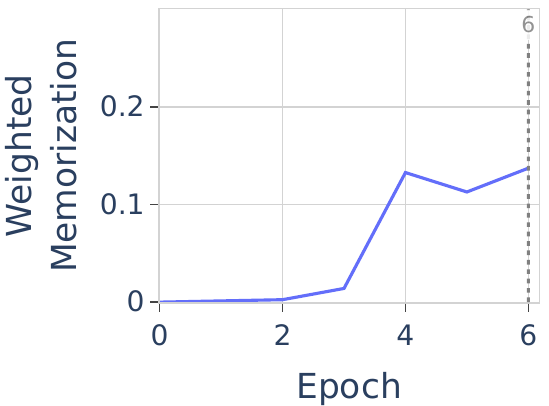}
    }

    \caption{Same as Figure~\ref{fig:mat_threshold_dynamics_l2}, but on language $L_4$.}
    \label{fig:mat_threshold_dynamics_l4}

\end{figure}

\begin{figure}[h]
    \raggedleft

    \subfloat[Base]{
        \includegraphics[scale=0.35]{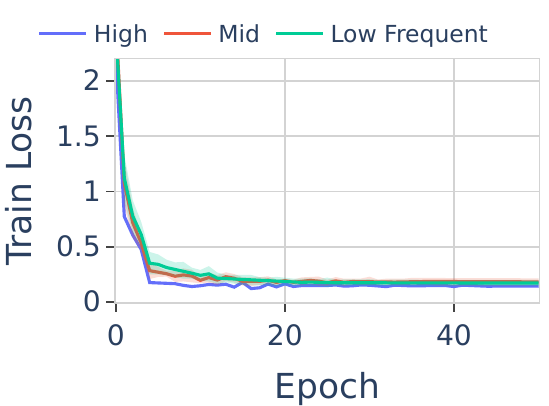}
    }
    \subfloat[Base]{
        \includegraphics[scale=0.35]{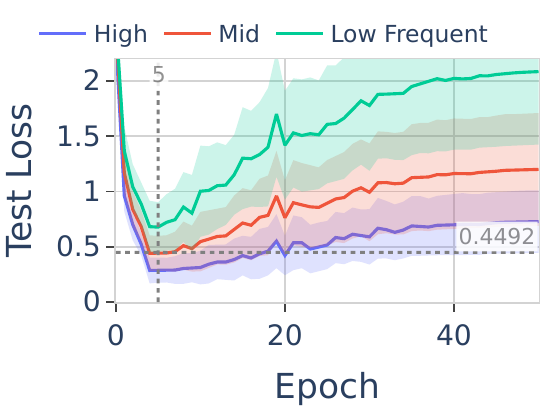}
    }
    \subfloat[Base]{
        \includegraphics[scale=0.35]{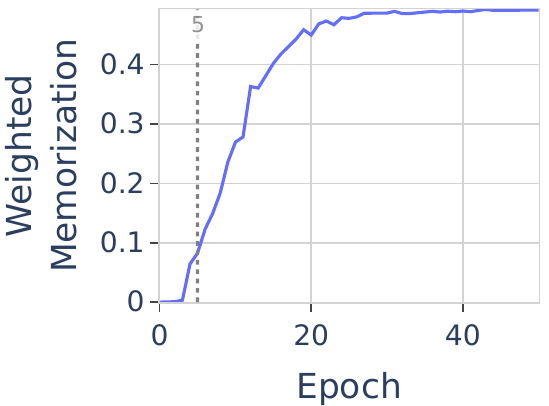}
    }

    \subfloat[$0.5x$]{
        \includegraphics[scale=0.35]{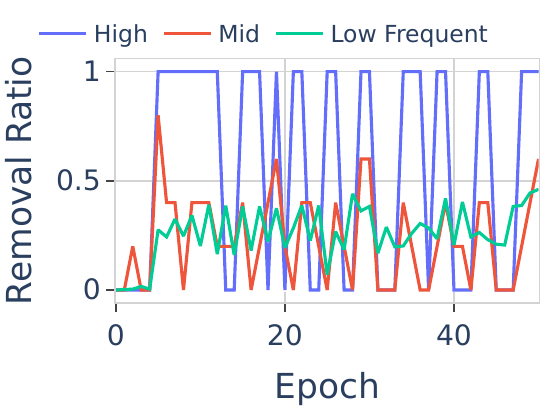}
    }
    \subfloat[$0.5x$]{
        \includegraphics[scale=0.35]{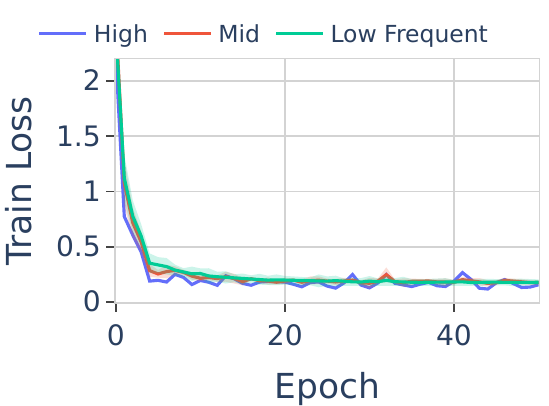}
    }
    \subfloat[$0.5x$]{
        \includegraphics[scale=0.35]{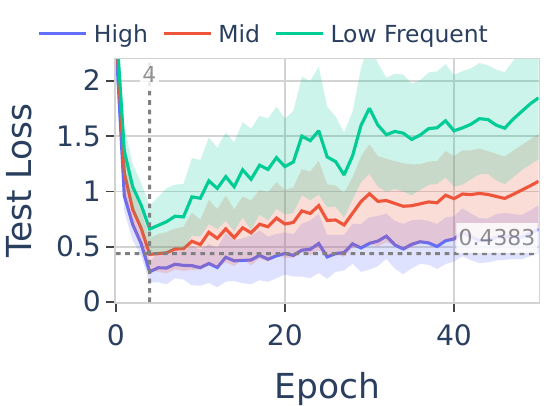}
    }
    \subfloat[$0.5x$]{
        \includegraphics[scale=0.35]{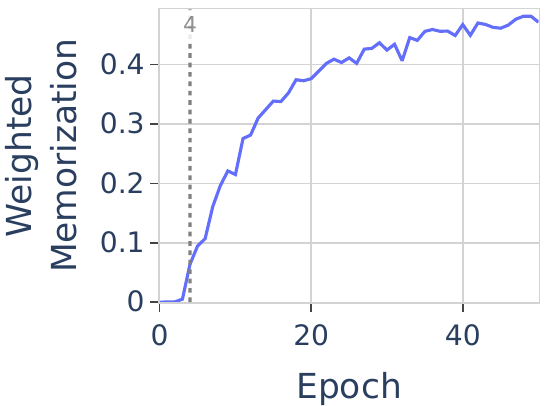}
    }

    \subfloat[$x$]{
        \includegraphics[scale=0.35]{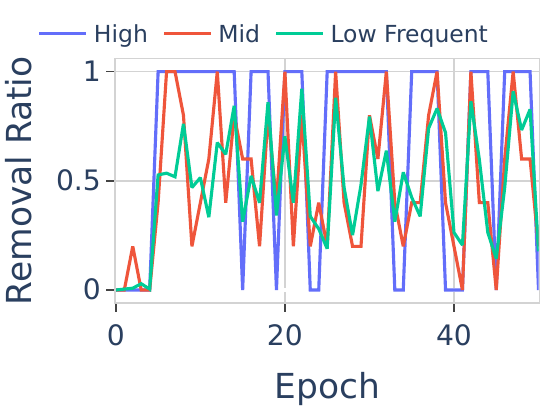}
    }
    \subfloat[$x$]{
        \includegraphics[scale=0.35]{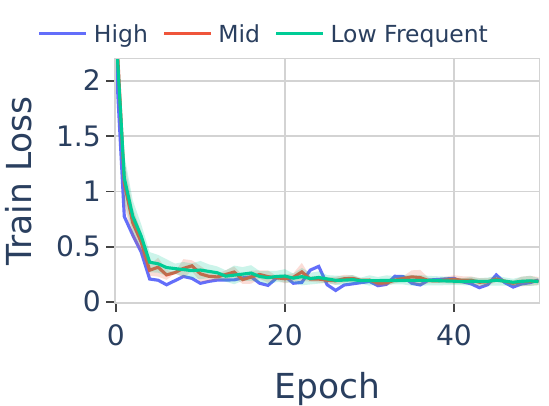}
    }
    \subfloat[$x$]{
        \includegraphics[scale=0.35]{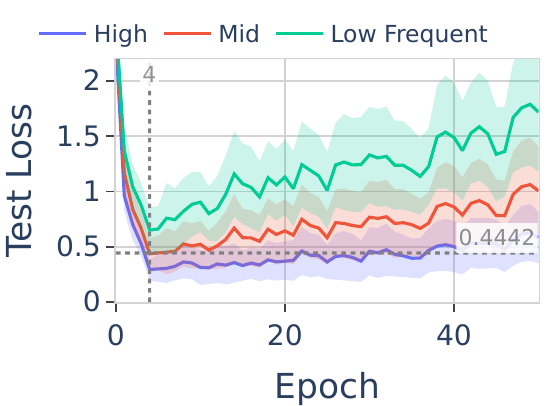}
    }
    \subfloat[$x$]{
        \includegraphics[scale=0.35]{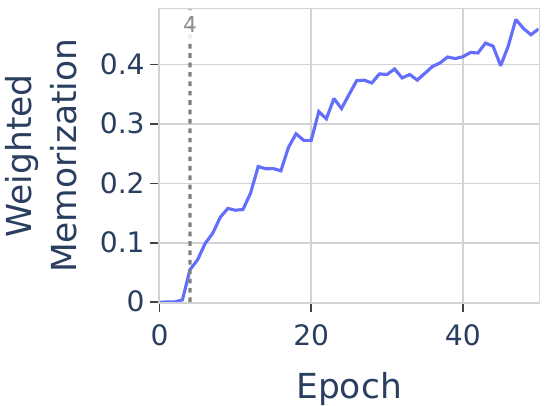}
    }

    \subfloat[$2x$]{
        \includegraphics[scale=0.35]{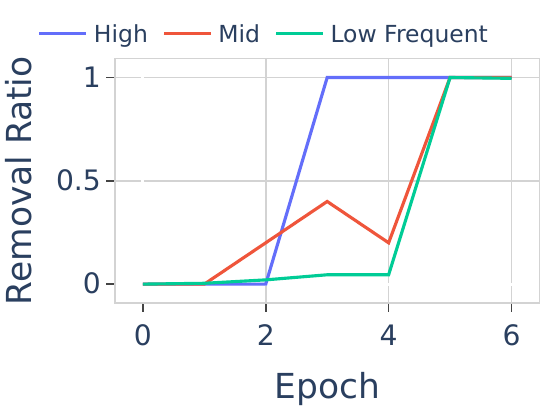}
    }
    \subfloat[$2x$]{
        \includegraphics[scale=0.35]{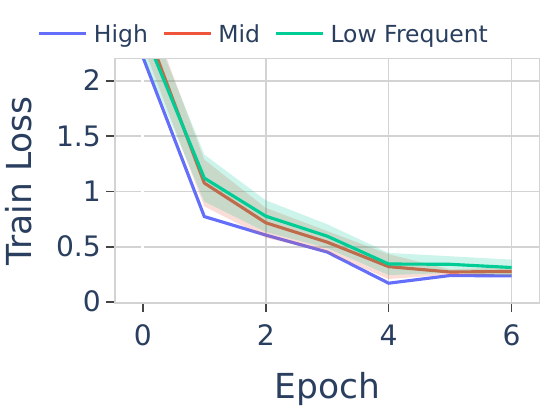}
    }
    \subfloat[$2x$]{
        \includegraphics[scale=0.35]{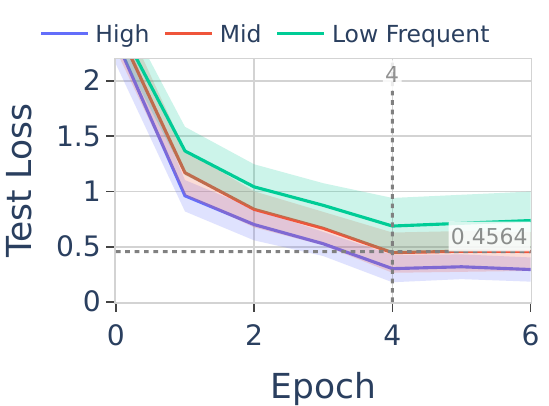}
    }
    \subfloat[$2x$]{
        \includegraphics[scale=0.35]{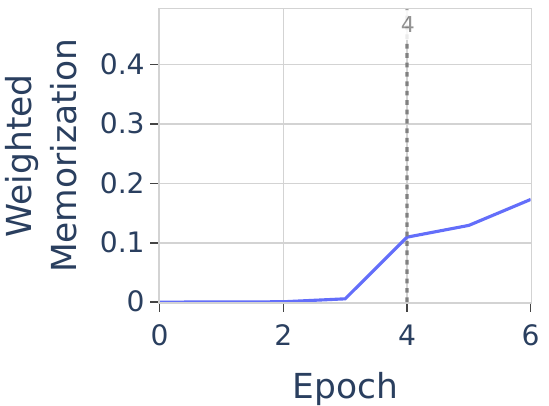}
    }

    \caption{Same as Figure~\ref{fig:mat_threshold_dynamics_l2}, but on language $L_9$.}
    \label{fig:mat_threshold_dynamics_l9}

\end{figure}

\begin{figure}[h]
    \centering

    \subfloat[$L_2$]{
        \includegraphics[scale=0.45]{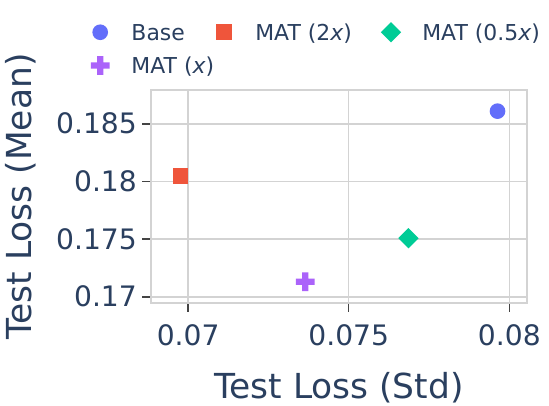}
    }
    \subfloat[$L_2$]{
        \includegraphics[scale=0.45]{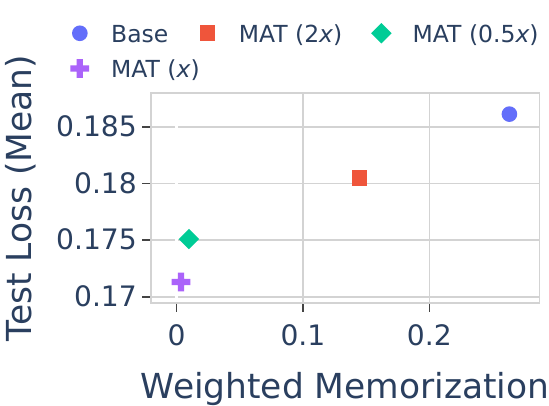}
    }
    \subfloat[$L_2$]{
        \includegraphics[scale=0.45]{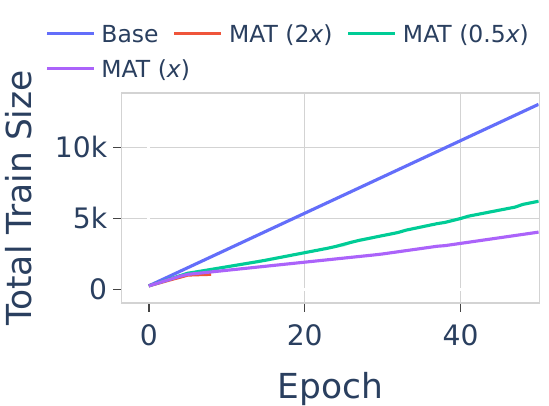}
    }

    \subfloat[$L_4$]{
        \includegraphics[scale=0.45]{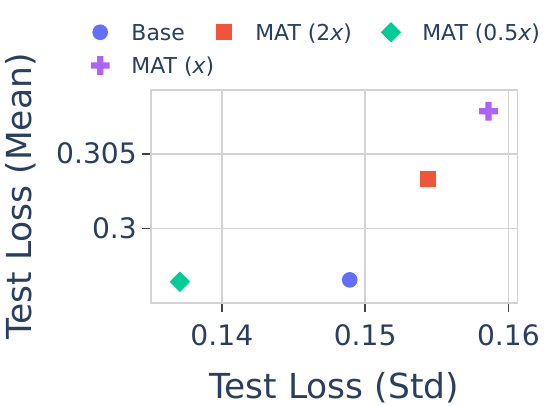}
    }
    \subfloat[$L_4$]{
        \includegraphics[scale=0.45]{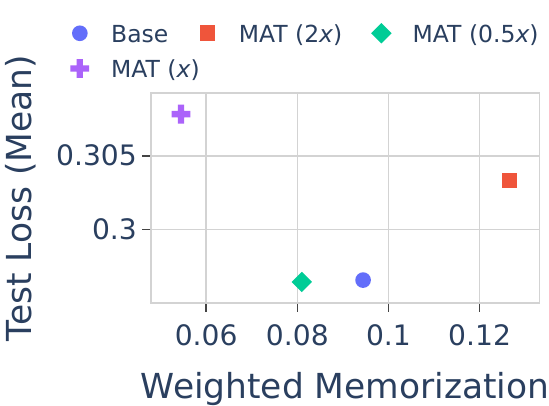}
    }
    \subfloat[$L_4$]{
        \includegraphics[scale=0.45]{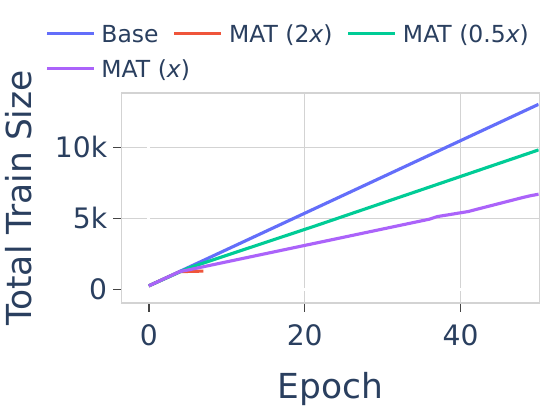}
    }

    \subfloat[$L_9$]{
        \includegraphics[scale=0.45]{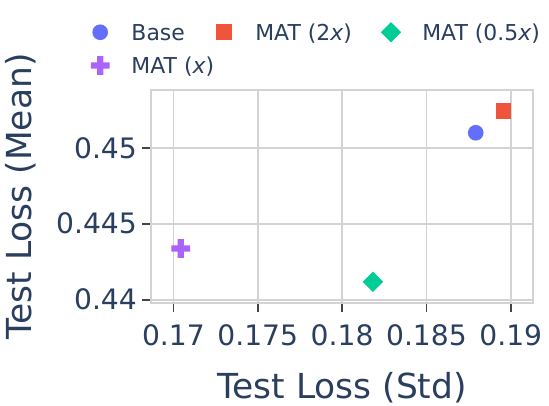}
    }
    \subfloat[$L_9$]{
        \includegraphics[scale=0.45]{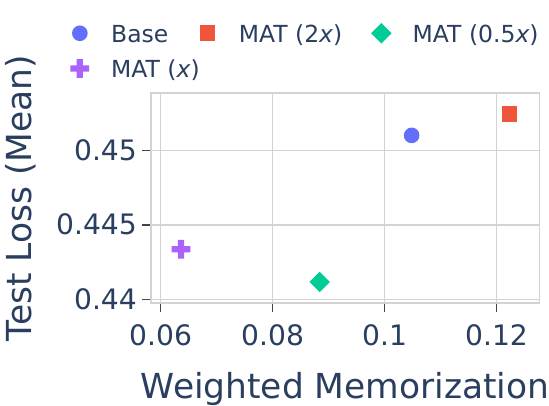}
    }
    \subfloat[$L_9$]{
        \includegraphics[scale=0.45]{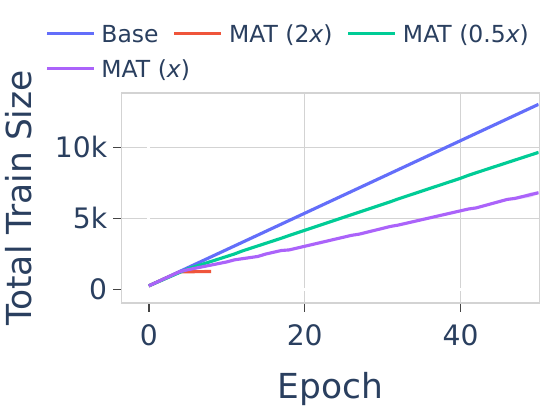}
    }

    \caption{Summary of MAT's threshold sensitivity (Base, $0.5x$, $x$, $2x$) across languages $L_2$, $L_4$, $L_9$ (top to bottom) using Mistral-$7$B, averaged over three seeds. All statistics are measured at the epoch of optimal learning (lowest test loss) for each variant, not at the end of training. $x$ wins on memorization in every language tested. $0.5x$ occasionally improves stability or mean test loss, most notably in $L_4$, where it beats $x$ on both std and mean, but always at a cost in memorization ($L_4$: 0.081 vs. $x$'s 0.054). $L_4$ is the one setting where $x$ trades away some stability for its memorization gains; in $L_2$ and $L_9$, $x$ is best or near-best on both fronts. From an explainability perspective, $x$ remains the simplest to justify -- it down-samples exactly the fraction of training strings matching the fraction of test strings whose loss worsened.}
    \label{fig:mat_threshold_summary}

\end{figure}

\clearpage
\subsection{MAT with a Third Validation Set}
\label{sec:mat_val_ablation}

{
In the main paper, MAT's down-sampling fraction $x$ is derived from the test set. To confirm this is not required in practice, we re-run MAT with a proper three-way split: train/validation/test, derive $x$ from the held-out validation set instead (`MAT (w/ val)'), and use the test set to evaluate learning. We compare Base, MAT (test-derived, as in the main paper), and MAT (w/ val) on $L_2$, $L_4$, $L_9$ using Mistral-$7$B. Figure~\ref{fig:mat_val_dynamics} shows test-loss dynamics (single seed); Figure~\ref{fig:mat_val_summary} summarizes memorization, learning stability, and training cost across all three schemes, averaged over three seeds.
}

\begin{figure}[h]
    \centering

    \subfloat[$L_2$, train]{
        \includegraphics[scale=0.35]{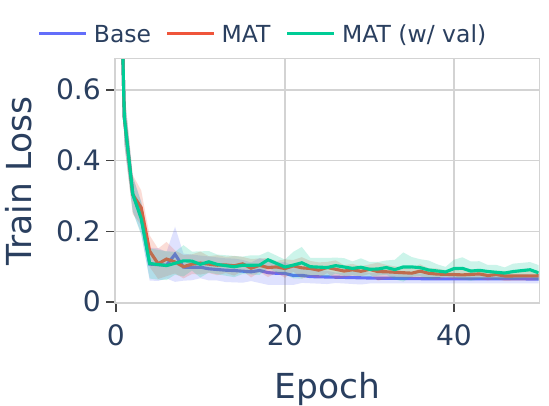}
    }\hspace{1em}
    \subfloat[$L_2$, test]{
        \includegraphics[scale=0.35]{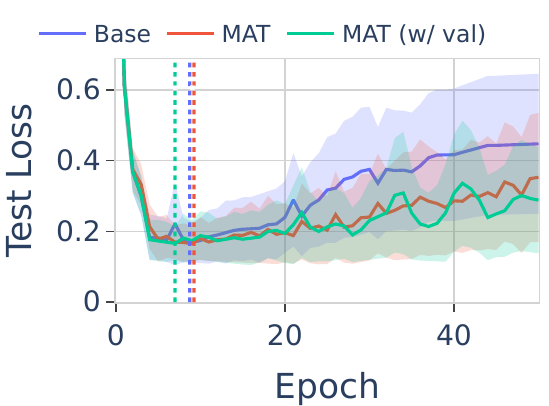}
    }\hspace{1em}
    \subfloat[$L_2$, contextual]{
        \includegraphics[scale=0.35]{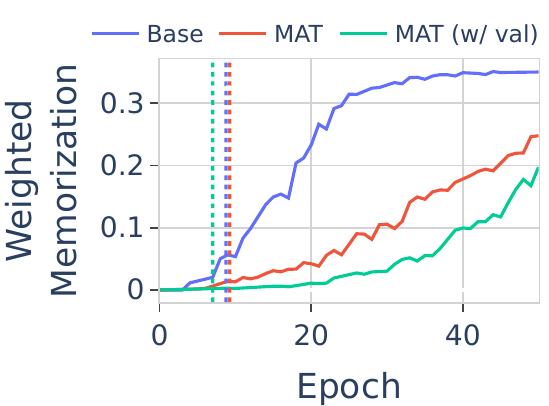}
    }

    \subfloat[$L_4$, train]{
        \includegraphics[scale=0.35]{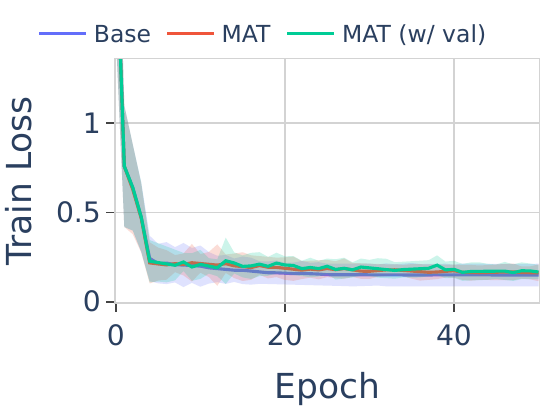}
    }\hspace{1em}
    \subfloat[$L_4$, test]{
        \includegraphics[scale=0.35]{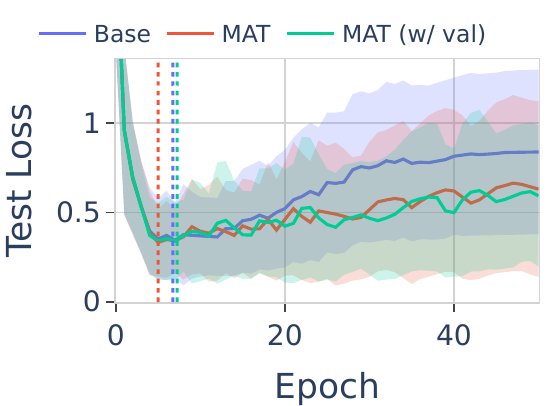}
    }\hspace{1em}
    \subfloat[$L_4$, contextual]{
        \includegraphics[scale=0.35]{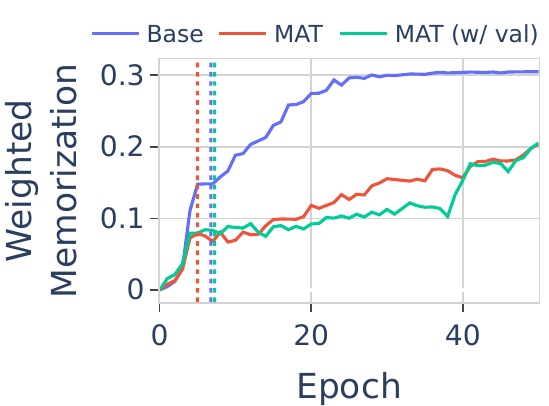}
    }

    \subfloat[$L_9$, train]{
        \includegraphics[scale=0.35]{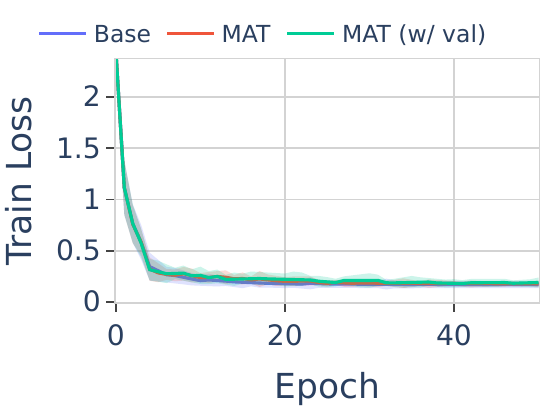}
    }\hspace{1em}
    \subfloat[$L_9$, test]{
        \includegraphics[scale=0.35]{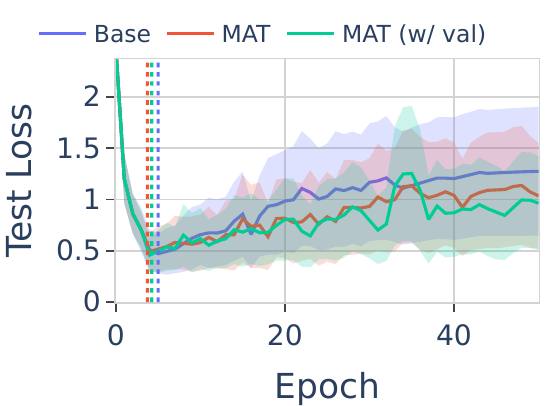}
    }\hspace{1em}
    \subfloat[$L_9$, contextual]{
        \includegraphics[scale=0.35]{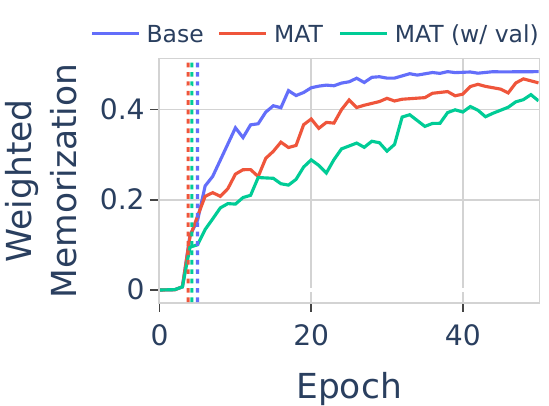}
    }

    \caption{Training dynamics under Base, MAT (test-derived), and MAT (w/ val) on $L_2$, $L_4$, $L_9$ using Mistral-$7$B (seed $0$).}
    \label{fig:mat_val_dynamics}

\end{figure}

\begin{figure}[h]
    \centering

    \subfloat[$L_2$]{
        \includegraphics[scale=0.45]{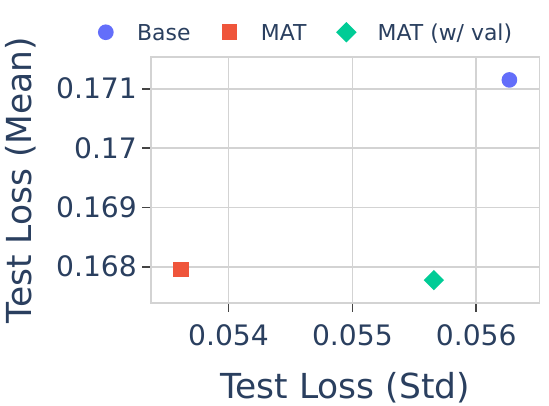}
    }
    \subfloat[$L_2$]{
        \includegraphics[scale=0.45]{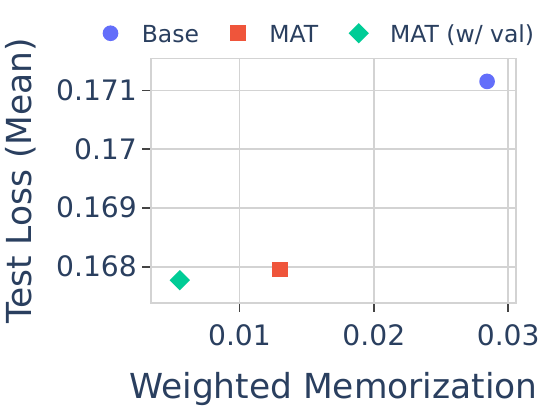}
    }
    \subfloat[$L_2$]{
        \includegraphics[scale=0.45]{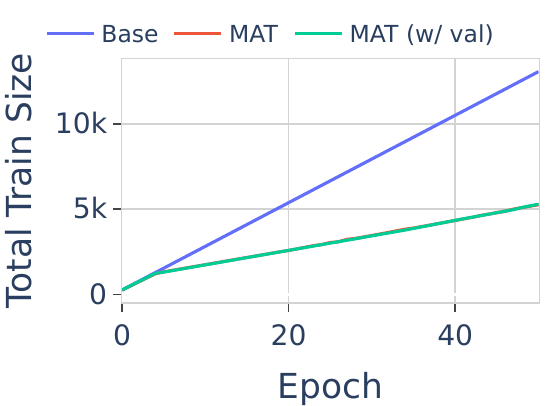}
    }

    \subfloat[$L_4$]{
        \includegraphics[scale=0.45]{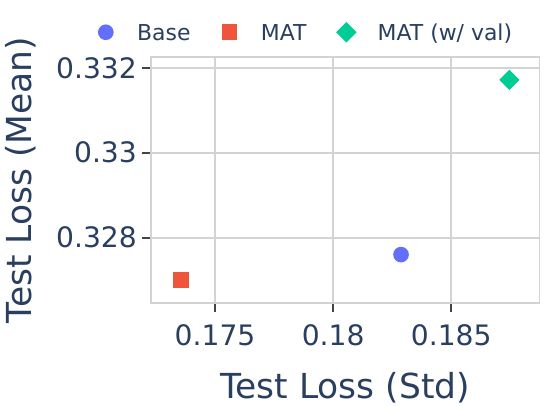}
    }
    \subfloat[$L_4$]{
        \includegraphics[scale=0.45]{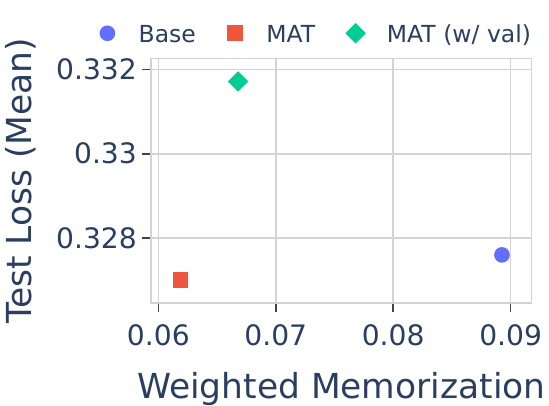}
    }
    \subfloat[$L_4$]{
        \includegraphics[scale=0.45]{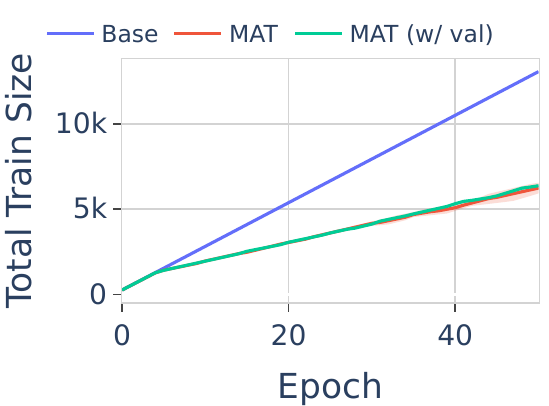}
    }

    \subfloat[$L_9$]{
        \includegraphics[scale=0.45]{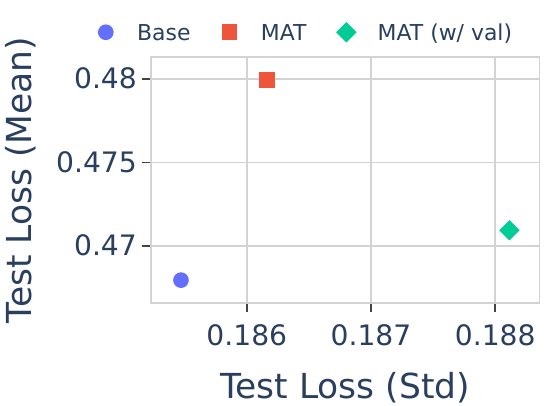}
    }
    \subfloat[$L_9$]{
        \includegraphics[scale=0.45]{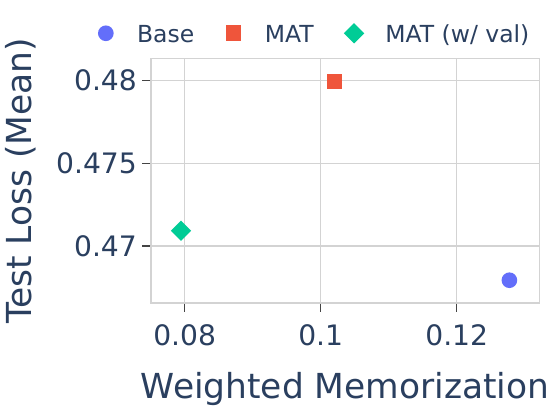}
    }
    \subfloat[$L_9$]{
        \includegraphics[scale=0.45]{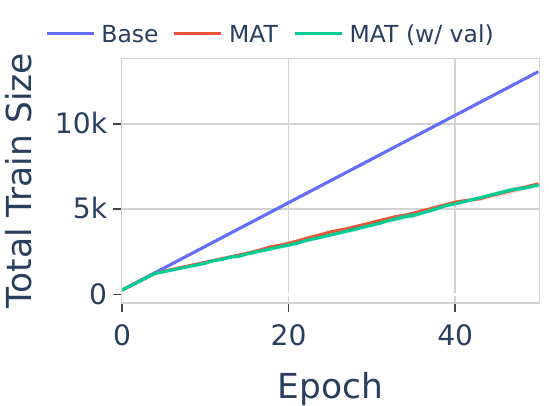}
    }

    \caption{When comparing test-driven split MAT and validation-driven split MAT (w/ val), memorization is comparable or better under the validation split. It improves further in $L_2$ and $L_9$, and stays close in $L_4$. Learning stability (std) is slightly better under the test-derived split in all three languages, though the gap is small in $L_2$ and $L_9$ and only visible in $L_4$. Thus, the memorization benefit transfers cleanly to a realistic held-out-validation setup, at only a small cost in stability.}
    \label{fig:mat_val_summary}

\end{figure}

\clearpage
\subsection{Comparison of MAT and MAT* with Prior Data-pruning Approaches~\cite{marion2023less}}
\label{sec:mat_vs_data_pruning}

Following~\citet{marion2023less}, we first train a reference model on the full training dataset and select the checkpoint with the lowest test loss. This reference model is used only to rank training strings by perplexity. We then retain $ 33\% $ of the training strings from either the bottom, middle, or top of this ranking, and down-sample the remaining strings. For example, bottom-ranked strings have the lowest training loss and are therefore easier for the reference model. For each configuration (bottom, middle, or top), we train a separate pruned model on the reduced dataset and compare its performance with the others.

\begin{figure}[h]
    \centering

    \subfloat[Base Training]{\includegraphics[scale=0.35]{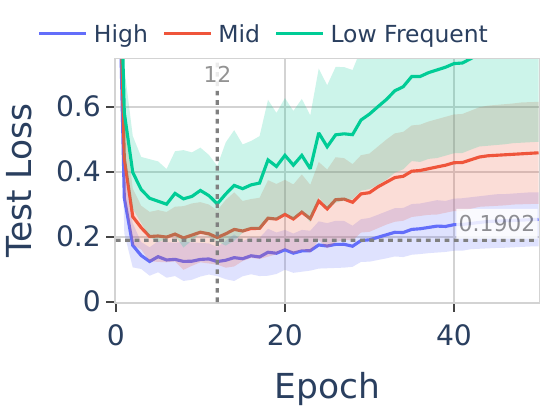}}
    \subfloat[Base Training]{\includegraphics[scale=0.35]{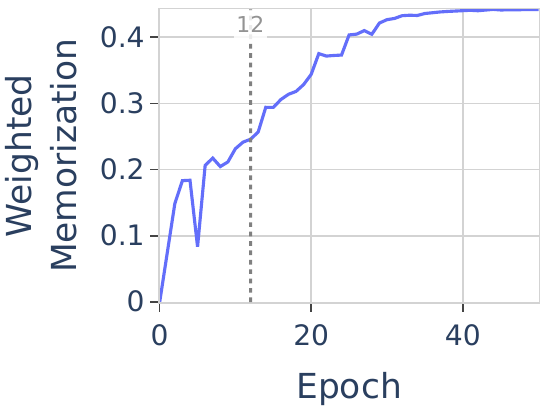}}
    \subfloat[Keep Bottom]{\includegraphics[scale=0.35]{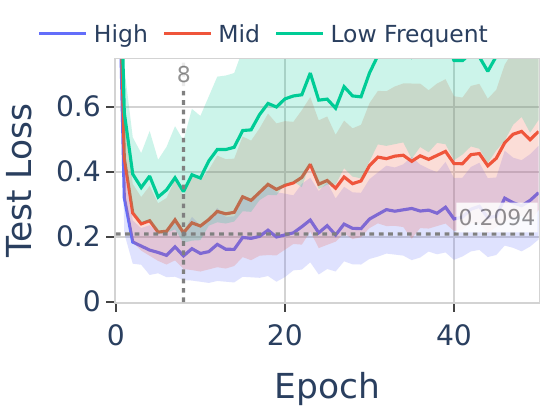}}
    \subfloat[Keep Bottom]{\includegraphics[scale=0.35]{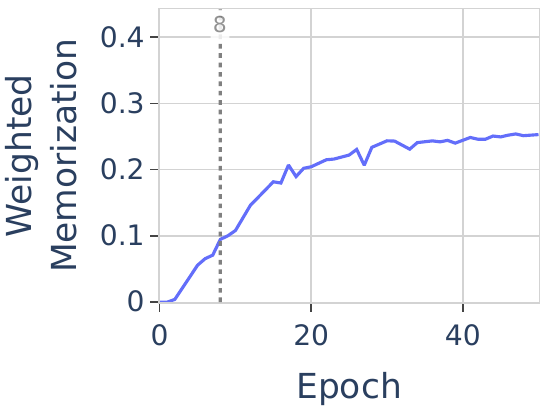}}

    \subfloat[Keep Middle]{\includegraphics[scale=0.35]{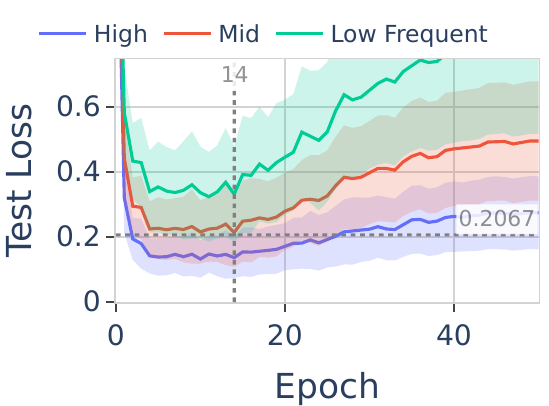}}
    \subfloat[Keep Middle]{\includegraphics[scale=0.35]{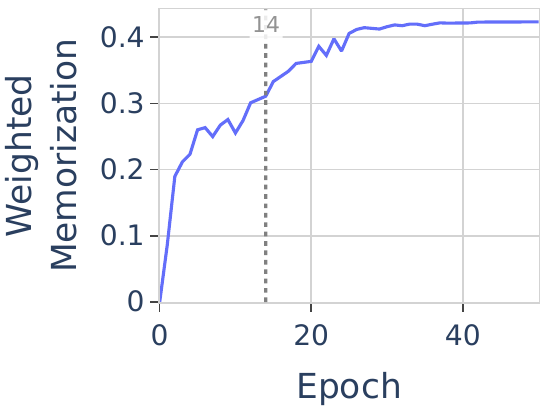}}    
    \subfloat[Keep Top]{\includegraphics[scale=0.35]{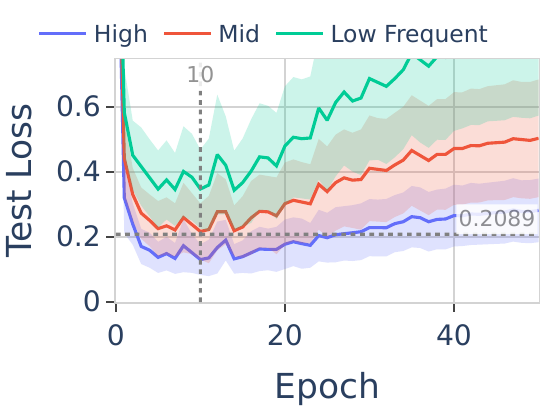}}
    \subfloat[Keep Top]{\includegraphics[scale=0.35]{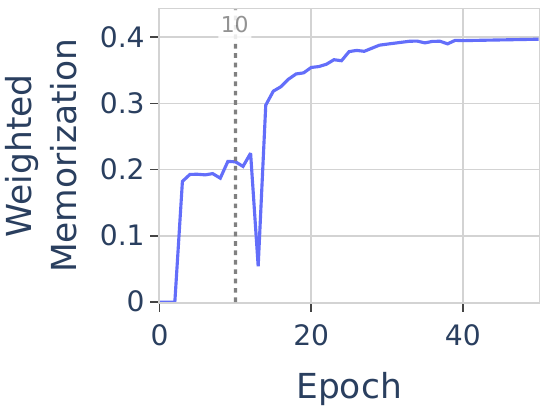}}

    \subfloat[MAT]{\includegraphics[scale=0.35]{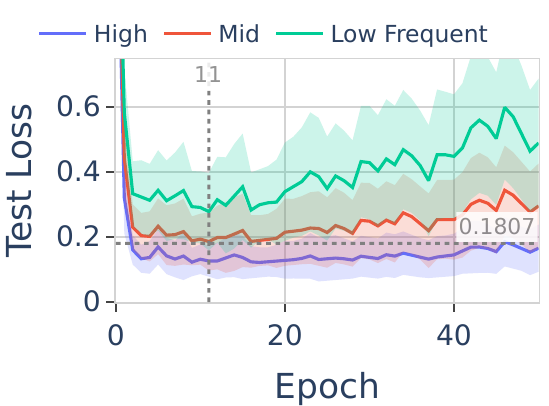}}
    \subfloat[MAT]{\includegraphics[scale=0.35]{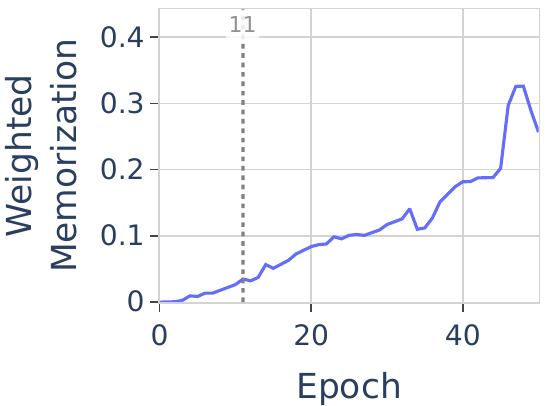}}    
    \subfloat[MAT*]{\includegraphics[scale=0.35]{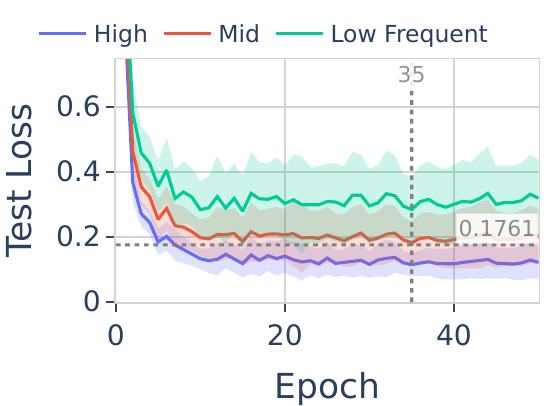}}
    \subfloat[MAT*]{\includegraphics[scale=0.35]{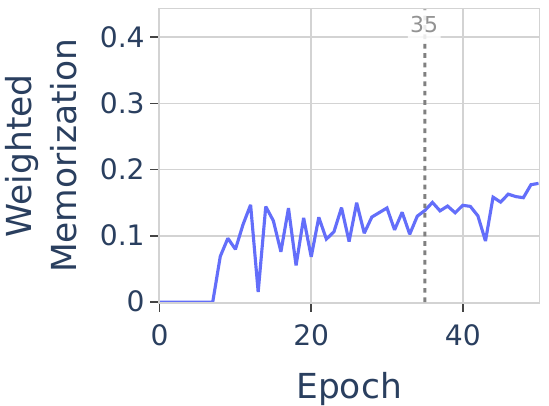}}

    \caption{(Column 1 and 3): Test loss of strings from different frequency groups in a language. (Column 2 and 4): Contextual memorization of training strings. Vertical line indicates the lowest test loss of all test strings. Results are obtained using Mistral-$ 7 $B on $ L_2 $.}

\end{figure}

\begin{figure*}
    \centering
    
    \subfloat{\includegraphics[scale=0.5]{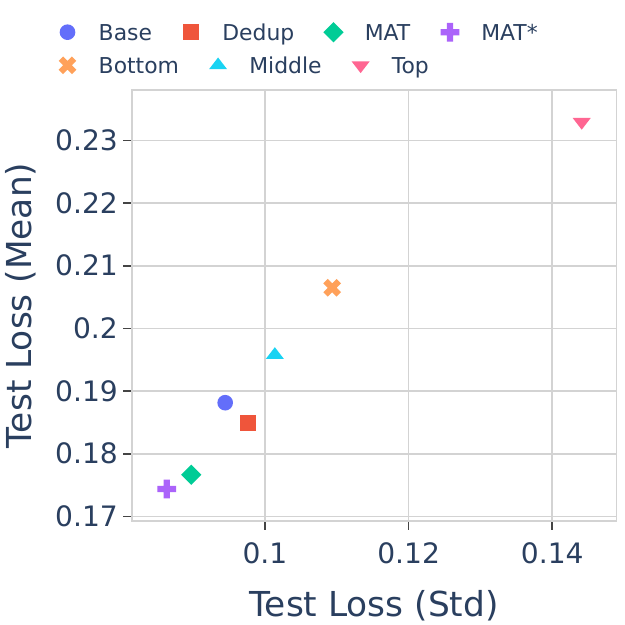}}\hfil
    \subfloat{\includegraphics[scale=0.5]{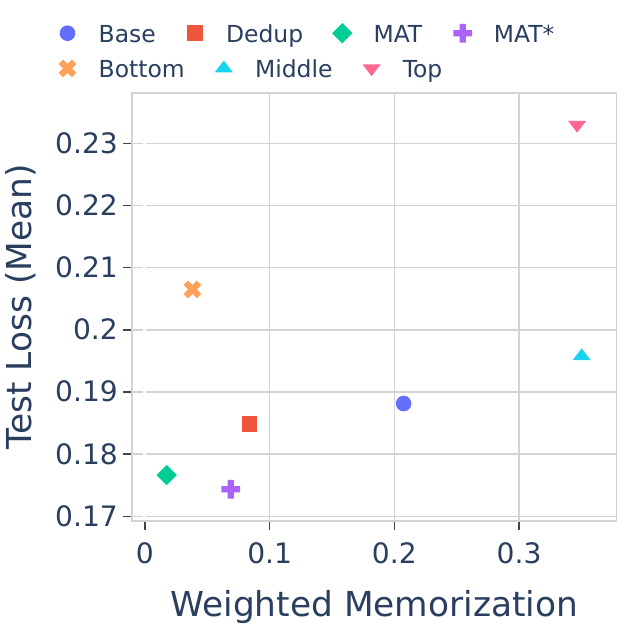}}\hfil
    \subfloat{\includegraphics[scale=0.5]{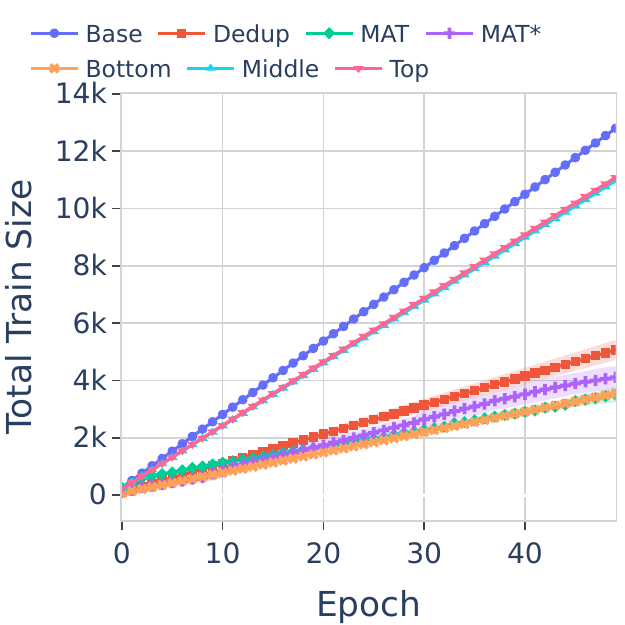}}

    \caption{Impact of different training schemes on learning, memorization, and training cost. A scheme on the lower-left side of the first two plots is better. In the third plot, lower training size is better. For methods (keep) bottom, middle, and top, we keep $ 33\% $ training strings, ranked by loss with respect to the `base' (default) training scheme at respective optimal epoch. Compared to down-sampling heuristics of~\citet{marion2023less}, both MAT and MAT* result in lower mean and standard deviation of test loss and lower contextual memorization. Result is the average of $ 3 $ runs of Mistral-$ 7 $B on $ L_2 $. }
\end{figure*}

\clearpage
\subsection{MAT on Pre-training}
\label{sec:mat_pretraining}

\begin{figure*}[h]
    \centering
    \subfloat[]{\includegraphics[scale=0.45]{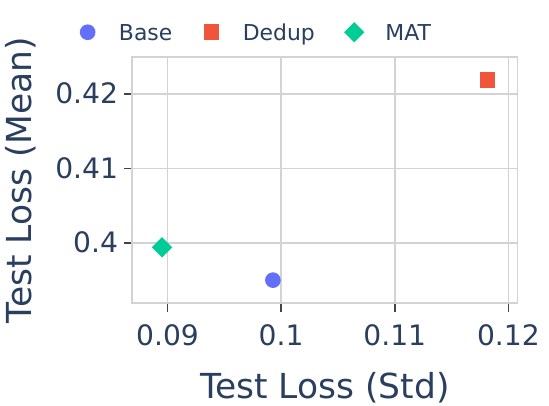}}
    \subfloat[]{\includegraphics[scale=0.45]{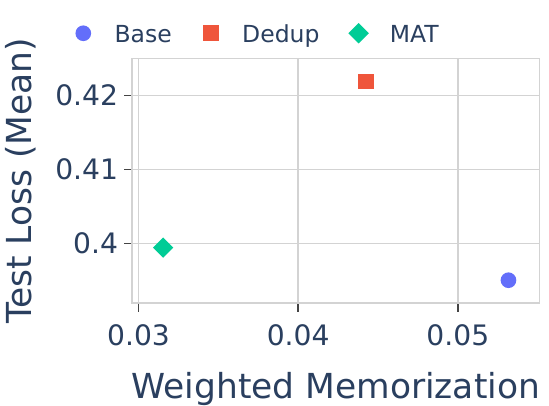}}
    \subfloat[]{\includegraphics[scale=0.45]{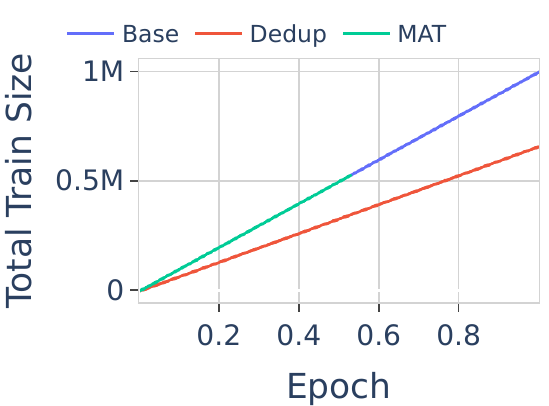}}

    \caption{Learning, memorization, and training cost efficiency in a large training setup, where \textit{new} training strings are added after each epoch, shown for $ L_{10} $. Therefore, the experiment mimics a \textbf{pre-training} setup, where the LLM passes over the training dataset only once and there is no repetition. Yet, MAT is effective in learning, memorization, and training cost efficiency.}
\end{figure*}

\clearpage
\subsection{MAT on Pre-training on a Natural Language Dataset}
\label{sec:mat_nlp_appendix}

\begin{figure}[h]
    \centering

    \subfloat[Tiny Shakespeare]{
        \includegraphics[scale=0.45]{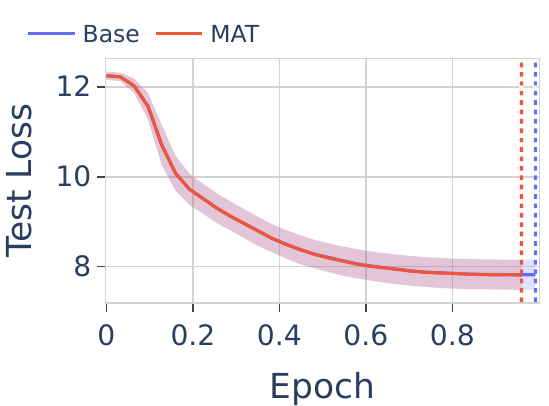}
        \label{fig:epoch_mat}
    }\hfil
    \subfloat[Tiny Shakespeare]{
        \includegraphics[scale=0.45]{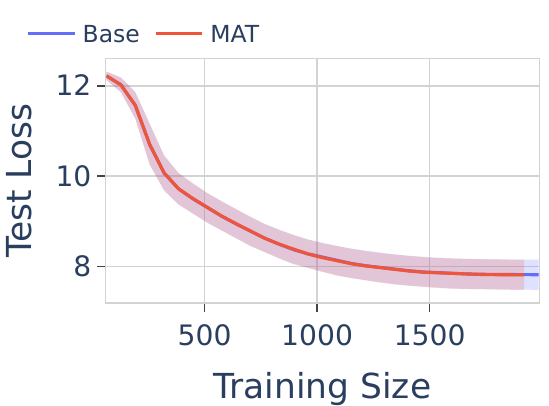}
        \label{fig:size_mat}
    }\hfil
    \subfloat[Tiny Shakespeare]{
        \includegraphics[scale=0.45]{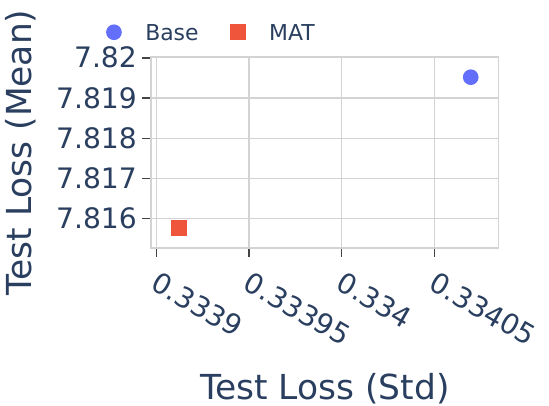}
        \label{fig:tradeoff_mat}
    }

    \subfloat[FineWeb-Edu (English)]{
        \includegraphics[scale=0.45]{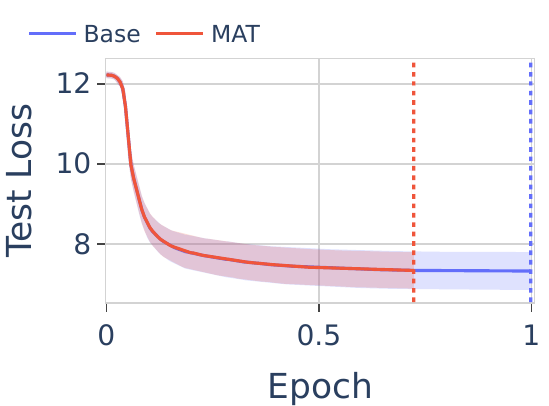}
        \label{fig:epoch_mat_fw}
    }\hfil
    \subfloat[FineWeb-Edu (English)]{
        \includegraphics[scale=0.45]{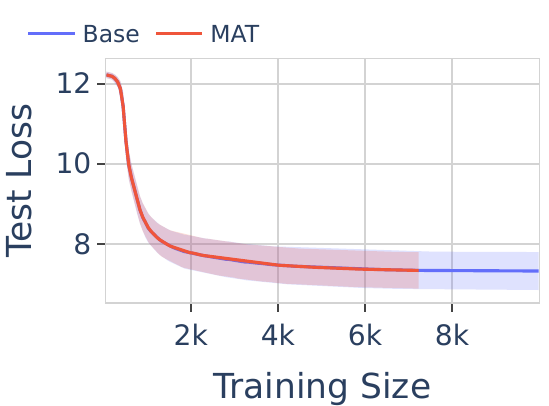}
        \label{fig:size_mat_fw}
    }\hfil
    \subfloat[FineWeb-Edu (English)]{
        \includegraphics[scale=0.45]{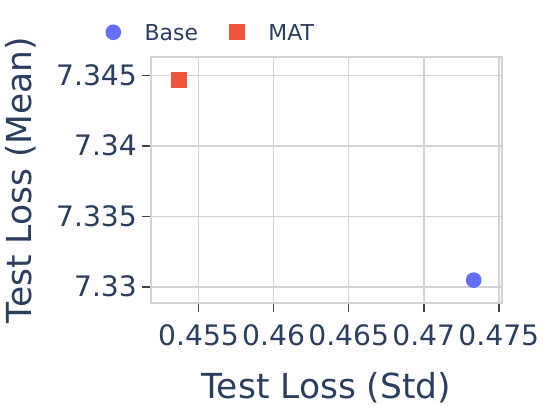}
        \label{fig:tradeoff_mat_fw}
    }

    \subfloat[German]{
        \includegraphics[scale=0.45]{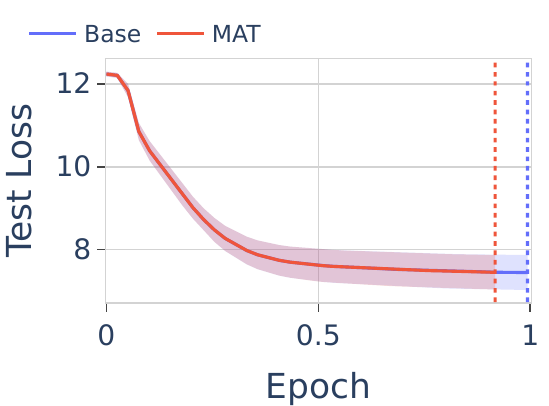}
        \label{fig:epoch_mat_de}
    }\hfil
    \subfloat[German]{
        \includegraphics[scale=0.45]{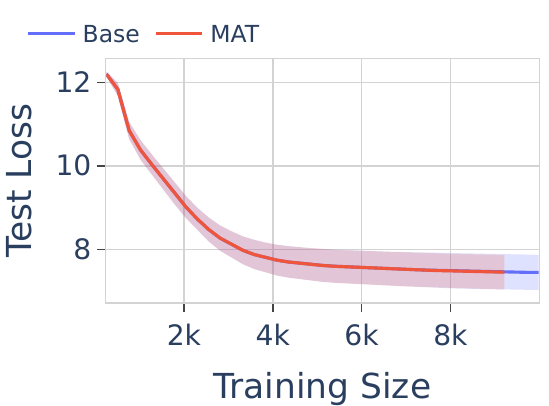}
        \label{fig:size_mat_de}
    }\hfil
    \subfloat[German]{
        \includegraphics[scale=0.45]{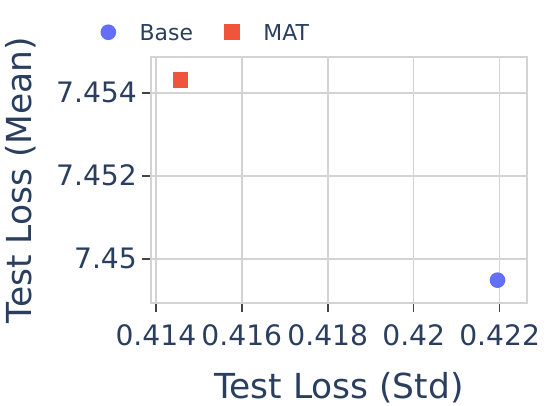}
        \label{fig:tradeoff_mat_de}
    }

    \caption{
        \textbf{MAT on natural-language pre-training (1/2).} We pre-train a Qwen-$2.5$-$1.5$B model from scratch without any data repetition (i.e., a single epoch), independently on Tiny Shakespeare, FineWeb-Edu~\cite{lozhkov2024fineweb-edu}, and the German split of FineWeb2~\cite{penedo2025fineweb2}. Columns show, left to right: test loss vs.\ epoch, test loss vs.\ training size, and the mean/variance test-loss trade-off at the optimal learning point (when mean test loss is minimized). In all three datasets, Base and MAT track closely by epoch and training size, with MAT terminating earlier, and MAT always reaches a lower test-loss variance than Base. On Tiny Shakespeare this comes with a lower mean test loss as well; on FineWeb-Edu and German it comes at the cost of a marginally higher mean -- the same pattern continues for Chinese, Swahili, and Bengali in Figure~\ref{fig:mat_nlp_part2}.
    }
    \label{fig:mat_nlp_part1}
\end{figure}

\begin{figure}[h]
    \centering

    \subfloat[Chinese]{
        \includegraphics[scale=0.45]{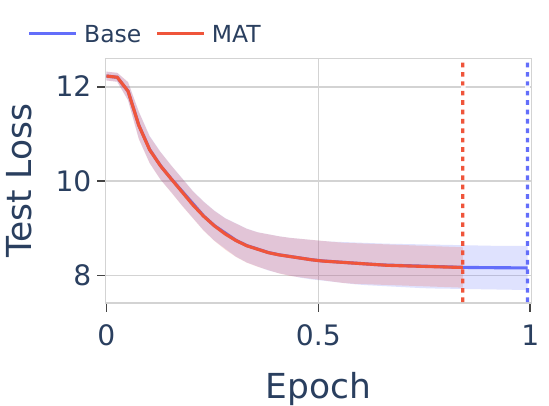}
        \label{fig:epoch_mat_zh}
    }\hfil
    \subfloat[Chinese]{
        \includegraphics[scale=0.45]{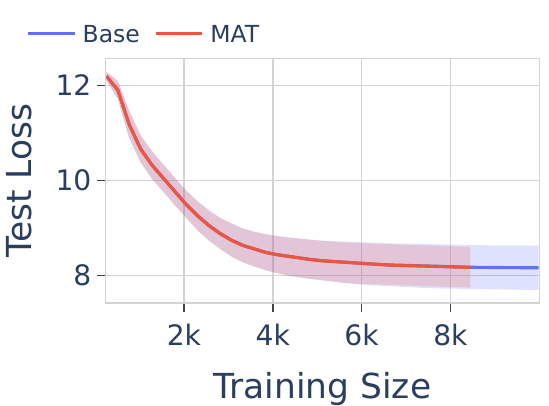}
        \label{fig:size_mat_zh}
    }\hfil
    \subfloat[Chinese]{
        \includegraphics[scale=0.45]{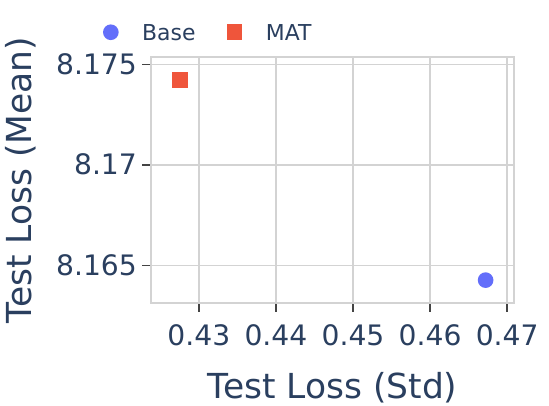}
        \label{fig:tradeoff_mat_zh}
    }

    \subfloat[Swahili]{
        \includegraphics[scale=0.45]{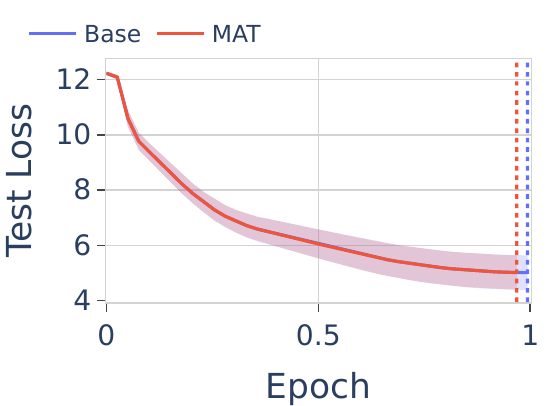}
        \label{fig:epoch_mat_sw}
    }\hfil
    \subfloat[Swahili]{
        \includegraphics[scale=0.45]{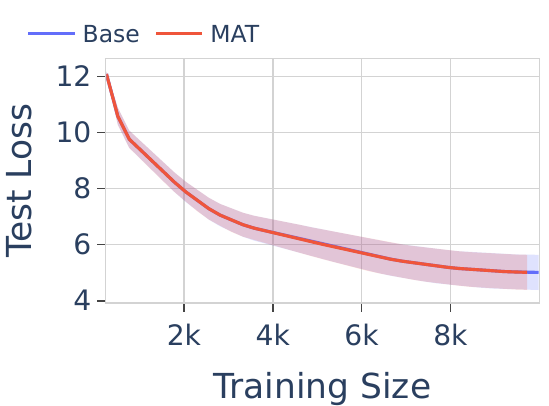}
        \label{fig:size_mat_sw}
    }\hfil
    \subfloat[Swahili]{
        \includegraphics[scale=0.45]{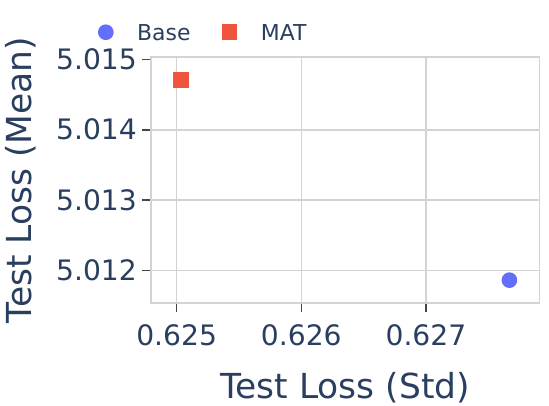}
        \label{fig:tradeoff_mat_sw}
    }

    \subfloat[Bengali]{
        \includegraphics[scale=0.45]{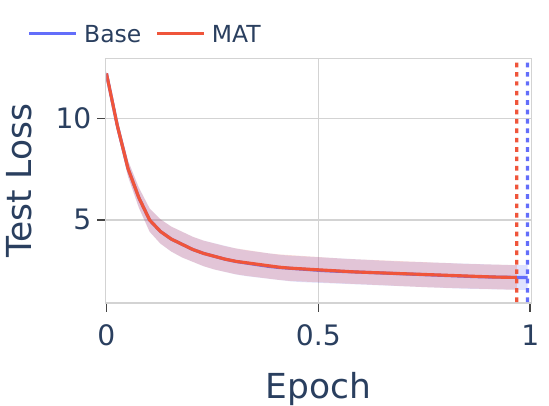}
        \label{fig:epoch_mat_bn}
    }\hfil
    \subfloat[Bengali]{
        \includegraphics[scale=0.45]{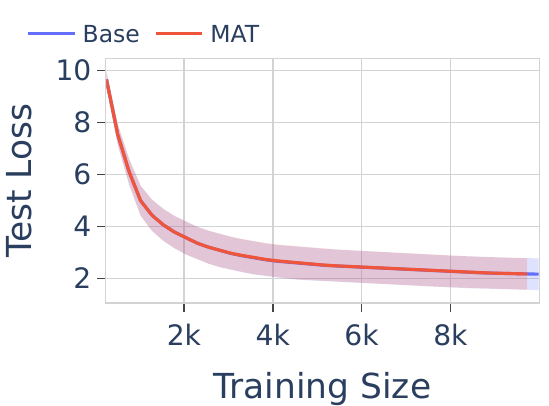}
        \label{fig:size_mat_bn}
    }\hfil
    \subfloat[Bengali]{
        \includegraphics[scale=0.45]{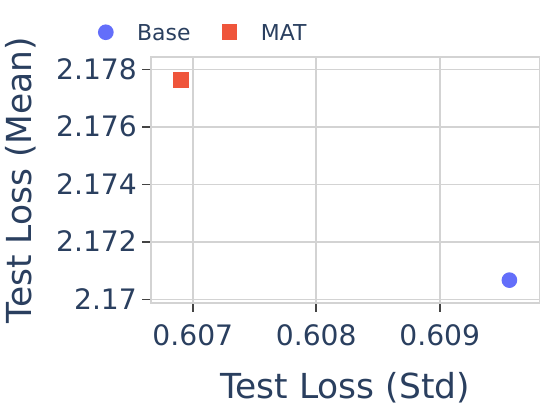}
        \label{fig:tradeoff_mat_bn}
    }

    \caption{
        \textbf{MAT on natural-language pre-training (2/2), continued from Figure~\ref{fig:mat_nlp_part1}.} Same setup, on the Chinese, Swahili, and Bengali splits of FineWeb2~\cite{penedo2025fineweb2}. As with German, Base and MAT track closely by epoch and training size, with MAT terminating earlier, and MAT reaches a lower test-loss variance at the cost of a marginally higher mean in every language here. \textit{Therefore, MAT is effective in reducing test loss variance and training size in natural language as well.}
    }
    \label{fig:mat_nlp_part2}
\end{figure}

\clearpage
\subsection{Distance-Dependent Influence of Training Strings on Test Loss}
\label{sec:distance_dependent_influence}

We examine how the inclusion or exclusion of a training string $s$ from $D_{\text{train}}$ influences (1) the model's loss on $s$ itself, and (2) the loss on structurally related test strings at varying tree distances. For each training string, we compare loss curves under two conditions: $s \in D_{\text{train}}$ (solid line) versus $s \notin D_{\text{train}}$ (dotted line). The first subplot in each figure captures the self-influence: a training string achieves significantly lower loss when included in training, eventually becoming contextually memorized (Figure~\ref{fig:intro_contextual_memorization}). The remaining subplots trace this influence on test strings at increasing tree distances (close: $2$--$10$, medium: $14$--$28$, far: $36$--$68$).

A consistent pattern emerges: structurally close test strings benefit substantially from the presence of the training string, exhibiting noticeably lower loss under inclusion. As tree distance grows, this benefit weakens and can even invert: distant test strings occasionally show \emph{higher} loss when the training string is included. This distance-dependent influence underpins MAT's adaptive down-sampling strategy: because a memorized training string provides diminishing generalization value to structurally distant test strings, removing it is nearly costless for those strings, letting MAT achieve learning stability (test loss remaining low after reaching local minimum) and uniformity (similar loss on all test strings, i.e., lower variance).

\begin{figure}[h]
    \centering
    \subfloat[Training string (self)]{
        \includegraphics[scale=0.35]{
            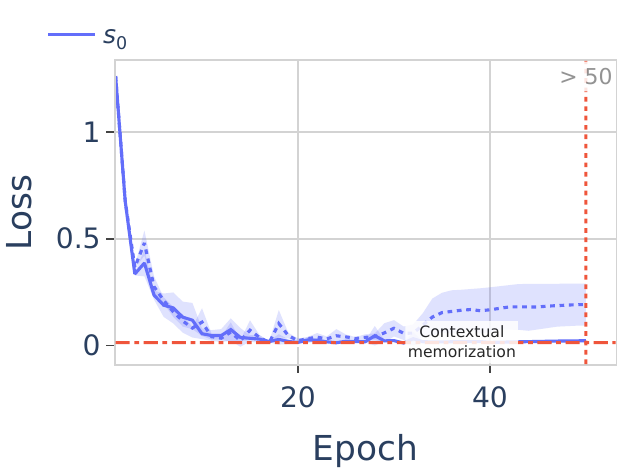
        }
    }

    \subfloat[Tree distance: $ 2 $, test ID: $ 27 $]{
        \includegraphics[scale=0.35]{
            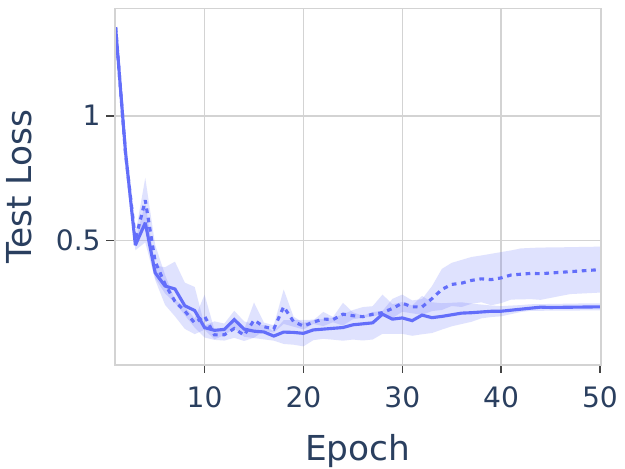
        }
    }
    \subfloat[Tree distance: $ 4 $, test ID: $ 16 $]{
        \includegraphics[scale=0.35]{
            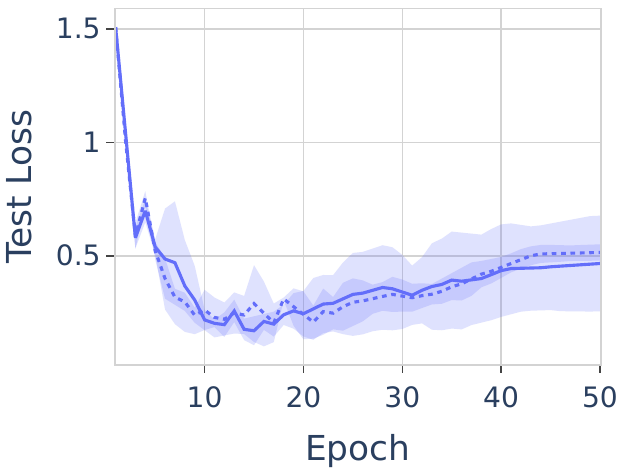
        }
    }
    \subfloat[Tree distance: $ 8 $, test ID: $ 5 $]{
        \includegraphics[scale=0.35]{
            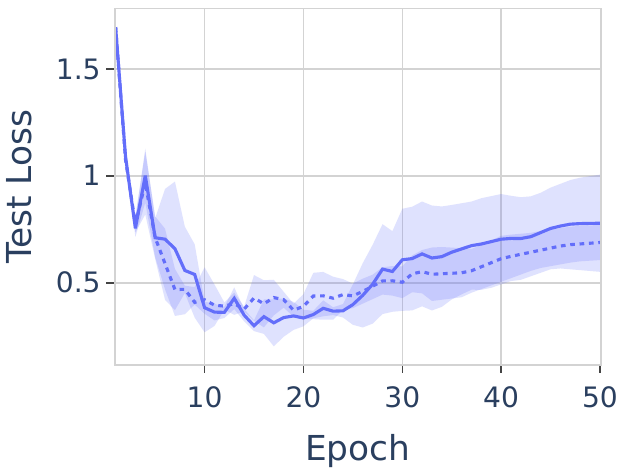
        }
    }

    \subfloat[Tree distance: $ 10 $, test ID: $ 34 $]{
        \includegraphics[scale=0.35]{
            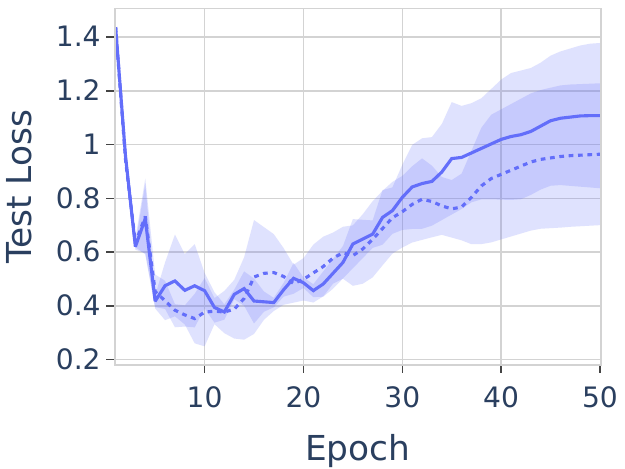
        }
    }
    \subfloat[Tree distance: $ 18 $, test ID: $ 36 $]{
        \includegraphics[scale=0.35]{
            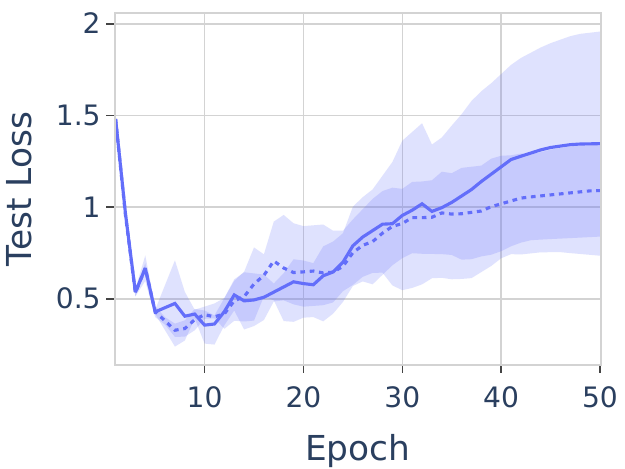
        }
    }
    \subfloat[Tree distance: $ 26 $, test ID: $ 60 $]{
        \includegraphics[scale=0.35]{
            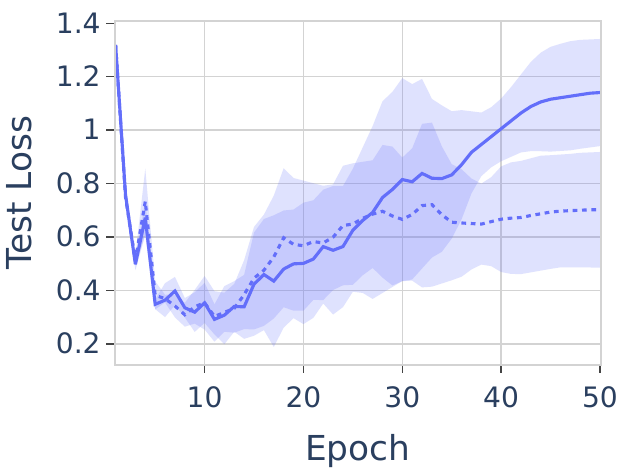
        }
    }

    \subfloat[Tree distance: $ 36 $, test ID: $ 40 $]{
        \includegraphics[scale=0.35]{
            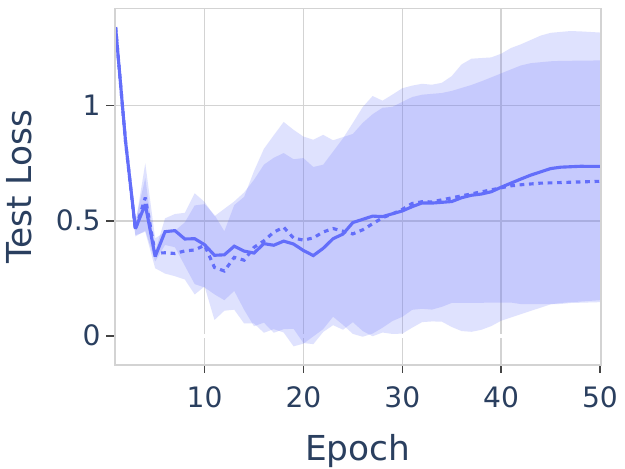
        }
    }
    \subfloat[Tree distance: $ 47 $, test ID: $ 118 $]{
        \includegraphics[scale=0.35]{
            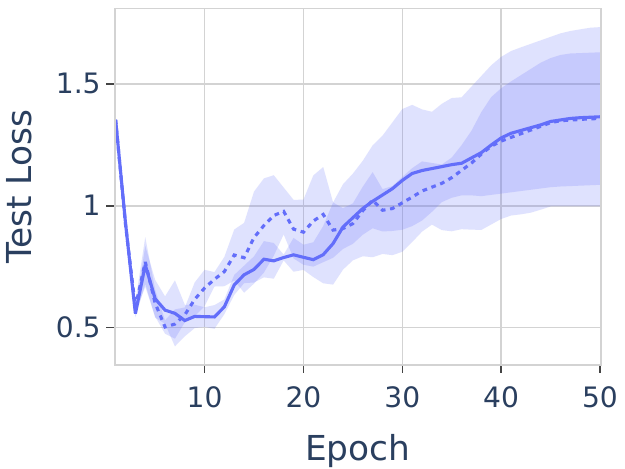
        }
    }
    \subfloat[Tree distance: $ 62 $, test ID: $ 14 $]{
        \includegraphics[scale=0.35]{
            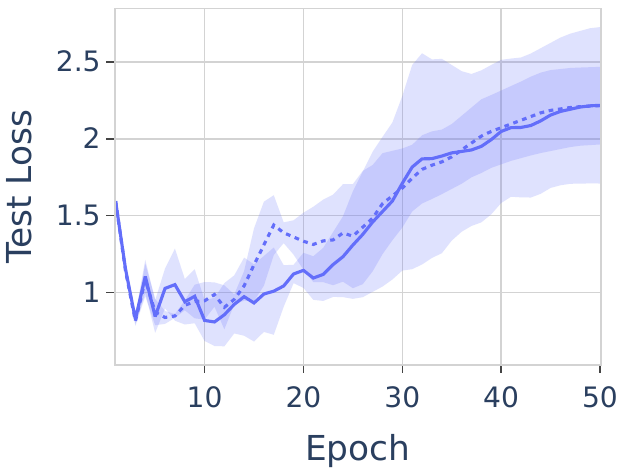
        }
    }

    \caption{Counterfactual influence of training string $ s_0 $ on its own loss~(a) and on test strings at increasing tree distances~(b)--(j). Solid lines ($s \in D_{\text{train}}$) versus dotted lines ($s \notin D_{\text{train}}$) reveal a distance-dependent pattern: nearby test strings exhibit substantially lower loss under inclusion, while the benefit diminishes, and occasionally reverses, for structurally distant test strings.}
\label{fig:distance_dependent_s0}
\end{figure}

\begin{figure}
    \centering
    \subfloat[Training string (self)]{
        \includegraphics[scale=0.35]{
            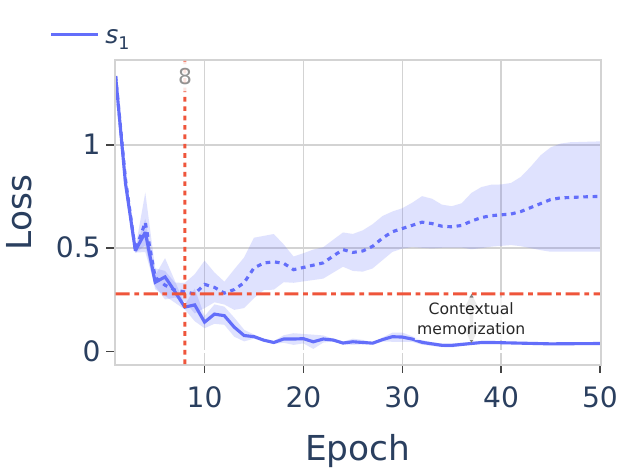
        }
    }

    \subfloat[Tree distance: $ 2 $, test ID: $ 34 $]{
        \includegraphics[scale=0.35]{
            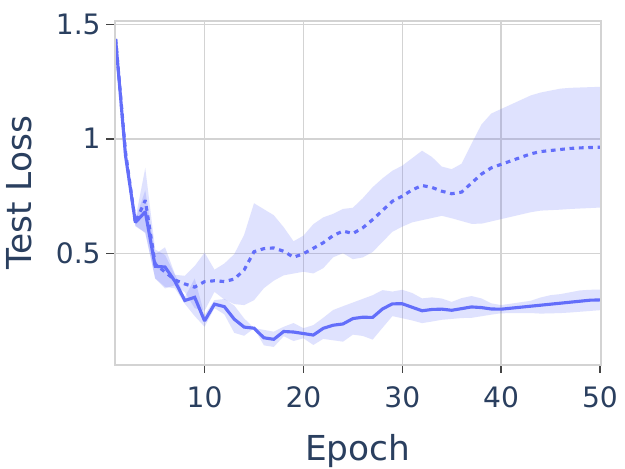
        }
    }
    \subfloat[Tree distance: $ 6 $, test ID: $ 66 $]{
        \includegraphics[scale=0.35]{
            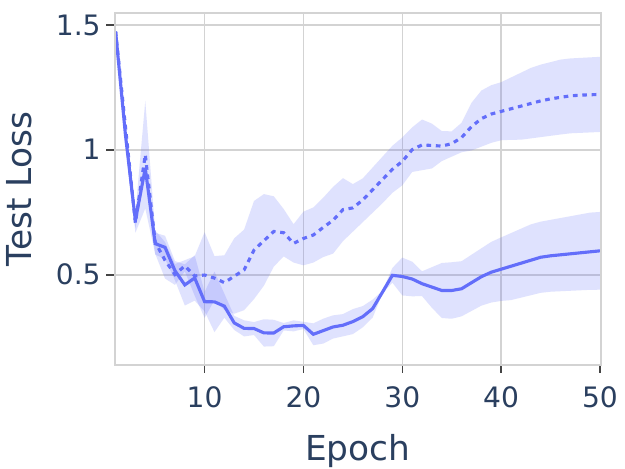
        }
    }
    \subfloat[Tree distance: $ 10 $, test ID: $ 27 $]{
        \includegraphics[scale=0.35]{
            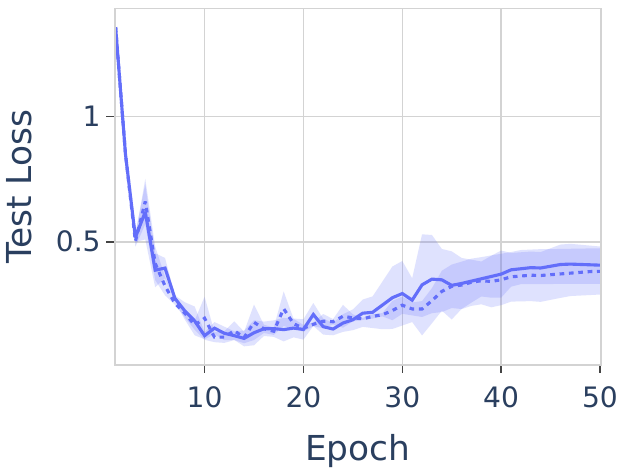
        }
    }

    \subfloat[Tree distance: $ 14 $, test ID: $ 80 $]{
        \includegraphics[scale=0.35]{
            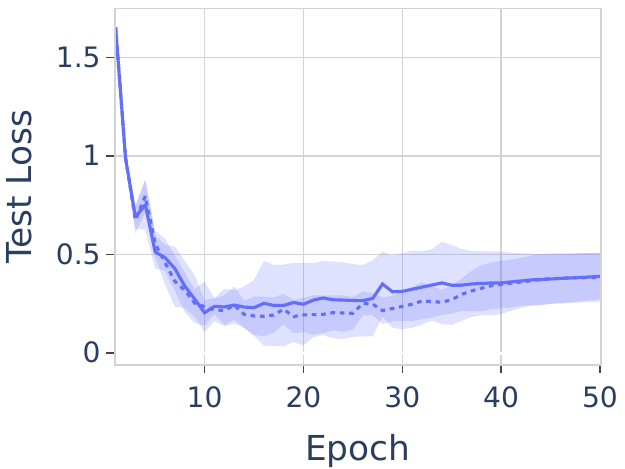
        }
    }
    \subfloat[Tree distance: $ 20 $, test ID: $ 18 $]{
        \includegraphics[scale=0.35]{
            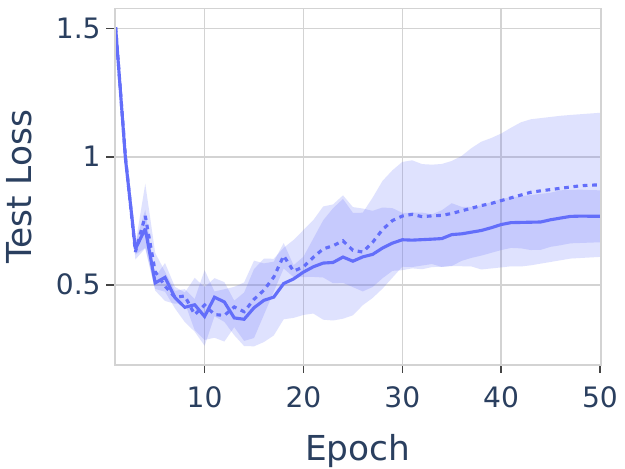
        }
    }
    \subfloat[Tree distance: $ 28 $, test ID: $ 10 $]{
        \includegraphics[scale=0.35]{
            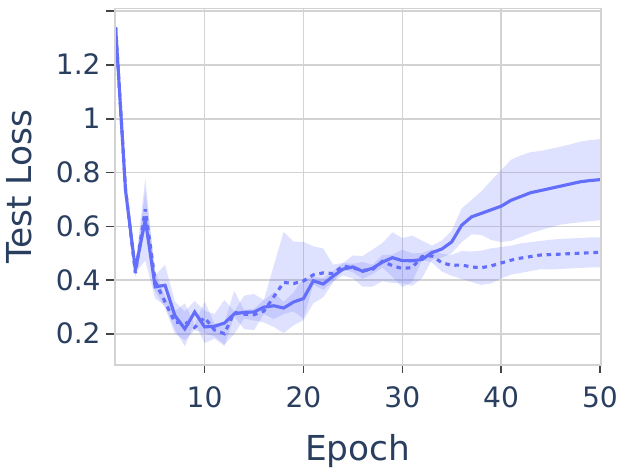
        }
    }

    \subfloat[Tree distance: $ 37 $, test ID: $ 115 $]{
        \includegraphics[scale=0.35]{
            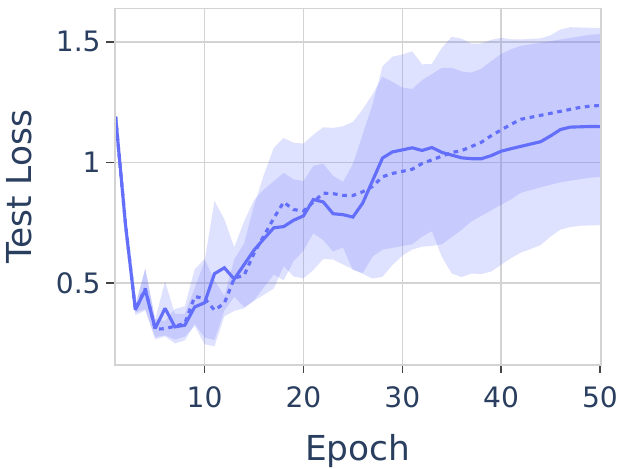
        }
    }
    \subfloat[Tree distance: $ 47 $, test ID: $ 118 $]{
        \includegraphics[scale=0.35]{
            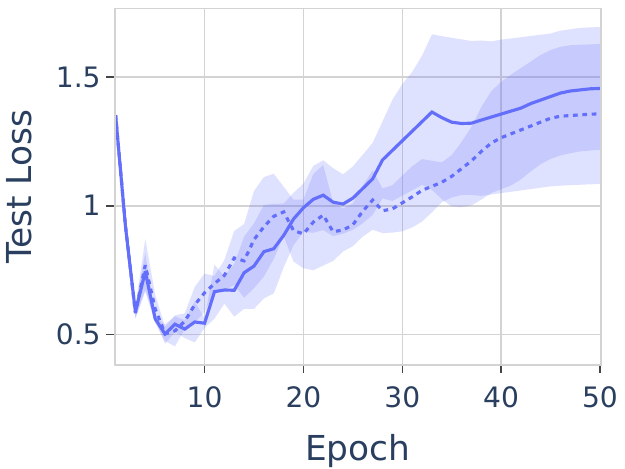
        }
    }
    \subfloat[Tree distance: $ 59 $, test ID: $ 14 $]{
        \includegraphics[scale=0.35]{
            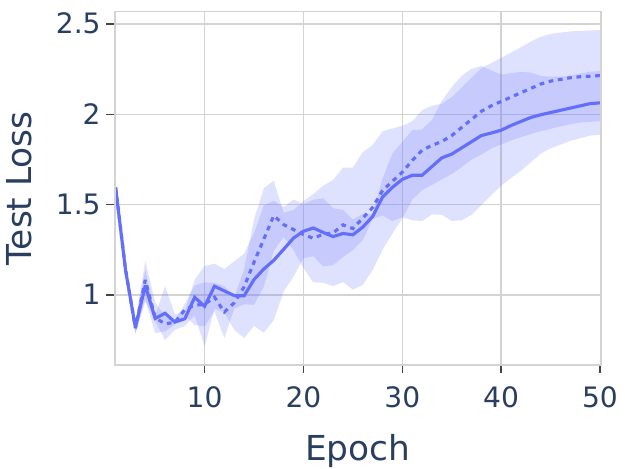
        }
    }

    \caption{Counterfactual influence of training string $ s_1 $ on its own loss~(a) and on test strings at increasing tree distances~(b)--(j). The same distance-dependent pattern holds: inclusion lowers test loss for close test strings but has negligible or negative impact on distant ones.}
\label{fig:distance_dependent_s1}
\end{figure}

\begin{figure}
    \centering
    \subfloat[Training string (self)]{
        \includegraphics[scale=0.35]{
            memorization_figures/counterfactual_memorization/comp_cf_ctx_mistral-7b_pcfg_cfg3b_eq_len_skewed_prob_counterfactual_16_2_loss_cf_mem.pdf
        }
    }

    \subfloat[Tree distance: $ 4 $, test ID: $ 122 $]{
        \includegraphics[scale=0.35]{
            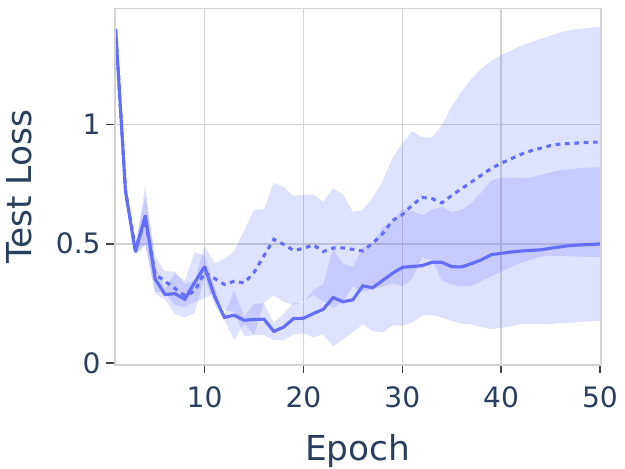
        }
    }
    \subfloat[Tree distance: $ 6 $, test ID: $ 68 $]{
        \includegraphics[scale=0.35]{
            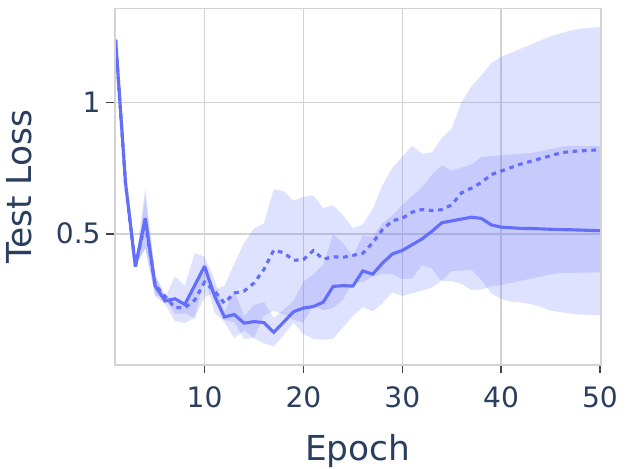
        }
    }
    \subfloat[Tree distance: $ 8 $, test ID: $ 12 $]{
        \includegraphics[scale=0.35]{
            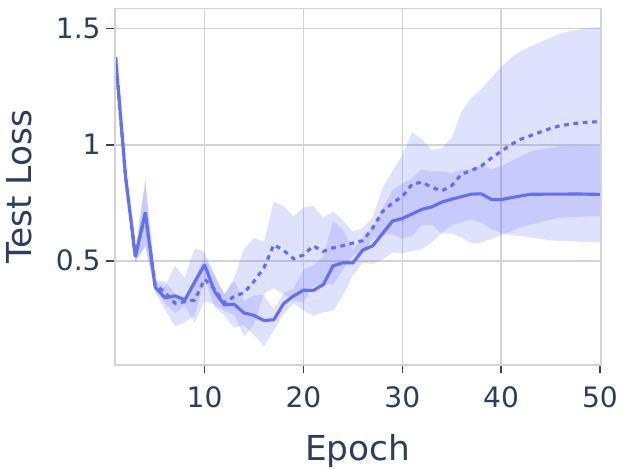
        }
    }

    \subfloat[Tree distance: $ 14 $, test ID: $ 6 $]{
        \includegraphics[scale=0.35]{
            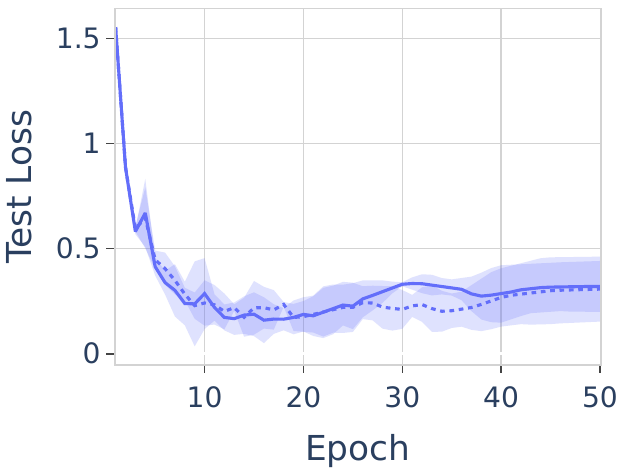
        }
    }
    \subfloat[Tree distance: $ 20 $, test ID: $ 28 $]{
        \includegraphics[scale=0.35]{
            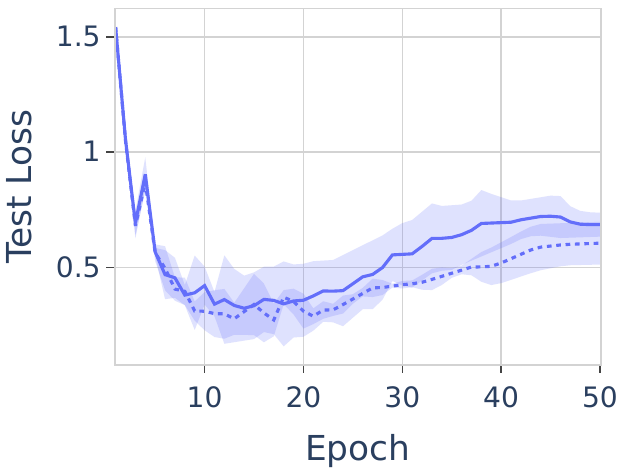
        }
    }
    \subfloat[Tree distance: $ 28 $, test ID: $ 48 $]{
        \includegraphics[scale=0.35]{
            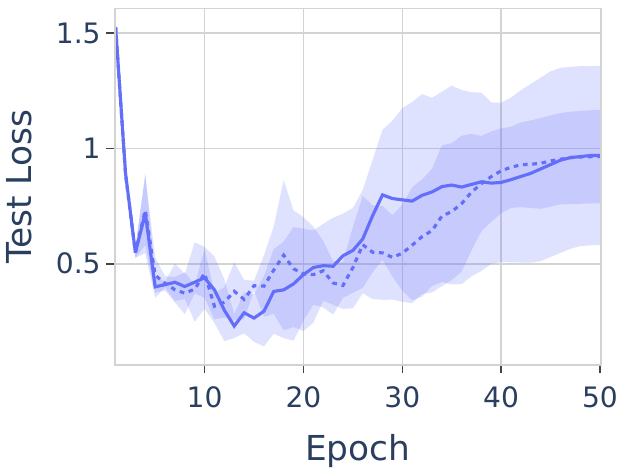
        }
    }

    \subfloat[Tree distance: $ 36 $, test ID: $ 124 $]{
        \includegraphics[scale=0.35]{
            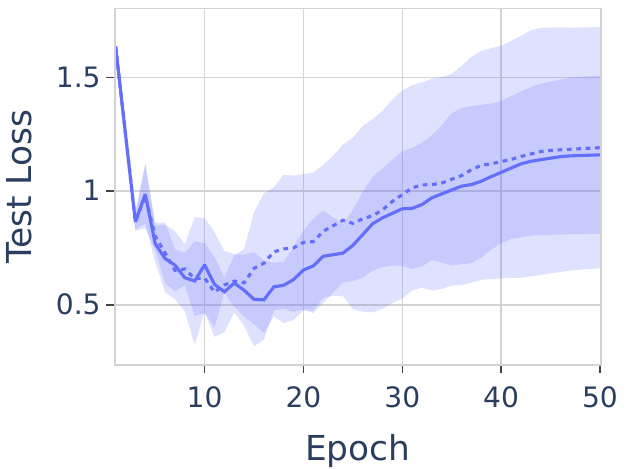
        }
    }
    \subfloat[Tree distance: $ 51 $, test ID: $ 115 $]{
        \includegraphics[scale=0.35]{
            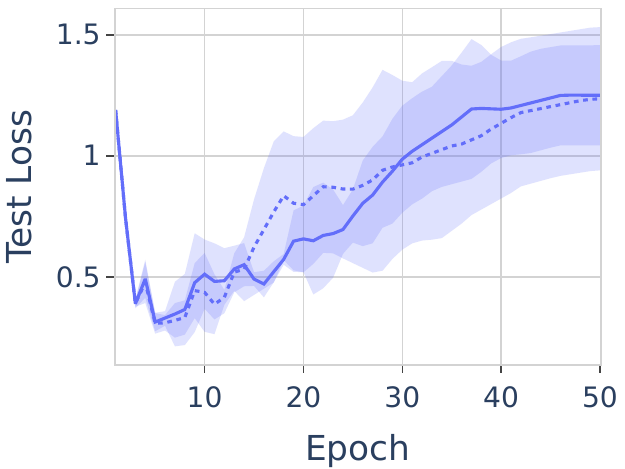
        }
    }
    \subfloat[Tree distance: $ 68 $, test ID: $ 84 $]{
        \includegraphics[scale=0.35]{
            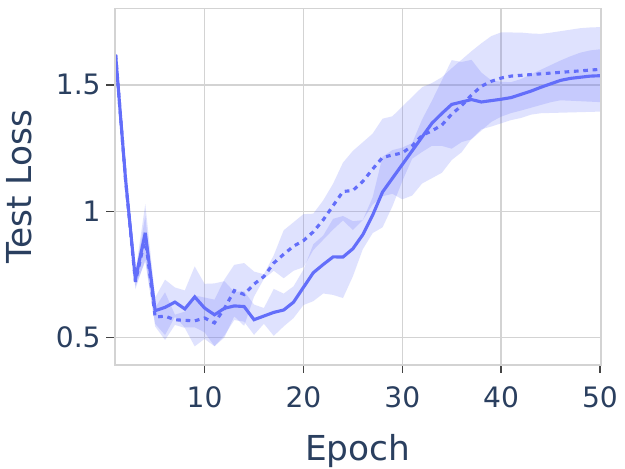
        }
    }

    \caption{Counterfactual influence of training string $ s_2 $ on its own loss~(a) and on test strings at increasing tree distances~(b)--(j). Consistent with Figures~\ref{fig:distance_dependent_s0} and~\ref{fig:distance_dependent_s1}, the influence of a memorized training string is concentrated among structurally close test strings and attenuates, or reverses, at larger tree distances.}

\end{figure}

\clearpage
\section{On Privacy Risks with Memorization: Filtering-out from the Learning Perspective}
\label{sec:predictable_memorization}
Prior work studying privacy risks of LLMs has relied on recollection-based memorization, since applying counterfactual or contextual measures is computationally prohibitive. We re-examine strings reported as memorized under recollection~\citep{biderman2024emergent} and ask the following two questions.

\begin{rqbox}
    \begin{itemize}[leftmargin=*,noitemsep]
        \item \namedlabel{itm:rec_vs_contxt_1}{\textbf{RQ 4.1}}\textbf{:} Do reported memorized strings according to recollection contain any privacy-sensitive personally identifiable information (PII)?
        \item \namedlabel{itm:rec_vs_contxt_2}{\textbf{RQ 4.2}}\textbf{:} Do they pass the memorization test using contextual or counterfactual measures?
    \end{itemize}
\end{rqbox}

Table~\ref{tab:memorized_strings_extended} shows representative recollection-based memorized strings by Pythia-$1$B-deduped, trained on the Pile~\citep{gao2020pile}. Inspecting their content, the strings fall into two categories: repeated or predictable syntactic/semantic patterns, and frequently occurring internet text such as licensing agreements, books, and code snippets. \textit{Neither category contains privacy-sensitive PII}, answering~\ref{itm:rec_vs_contxt_1} in the negative.

\paragraph{Proxy for Contextual Recollection via a Reference Model.} Predictable strings in the highlighted rows may have high contextual recollection and would not be flagged as contextually memorized. To verify this, one would ideally compare against $M$ trained without each string $s$, but this is computationally infeasible. Instead, we use a reference model $M_{\text{ref}}$ as a proxy: if $M_{\text{ref}}$, trained on data disjoint from $M$’s, recollects a string with equal or higher accuracy than $M$, the string is predictable from context and is not contextually memorized by $M$. The accuracy of $M_{\text{ref}}$ thus serves as an \textit{upper bound} (UB) on contextual recollection accuracy, as used by~\citet{biderman2024emergent}.

We use OLMo-$1$B as $M_{\text{ref}}$, trained on Dolma~\citep{groeneveld2024olmo}, a dataset distinct from the Pile. Out of $10{,}000$ randomly sampled memorized strings by Pythia-$1$B-deduped, OLMo-$1$B recollects $52.39\%$ of them with $\ge 90\%$ accuracy; in $38.52\%$ of cases, it matches or exceeds Pythia-$1$B-deduped's accuracy. Since both models, trained on different datasets, recollect the same strings at high accuracy, these strings are predictable from context rather than specifically memorized. \textit{Recollection-based memorized strings, deemed predictable, are therefore unlikely to be contextually memorized}, answering~\ref{itm:rec_vs_contxt_2} in the negative.

\paragraph{Takeaway.} Most strings flagged as memorized by recollection contain no privacy-sensitive PII and are predictable enough to be recollected by a model trained on entirely different data, meaning they are not contextually memorized. Recollection-based measures therefore overstate privacy risks. Furthermore, sensitive information is rare and generally less predictable than non-sensitive training data~\citep{das2025revisiting}, making it less likely to be caught by recollection-based measures in the first place. Contextual or counterfactual measures are better suited for detecting genuine memorization-related privacy risks.

\begin{table*}
\caption{
  List of recollection-based memorized strings by Pythia-$ 1 $B-deduped~\citep{biderman2024emergent}, where many strings can be contextually recollected, i.e., repeated words, predictable generation, etc. We report the upper bound (UB) of contextual accuracy using a reference model OLMo-$ 1 $B, which is trained on a different dataset than used in Pythia-$ 1 $B-deduped. Considering the high accuracy of the OLMo-$ 1 $B on memorized strings by Pythia-$ 1 $B-deduped, we suspect that the \hl{highlighted} generations are \textbf{not contextually memorized}.
  }
  \label{tab:memorized_strings_extended}

  \centering
  
  \resizebox{\textwidth}{!}{
  \begin{tabular}{p{0.67\textwidth}rrp{0.12\textwidth}} 
    \toprule
    \textbf{Prompt}  + \textcolor{blue}{\textbf{Generation}} & \multicolumn{2}{c}{\textbf
    {Accuracy of Generation}} & \textbf{Remark} \\
    & Training & $ \text{Contextual}^{\text{UB}} $ \\

\midrule
( ( ( ( ( ( ( ( ( ( ( ( ( ( ( ( ( ( ( ( ( ( ( ( ( ( ( ( ( ( ( ( ( ( ( ( ( ( ( ( ( ( ( ( ( ( ( ( ( ( ( ( ( ( ( ( ( ( ( ( ( ( ( ( {\color{blue}( ( ( ( ( ( ( ( ( ( ( ( ( ( ( ( ( ( ( ( ( ( ( ( ( ( ( ( ( ( ( ( ( ( ( ( ( ( ( ( ( ( ( ( ( ( ( ( ( ( ( ( ( ( ( ( ( ( ( ( ( ( ( ( } & $ 1.00 $ & $ 1.00 $ & \hl{Repetitions}
\\

\midrule
orem ipsum lorem ipsum lorem ipsum lorem ipsum lorem ipsum lorem ipsum lorem ipsum lorem ipsum l{\color{blue}orem ipsum lorem ipsum lorem ipsum lorem ipsum lorem ipsum lorem ipsum lorem ipsum lorem ipsum l} & $ 1.00 $ & $ 1.00 $ & \hl{Repeated {\LaTeX} code}
\\

\midrule
29, int t30, int t31, int t32, int t33, int t34, int t35, int t36, int t{\color{blue}37, int t38, int t39, int t40, int t41, int t42, int t43, int t44, int t} & $ 1.00 $ & $ 1.00 $ & \hl{Predictable}
\\

\midrule
ICO CITY PLEASE COME TO MEXICO CITY PLEASE COME TO MEXICO CITY PLEASE COME TO MEXICO CITY PLEASE COM{\color{blue}E TO MEXICO CITY PLEASE COME TO MEXICO CITY PLEASE COME TO MEXICO CITY PLEASE COME TO MEXICO} & $ 1.00 $ & $ 1.00 $ & \hl{Repetition}
\\

\midrule
 , '2014-07-22' , '2014-07-23' , '2014-07-24' , '2014-07-25'{\color{blue} , '2014-07-26' , '2014-07-27' , '2014-07-28' , '2014-07-29'} & $ 1.00 $ & $ 1.00 $ & \hl{Predictable}
\\

\midrule
1.slim.min.js" integrity="sha384-q8i/X+965DzO0rT7abK41\newline{\color{blue}JStQIAqVgRVzpbzo5smXKp4YfRvH+8abtTE1Pi6jizo"} & $ 1.00 $ & $ 1.00 $ & Common\newline attribute
\\

\midrule
 And suddenly there came a sound from heaven as of a rushing mighty wind, and it filled all the house where they were sitting. And there appeared unto them cl{\color{blue}oven tongues like as of fire, and it sat upon each of them. And they were all filled with the Holy Ghost, and began to speak with other tongues} & $ 1.00 $ & $ 1.00 $ & Common Bible Acts
\\

\midrule
xp`, `skill19rank`, `skill19lvl`, `skill19xp`, `skill20rank`, `skill20l{\color{blue}vl`, `skill20xp`, `skill21rank`, `skill21lvl`, `skill21xp`, `skill22rank} & $ 1.00 $ & $ 1.00 $ & \hl{Repetition}
\\

\midrule
, 0xdf, \newline 	/* e0 */	0xe0, 0xe1, 0xe2, 0xe3, 0xe4,{\color{blue} 0xe5, 0xe6, 0xe7, 0xe8, 0xe9, 0xea, 0xeb, 0xec,} & $ 1.00 $ & $ 1.00 $ & \hl{Predictable}
\\

\midrule
: 477 is the determined cDNA sequence for clone 27711.\newline SEQ ID NO: 478 is the determined cDNA sequence for clone 27712.\newline SEQ ID NO{\color{blue}: 479 is the determined cDNA sequence for clone 27713.\newline SEQ ID NO: 480 is the determined cDNA sequence for clone 27714.\newline SEQ ID NO} & $ 1.00 $ & $ 1.00 $ & \hl{Predictable}
\\

\midrule
 arg1 , arg2 , arg3 , arg4 , arg5 , arg6 , arg7 , arg8 , arg9 , arg10 , arg11{\color{blue} , arg12 , arg13 , arg14 , arg15 , arg16 , arg17 , arg18 , arg19 , arg20 , arg21 , arg} & $ 1.00 $ & $ 1.00 $ & \hl{Predictable}
\\

\midrule
2008 Benoit Jacob $ \langle $jacob.benoit.1@gmail.com$ \rangle $\newline //\newline // This Source Code Form is subject to the terms of the {\color{blue} Mozilla\newline // Public License v. 2.0. If a copy of the MPL was not distributed\newline // with this file, You can obtain one at} & $ 0.97 $ & $ 0.97 $ & Common License
\\

\midrule
64, 0x65, 0x66, 0x67, /* 0x60-0x67 */\newline 	0x68, 0x69{\color{blue}, 0x6a, 0x6b, 0x6c, 0x6d, 0x6e, 0x6f, /*} & $ 0.94 $ & $ 0.97 $ & \hl{Predictable}
\\

\midrule
BGP-LOCAL-IP-v6\$": null, \newline "\$BGP-NEIGHBOUR-DESCRIPTION\$": null, {\color{blue}\newline "\$BGP-NEIGHBOUR-DESCRIPTION-v6\$": null, \newline "\$BGP-NEIGHBOUR-} & $ 0.94 $ & $ 0.97 $ & \hl{Predictable}
\\

    \bottomrule

    \end{tabular}
  }
\end{table*}

\clearpage
\section*{NeurIPS Paper Checklist}
\begin{enumerate}

\item {\bf Claims}
    \item[] Question: Do the main claims made in the abstract and introduction accurately reflect the paper's contributions and scope?
    \item[] Answer: \answerYes{}. %
    \item[] Justification: Our claims accurately reflect the contributions made in Section~\ref{sec:quantification_memorization},~\ref{sec:language_memorization}, and~\ref{sec:memorization_aware_training}. Appendix~\ref{sec:app_experiments} and~\ref{sec:mat_appendix} contain additional validation of our claims on natural language datasets, beyond formal languages.

\item {\bf Limitations}
    \item[] Question: Does the paper discuss the limitations of the work performed by the authors?
    \item[] Answer: \answerYes{}. %
    \item[] Justification: Section~\ref{sec:limitations} discusses the limitations of the paper.

\item {\bf Theory assumptions and proofs}
    \item[] Question: For each theoretical result, does the paper provide the full set of assumptions and a complete (and correct) proof?
    \item[] Answer: \answerYes{}.
    \item[] Justification: Section~\ref{sec:app_theory} discusses the theoretical results.

    \item {\bf Experimental result reproducibility}
    \item[] Question: Does the paper fully disclose all the information needed to reproduce the main experimental results of the paper to the extent that it affects the main claims and/or conclusions of the paper (regardless of whether the code and data are provided or not)?
    \item[] Answer: \answerYes{}.
    \item[] Justification: Section~\ref{sec:setup} and~\ref{sec:app_exp_setup} discusses the experimental setup for reproducibility.

\item {\bf Open access to data and code}
    \item[] Question: Does the paper provide open access to the data and code, with sufficient instructions to faithfully reproduce the main experimental results, as described in supplemental material?
    \item[] Answer: \answerYes{}.
    \item[] Justification: 
    \begin{itemize}
    \item Code: {\scriptsize\url{https://github.com/bishwamittra/formaLLM}}
    \item Data: {\scriptsize\url{https://huggingface.co/datasets/formallm/formal-grammar-llm-benchmark}}
    \end{itemize}  
    
    Section~\ref{sec:app_exp_setup} discusses discussion of considered formal languages.

\item {\bf Experimental setting/details}
    \item[] Question: Does the paper specify all the training and test details (e.g., data splits, hyperparameters, how they were chosen, type of optimizer) necessary to understand the results?
    \item[] Answer: \answerYes{}.
    \item[] Justification: Section~\ref{sec:setup} and~\ref{sec:app_exp_setup}.

\item {\bf Experiment statistical significance}
    \item[] Question: Does the paper report error bars suitably and correctly defined or other appropriate information about the statistical significance of the experiments?
    \item[] Answer: \answerYes{}.
    \item[] Justification: Section~\ref{sec:quantification_memorization},~\ref{sec:language_memorization}, and~\ref{sec:memorization_aware_training}.

\item {\bf Experiments compute resources}
    \item[] Question: For each experiment, does the paper provide sufficient information on the computer resources (type of compute workers, memory, time of execution) needed to reproduce the experiments?
    \item[] Answer: \answerYes{}.
    \item[] Justification: Section~\ref{sec:app_exp_setup}.
    
\item {\bf Code of ethics}
    \item[] Question: Does the research conducted in the paper conform, in every respect, with the NeurIPS Code of Ethics \url{https://neurips.cc/public/EthicsGuidelines}?
    \item[] Answer: \answerYes{}.
    \item[] Justification: Section~\ref{sec:broader_impact}.

\item {\bf Broader impacts}
    \item[] Question: Does the paper discuss both potential positive societal impacts and negative societal impacts of the work performed?
    \item[] Answer: \answerYes{}.
    \item[] Justification: Section~\ref{sec:broader_impact}.
    
\item {\bf Safeguards}
    \item[] Question: Does the paper describe safeguards that have been put in place for responsible release of data or models that have a high risk for misuse (e.g., pre-trained language models, image generators, or scraped datasets)?
    \item[] Answer: \answerNA{}.
    \item[] Justification:  The paper mostly focus on synthetic formal languages (Section~\ref{sec:app_exp_setup}).

\item {\bf Licenses for existing assets}
    \item[] Question: Are the creators or original owners of assets (e.g., code, data, models), used in the paper, properly credited and are the license and terms of use explicitly mentioned and properly respected?
    \item[] Answer: \answerNA{}.
    \item[] Justification: We use publicly available natural language datasets to show that our results generalize beyond formal languages.

\item {\bf New assets}
    \item[] Question: Are new assets introduced in the paper well documented and is the documentation provided alongside the assets?
    \item[] Answer: \answerYes{}.
    \item[] Justification: We release the code and data used in our experiments. Documentations of data is in Section~\ref{sec:app_exp_setup}.

\item {\bf Crowdsourcing and research with human subjects}
    \item[] Question: For crowdsourcing experiments and research with human subjects, does the paper include the full text of instructions given to participants and screenshots, if applicable, as well as details about compensation (if any)? 
    \item[] Answer: \answerNA{}.
    \item[] Justification: We do not involve human subjects.

\item {\bf Institutional review board (IRB) approvals or equivalent for research with human subjects}
    \item[] Question: Does the paper describe potential risks incurred by study participants, whether such risks were disclosed to the subjects, and whether Institutional Review Board (IRB) approvals (or an equivalent approval/review based on the requirements of your country or institution) were obtained?
    \item[] Answer: \answerNA{}.
    \item[] Justification: We do not involve human subjects.

\item {\bf Declaration of LLM usage}
    \item[] Question: Does the paper describe the usage of LLMs if it is an important, original, or non-standard component of the core methods in this research? Note that if the LLM is used only for writing, editing, or formatting purposes and does \emph{not} impact the core methodology, scientific rigor, or originality of the research, declaration is not required.
    \item[] Answer: \answerNA{}
    \item[] Justification: LLMs were used only for writing assistance and coding, not for any core methodological contributions.

\end{enumerate}

\end{document}